\documentclass[11pt, a4paper, logo, copyright, table]{googledeepmind}

\usepackage[authoryear, sort&compress, round]{natbib}
\usepackage[utf8]{inputenc} 
\usepackage[T1]{fontenc}    
\usepackage{hyperref}       
\usepackage{url}            
\usepackage{booktabs}       
\usepackage{amsfonts}       
\usepackage{nicefrac}       
\usepackage{microtype}      
\definecolor{darkblue}{rgb}{0, 0, 0.7}
\hypersetup{colorlinks=true, citecolor=darkblue, linkcolor=darkblue, urlcolor=darkblue}

\usepackage{enumitem}
\usepackage[noend]{algpseudocode}
\usepackage{algorithm}
\usepackage{algorithmicx}
\usepackage{mathtools}
\usepackage{nccmath}
\usepackage{multirow}
\usepackage{bigdelim}
\usepackage{color, colortbl}

\usepackage{amsthm}
\usepackage{makecell}
\usepackage{times}
\usepackage{latexsym}
\usepackage{tabularx}
\usepackage{array}
\usepackage{multirow}
\usepackage{siunitx}
\usepackage[pdftex]{graphicx}
\usepackage{enumitem}
\usepackage{xspace}
\usepackage{caption}
\usepackage{wrapfig}
\usepackage{subcaption}

\usepackage{dcolumn}
\newcolumntype{d}{D{.}{.}{-1}}
\newcolumntype{z}{D{(}{\ (}{1.1}}
\usepackage{bbm}
\usepackage[procnames]{listings}
\newcommand{\cmark}{\ding{51}}
\newcommand{\xmark}{\ding{55}}
\definecolor{custom_green}{rgb}{0.0, 0.5, 0.0}
\definecolor{custom_red}{rgb}{1.0, 0.01, 0.24}

\usepackage{adjustbox}

\usepackage[many]{tcolorbox}

\title{Relaxed Recursive Transformers: \\Effective Parameter Sharing with Layer-wise LoRA}

\correspondingauthor{my-email@google.com}

\author[1,*]{Sangmin Bae}
\author[2]{Adam Fisch}
\author[3]{Hrayr Harutyunyan}
\author[3]{Ziwei Ji}
\author[2]{Seungyeon Kim}
\author[2,$\dagger$]{Tal Schuster}

\affil[1]{KAIST AI}
\affil[2]{Google DeepMind}
\affil[3]{Google Research}
\affil[*]{Work done during an internship at Google DeepMind}
\affil[$\dagger$]{Corresponding Author}

\begin{abstract}
Large language models (LLMs) are expensive to deploy. Parameter sharing offers a possible path towards reducing their size and cost, but its effectiveness in modern LLMs remains fairly limited. In this work, we revisit ``layer tying'' as form of parameter sharing in Transformers, and introduce novel methods for converting existing LLMs into smaller ``Recursive Transformers'' that share parameters across layers, with minimal loss of performance. Here, our Recursive Transformers are efficiently initialized from standard pretrained Transformers, but only use a single block of unique layers that is then repeated multiple times in a loop. We further improve  performance by introducing Relaxed Recursive Transformers that add flexibility to the layer tying constraint via depth-wise low-rank adaptation (LoRA) modules, yet still preserve the compactness of the overall model. We show that our recursive models (e.g., recursive Gemma 1B) outperform both similar-sized vanilla pretrained models (such as TinyLlama 1.1B and Pythia 1B) and knowledge distillation baselines---and can even recover most of the performance of the original ``full-size'' model (e.g., Gemma 2B with no shared parameters). Finally, we propose Continuous Depth-wise Batching, a promising new inference paradigm enabled by the Recursive Transformer when paired with early exiting. In a theoretical analysis, we show  that this has the potential to lead to significant (2-3$\times$) gains in inference throughput.

\end{abstract}

\begin{document}

\maketitle

\section{Introduction}
\label{sec:introduction}

Efficient deployment of large language models (LLMs) demands a balance between performance and resources~\citep{DBLP:journals/corr/abs-2404-02258, DBLP:conf/icml/LeviathanKM23, DBLP:journals/corr/abs-2408-00118, DBLP:journals/tmlr/Wan0LA0LQYZZC024, DBLP:journals/corr/abs-2404-14294}.
While larger models with more parameters consistently demonstrate superior performance~\citep{rosenfeld2020a,rae2021scaling,hoffmann2022trainingcomputeoptimallargelanguage}, their substantial memory and computational demands are expensive~\citep{MLSYS2023_c4be71ab}. Parameter sharing approaches~\cite[e.g.][]{DBLP:conf/iclr/DehghaniGVUK19, DBLP:conf/aaai/XiaHTTHQ19, DBLP:conf/iclr/LanCGGSS20, DBLP:conf/sustainlp/TakaseK23}, wherein weights are reused across model layers, can lower these costs by reducing memory footprint, and thereby allow for the use of fewer (or lower-grade) accelerators, or larger batch sizes for better throughput.
While parameter sharing has shown encouraging capabilities in previous work~\citep{DBLP:conf/iclr/LanCGGSS20,DBLP:conf/icml/GiannouRS0LP23}, its application to modern LLMs has yielded limited reported success.

In this work, we revisit parameter sharing for LLMs, and propose novel methodologies to \emph{convert} existing, unshared models into smaller, and more efficient,  Recursive Transformers. These models use a single block of unique layers that are recursively reused across multiple loops, yet still achieve impressive performance relative to their reduced size.
To mitigate the potential performance degradation associated with parameter sharing, we first initialize the shared block of layers based on the original model's pre-trained parameters, and then finetune the resulting recursive model for a limited number of ``uptraining'' steps. Importantly, we show that our initialization strategies allow us to achieve strong performance with minimal training time.
This is aligned with observations that model compression techniques such as layer skipping~\citep{DBLP:conf/acl/Zhang00S0CM24, DBLP:journals/corr/abs-2311-15436, DBLP:conf/iclr/FanGJ20, DBLP:conf/acl/ElhoushiSLHWL0A24}, pruning~\citep{DBLP:conf/iclr/FrankleC19, DBLP:conf/cvpr/RamanujanWKFR20} or nesting~\citep{devvrit2023matformernestedtransformerelastic} can preserve surprisingly high performance---further motivating our approach of compressing models to more compact yet performant architectures (here, repeated layers with low-rank adapters).

\begin{figure}[t!]
    \vspace{-6pt}
    \centering
    \includegraphics[width=0.98\textwidth]{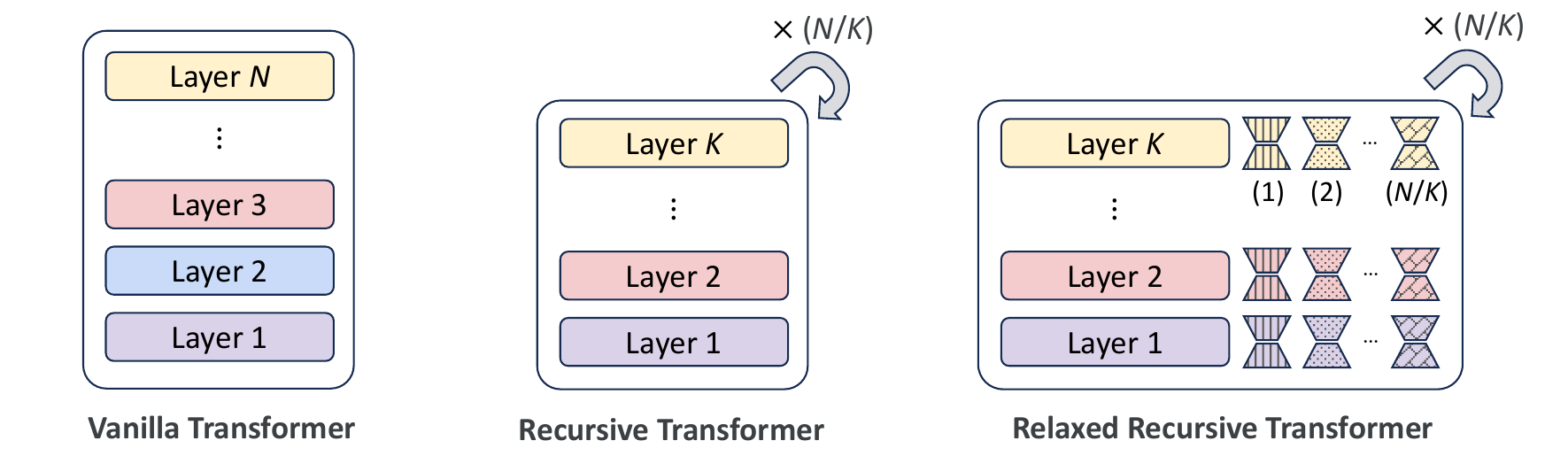}
    \vspace{-2pt}
    \caption{
    Overview of the conversion from a vanilla \textit{N}-layer Transformer to a Recursive Transformer with $N / K$ blocks of \textit{K} shared layers. The Recursive Transformer is obtained by repeating a single block of \textit{K} layers multiple times, resulting in a looped architecture. The Recursive Transformer can  also be converted into a Relaxed Recursive Transformer by adding  layer-specific LoRA modules. This  preserves many of the advantages of weight sharing, but also allows for better performance.
    }
    \label{fig:overview}
\end{figure}

As depicted in Figure\,\ref{fig:overview}, we further propose the Relaxed Recursive Transformer,
an extension of the Recursive Transformer in which the  weight tying across repeated layer blocks is slightly relaxed through the incorporation of multiple layer-specific, low-rank adaptation\,(LoRA) modules~\citep{DBLP:conf/iclr/HuSWALWWC22}. 
Despite its simplicity, this strategy offers several non-trivial advantages. First, it allows for low-rank deltas between shared layers, while only adding minimal  overhead.
Second, the rank of the LoRA matrices can be adjusted to control the degree of relaxation, which directly influences model capacity. 
Furthermore, since the relaxed model has the same overall shape as the original Transformer, we can efficiently initialize LoRA modules via truncated Singular Value Decomposition~\citep{hansen1987truncated} on the residual matrices between the original layer weights and the shared layer weights.
Hence, the rank values serve as a pivotal hyperparameter, enabling the Relaxed Recursive Transformer to seamlessly transition between the two extremes of the vanilla and Recursive Transformer architectures.

While the primary focus of this paper lies in how to formulate and train Recursive Transformers, we also highlight their potential to achieve significant throughput gains via a new batched inference paradigm, Continuous Depth-wise Batching, that their recursive nature enables. Prior work introduced continuous sequence-wise batching~\citep{DBLP:conf/osdi/YuJKKC22, DBLP:conf/sosp/KwonLZ0ZY0ZS23}, which leverages the fact that the computation performed to compute a new token is functionally the same (and uses the same model parameters) regardless of the token position within the sequence. This allows new requests to be continuously scheduled when slots within a batch become available. For example, when one response is completed, the start of the next response to be formed can immediately take the finished response's place in the batch, without waiting for the rest of the batch responses that might be longer. 
In our Recursive Transformer, parameter sharing occurs not only across different timesteps, but also across different depths (loop iterations). This enables an extra dimension of dynamic grouping: jointly computing different iterations of the looped layer blocks per individual responses within the same batch.\looseness=-1

Our key contributions are as follows:
\begin{itemize}[leftmargin=*]
\vspace{-8pt}
\item 
We introduce a framework for initializing and training Relaxed Recursive Transformers and demonstrate strong performance compared to non-recursive models of comparable size.
For example, when we uptrained a recursive Gemma 1B model converted from a pretrained Gemma 2B~\citep{team2024gemma}, we observed up to 13.5 absolute accuracy improvement (22\% error reduction) on few-shot tasks compared to a non-recursive Gemma 1B model (pretrained from scratch). Furthermore, we show that by incorporating knowledge distillation~\citep{DBLP:journals/corr/HintonVD15, DBLP:conf/emnlp/KimR16}, our recursive Gemma model, uptrained on 60 billion tokens, achieves performance on par with the full-size Gemma model trained on a massive 3 trillion token corpus (see \textsection\ref{sec:main_results} for  details).\looseness=-1

\item
Based on our Relaxed Recursive Transformer, we also evaluate a key use case for continuous depth-wise batching with early-exiting~\citep{DBLP:conf/emnlp/BaeKSY23, DBLP:conf/nips/SchusterFG0B0TM22, DBLP:conf/iclr/ElbayadGGA20, DBLP:journals/corr/Graves16}, which opportunistically makes predictions for samples with high confidence at earlier stages. From our simulation, Early Exits reveal a substantial throughput improvement of up to 2-3$\times$ compared to a vanilla Transformer with the same architecture.
Notably, the recursive Gemma model, which outperforms the vanilla Pythia model, can theoretically achieve a nearly 4$\times$ increase in throughput (see \textsection\ref{exp:hypothetical_generation_speedup} for details).
\end{itemize}
\section{Effective Model Compression with Recursive Patterns}
\label{sec:methods}

In this section, we present the main details of our method for converting a vanilla Transformer model into a parameter-shared model that outperforms models of equivalent size. We first provide a short overview of the Transformer architecture ($\S$\ref{sec:methods:background}).
Then, we introduce the Recursive Transformer and present effective techniques to initialize its looped layers by leveraging the weights of the original pretrained model ($\S$\ref{sec:methods:recursive}).
In $\S$\ref{sec:methods:relaxed}, we relax the parameter-sharing constraint in the model design, and add a limited set of layer-specific parameters to further improve the model's accuracy while maintaining compact representations. 
Finally, we show how, beyond reduced memory, Recursive Transformers readily support further throughput optimizations via a novel inference paradigm ($\S$\ref{sec:methods:early_exit}).

\subsection{Basic Transformer Architecture}
\label{sec:methods:background}

Large language models~\citep{DBLP:journals/corr/abs-2408-00118, DBLP:journals/corr/abs-2403-05530, DBLP:journals/corr/abs-2303-08774, DBLP:journals/corr/abs-2407-21783} typically leverage the Transformer architecture \citep{vaswani2017attention}. A Transformer consists of \textit{L} layers, where the hidden states at each time step \textit{t} are computed by running through the series of layers:\looseness=-1
\begin{align}
    \mathbf{h}^{\ell}_{t} = f(\mathbf{h}^{\ell - 1}_{t}; \,\Phi_\ell), \,\,\ell \in [1, L],
    \label{eq:full_size_model}
\end{align}
with $\mathbf{h}^{0}_{t}$ representing the embedding of the token $y_{t-1}$ from the previous time step, and $\Phi_\ell$ denoting the trainable parameters of the $\ell$-th layer.  

Each layer has two core components: a multi-head attention (MHA) mechanism and a feed-forward network (FFN). 
MHA employs multiple attention heads to capture diverse relationships within the input sequence via linear attention weights and scaled dot-product attention mechanisms. 
The FFN structure typically consists of two linear transformations, but different models exhibits distinct structural variations. 
See Appendix\,\ref{app:transformer} for further details.

\subsection{Recursive Transformer: Looped Layer Tying}
\label{sec:methods:recursive}

In this work, we revisit parameter sharing in the context of LLMs and propose the Recursive Transformer architecture.
Among various looping strategies (refer to Appendix\,\ref{app:looping_strategy}), we specifically adopt the CYCLE strategy~\citep{DBLP:conf/sustainlp/TakaseK23} for Recursive Transformers, wherein a single block of unique layers is recursively reused. This inherent design aligns seamlessly with early-exiting mechanisms, potentially offering substantial speedup.
The model's hidden states are computed as:
\begin{align}
    \textbf{h}_t^\ell = f(\textbf{h}_t^{\ell-1}; \,\Phi'_{((\ell-1) \bmod L/B) + 1}), \,\,\ell \in [1, L],
    \label{eq:recursive_transformer}
\end{align}
where the parameter-shared model is parameterized by $\Phi'$, and \textit{B} denotes the number of looping blocks (we restrict \textit{B} to be a factor of \textit{L}).
For example, Gemma 2B~\citep{team2024gemma} with 18 layers can be converted to a recursive variant with 2 blocks by storing weights for only the first 9 layers. The forward pass will loop twice through these 9 layers.
We tie all trainable parameters, including the weights of the linear layers in the Transformer blocks and the weights of the RMSNorm~\citep{DBLP:conf/nips/ZhangS19a}.\looseness=-1

\begin{figure}[t!]
    \centering
        \includegraphics[width=\textwidth]{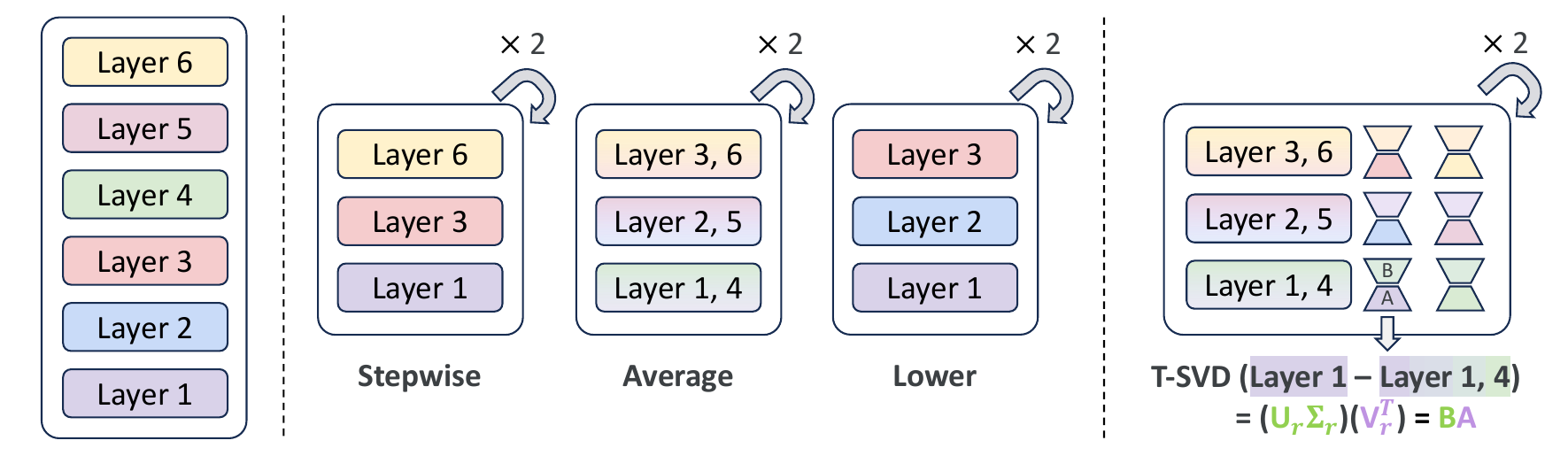}
    \caption{
    \textbf{Left:} An example of unshared, full-size model with 6 layers.
    \textbf{Middle:} Three proposed methodologies for initializing looped layers in a Recursive Transformer. 
    Each layer number indicates the source layer in the full-size model used for initialization.
    \textbf{Right:} 
    Example of a Relaxed Recursive Transformer initialized by SVD method. 
    Here, looped layers are initialized using the Average method.
    }
    \label{fig:looping_init_overview}
\end{figure}

\paragraph{Initialization techniques for looped layers}
To mitigate the potential performance drop associated with reduced capacity in parameter-shared models, we propose several novel initialization methodologies to facilitate effective knowledge transfer from unshared, pretrained models to Recursive Transformers.
Figure~\ref{fig:looping_init_overview} illustrates three such techniques.
The Stepwise method selects intermediate layers at specific intervals while keeping the first and last layer fixed. This is motivated by prior work~\citep{DBLP:conf/icml/LiuWDZY0S0TRC23, DBLP:conf/acl/Zhang00S0CM24, DBLP:journals/corr/abs-2311-15436, DBLP:conf/iclr/FanGJ20} showing minimal impact on generation quality when skipping a few layers in LLMs. 
The Average method initializes the shared weights among tied layers by averaging their weight matrices, whereas the Lower method directly uses weights from the first \textit{K} layers of the unshared model. 
We conducted a brief uptraining on 15 billion tokens to investigate the extent of performance recovery in these initialized models ($\S$\ref{sec:exp_initialization}) and found the Stepwise approach to perform best for Recursive Transformers. However, we found the Average method to perform best for Relaxed Recursive Transformers, discussed next.

\subsection{Relaxed Recursive Transformer: Multi-LoRA Layers}
\label{sec:methods:relaxed}

While full layer-tying is effective for compressing the model's size while maintaining strong capabilities, it has two noticeable limitations: (1) the set of possible model sizes is limited to scaling the number of layers, and (2) each model layer ends up having to serve multiple roles associated with different depths of the model.
To address this, we introduce Relaxed Recursive Transformers in which we incorporate independent adapter modules~\citep{DBLP:conf/iclr/HuSWALWWC22, DBLP:conf/icml/HoulsbyGJMLGAG19} for each layer, relaxing the strict parameter sharing.
While we experiment with various approaches like layer-specific prefixes~\citep{DBLP:journals/corr/abs-2110-07602} (see Appendix\,\ref{app:prefix_tuning_relaxation}), we find low-rank adaptation (LoRA) modules~\citep{DBLP:conf/iclr/HuSWALWWC22} to efficiently capture the subtle variations between tied layers. Specifically, we modify Eq.\,\ref{eq:recursive_transformer} to:
\begin{align}
    \textbf{h}_t^\ell = f(\textbf{h}_t^{\ell-1}; \,\Phi'_{((\ell-1) \bmod L/B) + 1}, \Delta \Phi'_\ell), \,\,\ell \in [1, L],
    \label{eq:relaxed_recursive_transformer}
\end{align}
where $\Delta \Phi'$ is the (small) set of parameters for the LoRA modules. 

In this relaxed model, each looped layer is augmented with multiple LoRA modules. For example, a recursive model with two loop iterations has a single block of shared layers, and two different LoRA modules are attached to each layer within this block. 
The first and second LoRA modules per layer are used during the first and second loop iterations, respectively. 
Functionally, these LoRA modules introduce low-rank deltas to all of the shared, linear weight matrices.
More concretely, for a base transformation $\textbf{h} = \textbf{W}'\textbf{x}$, our modified forward pass yields $\textbf{h} = \textbf{W}'\textbf{x} + \Delta \textbf{W}' \textbf{x} = \textbf{W}'\textbf{x} + \textbf{B}\textbf{A}\textbf{x}$, where $ \textbf {A}\in \mathbb{R}^{(r \times k)}$ and $\textbf {B} \in \mathbb{R}^{(d \times r)}$ denote the weight matrices of LoRA with rank \textit{r}.

\paragraph{LoRA initialization via truncated SVD}
Unlike typical LoRA finetuning setups that train only the LoRA parameters, here we train all model parameters to let the shared parameters learn an optimal centroid for all of the layer depths that they support. 
Therefore, instead of following standard zero initialization for adaptation to the frozen base model, we propose novel initialization methods, especially designed for Relaxed Recursive Transformers.
To effectively match the performance of the original full-size model after initializing the tied weights as described in $\S$\ref{sec:methods:recursive}, we aim for the sum of the tied weights ($\Phi'$) and LoRA weights ($\Delta \Phi'$) to approximately recover the full-size model's weights ($\Phi$). 
We exploit truncated Singular Value Decomposition (SVD)~\citep{hansen1987truncated} on residual matrices between original weights and tied weights:
\begin{align}
    \mathbf{U}_r^\ell, \mathbf{\Sigma}_r^\ell, \mathbf{V}_r^\ell = \text{Truncated SVD}(\mathbf{W}_\ell - \mathbf{W'}_{((\ell-1) \bmod L/B) + 1}; \,\,r), \,\,\ell \in [1, L],
    \label{eq:truncated_svd}
\end{align}
where outputs retain the first \textit{r} columns corresponding to the \textit{r} largest singular values. \textit{\textbf{W}} denotes the weight matrices of the full-size model, and $\text{\textit{\textbf{W}}}'$ denotes those of the Recursive Transformer.
We initialize the LoRA's weights with principal components in Eq.\,\ref{eq:truncated_svd}: \textit{\textbf{B}} as the product of $\text{\textit{\textbf{U}}}_r$ and $\mathbf{\Sigma}_r$, and \textit{\textbf{A}} as the transpose of the right singular vectors $\text{\textit{\textbf{V}}}_r$ (see Figure\,\ref{fig:looping_init_overview}).

\newtheorem*{remark*}{Remark}
\begin{remark*}
    By initializing LoRA weights through the proposed truncated SVD methodology, the rank of the LoRA modules serves as a pivotal hyperparameter, enabling the Relaxed Recursive Transformer to seamlessly transition between the two extremes of the vanilla and Recursive Transformer architectures. 
\end{remark*}

\vspace{-6pt}
With sufficiently large ranks, our Relaxed Recursive Transformer (Eq.\,\ref{eq:relaxed_recursive_transformer})  approximates the full-size vanilla model (Eq.\,\ref{eq:full_size_model}):
\begin{align}
    \mathbf{W}\textbf{x} \approx \mathbf{W}'\textbf{x} + (\mathbf{U}_r \mathbf{\Sigma}_r)(\mathbf{V}^\top_r)\textbf{x} = \mathbf{W}'\textbf{x} + \mathbf{B}\mathbf{A}\textbf{x} = \mathbf{W}'\textbf{x} + \Delta \mathbf{W}'\textbf{x},
    \label{eq:relaxed_to_vanilla}
\end{align}
Meanwhile, setting the rank to zero reduces the model to a Recursive Transformer, as the LoRA modules contribute no additional parameters, highlighting the flexibility of this relaxation approach.

\subsection{Continuous Depth-wise Batching and Early-Exiting}
\label{sec:methods:early_exit}

\begin{figure}[t!]
    \vspace{-8pt}
    \centering
    \begin{subfigure}[t]{0.35\textwidth}
        \includegraphics[width=\textwidth]{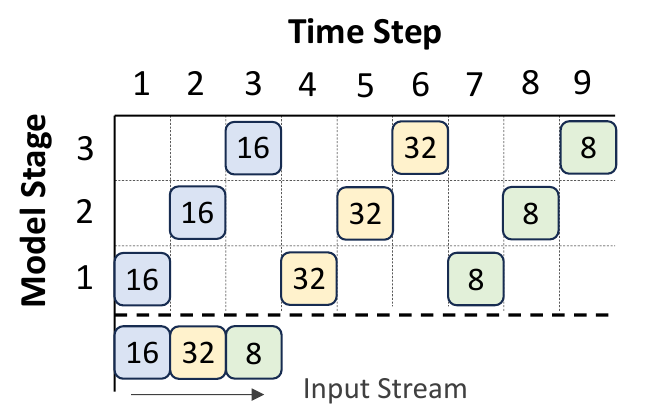}
        \subcaption{Vanilla Batching}
    \end{subfigure}
    \centering
    \begin{subfigure}[t]{0.30\textwidth}
        \includegraphics[width=\textwidth]{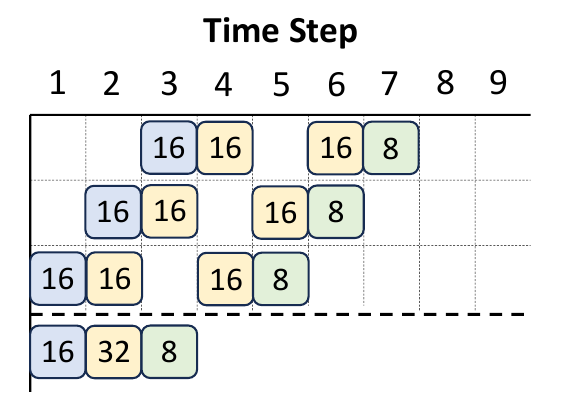}
        \subcaption{Depth-wise Batching}
    \end{subfigure}
    \centering
    \begin{subfigure}[t]{0.30\textwidth}
        \includegraphics[width=\textwidth]{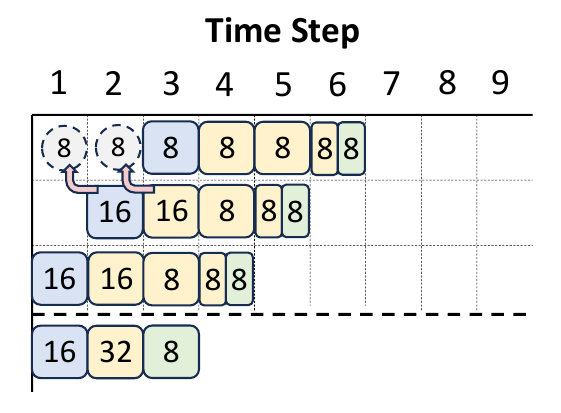}
        \subcaption{With Early-exiting}
        \label{fig:cdb_early_exiting}
    \end{subfigure}
    \caption{
    An illustrative example of a continuous depth-wise batching strategy together with early-exiting. We assume a maximum batch size of $32$, three model ``stages'' (e.g., layer blocks), and a stream of batched inputs that arrive sequentially in time. In \textbf{(a)},  all three model stages must complete for the first (non-maximal) batch of $16$ before the second batch of $32$ examples that arrives next can be started. In \textbf{(b)}, however, half of second batch of $32$ examples can share computation with the first batch of $16$ that is still finishing. Finally, \textbf{(c)} demonstrates a situation where some examples within each batch can early-exit after stage 2; their vacant slots in the batch are then immediately filled.
    }
    \label{fig:continuous_depthwise_batching}
\end{figure}

In real-world deployments, user requests arrive sequentially and asynchronously. Recent research has introduced continuous sequence-wise batching~\citep{DBLP:conf/osdi/YuJKKC22, DBLP:conf/sosp/KwonLZ0ZY0ZS23}, a serving strategy that allows new requests to immediately replace completed (terminated) sequence within a batch. This approach exploits the fact that the computation performed for a new token is functionally the same and utilize the same model parameters. By continuously scheduling requests in this manner, models can operate at their maximum batch capacity, thereby enhancing serving efficiency.

The repetitive structure of Recursive Transformers allows for the same function to be applied not just across sequences, but also across depths (loop iterations). This introduces a new dimension for continuous batching, which we call Continuous Depth-wise Batching. This technique enables the simultaneous computation of different iterations of the looped layer block for different samples (See Figure\,\ref{fig:continuous_depthwise_batching} for an example with a single forward pass; this easily extends to multiple decode iterations per request.) With a maximum batch size of 32, a standard Transformer must wait for all model stages to complete before processing new requests. In contrast, our Recursive Transformer, because it shares layer functions across all stages, can immediately schedule new incoming requests at timestep 2, maximizing batch size utilization. This strategy can yield a substantial speedup in generation and reduce the time to first token~\citep{DBLP:journals/corr/abs-2407-14057, DBLP:journals/corr/abs-2312-15234} through faster scheduling.

Throughput improvements from depth-wise batching are further amplified when combined with early-exiting~\citep{DBLP:conf/emnlp/BaeKSY23, DBLP:conf/nips/SchusterFG0B0TM22, DBLP:conf/iclr/ElbayadGGA20}. As depicted in Figure\,\ref{fig:cdb_early_exiting}, once some samples exit after certain looping iterations, queued requests can then be immediately scheduled. 
While Recursive Transformers leverage the speedup from early-exiting, they also inherently address a key challenge of batched inference in early-exiting approaches: the synchronization issue when serving large batches, as early-exited tokens might wait for others to complete processing through the entire model.
We demonstrate that Recursive Transformers, equipped with this dynamic sample scheduling at various depths, can theoretically allow up to 2-3$\times$ speedup on evaluated LLMs. 

\begin{table}[t!]
    \vspace{-1pt}
    \small
    \centering
    \resizebox{\textwidth}{!}{
    \setlength{\tabcolsep}{6.5pt}
    \begin{tabular}{l|cccccccc|ccc}
    \toprule
      &  \multicolumn{8}{c|}{\textbf{Model Architecture}} & \multicolumn{3}{c}{\textbf{Pretraining}} \\
    \cmidrule(l{2pt}r{2pt}){2-9} \cmidrule(l{2pt}r{2pt}){10-12}
     \textbf{Models} & N-emb & Emb & $N_L$ & $d_{model}$ & $N_{head}$ & $N_{KV}$ & $d_{head}$ & Vocab & Dataset & $N_{tok}$ & $L_{ctx}$  \\
    \midrule
    Gemma\,2B & 1.98B & 0.52B & 18 & 2048 & 8 & 1 & 256 & 256K & Unreleased & 3T & 8K \\[3pt]
    \multirow{2}{*}{TinyLlama\,1.1B} & \multirow{2}{*}{0.97B} & \multirow{2}{*}{0.13B} & \multirow{2}{*}{22} & \multirow{2}{*}{2048} & \multirow{2}{*}{32} & \multirow{2}{*}{4} & \multirow{2}{*}{64} & \multirow{2}{*}{32K} & SlimPajama\,+ & 73B$^\ast$ & \multirow{2}{*}{2K}  \\
     &  &  &  &  &  &  &  &  & Starcoderdata  & 32B  &  \\[3pt]
    Pythia\,1B & 0.81B & 0.21B & 16 & 2048 & 8 & 8 & 256 & 50K & Pile & 300B & 2K \\
    \bottomrule
    \end{tabular}
    }
    \caption{
    Key parameters and pretraining details of three models. 
    The sizes of each model refer to the number of embedding parameters (embedding matrices and classifier heads), and all other non-embedding parameters.
    Gemma and TinyLlama utilize Multi-Query~\citep{DBLP:journals/corr/abs-1911-02150} and Grouped-Query~\citep{DBLP:conf/emnlp/AinslieLJZLS23} attention mechanisms, which leads to a reduced number of key-value heads.
    $^\ast$We take an early TinyLlama checkpoint to study recursive conversions on top of an under-trained model on SlimPajama. The vanilla performance with longer pretraining is reported in Table~\ref{tab:pretrained_model_performance_app}.
    }
    \label{tab:model_arch}
\end{table}

\section{Experiments}
\label{sec:experiments}

\subsection{Experimental Setup}
\label{sec:exp_setup}

We evaluate our method on three popular pretrained LLMs: Gemma 2B\,\citep{team2024gemma}, TinyLlama 1.1B\,\citep{zhang2024tinyllama}, and Pythia 1B\,\citep{DBLP:conf/icml/BidermanSABOHKP23}. 
Table\,\ref{tab:model_arch} summarizes each model's architecture and pretraining recipes, and their few-shot performance is summarized in Appendix\,\ref{app:pretrained_model_performance}.
Unless stated otherwise, the number of looping blocks\,(\textit{B}) is set to 2 for all experiments. The results for Gemma with $\text{\textit{B}}=3$ is provided in the appendix.
After converting to Recursive Transformers, we uptrained models on the SlimPajama dataset\,\citep{cerebras2023slimpajama}.
We used the Language Model Evaluation Harness framework\,\citep{eval-harness} to evaluate accuracy on seven few-shot tasks, and averaged them for performance comparison.
Detailed experimental setup can be found in Appendix\,\ref{app:experimental_settings}.

\begin{table}[t!]
    \vspace{-3pt}
    \small
    \centering
    \resizebox{\textwidth}{!}{
    \setlength{\tabcolsep}{4pt}
    \begin{tabular}{l|c|cc|rrr|ccccccc|c}
    \toprule
      &  & \multicolumn{2}{c|}{\textbf{Uptrain}} & \multicolumn{3}{c|}{\textbf{Perplexity\,$\downarrow$}} & \multicolumn{8}{c}{\textbf{Few-shot Accuracy\,$\uparrow$}} \\
    \cmidrule(l{2pt}r{2pt}){3-4} \cmidrule(l{2pt}r{2pt}){5-7}  \cmidrule(l{2pt}r{2pt}){8-15} 
     \textbf{Models} & N-emb & PT & $N_{tok}$ & SlimP & RedP & PG19 & LD & HS & PQ & WG & ARC-e & ARC-c & OB & Avg  \\
    \midrule
    \multirow{3}{*}{Gemma\,2B} & 1.99B & \cmark & - & 11.46 & \textbf{8.18} & 13.52 & {63.1} & \textbf{71.4} & \textbf{78.1} & \textbf{65.0} & \textbf{72.3} & \textbf{41.9} & 40.2 & \textbf{61.7}  \\
     & 1.99B &  \cmark & 15B & 10.76 & 8.47 & 13.08 & \textbf{63.5} & 68.5 & 77.0 & 63.5 & 67.6 & 38.1 & {42.6} & 60.1  \\
      & 1.99B & \cmark & 60B & \textbf{10.58} & 8.44 & \textbf{12.71} & 60.3 & 67.9 & 76.9 & 63.5 & 64.9 & 37.2 & 39.6 & 58.6  \\
     \midrule
    \multirow{3}{*}{TinyLlama\,1.1B} & 0.97B & \cmark &  - & 12.26 & 9.37 & 11.94 & 43.3 & 42.2 & 66.8 & 53.4 & 44.7 & 23.2 & 29.2 & 43.3  \\
     & 0.97B  & \cmark  & 15B & 9.87 & 8.24 & 10.73 & 49.2 & 46.3 & \textbf{68.8} & 54.0 & 48.2 & 26.0 & 32.2 & 46.4 \\
     & 0.97B  & \cmark & 60B & \textbf{9.59} & \textbf{8.12} & \textbf{10.42} & \textbf{51.6} & \textbf{48.8} & 68.6 & \textbf{54.1} & \textbf{49.9} & \textbf{26.2} & \textbf{32.8} & \textbf{47.4}  \\
     \midrule
    \multirow{3}{*}{Pythia\,1B} & 0.81B  & \cmark & - & 15.68 & 9.90 & \textbf{12.05} & \textbf{57.5} & 49.1 & 70.4 & 52.8 & \textbf{51.9} & 26.7 & \textbf{33.4} & \textbf{48.8}  \\
     & 0.81B  & \cmark & 15B & 13.46 & 9.95 & 13.38 &  55.0 & 49.0 & 71.0 & 53.6 & 51.8 & \textbf{28.2} & 32.8 & \textbf{48.8} \\
     & 0.81B  & \cmark & 60B & \textbf{12.83} & \textbf{9.76} & 13.57 & 53.0 & \textbf{50.2} & \textbf{71.1} & \textbf{54.8} & \textbf{51.9} & 27.7 & 31.6 & 48.6  \\
    \bottomrule
    \end{tabular}
    }
    \caption{
    Uptraining the pretrained models on datasets that differ significantly in quality or distribution from their pretraining datasets can lead to decreased performance. 
    We evaluated models after uptraining on the SlimPajama dataset. 
    We measured perplexity on test sets of the  SlimPajama, RedPajama, and PG19, and few-shot accuracy on LAMBADA, HellaSwag, PIQA, WinoGrande, ARC-easy, ARC-challenge, and OpenBookQA benchmarks.
    }
    \label{tab:target_baseline_performance}
\end{table}

\subsection{Non-Recursive Model Baselines}
\label{sec:exp_baselines}

Given that we leveraged pretrained model weights for initialization and subsequently uptrained the models, it becomes crucial to define clear performance targets for our parameter-shared models. 

\vspace{-15pt}
\paragraph{Full-size model}
Our ultimate goal is for the Recursive Transformer to achieve performance comparable to the original, full-size pretrained model, without much uptraining.
However, we observed that the distribution divergence between the pretraining and uptraining datasets can hinder achieving the desired performance. 
In particular, uptraining on new datasets, particularly those of comparatively lower quality, sometimes led to performance degradation on certain benchmarks.
Table\,\ref{tab:target_baseline_performance} summarizes the evaluation results of full-size models based on the number of uptraining tokens. For instance, in the case of Gemma, where the pretraining dataset is unreleased but potentially well-curated\,\citep{team2024gemma}, all few-shot performance metrics gradually decreased after uptraining on the SlimPajama dataset. 
This suggests that the achievable upper bound performance with the SlimPajama dataset might be considerably lower than the original model performance.
Therefore, we set the target performance for Gemma and Pythia models as the performance achieved by uptraining a full-size pretrained model with an equivalent number of tokens.
Since TinyLlama was already pretrained on SlimPajama---which is the same dataset we use for uptraining (eliminating any distribution shift)---for slightly longer than our runs, we use the performance of the original checkpoint as reference.

\vspace{-12pt}
\paragraph{Reduced-size model}
To demonstrate the performance advantages of Recursive Transformers compared to models with an equivalent number of parameters, we introduce another baseline: reduced-size models. These models have either half or one-third the parameters of their full-sized counterparts, matching the parameter count of our recursive models. However, these reduced models are pretrained from scratch on the same training recipe (number of training tokens and distillation from full-size model), but without the benefits of the pretrained weights and the looping mechanism. This comparison serves to highlight the efficacy of our initialization techniques and the recursive function itself in attaining strong performance, even with a constrained model size.

\subsection{Main Results}
\label{sec:main_results}

\begin{figure}[t!]
    \centering
    \begin{subfigure}[t]{0.325\textwidth}
        \includegraphics[width=\textwidth]{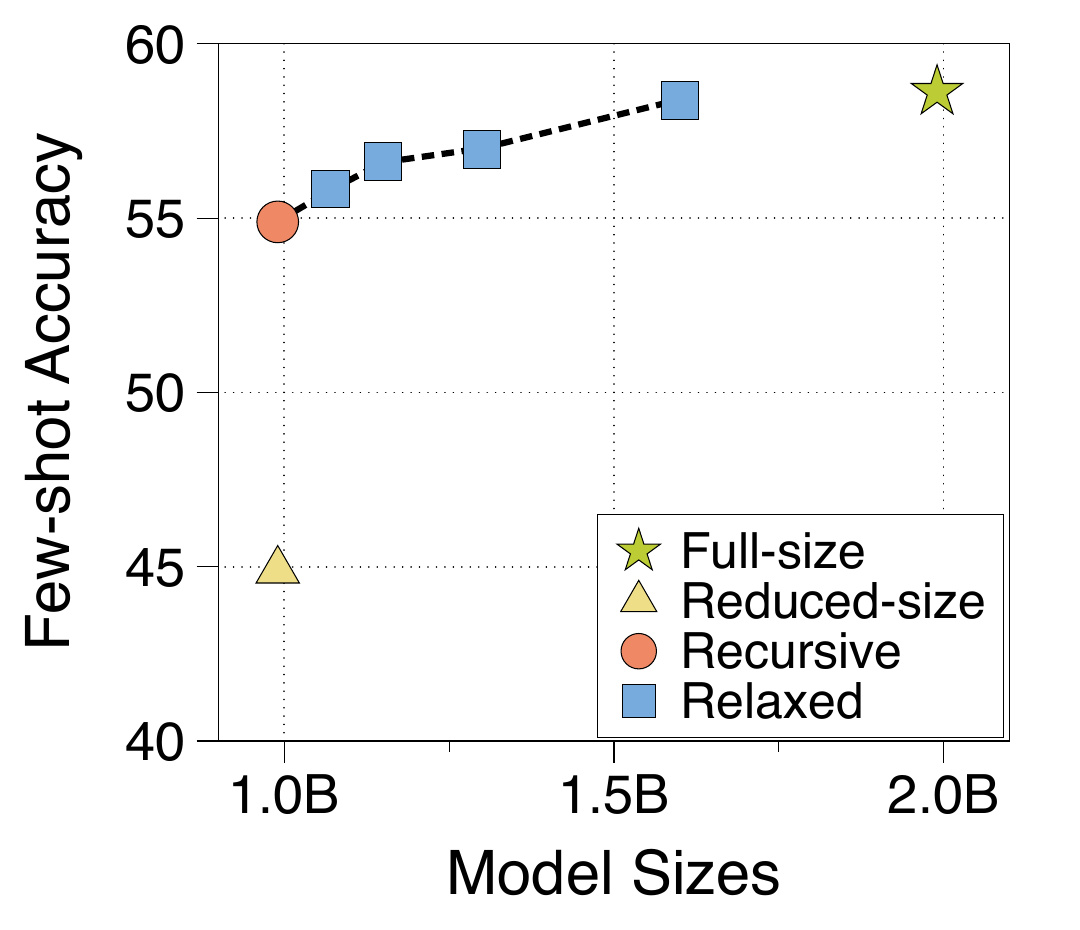}
        \subcaption{Gemma}
    \end{subfigure}
    \centering
    \begin{subfigure}[t]{0.325\textwidth}
        \includegraphics[width=\textwidth]{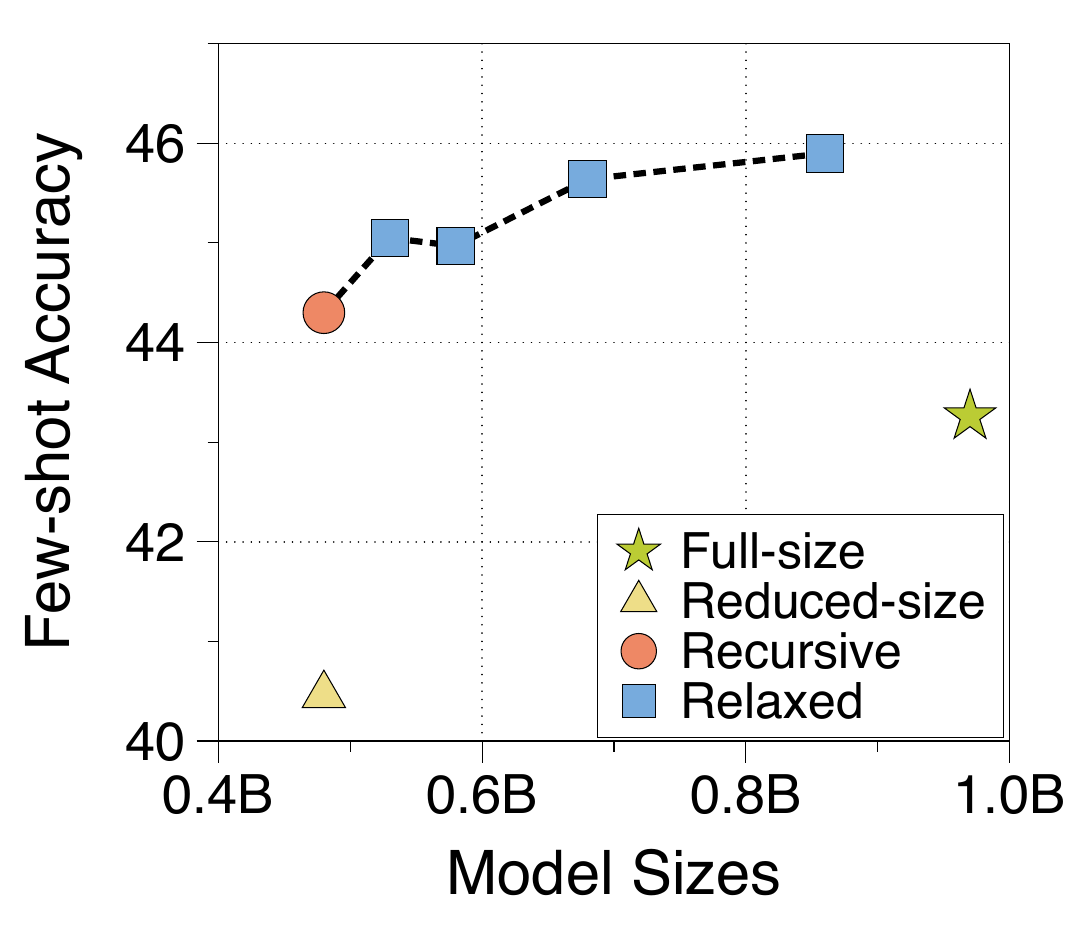}
        \subcaption{TinyLlama}
    \end{subfigure}
    \centering
    \begin{subfigure}[t]{0.325\textwidth}
        \includegraphics[width=\textwidth]{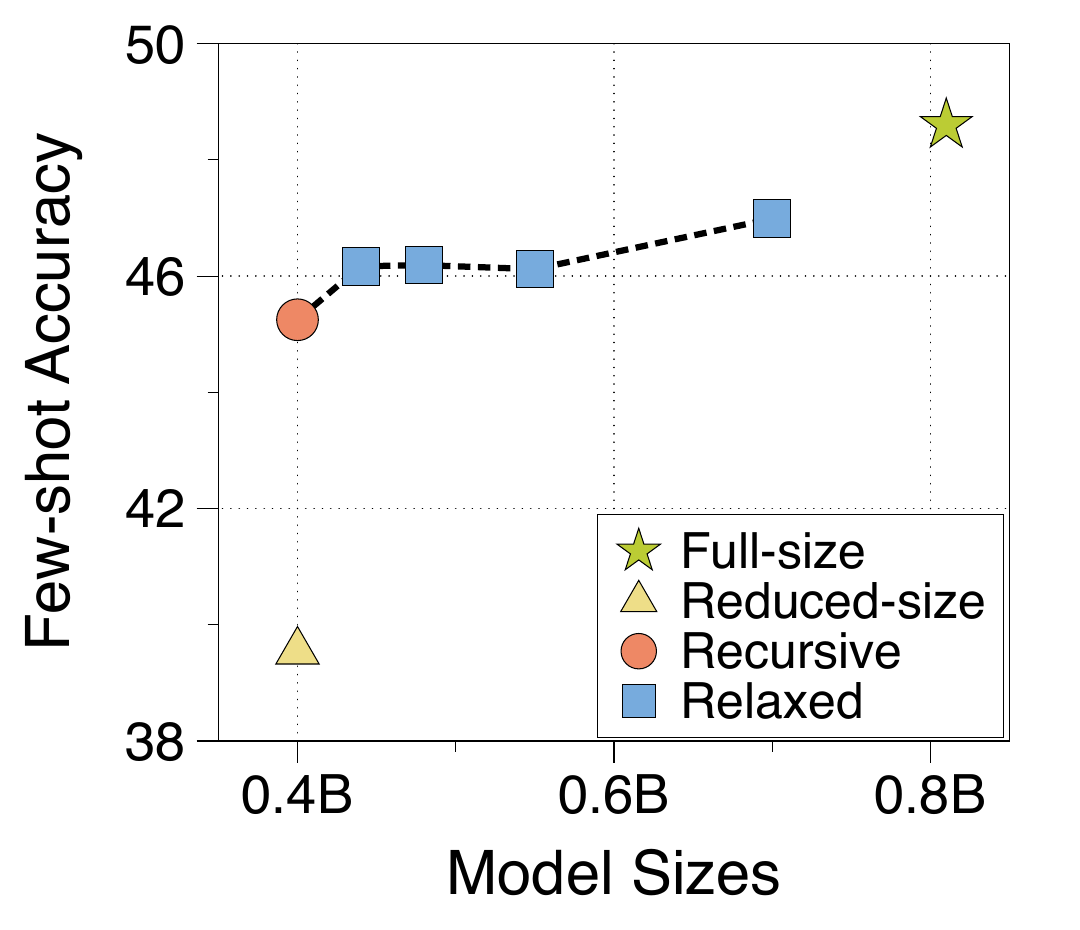}
        \subcaption{Pythia}
    \end{subfigure}
    \caption{
    Recursive and Relaxed Recursive Transformers achieve comparable performance to full-size models, and significantly outperform reduced-size models.
    Recursive models were initialized using the Stepwise method, while relaxed models utilized Average and SVD methods for looped layers and LoRA modules. We show the performance of four different rank values:  64, 128, 256, and 512.
    Recursive and reduced-size models were either uptrained (recursive model) and pretrained from scratch (reduced-size model) on 60 billion tokens using a knowledge distillation objective.
    }
    \label{fig:main_figures}
\end{figure}

Figure\,\ref{fig:main_figures} presents the few-shot performance of Recursive Transformers with two blocks and their relaxed variants. Recursive Transformers, even without relaxation, demonstrate remarkably high performance  despite having only half the parameters of the full-size model. The Gemma model achieved a 10\%p performance gain compared to the reduced-size model, which was also trained on 60 billion tokens using distillation loss.
Remarkably, the recursive TinyLlama model even surpassed the vanilla model's performance, even though the latter was pretrained on a larger corpus of 105 billion tokens. 
Our initialization techniques proved highly effective in achieving this superior result, along with the benefit of the uptraining dataset (SlimPajama) being the same as its pretraining dataset. 

The relaxed models effectively interpolate between the full-size model and the Recursive Transformer, depending on the LoRA rank. As the model size increases with larger LoRA modules, SVD initialization methods allow for a more precise approximation of full-rank matrices, resulting in improved performance. Notably, the relaxed Gemma model with a rank of 512 achieves performance on par with the original model pretrained on 3 trillion tokens (58.4\% vs. 58.6\%), despite using fewer parameters and uptraining on only 60 billion tokens. This trade-off provides flexibility in selecting the best configuration for various deployment scenarios. We believe that additional uptraining and higher-quality datasets could yield better performance with even more streamlined models. 

In the subsequent sections, we provide a comprehensive overview of extensive ablation studies conducted prior to achieving this final performance. In \S\ref{sec:exp_initialization}, we delve into the analysis of various initialization methodologies for Recursive Transformers. Insights into the relaxation model are detailed in \S\ref{sec:exp_relaxed_recursive_transformer}. Finally, we explore enhanced training strategies like knowledge distillation (\S\ref{sec:kd_long_training}).

\subsection{Initialization Techniques for Looped Layers}
\label{sec:exp_initialization}

\begin{figure}[t!]
    \centering
    \begin{subfigure}[t]{0.325\textwidth}
        \includegraphics[width=\textwidth]{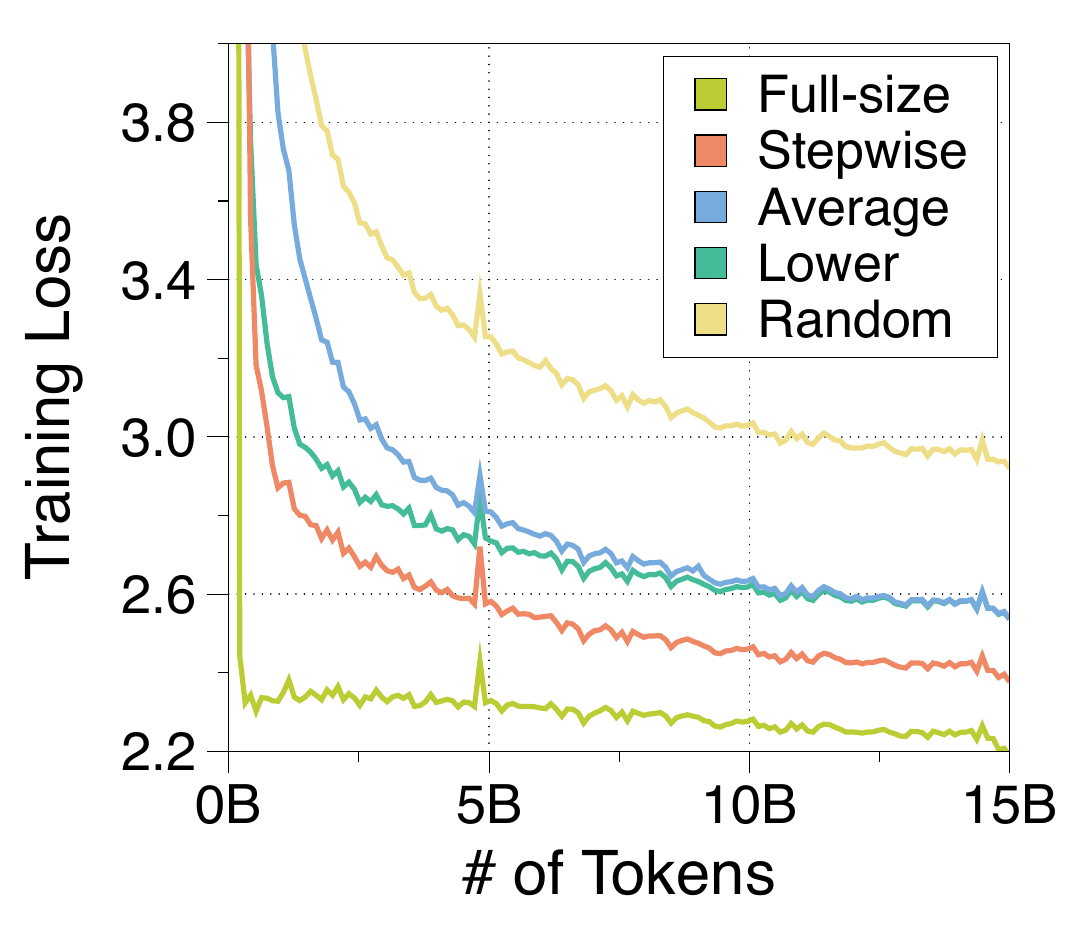}
        \subcaption{Loss curves for Gemma}
        \label{fig:initialization_gemma_training_loss}
    \end{subfigure}
    \centering
    \begin{subfigure}[t]{0.325\textwidth}
        \includegraphics[width=\textwidth]{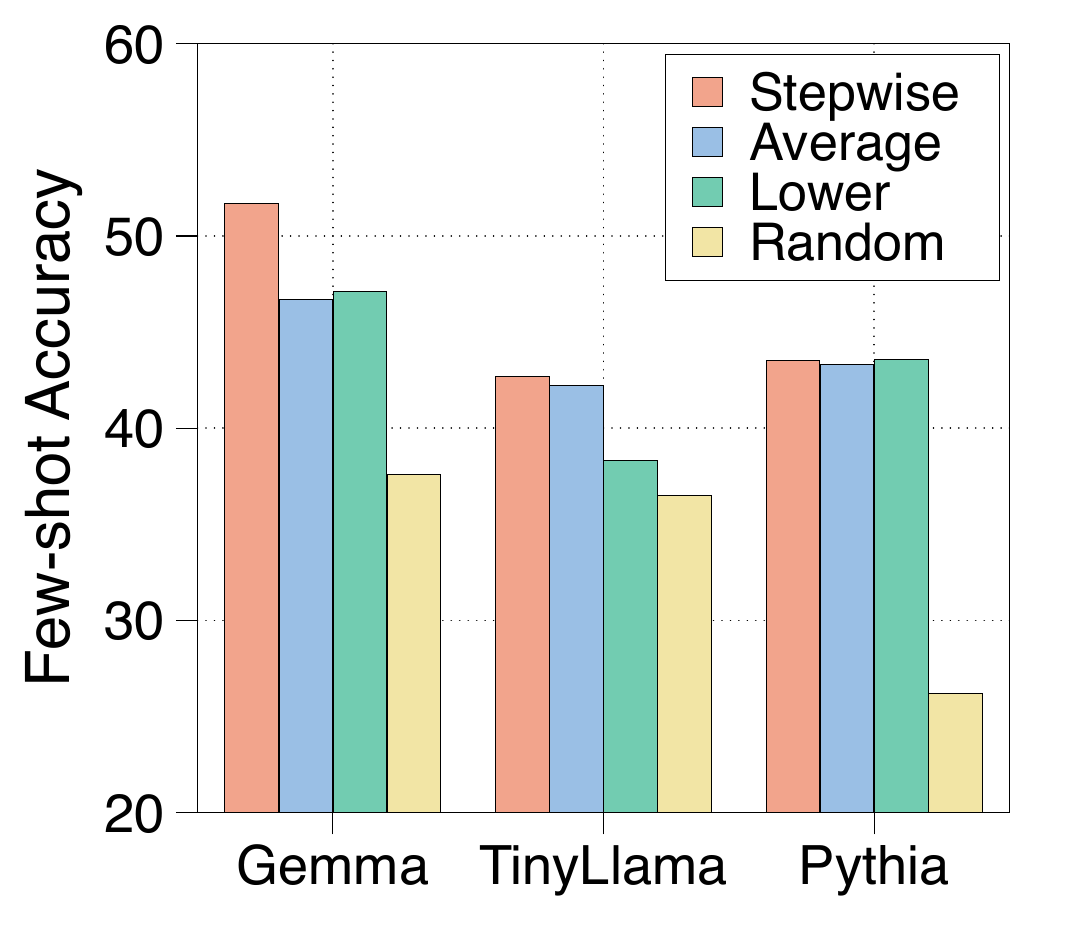}
        \subcaption{Average few-shot performance}
        \label{fig:initialization_avg_fewshot}
    \end{subfigure}
    \centering
    \begin{subfigure}[t]{0.325\textwidth}
        \includegraphics[width=\textwidth]{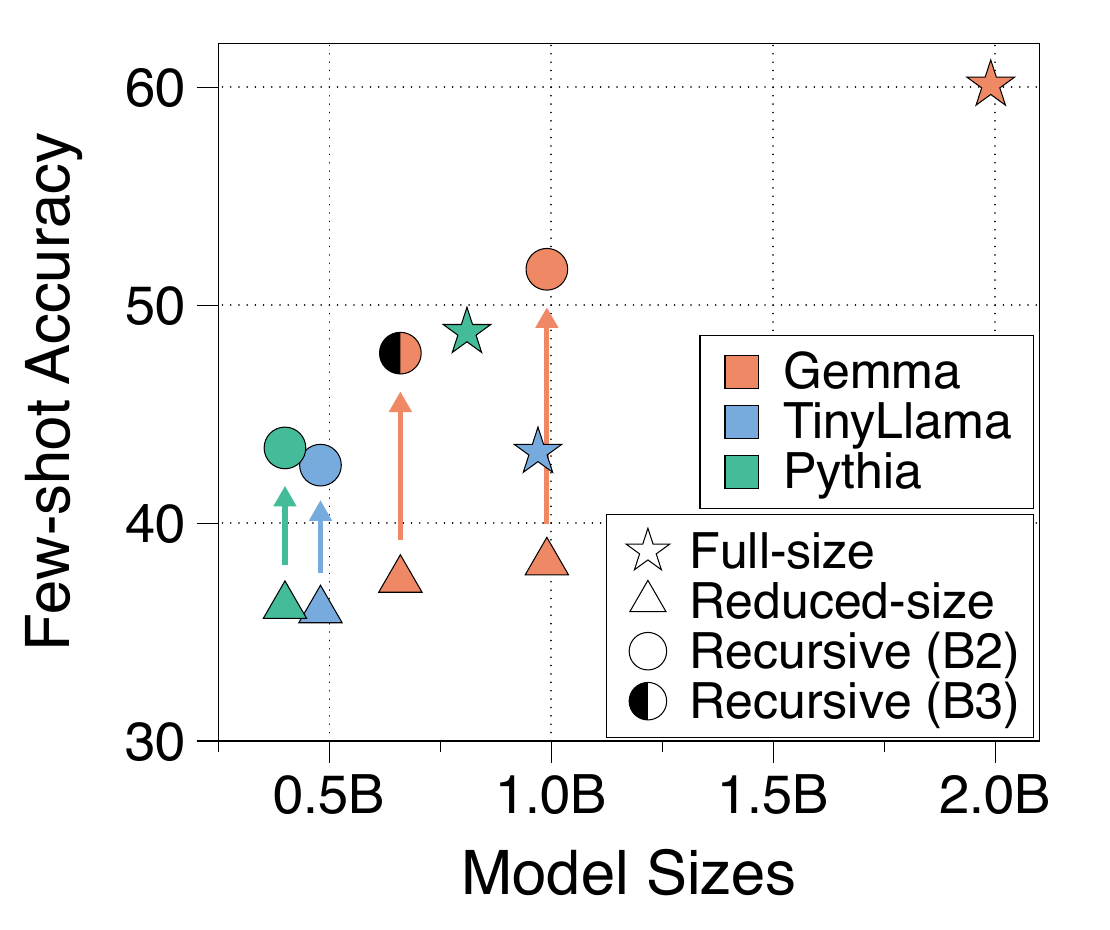} 
        \subcaption{Recursive model performance}
        \label{fig:initialization_model_size_performance}
    \end{subfigure}
    \caption{
    \textbf{(a)} Among the proposed methods, the Stepwise method obtains the lowest training loss on the SlimPajama dataset.
    \textbf{(b)} The Stepwise method consistently demonstrate the highest average few-shot accuracy across three architectures.
    \textbf{(c)} Recursive Transformers initialized with the Stepwise method demonstrated significant performance gains compared to non-recursive model baselines.
    }
    \label{fig:initialization_gemma}
\end{figure}

\paragraph{Stepwise initialization serves as the best initial point for Recursive Transformers}
We present the training loss of Gemma models initialized using three different methods in Figure\,\ref{fig:initialization_gemma_training_loss}, and their few-shot performance in Figure\,\ref{fig:initialization_avg_fewshot}.
Our proposed methods significantly outperformed random initialization, which simply adds recursion to a reduced-size model, suggesting that leveraging pretrained weights in any manner is beneficial for performance boost. Moreover, the Stepwise methodology consistently demonstrated best performance, aligning with insights that LLMs can preserve performance even with a few layers skipped\,\citep{DBLP:journals/corr/abs-2404-02258, DBLP:conf/acl/Zhang00S0CM24, DBLP:conf/acl/ElhoushiSLHWL0A24}. 
Interestingly, as summarized in Table~\ref{tab:initialization_total_app}, the recursive TinyLlama model, uptrained on only 15 billion tokens, yields few-shot performance comparable to the original model pretrained on 105 billion tokens.
This suggests that with sufficient training, even a recursive architecture can match the performance of a full-size pretrained model~\citep{DBLP:conf/iclr/DehghaniGVUK19, DBLP:conf/sustainlp/TakaseK23}.

\vspace{-12pt}
\paragraph{Recursive Gemma 1B outperforms both pretrained TinyLlama 1.1B and Pythia 1B}
The looped Gemma 1B model, utilizing our proposed Stepwise method, outperformed reduced-size baselines with equivalent parameter counts by up to 13.5 percentage points (51.7\% vs. 38.2\%). Furthermore, it even outperformed the full-size TinyLlama 1.1B and Pythia 1B models (see Figure\,\ref{fig:initialization_model_size_performance}). This is a noteworthy achievement given that Pythia was pretrained on 300 billion tokens, whereas the recursive Gemma was uptrained on only 15 billion tokens. 
Consequently, high-performing LLMs serve as a promising starting point, as their recursive counterparts readily outperform other ordinary vanilla models of similar size.
Further details can be found in Appendix \ref{app:initialization}.

\vspace{3pt}
\begin{tcolorbox}[
    enhanced,
    colback=blue!5!white, 
    colframe=black, 
    width=\textwidth,
    coltitle=white,
    title=\textbf{Takeaways for Recursive Transformer},
    top=3mm, bottom=2mm,
    attach boxed title to top left={yshift=-2.8mm, xshift=4mm},
    boxed title style={colback=black, colframe=black, boxrule=0mm, 
    toptitle=1mm, bottomtitle=1mm}
]
We find that converting well-pretrained models into Recursive Transformers leads to high-performing models with minimal uptraining. Notably, initializing looped layers via the Stepwise method yields the best results. With just 15 billion tokens of uptraining, a recursive Gemma 1B model outperforms even the full-size pretrained TinyLlama and Pythia models.
\end{tcolorbox}

\subsection{Relaxation of Strict Parameter Sharing via LoRA Modules}
\label{sec:exp_relaxed_recursive_transformer}

\begin{figure}[t!]
    \centering
    \begin{subfigure}[t]{0.325\textwidth}
        \includegraphics[width=\textwidth]{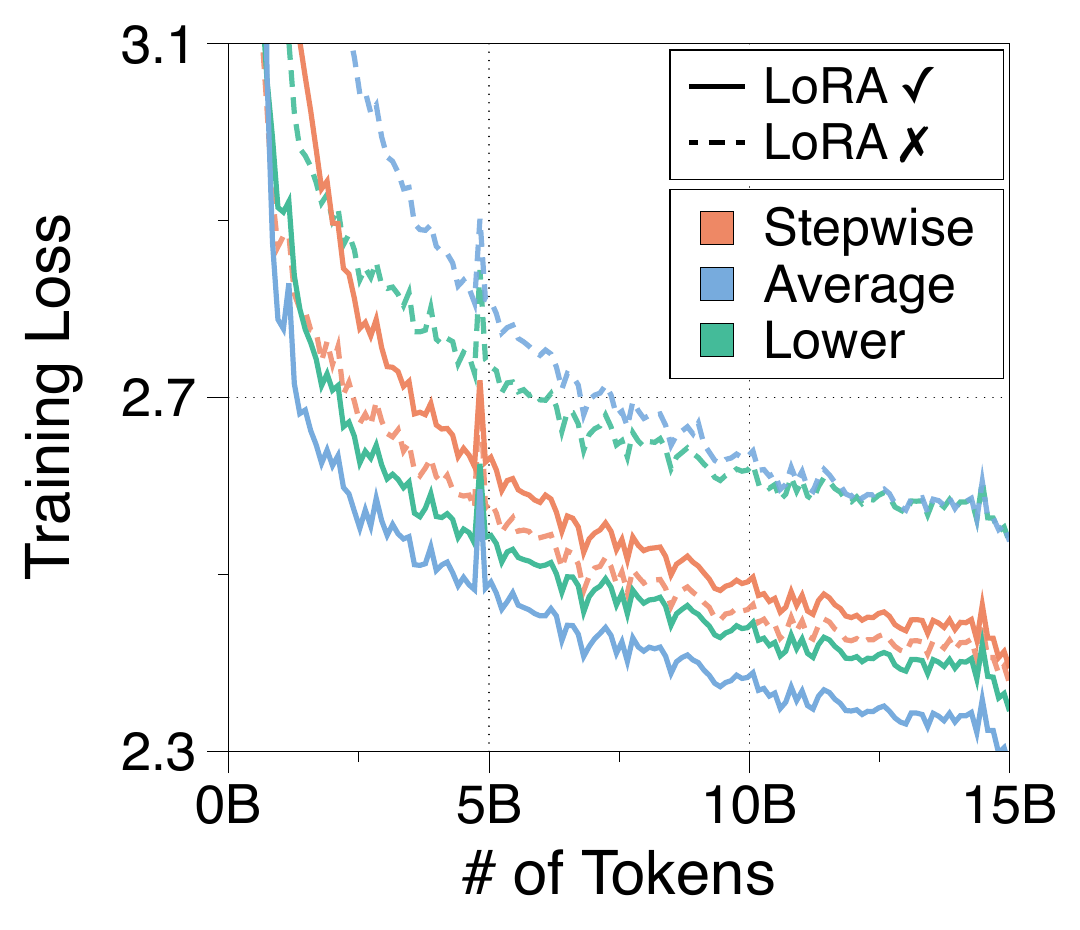}
        \subcaption{Loss changes in Gemma}
        \label{fig:lora_gemma_training_loss}
    \end{subfigure}
    \centering
    \begin{subfigure}[t]{0.325\textwidth}
        \includegraphics[width=\textwidth]{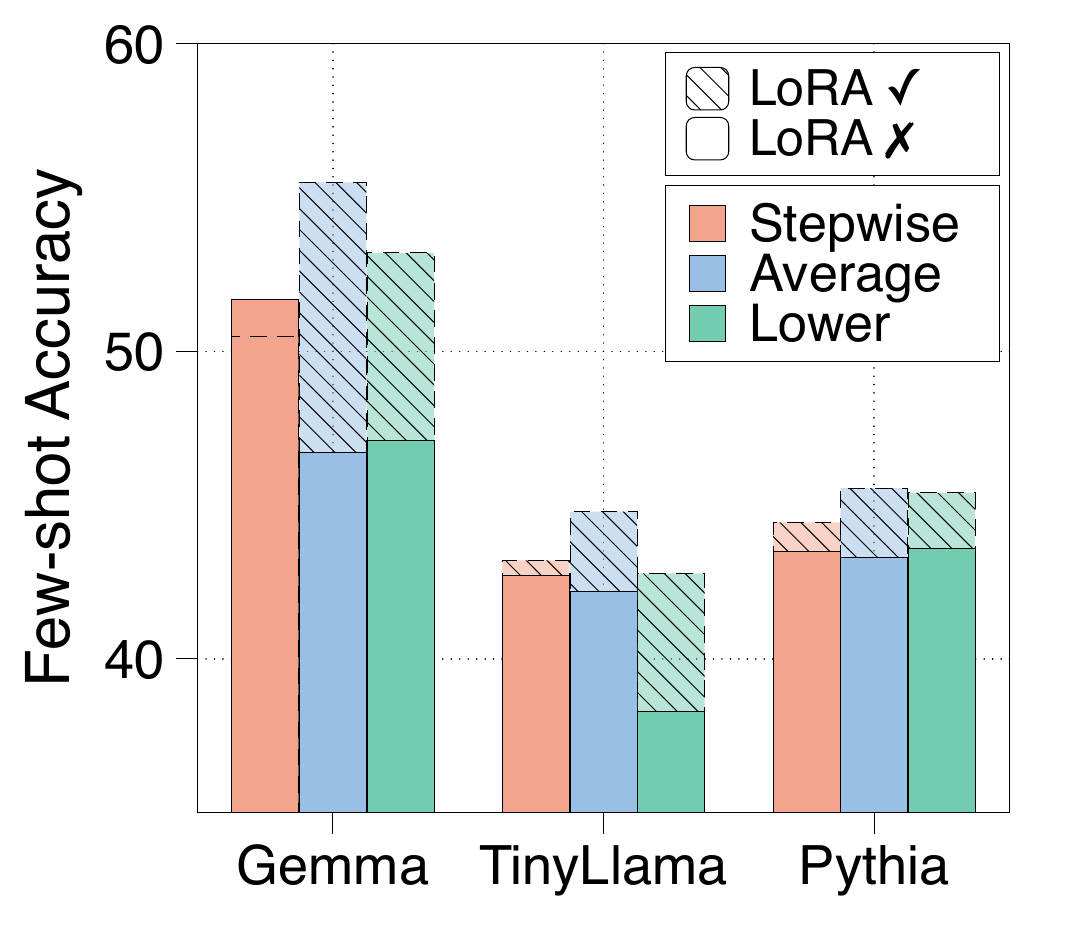}
        \subcaption{Accuracy gains from relaxation}
        \label{fig:lora_fewshot_accuracy}
    \end{subfigure}
    \centering
    \begin{subfigure}[t]{0.325\textwidth}
        \includegraphics[width=\textwidth]{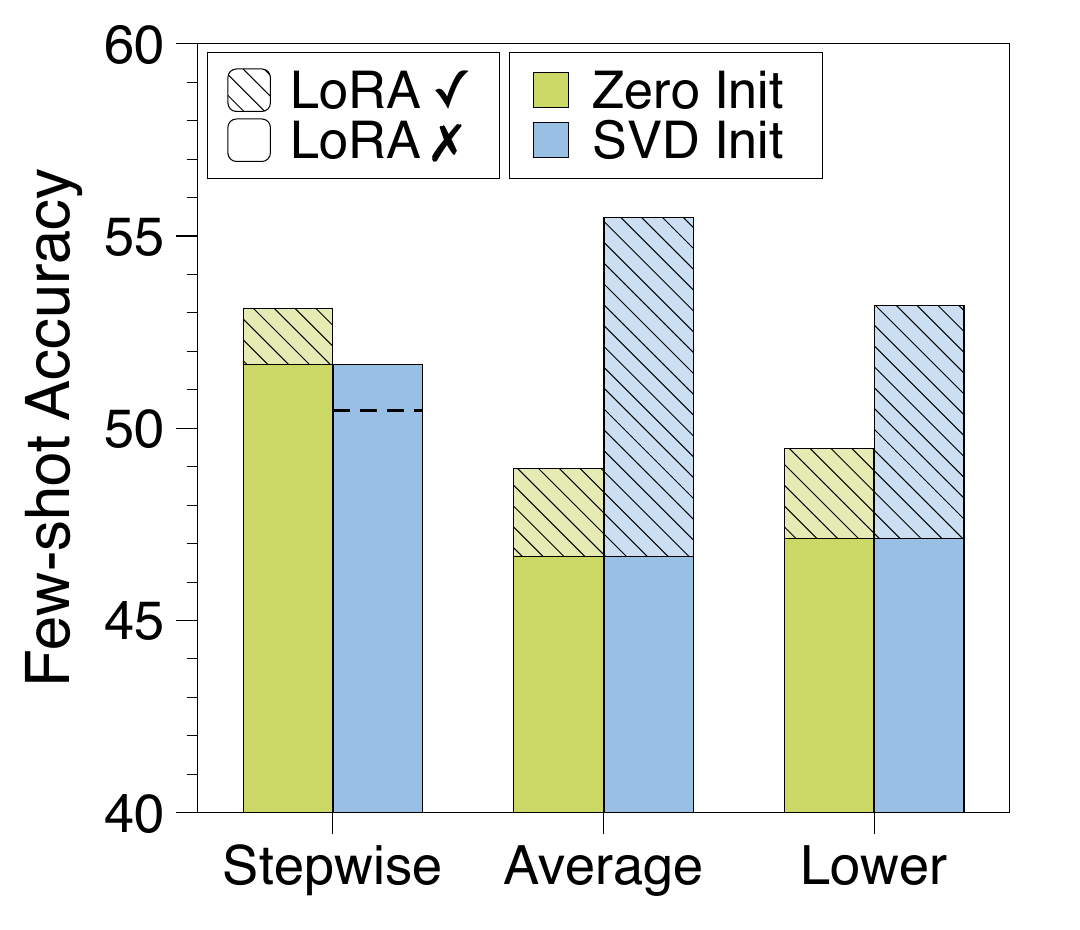} 
        \subcaption{Effects of SVD initialization}
        \label{fig:svd_zero_gemma}
    \end{subfigure}
    \caption{
    The Relaxed Recursive Transformer, with its looped layer initialized using Average method, achieved the best performance in terms of both (\textbf{a}) training loss and (\textbf{b}) few-shot accuracy. The models utilize two blocks, with the LoRA modules initialized using the SVD method at a rank of 512.
    (\textbf{c}) SVD initialization method significantly enhanced performance compared to zero initialization.
    }
    \label{fig:layer_wise_lora_main}
\end{figure}

\paragraph{Average initialization for looped layers is most compatible with Relaxed Recursive Transformer}
Figures\,\ref{fig:lora_gemma_training_loss} and \ref{fig:lora_fewshot_accuracy} illustrate the effect of relaxing parameter sharing via layer-wise LoRA modules.
Notably, initializing tied layers in relaxed models with Average method yielded substantial performance improvements, even outperforming the non-relaxed model initialized with Stepwise. Approximating residual matrices between averaged weights and their individual weights appears readily achievable using truncated SVD with low ranks.
In contrast, we observed an intriguing phenomenon where our models initialized with Stepwise occasionally showed performance degradation after relaxation. This is likely because capturing the nuances between entirely distinct layer weights is challenging with an insufficient rank, leading to a suboptimal solution.
Further details are provided in Appendix\,\ref{app:cross_layer_lora}. 

\vspace{-12pt}
\paragraph{SVD initialization to approximate pretrained weights outperforms zero initialization}
LoRA modules initialized with zero values guarantee that the model begins training from the same point as the non-relaxed model. Conversely, SVD initialization positions the model closer to either the full-size model (with full-rank) or the non-relaxed model (with small rank).
To emphasize the effectiveness of initializing near full-size model weights, we compared these two methods at a moderately large rank of 512, as shown in Figure\,\ref{fig:svd_zero_gemma}.
Our proposed SVD strategy demonstrated an impressive performance boost of up to 6.5 points, facilitating faster convergence by updating the principal low-rank matrices (aligned with findings in \citet{DBLP:journals/corr/abs-2404-02948}).
For results across other architectures, refer to Figure\,\ref{fig:zero_svd_init_app}.

\begin{figure}[t!]
\vspace{-7pt}
    \centering
    \begin{subfigure}[t]{0.325\textwidth}
        \includegraphics[width=\textwidth]{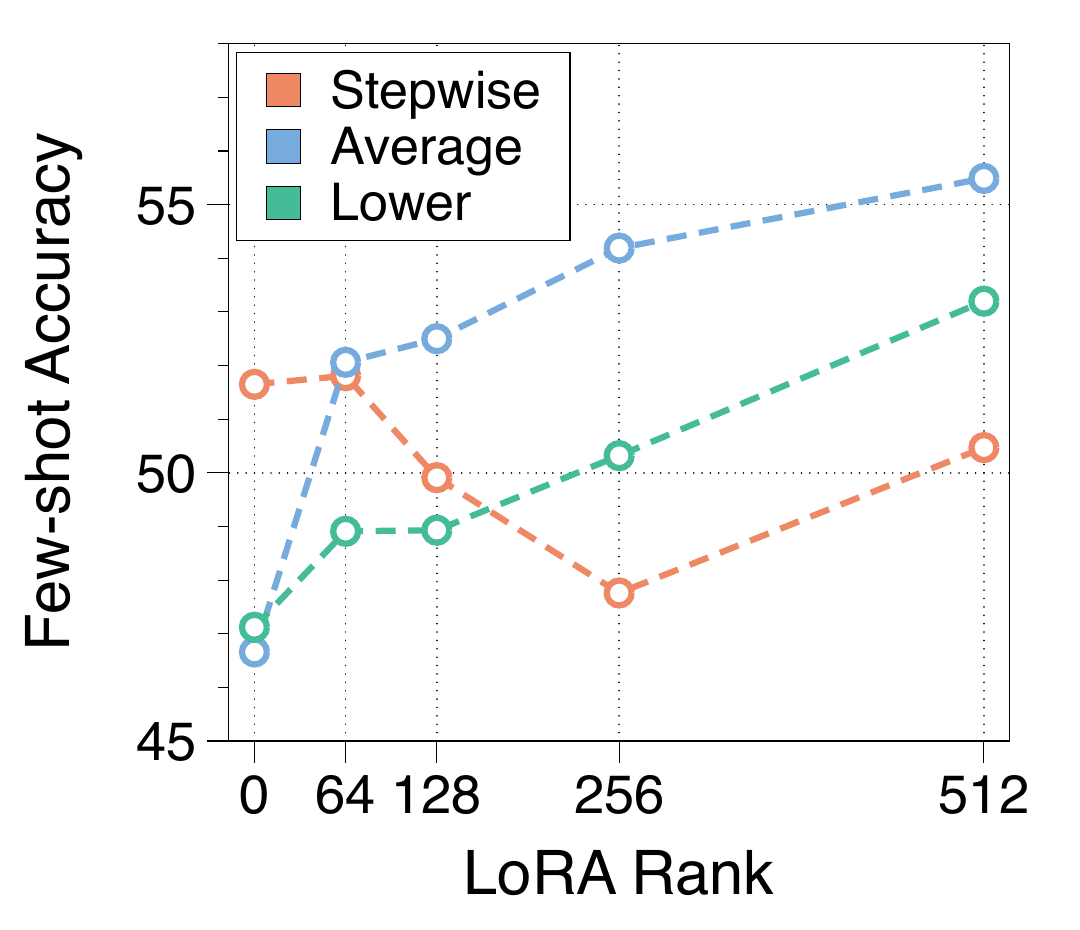}
        \subcaption{Performance by LoRA rank}
        \label{fig:lora_rank}
    \end{subfigure}
    \centering
    \begin{subfigure}[t]{0.325\textwidth}
        \includegraphics[width=\textwidth]{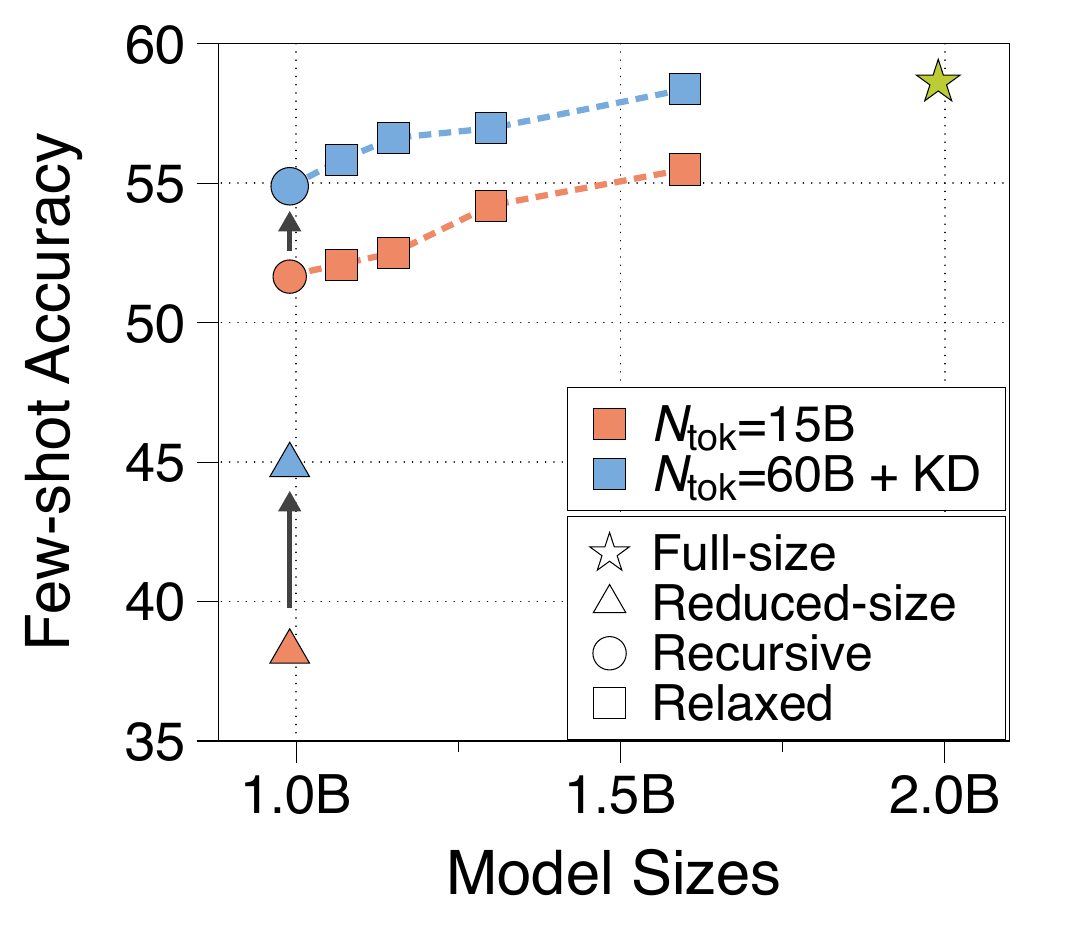} 
        \subcaption{Gains from longer KD training}
        \label{fig:kd_long_gemma}
    \end{subfigure}
    \centering
    \begin{subfigure}[t]{0.325\textwidth}
        \includegraphics[width=\textwidth]{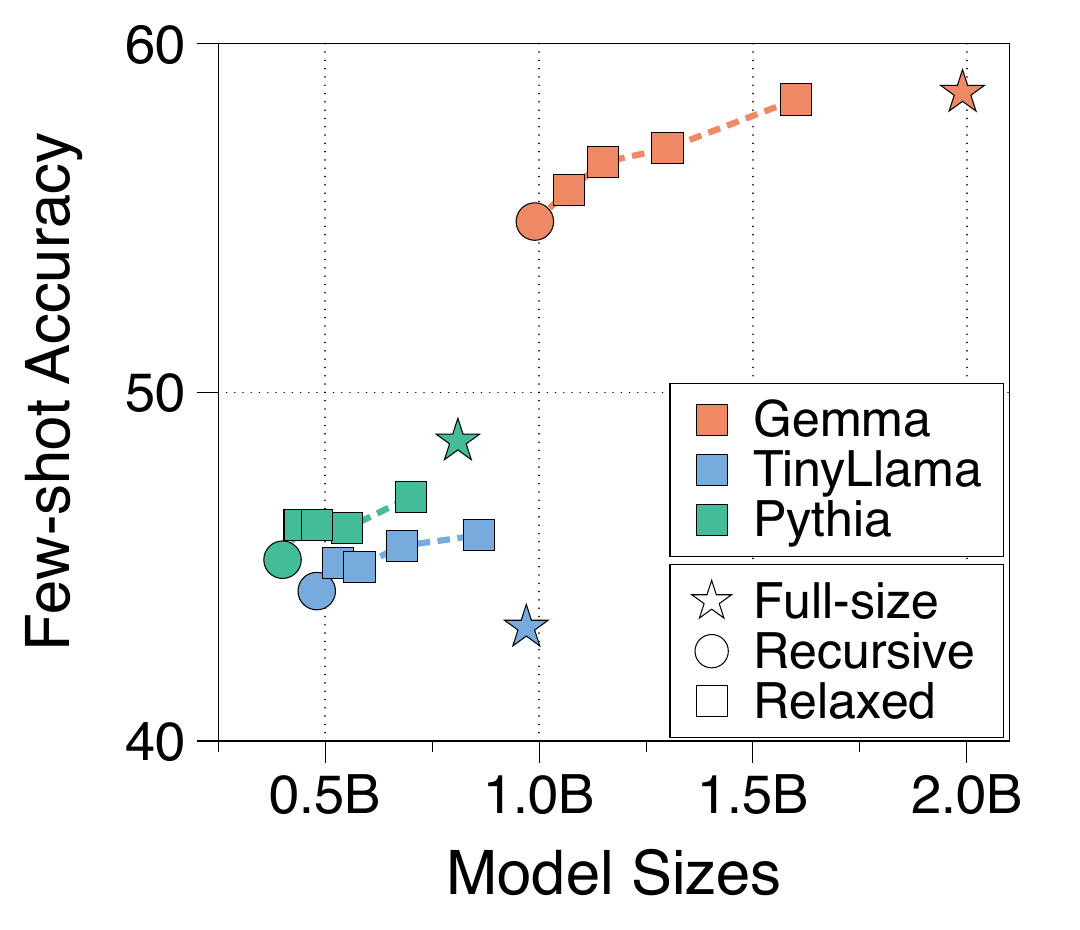}
        \subcaption{Overall performance}
        \label{fig:overall_performance}
    \end{subfigure}
    \label{fig:lora_rank_long_training}
    \caption{
    (\textbf{a}) Increasing the LoRA rank typically leads to improved performance in relaxed Gemma models, attributed to the use of SVD initialization.
    (\textbf{b}) Extended uptraining and knowledge distillation yielded substantial accuracy gains for Gemma models.
    Note that the full-size model is a pretrained model that is further uptrained on 60 billion tokens.
    (\textbf{c}) 
    Recursive and Relaxed Recursive Transformers achieve a compelling Pareto frontier with respect to model size and performance.
    Recursive and relaxed models used Stepwise and Average method to initialize looped layers, respectively.
    }
\end{figure}

\vspace{-12pt}
\paragraph{Higher rank enhances recovery of original pretrained weights}
At full rank, relaxed models can perfectly match full-size pretrained models.
Consequently, as illustrated in Figure\,\ref{fig:lora_rank}, performance generally improves with increasing rank, resulting in a clear Pareto frontier between model size and performance.
However, only Stepwise initialization showed a U-shaped performance trend: a middle-range rank resulted in poor approximation, whereas very low ranks (akin to random initialization for LoRA modules) yielded better performance.
The overall results are summarized in Table\,\ref{tab:lora_total_app}.

\vspace{3pt}
\begin{tcolorbox}[
    enhanced,
    colback=blue!5!white, 
    colframe=black, 
    width=\textwidth,
    coltitle=white,
    title=\textbf{Takeaways for Relaxed Recursive Transformer},
    top=3mm, bottom=2mm,
    attach boxed title to top left={yshift=-2.8mm, xshift=4mm},
    boxed title style={colback=black, colframe=black, boxrule=0mm, 
    toptitle=1mm, bottomtitle=1mm}
]
Adjusting the LoRA rank in the Relaxed Recursive Transformer, together with our SVD-based initialization technique, allows for a smoother trade-off between a fully weight-tied recursive model and a vanilla model. Furthermore, we find that initializing the shared weights in the looped layers with the Average method leads to the best performance in this setting.
\end{tcolorbox}

\subsection{Extended Uptraining and Knowledge Distillation}
\label{sec:kd_long_training}

We further enhanced the performance of our low-rank models by introducing two techniques: uptraining on an extended corpus and knowledge distillation from the full-sized model. Specifically, we increased the number of uptraining tokens from 0.5\% to 2\% of the total 3 trillion tokens used for pretraining Gemma models, resulting in a total of 60 billion tokens. Additionally, we regularized the losses using a forward Kullback-Leibler divergence\,\citep{DBLP:journals/corr/HintonVD15, DBLP:conf/emnlp/KimR16}, which exhibited the best performance gains among the examined distillation losses. Table~\ref{tab:kd_ablation_study} summarizes the results of various ablation studies conducted to investigate the impact of these two techniques. 

The combined effect of these techniques is presented in Figure\,\ref{fig:kd_long_gemma}, demonstrating an improvement of up to 4.1 percentage points in few-shot accuracy compared to the previous 15 billion token uptraining results. Notably, the relaxed Gemma model with a rank of 512 nearly matched the performance of the full-size model.
We also expect that further performance gains can be achieved with a much lighter recursive model by utilizing a superior teacher model or conducting more extensive training on high-quality data.
Figure\,\ref{fig:overall_performance} illustrates the Pareto frontier achieved by the final models. All models exhibit competitive performance compared to the full-size model. Moreover, the superior performance of the recursive Gemma model strongly highlights the advantages of converting high-performing LLMs to a recursive architecture. Additional details can be found in Appendix\,\ref{app:further_techniques}.

\begin{table}[t!]
        \small
    \centering
    \renewcommand{\arraystretch}{0.9}
    \resizebox{\textwidth}{!}{
    \setlength{\tabcolsep}{4pt}
    \begin{tabular}{c|cc|cc|ccc|ccccccc|cc}
    \toprule
     & \multicolumn{2}{c|}{\textbf{Uptrain}} & \multicolumn{2}{c|}{\textbf{Looping}} &  \multicolumn{3}{c|}{\textbf{Early-Exit Train}} & \multicolumn{9}{c}{\textbf{Few-shot Accuracy\,$\uparrow$}} \\
    \cmidrule(l{2pt}r{2pt}){2-3} \cmidrule(l{2pt}r{2pt}){4-5}  \cmidrule(l{2pt}r{2pt}){6-8}  \cmidrule(l{2pt}r{2pt}){9-17} 
     N-emb & PT & $N_{tok}$ & Block & Init  &  $N_{tok}$ & CE & KD  & LD & HS & PQ & WG & ARC-e & ARC-c & OB & Avg & $\Delta$  \\
    \midrule
    0.99B & \cmark & 15B  & 2 & Step &  - &- & - & {53.0} & {57.3} & {73.2} & {56.2} & {56.1} & {29.2} & {36.6} & {51.7} & - \\
    \midrule
     \multirow{2}{*}{0.99B} & \multirow{2}{*}{\cmark} & \multirow{2}{*}{15B} & \multirow{2}{*}{2} & \multirow{2}{*}{Step} &  \multirow{2}{*}{15B}  &  \multirow{2}{*}{Weighted} &  \multirow{2}{*}{\xmark}  &   48.9 &  55.5 &  72.7 &  55.3 &  54.9 &  30.1 &  36.0 &  50.5 & \textcolor{custom_red}{\textbf{--\,1.2}} \\
     &  &  &  & &  &   & &   49.5 &  54.8 &  72.0 &  53.4 &  54.1 &  29.1 &  35.6 &  49.8 & - \\
     \addlinespace[-1pt]
     \cmidrule(l{2pt}r{2pt}){9-17} 
     \addlinespace[-1pt]
      &  &  &  &  &  &    &  &   53.0 &  59.1 &  73.9 &  55.4 &  57.4 &  30.6 &  37.8 &  52.5 & \textcolor{custom_green}{\textbf{+0.8}} \\
     \multirow{-2}{*}{0.99B} & \multirow{-2}{*}{\cmark} &  \multirow{-2}{*}{15B}  & \multirow{-2}{*}{2} & \multirow{-2}{*}{Step}   & \multirow{-2}{*}{15B}  & \multirow{-2}{*}{Agg\,(0.1)} & \multirow{-2}{*}{\xmark}   &  45.9 &  51.2 &  71.4 &  54.5 &  48.1 &  26.8 &  32.0 &  47.1 & - \\
     \midrule
     \multirow{2}{*}{0.99B} & \multirow{2}{*}{\cmark} & \multirow{2}{*}{15B} & \multirow{2}{*}{2} & \multirow{2}{*}{Step} &   \multirow{2}{*}{15B}   &  \multirow{2}{*}{Weighted} &  \multirow{2}{*}{\cmark}  &   47.7 &  55.1 &  73.2 &  55.6 &  54.5 &  29.1 &  37.2 &  50.4 & \textcolor{custom_red}{\textbf{--\,1.3}} \\
     &  &  &  & &  &  &  &  48.3 &  54.9 &  72.1 &  55.9 &  54.3 &  28.4 &  35.4 &  49.9 & - \\
     \addlinespace[-1pt]
     \cmidrule(l{2pt}r{2pt}){9-17} 
     \addlinespace[-1pt]
     \rowcolor[gray]{0.9}
      &  &  &  & &  &    &  &    52.9 &  58.9 &  73.7 &  55.7 &  57.5 &  31.1 &  38.2 &  52.6 & \textcolor{custom_green}{\textbf{+0.9}} \\
    \rowcolor[gray]{0.9}
     \multirow{-2}{*}{0.99B} & \multirow{-2}{*}{\cmark} &  \multirow{-2}{*}{15B}  & \multirow{-2}{*}{2} & \multirow{-2}{*}{Step}   & \multirow{-2}{*}{15B}  & \multirow{-2}{*}{Agg\,(0.1)} & \multirow{-2}{*}{\cmark} &  46.3 &  52.1 &  71.6 &  55.3 &  49.2 &  28.5 &  32.6 &  48.0 & - \\
    \bottomrule
    \end{tabular}
    }
    \caption{
    A small loss coefficient to the first loop output (intermediate output) can significantly improve intermediate performance without compromising the final performance.
    Performance was evaluated under a static-exiting scenario\,\citep{DBLP:conf/nips/SchusterFG0B0TM22}, where all tokens exit at either first or second loop.
    We further trained the previously uptrained Gemma models on 15 billion tokens (post-training).
    Delta\,($\Delta$) denotes the performance changes in the final outputs after early-exit training. 
    }
    \label{tab:early_exit_ablation}
\end{table}

\subsection{Early-Exit Training Strategy for Recursive Transformer}

The throughput of Recursive Transformers can be amplified by an early-exiting framework. Hence, we further train intermediate representations from fewer looping iterations to enable token prediction.
We conducted an ablation study on various strategies, as summarized in Table\,\ref{tab:early_exit_ablation} (more detailed results are presented in Table\,\ref{tab:early_exit_ablation_app}). 
Directly applying the weighted CE loss ($\mathcal{L} = \sum_{i=1}^{B} \alpha_{i} \mathcal{L}_{i} \text{\; where \;} \alpha_{i} = i / \sum_{i} i$) commonly used in prior works~\citep{DBLP:conf/nips/SchusterFG0B0TM22, DBLP:conf/emnlp/BaeKSY23} led to an overemphasis on the training of intermediate representations.
To address this, we employ an aggressive coefficient strategy that aggressively reduces the loss coefficient for intermediate outputs while maintaining a coefficient of 1 for the final output.
Our experiments demonstrated that an aggressive coefficient of 0.1, utilizing knowledge distillation from the detached final outputs~\citep{DBLP:conf/emnlp/BaeKSY23}, effectively preserves final performance while enhancing intermediate performance. Notably, the first loop output yielded only a difference of 4.6 percentage points in accuracy compared to the final output.
This underscores the potential to maximize the benefits of early-exiting in parameter-shared LLMs.

We applied this post-training strategy for early-exiting to our final uptrained models (shown in \S\ref{sec:main_results}), with all experimental results detailed in Appendix\,\ref{app:early_exit}.
The aggressive coefficient strategy, combined with self-distillation, consistently achieved the best performance for intermediate outputs while maintaining strong performance for the final loop output across all models.
However, as the optimal strategy derived from the non-relaxed models was directly applied to the relaxed models, a more tailored training approach might further enhance the performance of intermediate loop outputs in Relaxed Recursive Transformers.

\subsection{Hypothetical Generation Speedup via Continuous Depth-wise Batching}
\label{exp:hypothetical_generation_speedup}

\begin{figure}[t!]
    \centering
    \begin{tabular}{cc}
        \raggedleft
        \begin{subfigure}[b]{0.49\textwidth}
            \includegraphics[width=\textwidth]{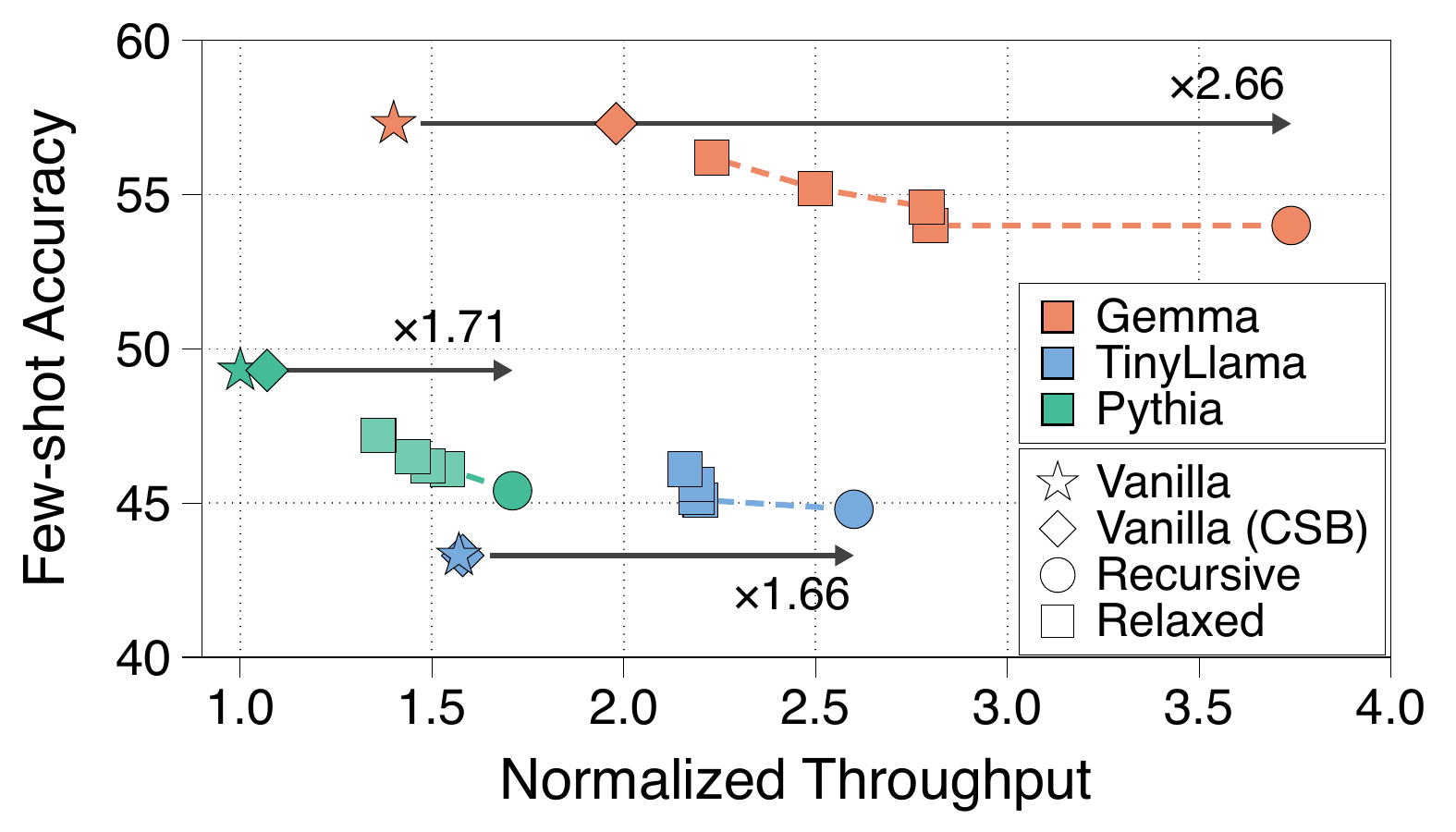}
            \vfill
        \end{subfigure} &
        \hspace{-10pt}
        \begin{minipage}[t]{0.49\textwidth}
        \vspace{-43mm}
            \raggedright
            \addtolength{\tabcolsep}{-3pt}
            \resizebox{1.0\textwidth}{!}{
            \begin{tabular}{ccc|cc|c|ccc}
            \toprule
            N-emb & Loop & LoRA & Batch & Exit & Acc. & Thr. & $\Delta_{V}$ & $\Delta_{Seq}$ \\
            \hline
            1.99B & - & - & -  & \xmark & 57.3 & 1080 & \textcolor{gray}{{$\mathbf{\times 1.00}$}} & \textcolor{custom_red}{{$\mathbf{\times 0.71}$}} \\ 
            1.99B & - & - & CSB & \xmark & 57.3 &  1528 & \textcolor{custom_green}{{$\mathbf{\times 1.41}$}} & \textcolor{gray}{{$\mathbf{\times 1.00}$}} \\ 
            \hline
            0.99B & 2 & - & CDB & \cmark  & 54.0 & 2877 & \textcolor{custom_green}{{$\mathbf{\times 2.66}$}} & \textcolor{custom_green}{{$\mathbf{\times 1.88}$}} \\ 
            1.07B & 2 & 64 & CDB & \cmark  & 54.0 & 2157 & \textcolor{custom_green}{{$\mathbf{\times 2.00}$}} & \textcolor{custom_green}{{$\mathbf{\times 1.41}$}} \\ 
            1.15B & 2 & 128 & CDB & \cmark  & 54.6 & 2149 & \textcolor{custom_green}{{$\mathbf{\times 1.99}$}} & \textcolor{custom_green}{{$\mathbf{\times 1.41}$}} \\ 
            1.30B & 2 & 256 & CDB & \cmark  & 55.2 & 1921 & \textcolor{custom_green}{{$\mathbf{\times 1.78}$}} & \textcolor{custom_green}{{$\mathbf{\times 1.26}$}} \\ 
            1.60B & 2 & 512 & CDB & \cmark  & 56.2 & 1719 & \textcolor{custom_green}{{$\mathbf{\times 1.59}$}} & \textcolor{custom_green}{{$\mathbf{\times 1.13}$}} \\ 
            \bottomrule
            \end{tabular}
            }
        \end{minipage}
    \end{tabular}
    \caption{
    Continuous depth-wise batching (CDB) with early exiting enables Recursive Transformers to theoretically achieve significant throughput improvements.
    Throughput\,(tokens/sec) was averaged across SlimPajama, RedPajama, and PG19, and then normalized to the throughput of the vanilla Pythia model. The accompanying table gives detailed throughout and performance measurements  for Gemma. $\Delta_V$ measures throughput relative to the vanilla Gemma model, while $\Delta_{Seq}$ measures throughput relative to the vanilla Gemma model with continuous sequence-wise batching (CSB).
    }
    \label{fig:early_exit_throughput}
\end{figure}

\paragraph{How we theoretically approximate actual throughput}
As developing practical early-exiting algorithms is beyond the scope of this work, we present hypothetical throughput improvements based on an oracle-exiting approach\,\citep{DBLP:conf/nips/SchusterFG0B0TM22, DBLP:conf/emnlp/BaeKSY23}. 
This assumes that tokens exit at the earliest looping block where their prediction aligns with the final loop's prediction.
We simulated the generation of language modeling datasets as if they were generated by our models, to obtain the exit trajectory for each token.
Then, we measured the average per-token generation time under specific constraints, such as different memory limit or context lengths. 
Using these measurements and the exit trajectory data, we conducted simulations to estimate theoretical throughput.
Detailed explanations and limitations are discussed in Appendix\,\ref{app:hypothetical_generation_speedup}.

\vspace{-12pt}
\paragraph{Continuous depth-wise batching paired with early-exiting can substantially boost throughput}
Figure\,\ref{fig:early_exit_throughput} illustrates the throughput of our proposed models and the vanilla Transformer across three architectures. We consistently achieve higher speeds than the vanilla models by combining continuous depth-wise batching with early-exiting, even surpassing those with continuous sequence-wise batching\,\citep{DBLP:conf/osdi/YuJKKC22, DBLP:conf/sosp/KwonLZ0ZY0ZS23}. 
In particular, Recursive models demonstrate up to 2.66$\times$ speedup in generation compared to vanilla counterparts. 
Additionally, the recursive Gemma model significantly outperforms the vanilla pretrained Pythia model, with nearly 4$\times$ improvement in throughput.
Relaxed recursive models show a clear trade-off between achievable few-shot performance and throughput, modulated by the degree of relaxation through the LoRA ranks. This characteristic enables flexible model selection tailored to specific deployment scenarios.
Comprehensive results are presented in Tables\,\ref{tab:final_performance_throughput} and \ref{tab:final_performance_throughput_16gb}.

\vspace{3pt}
\begin{tcolorbox}[
    enhanced,
    colback=blue!5!white, 
    colframe=black, 
    width=\textwidth,
    coltitle=white,
    title=\textbf{Takeaways for Continuous Depth-wise Batching},
    top=3mm, bottom=2mm,
    attach boxed title to top left={yshift=-2.8mm, xshift=4mm},
    boxed title style={colback=black, colframe=black, boxrule=0mm, 
    toptitle=1mm, bottomtitle=1mm}
]
We analyze the potential for throughput improvement in the Recursive Transformer via continuous depth-wise batching, a novel inference paradigm. In theory, we find that we can achieve up to 2-3$\times$ speedup compared to a vanilla Transformer. This even outperforms the throughput gain achieved by existing continuous sequence-wise batching methods in vanilla models.
\end{tcolorbox}

\section{Related Work}

Cross-layer parameter sharing has proven to be an effective method for achieving parameter efficiency in deep learning models such as RNNs~\citep{DBLP:journals/corr/abs-1808-03314, graves2016adaptive}, CNNs~\citep{DBLP:journals/corr/EigenRFL13, DBLP:conf/iclr/SavareseM19, DBLP:conf/cvpr/GuoYWLQY19, DBLP:conf/eccv/ShenLX22}, and the popular Transformer architecture. The Universal Transformer~\citep{DBLP:conf/iclr/DehghaniGVUK19}, a recurrent self-attentive model, demonstrated superior performance to non-recursive counterparts with significantly fewer parameters. This cross-layer parameter sharing approach has subsequently been explored in various tasks, including language understanding ~\citep{DBLP:conf/iclr/LanCGGSS20}, language modeling~\citep{DBLP:conf/nips/BaiKK19, mohtashami2023cotformer, DBLP:conf/icml/Liu0ILTFXCSKLC24, csordas2024moeut, DBLP:journals/corr/abs-2405-16712}, and machine translation~\citep{DBLP:conf/aaai/DabreF19, milbauer-etal-2023-lait, DBLP:conf/aaai/XiaHTTHQ19, DBLP:conf/sustainlp/TakaseK23, DBLP:conf/emnlp/GeCW22}. These methods often claim to achieve comparable performance with more compact models and increased computational speed, while also setting the ground for effective adaptive compute solutions~\citep{DBLP:conf/iclr/DehghaniGVUK19,graves2016adaptive,schuster-etal-2021-consistent}.

Concurrently, there has been growing interest in exploiting recurrent architectures for algorithmic or logical reasoning tasks~\citep{saunshi2024inductive}. 
Prior research~\citep{DBLP:conf/nips/SchwarzschildBG21, mcleish2022re} has shown that recurrent networks can extrapolate reasoning strategies learned on simple problems to harder, larger problems through additional recurrences during inference.
The looped Transformer structure has also been employed to emulate basic computing blocks for program simulation~\citep{DBLP:conf/icml/GiannouRS0LP23}, to learn iterative algorithms for data-fitting problems~\citep{DBLP:conf/iclr/Yang0NP24}, to achieve length generalization in algorithmic tasks~\citep{fan2024looped}, and promising theoretical potential for few-shot learning~\citep{gatmiry2024looped}.

However, previous work has predominantly focused on relatively small Transformer models, trained from scratch without leveraging pretrained model weights. 
Our work distinguishes itself by investigating parameter sharing in the context of LLMs and proposing effective initialization strategies that leverage the knowledge embedded within existing LLMs. 
To the best of our knowledge, we are the first to propose a generalized framework for parameter-shared models, enabling relaxation in weight tying constraints through layer-specific modules.

\clearpage
In this paper, we also discuss how Recursive Transformers can be well suited for early-exiting techniques to accelerate decoding in LLMs. The inherent recursive structure readily enables early-exiting for individual responses within a large serving batch, which is often a practical limitation of such techniques.
Vanilla Transformers encounter a synchronization issue with early-exiting, where the model must forward all layers if even a single token in a batch requires full processing (exited tokens must wait for them).
Several approaches attempt to exploit this idle time by computing missing KV caches for exited tokens in later layers, which are essential for subsequent sequence generation. These techniques include state propagation~\citep{DBLP:conf/nips/SchusterFG0B0TM22, DBLP:conf/iclr/ElbayadGGA20}, SkipDecode~\citep{del2023skipdecode}, and parallel decoding (which can be combined with Speculative Decoding)~\citep{DBLP:conf/emnlp/BaeKSY23, DBLP:conf/acl/ElhoushiSLHWL0A24, liu2024kangaroo, DBLP:conf/icml/ChenPLDZ24, DBLP:conf/naacl/TangZLAMM24}. 
Nevertheless, the heterogeneous parameters across varying model depths still hinder the efficient progression of exited tokens to subsequent sequences. 
In contrast, our Recursive Transformers enable parallel computation for tokens at different depths and sequences (in a continuous depth-wise batching paradigm)---also allow for parallel computation of missing KV caches with minimal overhead during the memory-bounded decoding phase.

\section{Conclusion and Future Work}

In this work, we introduced Recursive Transformers, in which we compress LLMs via parameter sharing across recursively looped blocks of layers. Additionally, we presented a novel relaxation strategy that allows for low-rank deltas between shared layers by integrating layer-specific LoRA modules into the fully-tied structure. Through novel initialization techniques for looped layers and LoRA modules, we achieved significant performance improvements that closely approximate the original pretrained model. 
Finally, by exploiting the recursive patterns and an early-exiting approach, we propose a continuous depth-wise batching paradigm tailored for efficient serving systems of Recursive Transformers. We theoretically demonstrated that an oracle-exiting strategy can yield substantial throughput gains, reaching up to 2-3$\times$ speedup. 
This work motivates further research on recursive patterns in modern LLMs such as:

\vspace{-12pt}
\paragraph{Compatibility with sparse designs}
Sparsity-based approaches, such as pruning~\citep{DBLP:journals/corr/HanPTD15}, quantization~\citep{DBLP:conf/cvpr/JacobKCZTHAK18}, or layer-skipping mechanisms~\citep{DBLP:journals/corr/abs-2404-02258}, recently also give promising model compression results. In fact, many of these techniques are complementary to our approach: for example, we can seamlessly have a recursive, \textit{sparse} architecture. In this work, we rather choose to focus on recursive dense designs (a domain that remains relatively unexplored) that also have very promising, practical performance traits (i.e., allowing for continuous depth-wise batching for faster throughput). 
That said, while in this work we take the first step at studying Relaxed Recursive Transformer with dense Transformer layers, we do believe that incorporating Mixture-of-expert~\citep{DBLP:journals/jmlr/FedusZS22}, activation-skipping~\citep{DBLP:conf/icml/LiuWDZY0S0TRC23} and SSM components~\citep{DBLP:journals/corr/abs-2405-16712} within the looped blocks are promising directions for future research.

\vspace{-12pt}
\paragraph{Latent Reasoning via Recurrent Depth}
Beyond efficiency gains through down-scaling materialized parameters with recursive patterns, an alternative research direction lies in scaling-up recurrent depth to facilitate latent reasoning. Specifically, recurrent computation can manifest thinking vertically by processing internal hidden states at each depth. One promising approach involves leveraging contemplation tokens~\citep{pfau2024let, DBLP:conf/iclr/GoyalJRMKN24} or latent (continuous) space representations~\citep{hao2024training, cheng2024compressed} to enhance reasoning in mathematical and code generation tasks. Another valuable direction focuses on enhancing the efficiency and training stability of approaches that recursively scale-up depth, building upon concepts of deep thinking~\citep{DBLP:conf/nips/SchwarzschildBG21, geiping2025scaling}.

\vspace{-12pt}
\paragraph{Scaling up Recursive Transformers}
Scaling our approach to larger LLMs (7B and beyond) is a promising avenue for future research. While our methodology is expected to remain effective, achieving comparable performance may require significantly higher uptraining costs. Increased model size offers the potential for a reduced memory footprint from recursive patterns; however, it is unclear whether this translates to larger batch sizes, given the corresponding increase in hidden dimensions. Nevertheless, our continuous depth-wise batching will yield considerable gains in serving efficiency.

\vspace{-12pt}
\paragraph{Beyond hypothetical generation speedup}
Our oracle-exiting approach assumes any intermediate prediction matching the final output can be exited. However, accurate throughput measurement requires confidence-based early-exiting algorithms~\citep{DBLP:conf/nips/SchusterFG0B0TM22, DBLP:conf/emnlp/BaeKSY23}. Moreover, practical deployment needs to address decoding bottlenecks like key-value cache computation for exited tokens in remaining loops. 
Nevertheless, there are potential solutions: for example, the missing KV cache computations can be addressed by leveraging continuous depth-wise batching, allowing the KV cache for exited positions in subsequent loops to be performed in parallel with the computations for the next sequence sample.
Moreover, we can explore key-value cache sharing strategies~\citep{DBLP:journals/corr/abs-2405-05254, DBLP:journals/corr/abs-2405-12981} for future work.\looseness=-1

\vspace{-12pt}
\paragraph{Efficient serving of multi-LoRA layers}
Relaxed models require the computation of distinct LoRA modules during batched inference, akin to multi-task learning~\citep{DBLP:conf/coling/FengHZHW24, wang2023customizable}, hindering parallel computation.
We concatenated LoRA weights into a single weight to improve efficiency over sequential computation, yet it introduces redundancy. 
To mitigate this, we can explore optimized CUDA kernels for LoRA serving~\citep{DBLP:journals/corr/abs-2311-03285, chen2024punica} and parallelization across accelerators, inspired by distributed training for Mixture of Experts~\citep{DBLP:journals/jmlr/FedusZS22, gale2023megablocks}.\looseness=-1

\section*{Acknowledgements}
We thank Jacob Eisenstein for valuable feedback on an earlier version of the paper. We thank Jiyoun Ha, Alfred Piccioni, Dayeong Lee for the support with setting up the experimental environment. We also thank Donald Metzler, Ivan Korotkov, Jai Gupta, Sanket V. Mehta, Vinh Q. Tran, Brennan Saeta, Jean-François Kagy, Zhen Qin, and Jing Lu for helpful conversations. Finally, we thank the Google Cloud Platform for awarding Google Cloud credits for this project.

{
\bibliography{googledeepmind}
}

\clearpage
\counterwithin{figure}{section}
\counterwithin{table}{section}
\appendix


{
\section{Components in Transformer Architecture}
\label{app:transformer}

The Transformer block consists of two core components: a multi-head attention (MHA) mechanism and a feed-forward network (FFN). MHA utilizes multiple attention heads to capture diverse relationships within the input sequence. The computation within each attention head is formulated as:
\begin{align*}
\text{Attention}(\mathbf{Q}, \mathbf{K}, \mathbf{V}) = \text{softmax} \left( \frac{\mathbf{Q}\mathbf{K}^T}{\sqrt{d_k}} \right) \mathbf{V},
\end{align*}
where $\mathbf{Q}$, $\mathbf{K}$, and $\mathbf{V}$ are linear projections of the input, parameterized by learned weight matrices $\mathbf{W}_\ell^Q$, $\mathbf{W}_\ell^K$, and $\mathbf{W}_\ell^V$, respectively. The outputs from each head of the multi-head attention are concatenated and then projected back to the original hidden size using a learned weight matrix $\mathbf{W}_\ell^{out}$. 

While the FFN structure typically consists of two linear transformations, in the Gemma model, it deviates from this standard architecture as follows:
\begin{align*}
\text{FFN}(\mathbf{x}) = \mathbf{W}_\ell^{down} (\text{GELU}(\mathbf{x} \mathbf{W}_\ell^{gate}) * \mathbf{x} \mathbf{W}_\ell^{up})
\end{align*}
with three learned linear weight matrices and a GeGLU activation~\citep{DBLP:journals/corr/abs-2002-05202}.

\section{Parameter Sharing Strategy}
\label{app:looping_strategy}

\citet{DBLP:conf/sustainlp/TakaseK23} discuss three strategies for partial layer tying in Transformer models, as depicted in Figure\,\ref{fig:sharing_strategy}. The SEQUENE strategy is the simplest, assigning  the same parameters to consecutive layers. The CYCLE strategy repeatedly stacks a single block of unique layers to achieve the desired depth.
Meanwhile, the CYCLE (REV) strategy stacks the lower layers in reverse order for the remaining layers.

In the comparative analysis of SEQUENCE and CYCLE strategies~\citep{DBLP:conf/icml/Liu0ILTFXCSKLC24}, CYCLE demonstrated marginally superior zero-shot performance. 
Although the SEQUENCE approach, which caches shared weights (the capacity of SRAM is typically sufficient to hold a single transformer block) and computes them iteratively, has the potential to mitigate the weight transfer bottleneck between SRAM and DRAM, we prioritized compatibility with early-exiting.
Consequently, we specifically employed the CYCLE strategy, which enables continuous depth-wise batching and thereby maximizes the throughput of Recursive Transformers.

\begin{figure}[h!]
    \centering
    \begin{subfigure}[t]{0.322\textwidth}
        \includegraphics[width=\textwidth]{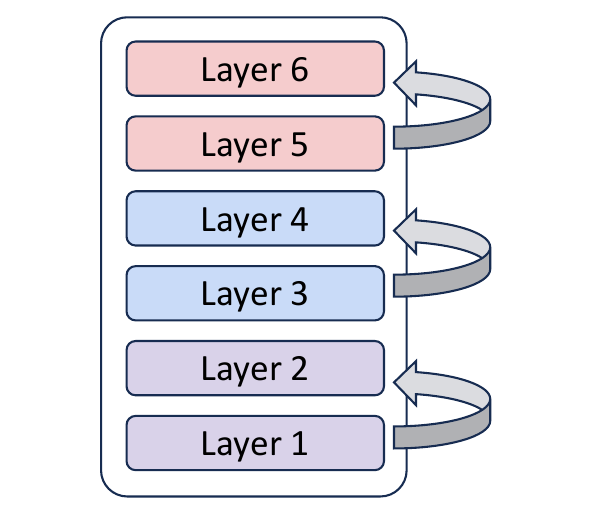}
        \subcaption{SEQUENCE}
    \end{subfigure}
    \centering
    \begin{subfigure}[t]{0.322\textwidth}
        \includegraphics[width=\textwidth]{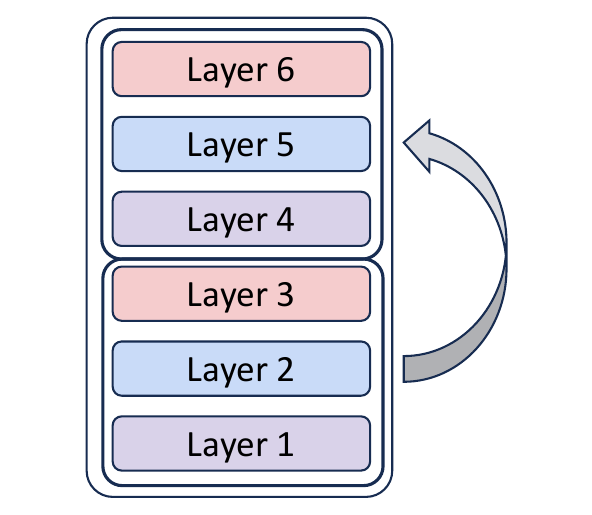}
        \subcaption{CYCLE}
    \end{subfigure}
    \centering
    \begin{subfigure}[t]{0.322\textwidth}
        \includegraphics[width=\textwidth]{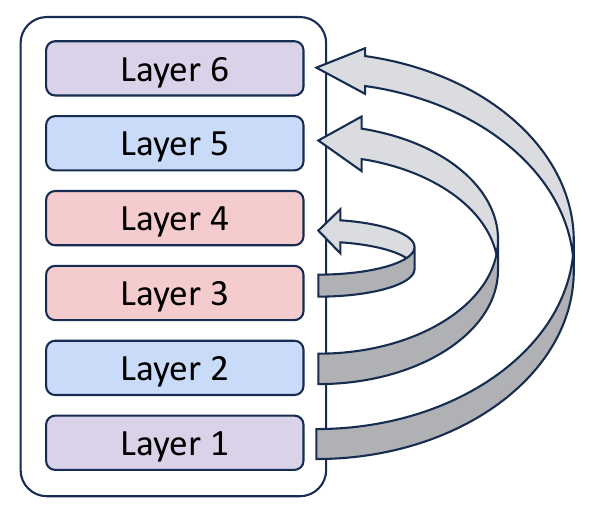}
        \subcaption{CYCLE\,(REV)}
    \end{subfigure}
    \caption{
    Three strategies for parameter sharing \citep{DBLP:conf/sustainlp/TakaseK23}. The examples utilize models with six layers, where identical colors represent shared weights. 
    }
    \label{fig:sharing_strategy}
\end{figure}

\clearpage
\section{Illustrative Examples of SVD Initialization in Relaxed Recursive Transformer}
\label{app:svd_init_relaxed_rets}

We propose an SVD initialization approach for LoRA modules within a Relaxed Recursive Transformer, effectively steering the summation of base and LoRA weights towards the pretrained weights of their corresponding depth. Figure\,\ref{fig:lora_init_overview_app} illustrates an overview of how the LoRA module is initialized under three different initialization techniques (Stepwise, Average, and Lower) for looped layers.
One crucial point is that if the initialized looped layer's weights match those of the original pretrained model, its corresponding LoRA module undergoes standard zero initialization: random Gaussian for matrix \textit{\textbf{A}} and zero for \textit{\textbf{B}}. For example, with the Stepwise method, the first loop's LoRA module receives standard zero initialization, while the second loop's LoRA is initialized using our proposed initialization.

\begin{figure}[ht]
    \centering
        \includegraphics[width=\textwidth]{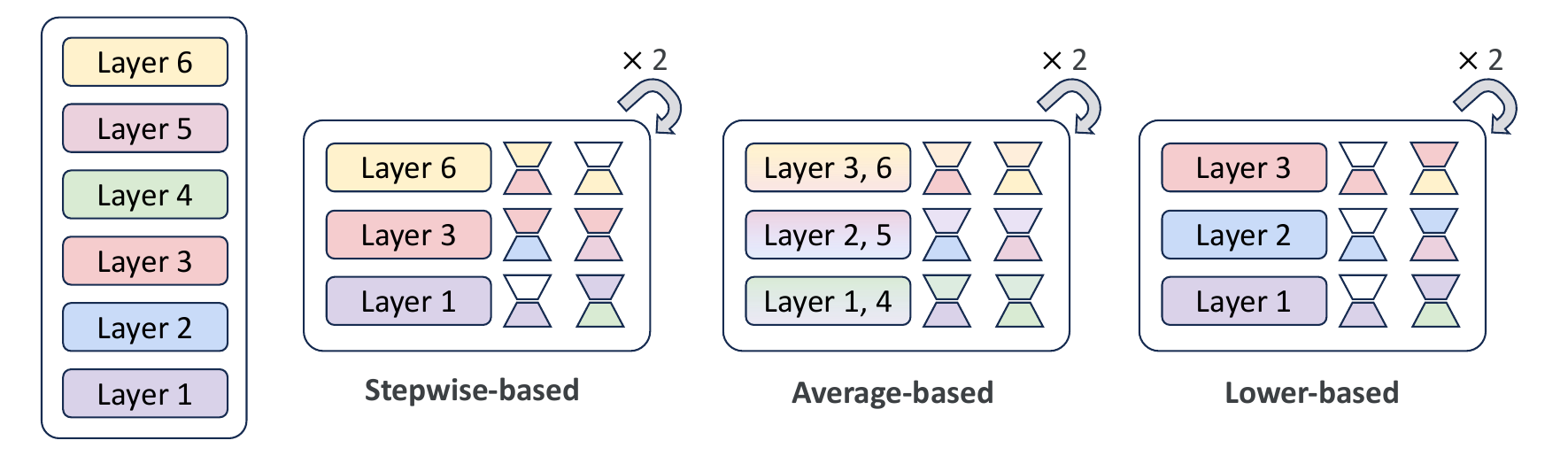}
    \caption{
    We visualize LoRA modules to show which residual matrices they target for initialization
    under three different looping initialization methods, assuming a full-size model with six layers and two looping blocks. For ease of understanding, \textit{{\textbf{A}}} matrices are colored according to the full-size model weights at the corresponding depth, while {\textit{\textbf{B}}} matrices are colored based on the looped layer weights. White \textit{{\textbf{B}}} matrices indicate cases where the full-size model and looped model weights are identical, resulting in standard zero initialization. 
    }
    \label{fig:lora_init_overview_app}
\end{figure}

\section{Overview of Three Pretrained LLMs}
\label{app:pretrained_model_performance}

We utilized three pretrained models---Gemma 2B\,\citep{team2024gemma}, TinyLlama 1.1B\,\citep{zhang2024tinyllama}, and Pythia 1B\,\citep{DBLP:conf/icml/BidermanSABOHKP23}---and converted them into Recursive Transformers. 
Their corresponding few-shot performance results are presented in Table\,\ref{tab:pretrained_model_performance_app}.

\begin{table}[h]
    \small
    \centering
    \resizebox{\textwidth}{!}{
    \setlength{\tabcolsep}{6pt}
    \begin{tabular}{l|rcr|ccccccc|c}
    \toprule
      &  &  & & \multicolumn{8}{c}{\textbf{Few-shot Accuracy\,$\uparrow$}} \\
    \cmidrule(l{2pt}r{2pt}){5-12} 
     \textbf{Models} & N-emb & Dataset & $N_{token}$ & LD & HS & PQ & WG & ARC-e & ARC-c & OB & Avg  \\
    \midrule
    Gemma\,2B & 1.99B & Unreleased & 3T & {63.13} & {71.38} & {78.13} & {65.04} & {72.26} & {41.89} & 40.20 & {61.72}  \\
     \midrule
    \multirow{5}{*}{TinyLlama\,1.1B} & \multirow{5}{*}{0.97B} & \multirow{4}{*}{SlimPajama\,+} & 105B & 43.26 & 42.23 & 66.81 & 53.35 & 44.74 & 23.21 & 29.20 & 43.26  \\
    &  & \multirow{4}{*}{Starcoderdata} & 503B & 48.92 & 49.56 & 69.42 & 55.80 & 48.32 & 26.54 & 31.40 & 47.14  \\
    & & & 1T & 53.00 & 52.52 & 69.91 & 55.96 & 52.36 & 27.82 & 33.40 & 49.28  \\
    & & & 2T & 53.33 & 54.63 & 70.67 & 56.83 & 54.67 & 28.07 & 33.40 & 50.23  \\
    & & & 3T & 58.82 & 59.20 & 73.29 & 59.12 & 55.35 & 30.12 & 36.00 & 53.13  \\
     \midrule
    Pythia\,1B & 0.81B & Pile & 300B &  57.52 & 49.10 & 70.40 & 52.80 & {51.89} & 26.71 & {33.40} & {48.83}  \\
    \bottomrule
    \end{tabular}
    }
    \caption{
    Few-shot performance of three pretrained models. Few-shot accuracy is measured on the LAMBADA, HellaSwag, PIQA, WinoGrande, ARC-easy, ARC-challenge, and OpenBookQA benchmarks. We evaluated intermediate checkpoints up to the fully trained checkpoint for TinyLlama 1.1B. Among these, we utilized the 105B intermediate checkpoint to study an under-trained model.
    }
    \label{tab:pretrained_model_performance_app}
\end{table}

This diversity offers several benefits. First, with three versions of recursive models, we can compare their performance based on the number of trainable parameters. Notably, the comparison between the recursive Gemma and the pretrained TinyLlama and Pythia models highlights that leveraging well-trained model weights can lead to a superior Recursive Transformer of equivalent size, even with substantially lower uptraining costs.
Second, by utilizing models ranging from under-trained (e.g., TinyLlama) to significantly over-trained (e.g., Gemma), we can gain insights into the uptraining costs required for Recursive Transformers to closely match the performance of pretrained models. 
Finally, the diversity in pretraining datasets allows us to observe how Recursive Transformers perform when faced with distribution shifts in the uptraining dataset. Table\,\ref{tab:target_baseline_performance} presents the evaluation results obtained after uptraining each of the pretrained models. While TinyLlama readily improves its performance due to uptraining on the same dataset, Gemma and Pythia show a decline in few-shot performance with SlimPajama uptraining, which can be attributed to the differences in data distribution and the lower quality of the uptraining dataset.

\section{Experimental Setup}
\label{app:experimental_settings}

\paragraph{Uptraining setting}
To convert vanilla Transformers into Recursive Transformers, we conducted further uptraining on either 15 billion or 60 billion tokens from the SlimPajama dataset\,\citep{cerebras2023slimpajama}. SlimPajama is an open-source dataset designed for training large language models, which is created by cleaning and deduplicating the RedPajama dataset\,\citep{together2023redpajama}. The source data primarily consists of web-crawled data, along with data from Github, books, Arxiv, Wikipedia, and StackExchange. 
We employed the HuggingFace training framework\,\citep{wolf2020transformers} and enhanced memory efficiency through the Zero Redundancy Optimizer (ZeRO)\,\citep{rajbhandari2020zero} from the DeepSpeed library\,\citep{rasley2020deepspeed}, along with mixed precision training.
The context length was set to 2048, and the batch size was approximately 2 million tokens. We used the AdamW optimizer\,\citep{DBLP:conf/iclr/LoshchilovH19} with a learning rate of 2e-4, utilizing a cosine annealing learning rate scheduler~\citep{DBLP:conf/iclr/LoshchilovH17}. Additionally, we set warmup steps to 200 for 15 billion token training and 800 for 60 billion token training. Eight H100 GPUs were used for the training.

\paragraph{Early-exit training setting}
Similar to the uptraining process, we used the SlimPajama dataset to enable models to predict next tokens at intermediate loops. Models with two looping blocks underwent additional training on a total of two exit points, whereas models with three blocks were trained on three exit points. We explored various strategies, but by default, we continued training on an additional 15 billion tokens (SlimPajama dataset), starting from the uptrained Recursive Transformers. We also utilized eight H100 GPUs and maintained consistent configurations with the uptraining settings, including batch size, context length, and learning rates.

\paragraph{Evaluation setting}
We evaluated perplexity on test sets from three language modeling datasets: SlimPajama, RedPajama, and PG19\,\citep{rae2019compressive}. Additionally, we used the Language Model Evaluation Harness framework\,\citep{eval-harness} to evaluate accuracy on seven few-shot tasks: LAMBADA\,\citep{paperno2016lambada}, HellaSwag\,\citep{zellers2019hellaswag}, PIQA\,\citep{bisk2020piqa}, WinoGrande\,\citep{DBLP:conf/aaai/SakaguchiBBC20}, ARC-easy and ARC-challenge\,\citep{clark2018think}, and OpenBookQA\,\citep{DBLP:conf/emnlp/MihaylovCKS18}. We adhered to the standard number of shots specified by the evaluation framework for each dataset. For few-shot datasets, excluding LAMBADA and WinoGrande, we normalized accuracy by the byte length of the target string. All evaluation performance measurements were conducted using a single H100 GPU.

\paragraph{Throughput measurement settings}
To present the hypothetical generation speeds of our Recursive Transformers, we prepared two key elements: per-token generation time and exit trajectory datasets.
Firstly, we measured the generation time under various model configurations using dummy weights and inputs. We measured the time for each component, such as embedding matrices, Transformer blocks, and the classifier head 
(final throughput comparisons were based solely on the time spent within Transformer blocks.)
We tested two settings of prefix and decoding lengths (512\,/\,2048 and 64\,/\,256), calculating the per-token time by dividing the total elapsed time by the decoding length. 
Using a single A100 40GB GPU, we recorded these decoding times across different batch sizes, until an out-of-memory error occurred or under a specific memory constraint was reached.

To obtain exit trajectory data, we assumed an oracle-exiting approach, where all tokens could exit at intermediate loops if intermediate predictions matched the final loop's prediction.
Since our models are not finetuned on any specific downstream tasks, we did simulation with three language modeling datasets (SlimPajama, RedPajama, and PG19) as if they were generated by our models. 
For simplicity, we assumed a queue of 20K samples, rather than considering their arrival in static or dynamic time intervals.
We then recorded the exit loop of each token in these samples using the oracle-exiting algorithm.
With these two measurement (per-token generation time and exit trajectories), we present the hypothetical throughput of Recursive Transformers under various simulation scenarios.

\section{Expanded Results of Initialization Methods for Looped Layers}
\label{app:initialization}

\paragraph{Ablation study of Stepwise method}
We initially hypothesized that the Stepwise method's performance could be significantly influenced by the specific rule used for layer selection from the pretrained model. To investigate this, we conducted a controlled experiment (illustrated in Figure\,\ref{fig:stepwise_abl_app}), where layers were selected at certain intervals starting from the first layer. We then varied whether the final layer of the pretrained model was included in the initialization or not. While a Pythia model showed no significant differences in training loss or few-shot performance, other models like Gemma exhibited superior results when both the first and last layers were preserved. This observation aligns well with prior work suggesting that maintaining the weights of the first and last layers during depth up-scaling for LLMs can yield performance benefits~\citep{DBLP:conf/naacl/KimKPLSKKKLKAYLPGCLK24}.

\paragraph{Ablation study of Average method}
The Average initialization method exhibited notably poor performance, particularly when applied to the Gemma model. We hypothesized that this could be attributed to instability in the model's learned distribution, potentially arising from averaging of normalization layer weights. 
Relatedly, several studies~\citep{csordas2024moeut, shim2024leveraging, mohtashami2023cotformer} have explored the careful design of layer normalization in parameter-shared models.
To investigate this further, we experimented with three different methods for initializing normalization weights, as outlined in Figure\,\ref{fig:average_abl_app}: averaging weights (Norm-avg), selecting weights from a single layer (Norm-choice), and zero initialization (Norm-zero).
The performance trend observed among these methods varied across different model architectures. However, zero initialization of normalization layers resulted in a huge performance drop in certain architectures like TinyLlama and Pythia. Conversely, we observed no big difference between averaging and single-layer selection, suggesting that any form of distillation of the normalization weights appears to be sufficient for maintaining performance.

\clearpage

\begin{figure}[h]
    \centering
    \begin{subfigure}[t]{0.6\textwidth}
        \includegraphics[width=\textwidth]{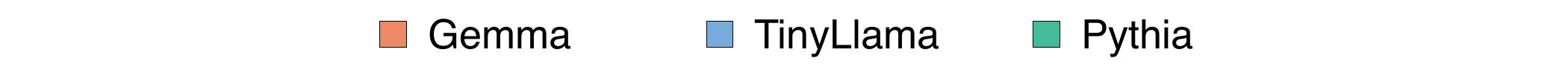}
    \end{subfigure}
    \\
    \centering
    \begin{subfigure}[t]{0.327\textwidth}
        \includegraphics[width=\textwidth]{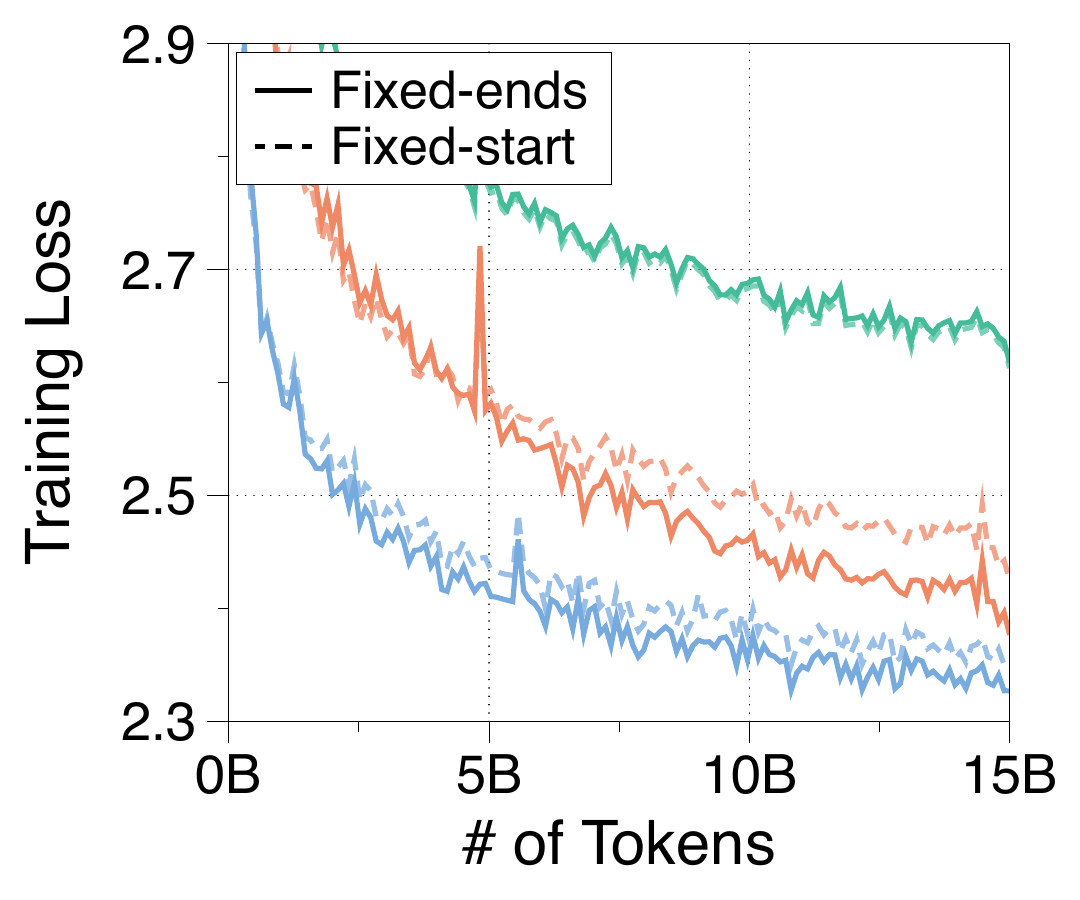}
        \subcaption{Stepwise ablations}
        \label{fig:stepwise_abl_app}
    \end{subfigure}
    \hspace{10pt}
    \centering
    \begin{subfigure}[t]{0.3\textwidth}
        \includegraphics[width=\textwidth]{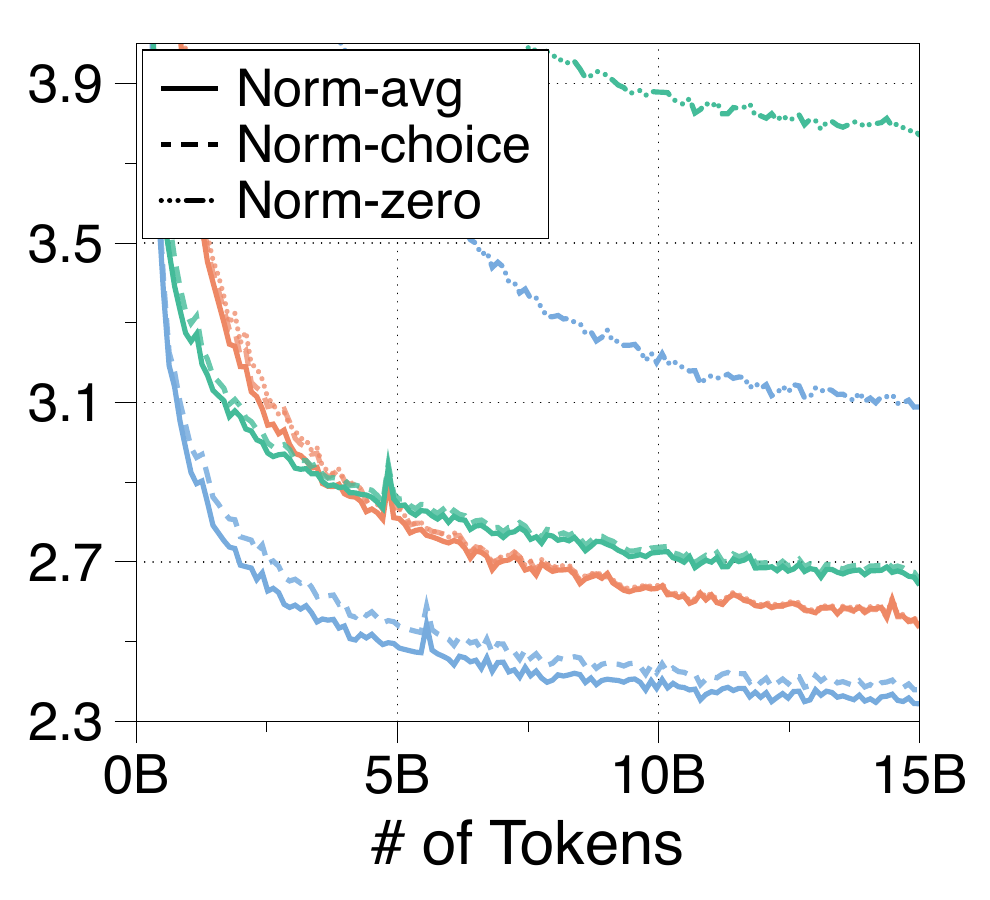}
        \subcaption{Average ablations}
        \label{fig:average_abl_app}
    \end{subfigure}
    \caption{
    Training loss curves of Stepwise and Average initialization variants across three models with two blocks. \textbf{(a)} ``Fixed-start'' indicates that the first layer of the pretrained model is selected initially, and subsequent layers are repeatedly chosen at a fixed interval. ``Fixed-ends'' means that the first and last layers are included, and intermediate layers are selected at specific step intervals. 
    \textbf{(b)} When initializing the weights of normalization layer (RMSNorm in Gemma and TinyLlama, and LayerNorm in Pythia), we consider whether to average the weights\,(Norm-avg), select a single layer's weights\,(Norm-choice), or use zero initialization\,(Norm-zero). 
    }
    \label{fig:initialization_ablation_app}
\end{figure}

\paragraph{Overall comparison of training perplexity}
Figure\,\ref{fig:training_loss_app} presents a comparative analysis of training loss across three model architectures and varying looping blocks, incorporating our proposed initialization methodologies. To set an upper bound on performance, we utilized a full-size model further uptrained on SlimPajama, accounting for the distribution shift between uptraining and pretraining data. Additionally, we trained a Recursive Transformer with a random initialization, ensuring its exclusive reliance on the recursive architecture without leveraging any pretrained weights.
While some variance was observed across architectures, all proposed methods utilizing pretrained model weights demonstrated significantly superior performance compared to random initialization. Notably, the Stepwise method consistently achieved the best performance across diverse settings. Although the full-size model's performance was considerably higher, bridging this gap with only 15 billion tokens of uptraining represents a remarkable achievement.

\begin{figure}[ht]
    \centering
    \begin{subfigure}[t]{\textwidth}
        \includegraphics[width=\textwidth]{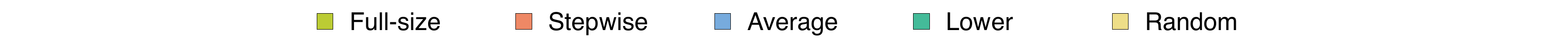}
    \end{subfigure}
    \centering
    \begin{subfigure}[t]{0.262\textwidth}
        \includegraphics[width=\textwidth]{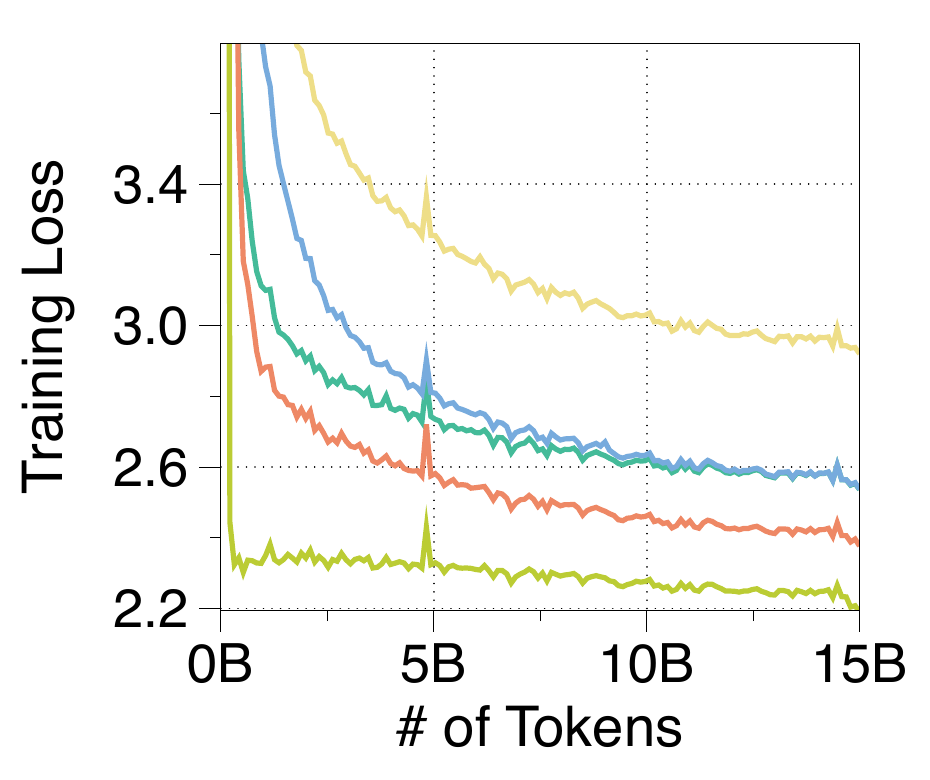}
        \subcaption{Gemma (2 blocks)}
    \end{subfigure}
    \centering
    \begin{subfigure}[t]{0.24\textwidth}
        \includegraphics[width=\textwidth]{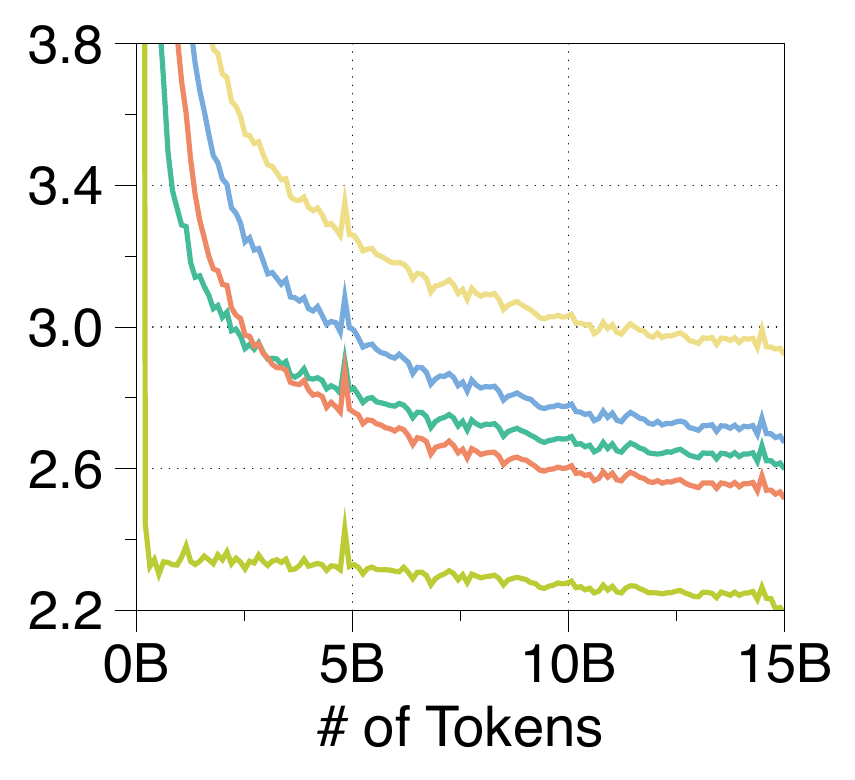}
        \subcaption{Gemma (3 blocks)}
    \end{subfigure}
    \centering
    \begin{subfigure}[t]{0.24\textwidth}
        \includegraphics[width=\textwidth]{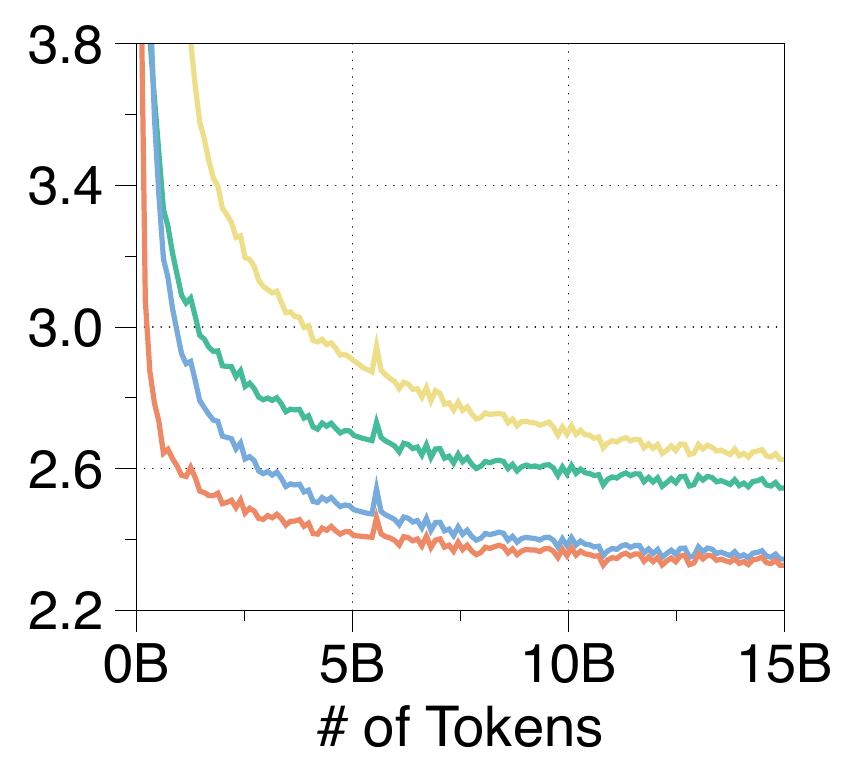}
        \subcaption{TinyLlama (2 blocks)}
    \end{subfigure}
    \centering
    \begin{subfigure}[t]{0.24\textwidth}
        \includegraphics[width=\textwidth]{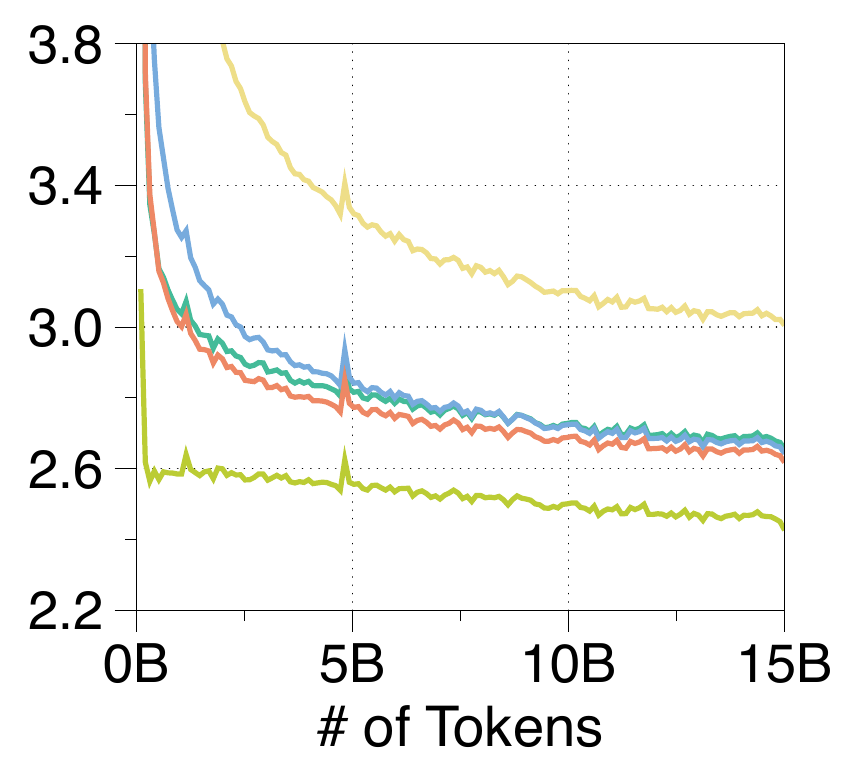}
        \subcaption{Pythia (2 blocks)}
    \end{subfigure}
    \caption{Training loss curves of Recursive Transformers using various initialization. We omitted a separate curve for the full-size TinyLlama model, as we used the original pretrained model as the full-size baseline because both pretraining and uptraining datasets are same as SlimPajama. 
    }
    \label{fig:training_loss_app}
\end{figure}

\paragraph{Overall comparison of few-shot performance}
Few-shot performance exhibited a consistent trend with training perplexity. Table\,\ref{tab:initialization_total_app} provides a comparative summary of the proposed looping initialization methods against the full-size model, the reduced-size model, and Recursive Transformers utilizing random initialization. Moreover, Figure\,\ref{fig:fewshot_bar_total_app} visually illustrates the performance differences across different few-shot datasets. Notably, the Stepwise method consistently demonstrated the best performance, showing a performance improvement of up to 14.1\%p compared to random initialization.

\begin{table}[ht!]
    \small
    \centering
    \resizebox{\textwidth}{!}{
    \setlength{\tabcolsep}{4pt}
    \begin{tabular}{l|c|cc|cc|rrr|ccccccc|cc}
    \toprule
     & & \multicolumn{2}{c|}{\textbf{Uptrain}} & \multicolumn{2}{c|}{\textbf{Looping}} & \multicolumn{3}{c|}{\textbf{Perplexity\,$\downarrow$}} & \multicolumn{9}{c}{\textbf{Few-shot Accuracy\,$\uparrow$}} \\
    \cmidrule(l{2pt}r{2pt}){3-4} \cmidrule(l{2pt}r{2pt}){5-6}  \cmidrule(l{2pt}r{2pt}){7-9}  \cmidrule(l{2pt}r{2pt}){10-18} 
     \textbf{Models} & N-emb & PT & $N_{tok}$ & Block & Init & SlimP & RedP & PG19 & LD & HS & PQ & WG & ARC-e & ARC-c & OB & Avg & $\Delta$  \\
    \midrule
    \multirow{11}{*}{Gemma} & 1.99B &  \cmark & 15B & - & - & 10.76 & 8.47 & 13.08 & 63.5 & 68.5 & 77.0 & 63.5 & 67.6 & 38.1 & {42.6} & 60.1  & - \\
     & 0.99B &  \xmark & 15B & - & - & 22.63 & 20.03 & 32.60 & 28.9 & 31.6 & 63.1 & 52.3 & 41.2 & 22.5 & 27.8 &  38.2 & -  \\
     & 0.66B &  \xmark & 15B & - & - & 24.44 & 21.69 & 36.03 & 27.2 & 30.6 & 63.8 & 50.5 & 40.6 & 22.0 & 27.0  & 37.4 & -   \\
     \cmidrule(l{2pt}r{2pt}){2-18} 
     \rowcolor[gray]{0.9}
     \cellcolor{white} & 0.99B & \cmark & 15B &  2 & Step & \textbf{12.85} & \textbf{10.29} & \textbf{16.21} & \textbf{53.0} & \textbf{57.3} & \textbf{73.2} & \textbf{56.2} & \textbf{56.1} & \textbf{29.2} & \textbf{36.6} & \textbf{51.7} & \textcolor{custom_green}{\!\!\!\textbf{+14.1}} \\
     & 0.99B & \cmark & 15B &  2 & Avg & 15.15 & 12.57 & 19.86 & 43.6 & 47.4 & 70.4 & 52.6 & 50.5 & 27.8 & 34.4 & 46.7 & \textcolor{custom_green}{\textbf{+9.1}}  \\
     & 0.99B & \cmark & 15B &  2 & Lower & 15.03 & 12.46 & 19.63 & 42.5 & 48.0 & 71.0 & 54.6 & 52.2 & 27.7 & 33.8 & 47.1 & \textcolor{custom_green}{\textbf{+9.5}}  \\
     & 0.99B & \xmark & 15B &  2 & Rand & 22.66 & 20.06 & 32.86 & 27.4 & 31.6 & 63.4 & 50.5 & 39.7 & 21.9 & 28.8 & 37.6 & -  \\
     \cmidrule(l{2pt}r{2pt}){2-18} 
     \rowcolor[gray]{0.9}
     \cellcolor{white} & 0.66B & \cmark & 15B &  3 & Step & \textbf{14.75} & \textbf{12.10} & \textbf{19.32} & \textbf{45.0} & \textbf{49.9} & \textbf{69.8} & \textbf{55.8} & \textbf{52.7} & \textbf{27.9} & \textbf{33.6} & \textbf{47.8} & \textcolor{custom_green}{\textbf{+9.9}} \\
     & 0.66B & \cmark & 15B &  3 & Avg & 17.45 & 14.65 & 23.63 & 39.4 & 39.0 & 66.6 & 48.7 & 46.5 & 24.7 & 31.8 & 42.4 & \textcolor{custom_green}{\textbf{+4.5}} \\
     & 0.66B & \cmark & 15B &  3 & Lower & 15.96 & 13.24 & 20.90 & 41.9 & 43.2 & 70.0 & 52.6 & 49.5 & 26.6 & 31.6 & 45.0 & \textcolor{custom_green}{\textbf{+7.1}} \\
     & 0.66B & \xmark & 15B &  3 & Rand & 22.67 & 20.09 & 32.77 & 28.1 & 31.4 & 63.8 & 51.1 & 41.0 & 23.0 & 26.6 & 37.9 & -  \\
     \midrule
    & 0.97B & \cmark & - & - & - & 12.26 & 9.37 & 11.94 & 43.3 & 42.2 & 66.8 & 53.4 & 44.7 & 23.2 & 29.2 & 43.3 & -  \\
     & 0.48B &  \xmark & 15B & - & - & 16.61 & 15.66 & 20.27 & 22.3 & 30.0 & 60.9 & 50.6 & 37.0 & 23.0 & 28.0 & 36.0 & -  \\
     \cmidrule(l{2pt}r{2pt}){2-18} 
     \rowcolor[gray]{0.9}
     \cellcolor{white}TinyLlama &  0.48B  &   \cmark &  15B &   2 &   Step &   \textbf{11.61} &   \textbf{9.89}  &   \textbf{13.00} &  \textbf{ 39.6} &  \textbf{ 39.8} &   \textbf{66.5} &   \textbf{52.9} &   \textbf{44.3} &   \textbf{24.9} &   \textbf{30.6} &   \textbf{42.7} &   \textcolor{custom_green}{\textbf{+6.2}}  \\
     & 0.48B  & \cmark & 15B & 2 & Avg & 11.86 & 10.29  & 13.42 & 38.6 & 39.4 & 66.1 & 52.8 & 42.7 & 25.4 & \textbf{30.6} & 42.2 & \textcolor{custom_green}{\textbf{+5.7}} \\
     & 0.48B  & \cmark & 15B & 2 & Lower & 14.67 & 12.67  & 16.68 & 31.9 & 32.3 & 62.6 & 52.0 & 39.1 & 22.1 & 27.8 & 38.3 & \textcolor{custom_green}{\textbf{+1.8}} \\
     & 0.48B  & \xmark & 15B & 2 & Rand & 16.14 & 15.11 & 19.55 & 24.7 & 30.7 & 61.2 & 50.6 & 36.4 & 22.6 & 29.2 & 36.5 & - \\
     \midrule
    & 0.81B  & \cmark & 15B & - & - & 13.46 & 9.95 & 13.38 &  55.0 & 49.0 & 71.0 & 53.6 & 51.8 & {28.2} & 32.8 & {48.8} & - \\
     & 0.40B &  \xmark & 15B & - & - & 25.69 & 20.00 & 32.08 & 24.3 & 30.0 & 61.9 & 50.7 & 38.3 & 22.3 & 26.0  & 36.2 & -  \\
     \cmidrule(l{2pt}r{2pt}){2-18} 
     \rowcolor[gray]{0.9}
    \cellcolor{white}Pythia &  0.40B  &   \cmark &   15B &   2 &   Step &   \textbf{16.38} &   \textbf{12.37} &   \textbf{17.74} &   43.4 &   \textbf{40.5} &   67.4 &   50.8 &   \textbf{46.3} &   25.7 &   30.0 &   43.5 &   \textcolor{custom_green}{\textbf{+7.3}} \\
     & 0.40B  & \cmark & 15B & 2 & Avg & 16.76 & 12.76 & 18.63 & 43.6 & 39.1 & \textbf{68.2} & 51.9 & 45.4 & 25.1 & 29.8 & 43.3 & \textcolor{custom_green}{\textbf{+7.1}} \\
     & 0.40B  & \cmark & 15B & 2 & Lower & 17.04 & 12.62 & 18.44 & \textbf{43.9} & 39.2 & 66.3 & \textbf{53.4} & 45.4 & \textbf{25.8} & \textbf{31.2} & \textbf{43.6} & \textcolor{custom_green}{\textbf{+7.4}} \\
     & 0.40B  & \xmark & 15B & 2 & Rand & 24.45 & 18.93 & 29.63 & 25.2 & 30.2 & 62.1 & 51.1 & 39.2 & 22.4 & 23.6 & 36.2 & - \\
    \bottomrule
    \end{tabular}
    }
    \caption{
    Evaluation results of various initialization methods for looped layers. We indicate whether pretrained weights are used and the number of uptraining tokens. Perplexity is evaluated on test sets of three language modeling datasets, and accuracy is evaluated on seven few-shot benchmarks. Delta values\,($\Delta$) show improvements over random initialization.
    }
    \label{tab:initialization_total_app}
\end{table}

\begin{figure}[ht!]
    \centering
    \begin{subfigure}[t]{\textwidth}
        \includegraphics[width=\textwidth]{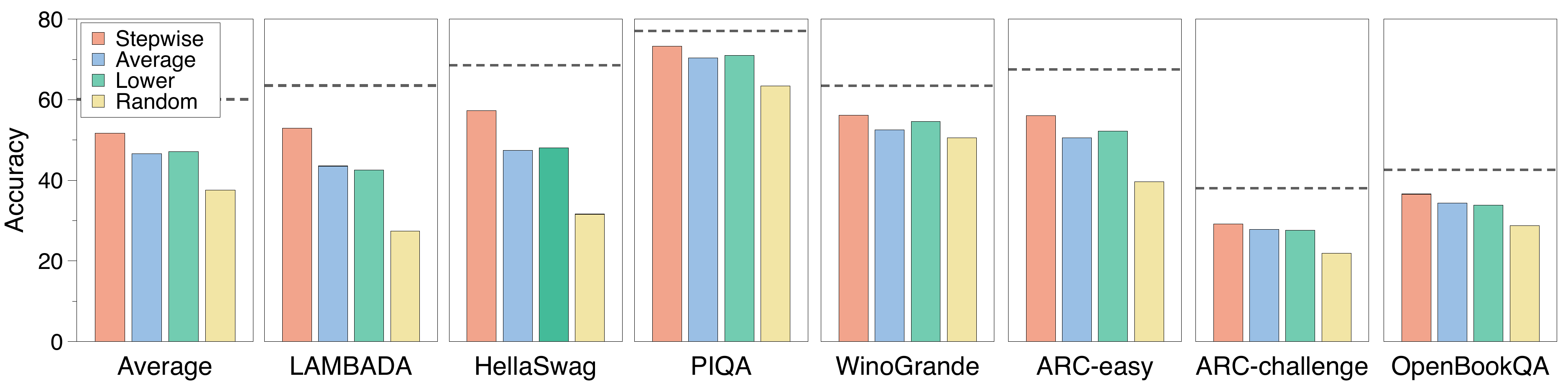}
        \caption{Recursive Gemma with 2 blocks}
        \vspace{5pt}
    \end{subfigure}
    \centering
    \begin{subfigure}[t]{\textwidth}
        \includegraphics[width=\textwidth]{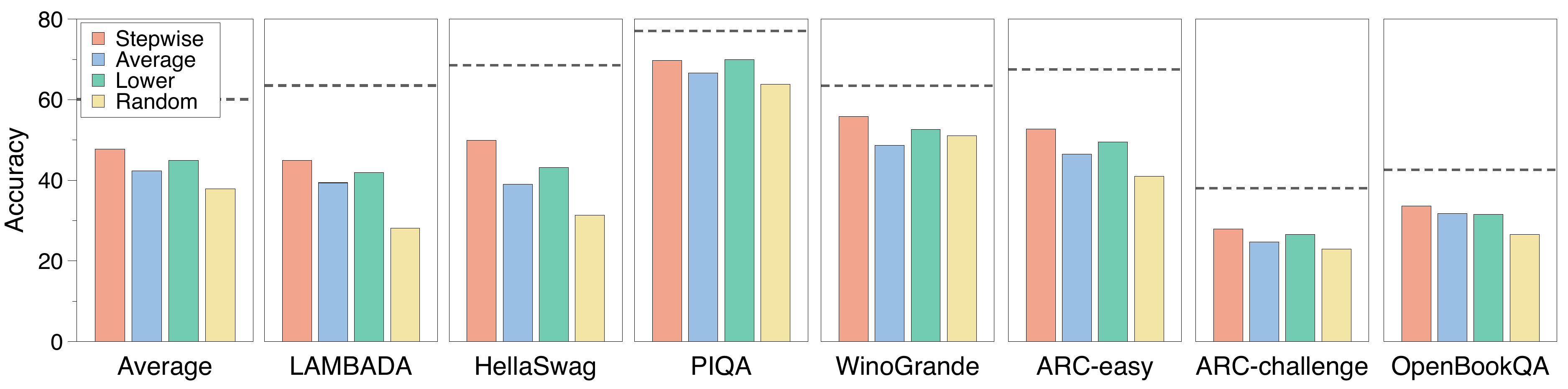}
        \caption{Recursive Gemma with 3 blocks}
        \vspace{5pt}
    \end{subfigure}
    \centering
    \begin{subfigure}[t]{\textwidth}
        \includegraphics[width=\textwidth]{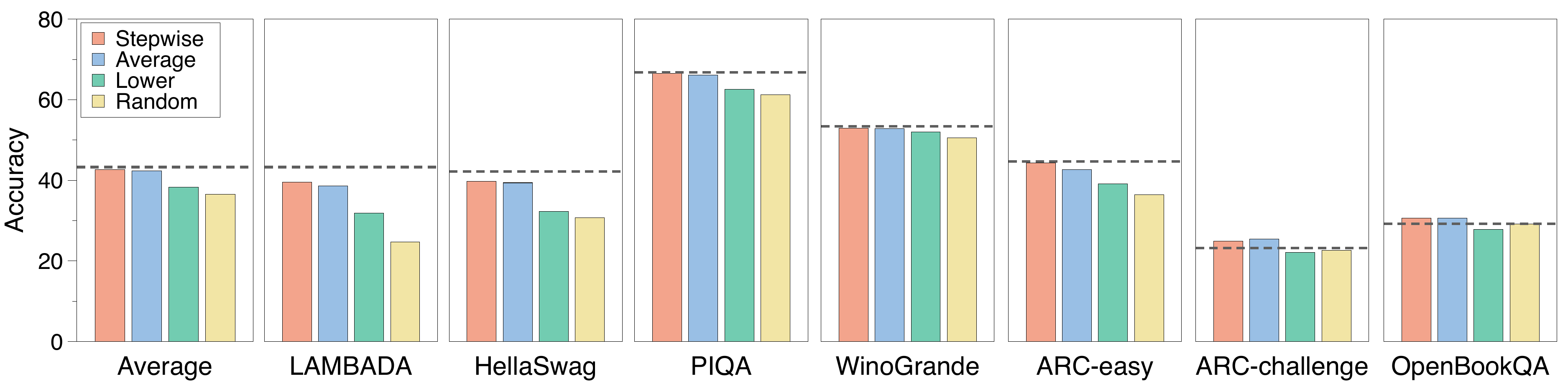}
        \caption{Recursive TinyLlama with 2 blocks}
        \vspace{5pt}
    \end{subfigure}
    \centering
    \begin{subfigure}[t]{\textwidth}
    \includegraphics[width=\textwidth]{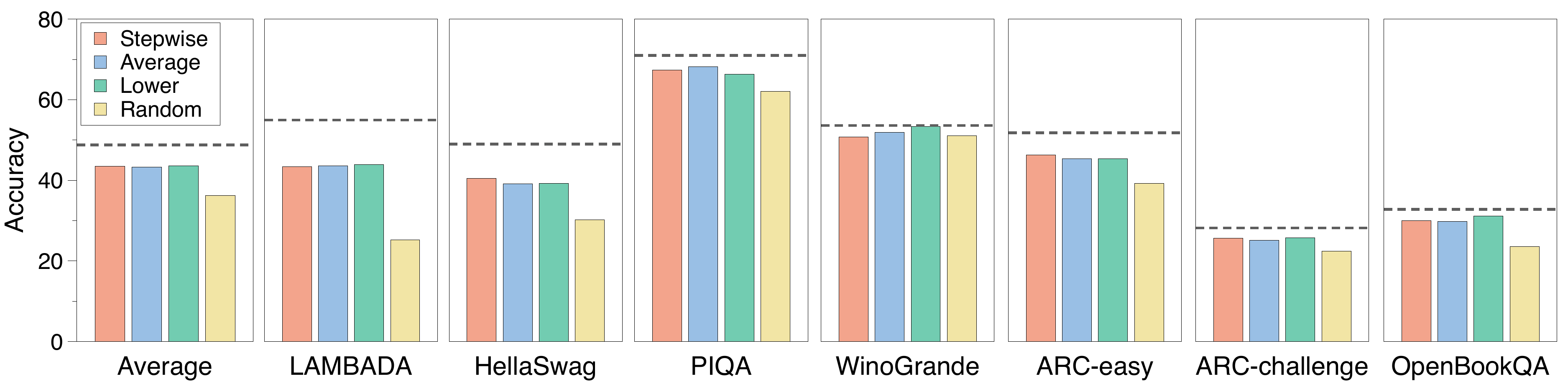}
        \caption{Recursive Pythia with 2 blocks}
        \vspace{5pt}
    \end{subfigure}
    \caption{Few-shot performance on seven benchmarks and their average accuracy based on four looping initialization methods. Full-size model performance is represented by a gray dotted line.
    }
    \label{fig:fewshot_bar_total_app}
\end{figure}

\clearpage
\paragraph{Comparison across various base model sizes}
We observed a consistent superiority of Stepwise initialization strategy for recursive conversion across both 1B and 2B model scales. To further evaluate on a wide range of base model sizes, we additionally experimented with two smaller model sizes, Pythia 410M and 160M.
Since we uptrained models on the Pile dataset~\citep{gao2020pile}, the pretraining corpus of the original Pythia model, we set the original model performance as the baseline for comparison.
The results in Table~\ref{tab:pythia_160_410} further validate the superior performance of the Stepwise method for looped layer initialization. These findings reinforce the robustness of our key observations regarding initialization methods for recursive conversion, complementing our original extensive experiments.\looseness=-1

\begin{table}[h]
    \small
    \centering
    \resizebox{\textwidth}{!}{
    \setlength{\tabcolsep}{4pt}
    \begin{tabular}{c|cc|cc|c|ccccccc|c}
    \toprule
     & \multicolumn{2}{c|}{\textbf{Uptrain}} & \multicolumn{2}{c|}{\textbf{Looping}} & \textbf{PPL}\,$\downarrow$  & \multicolumn{8}{c}{\textbf{Few-shot Accuracy\,$\uparrow$}} \\
    \cmidrule(l{2pt}r{2pt}){2-3} \cmidrule(l{2pt}r{2pt}){4-5}  \cmidrule(l{2pt}r{2pt}){6-6}  \cmidrule(l{2pt}r{2pt}){7-14} 
      N-emb & PT & $N_{tok}$ & Block & Init  & Pile & LD & HS & PQ & WG & ARC-e & ARC-c & OB & Avg  \\
    \midrule    
    300M & \cmark & - & - & - & - & 44.96 & 40.97 & 66.97 & 53.28 & 44.40 & 25.51 & 30.20 & 43.76 \\
    \midrule
    150M & \cmark & 15B & 2 & Step & \textbf{11.03} & \textbf{43.41} & \textbf{35.59} & \textbf{64.58} & \textbf{53.04} & 41.58 & 23.81 & \textbf{28.80} & \textbf{41.54} \\
    150M & \cmark & 15B & 2 & Lower & 11.47 & 42.98 & 34.32 & 63.93 & 52.41 & \textbf{42.34} & 24.15  & 25.00 & 40.73  \\
    150M & \cmark & 15B & 2 & Avg & 11.55 & 39.84  & 34.17  & 64.31  & 52.25 & 41.04  & \textbf{24.66} & 26.60 & 40.41  \\
    \midrule
    \,\,\,85M & \cmark & - & - & - & - & 13.53 & 30.67 & 58.22 & 48.62 & 36.62  & 25.00  & 28.60  & 34.47  \\
    \midrule
    \,\,\,43M & \cmark & 15B & 2 & Step & \textbf{15.93}  & 21.02  & 29.28  & 60.01  & 48.93 & 37.92 & \textbf{23.98} & \textbf{28.00} & 35.59 \\
    \,\,\,43M & \cmark & 15B & 2 & Lower & 16.19 & 21.46  &\textbf{ 29.61}  & 59.90  & \textbf{50.67} & \textbf{38.52} & 22.95  & \textbf{28.00}  & \textbf{35.87}  \\
    \,\,\,43M & \cmark & 15B & 2 & Avg & 16.12  & \textbf{22.36}  & 29.07  & \textbf{60.17}  & 49.96 & 37.24  & 23.29  & 26.60  & 35.53  \\
    \bottomrule
    \end{tabular}
    }
    \caption{
    Comparison between initialization methods for looped layers on Pythia 410M and 160M. Uptraining was performed using the Pile dataset, which was also used for pretraining the original Pythia model. In light of the inherent randomness in few-shot accuracy, a comparison based on the perplexity\,(PPL) would provide a more stable measure of performance.\looseness=-1
    }
    \label{tab:pythia_160_410}
\end{table}

\vspace{-8pt}
\paragraph{Individual contributions of leveraging pretrained weights and recursive patterns}

To understand the performance of our Recursive Transformer, we established two non-recursive baselines: full-size model and reduced-size model. The reduced size model performance is meant to serve as a lower bound which we can use to better judge the efficacy of (1) unique looping and parameter sharing techniques that are made possible by our approach and (2) leveraging pretrained layers. To further ablate the effect of each of two components, we conducted experiments using the Pythia 410M model presented in Table~\ref{tab:individual_effect}. Intuitively, we observed significant performance gains by leveraging pretrained layers, with further improvement achieved through recursion. We believe this additional experiment provides valuable insight into the performance contributions of the two approaches proposed for constructing Recursive Transformers.\looseness=-1

\begin{table}[h]
    \small
    \centering
    \resizebox{\textwidth}{!}{
    \setlength{\tabcolsep}{4pt}
    \begin{tabular}{c|cc|cc|c|ccccccc|c}
    \toprule
     & \multicolumn{2}{c|}{\textbf{Uptrain}} & \multicolumn{2}{c|}{\textbf{Looping}} & \textbf{PPL}\,$\downarrow$  & \multicolumn{8}{c}{\textbf{Few-shot Accuracy\,$\uparrow$}} \\
    \cmidrule(l{2pt}r{2pt}){2-3} \cmidrule(l{2pt}r{2pt}){4-5}  \cmidrule(l{2pt}r{2pt}){6-6}  \cmidrule(l{2pt}r{2pt}){7-14} 
      N-emb & PT & $N_{tok}$ & Block & Init  & Pile & LD & HS & PQ & WG & ARC-e & ARC-c & OB & Avg  \\
    \midrule    
    300M & \cmark & - & - & - & - & 44.96 & 40.97 & 66.97 & 53.28 & 44.40 & 25.51 & 30.20 & 43.76 \\
    \midrule
    150M & \xmark & 15B & - & - & 14.11 & 31.48 & 29.53 & 61.37 & \textbf{52.49} & 39.14 & 22.44  & 27.00 & 37.63 \\
    150M & \xmark & 15B & 2 & - &\textbf{ 13.81} & \textbf{31.55} & \textbf{29.94} & \textbf{62.30} & 50.88 & \textbf{40.28} & \textbf{23.98} & \textbf{28.20} & \textbf{38.02} \\
    \midrule
    150M & \cmark & 15B & - & Step & 11.48 & 40.48 & 34.19 & 63.42 & 50.99 & \textbf{41.84} & 23.12 & 28.40 & 40.35 \\
    150M & \cmark & 15B & 2 & Step & \textbf{11.03} & \textbf{43.41} & \textbf{35.59} & \textbf{64.58} & \textbf{53.04} & 41.58 & \textbf{23.81} & \textbf{28.80 }& \textbf{41.54} \\
    \bottomrule
    \end{tabular}
    }
    \caption{
    Performance of recursive and baseline models with Pythia 410M to investigate the individual contributions of pretrained weights and looping strategy. Uptraining was performed using the Pile dataset~\citep{gao2020pile}, which was also used for pretraining the original Pythia model.
    }
    \label{tab:individual_effect}
\end{table}

\clearpage

\section{Expanded Results of Relaxed Recursive Transformers}
\label{app:cross_layer_lora}

\paragraph{Training perplexity changes with LoRA modules}
Figure\,\ref{fig:lora_training_loss_app} illustrates the changes in training loss after incorporating the layer-wise LoRA modules. The Average and Lower initialization methods, when coupled with our proposed SVD-based initialization of LoRA modules, demonstrated significantly enhanced benefits. In particular, the Relaxed Recursive Transformer employing the Average method consistently outperformed the others. This suggests that it is considerably easier to learn the difference between the original pretrained weights and the averaged looped weights using low-rank matrices.

\begin{figure}[h]
    \centering
    \begin{subfigure}[t]{\textwidth}
        \includegraphics[width=\textwidth]{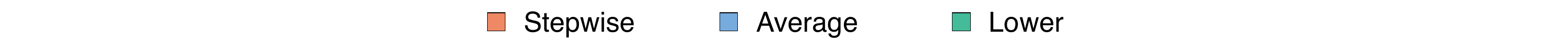}
    \end{subfigure}
    \centering
    \begin{subfigure}[t]{0.34\textwidth}
        \includegraphics[width=\textwidth]{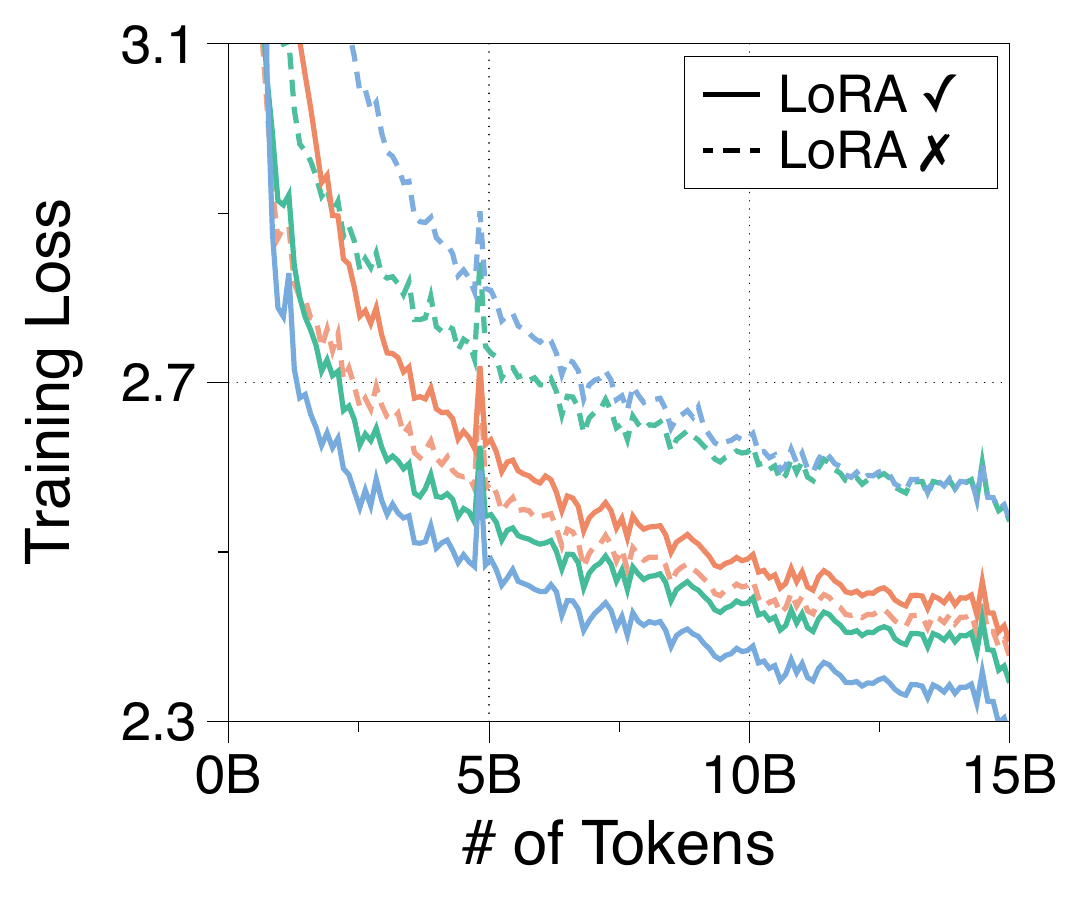}
        \subcaption{Gemma}
    \end{subfigure}
    \centering
    \begin{subfigure}[t]{0.31\textwidth}
        \includegraphics[width=\textwidth]{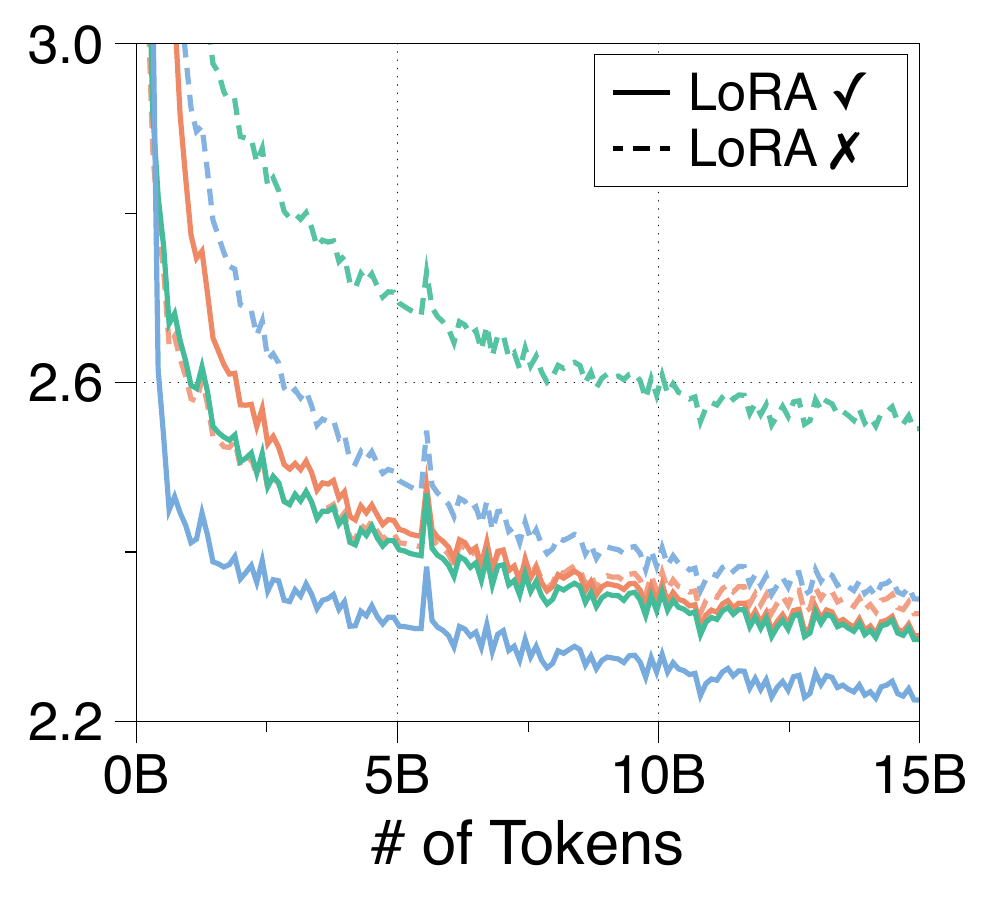}
        \subcaption{TinyLlama}
    \end{subfigure}
    \centering
    \begin{subfigure}[t]{0.31\textwidth}
        \includegraphics[width=\textwidth]{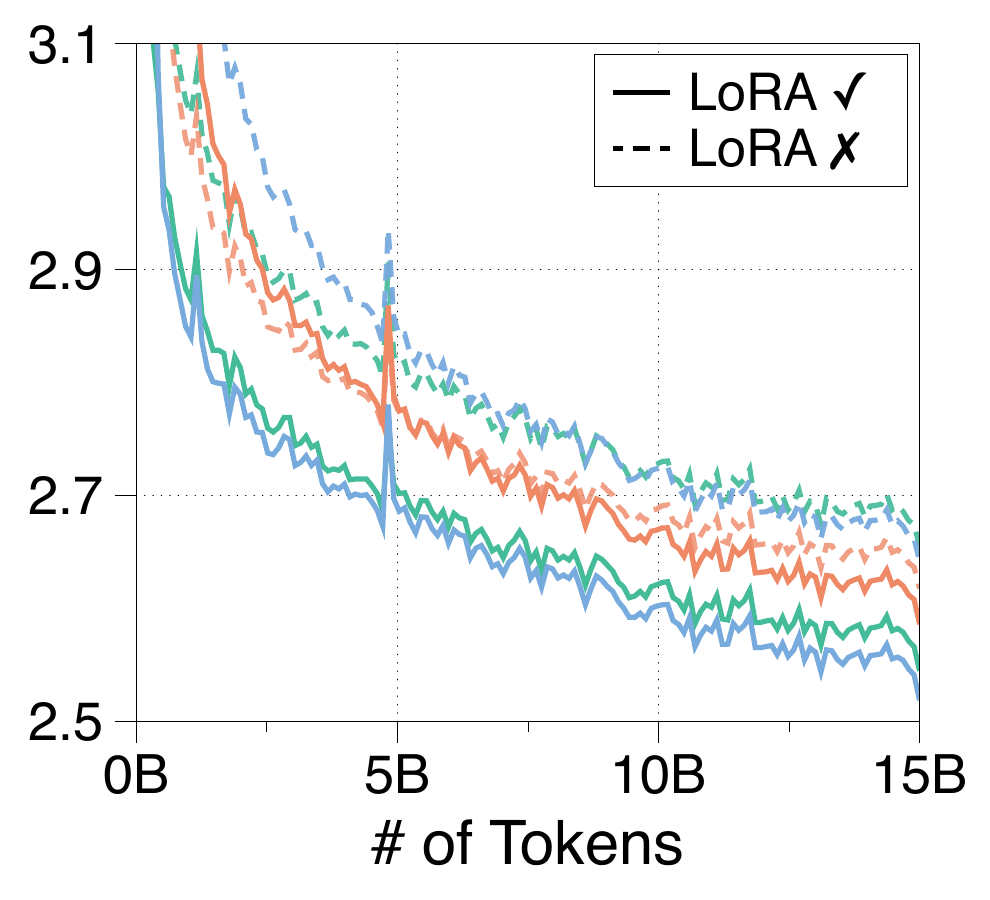}
        \subcaption{Pythia}
    \end{subfigure}
    \caption{
    Comparison of training loss for recursive and relaxed recursive models with two blocks. The LoRA rank is set to 512, and the SVD initialization method is used for LoRA modules.
    }
    \label{fig:lora_training_loss_app}
\end{figure}

\vspace{-10pt}
\paragraph{Comparison between SVD and zero initialization}

The utilization of layer-wise LoRA modules enhances model capacity by introducing additional parameters and relaxation, thereby potentially improving performance. 
As depicted in Figure\,\ref{fig:zero_svd_init_app}, SVD initialization significantly amplified these performance gains compared to standard zero initialization. However, an interesting exception was observed with the Stepwise method, where the SVD initialized LoRA module surprisingly led to a performance degradation. 
This appears to be attributed to LoRA ranks being insufficient to adequately approximate the low-rank deltas across layers, resulting in initialization at a sub-optimal point.

\begin{figure}[h]
    \centering
    \begin{subfigure}[t]{0.34\textwidth}
        \includegraphics[width=\textwidth]{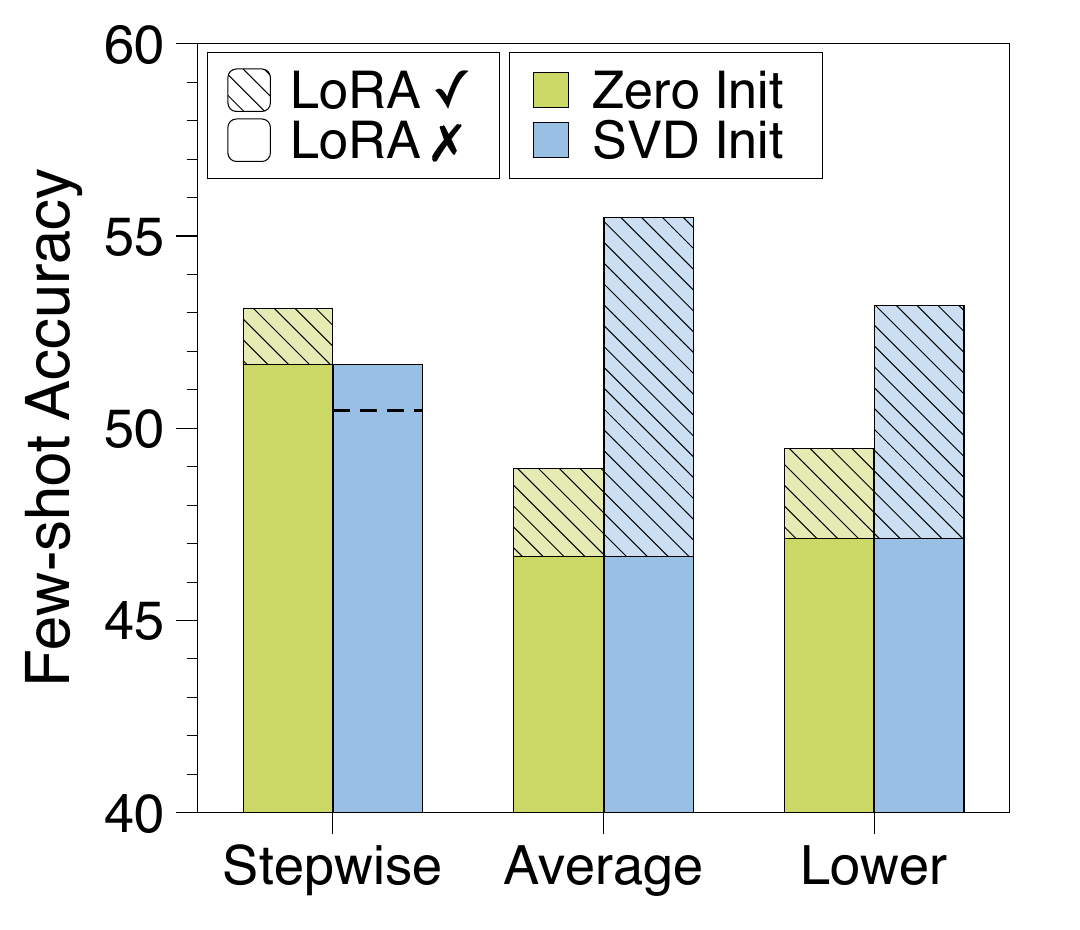}
        \subcaption{Gemma}
    \end{subfigure}
    \centering
    \begin{subfigure}[t]{0.31\textwidth}
        \includegraphics[width=\textwidth]{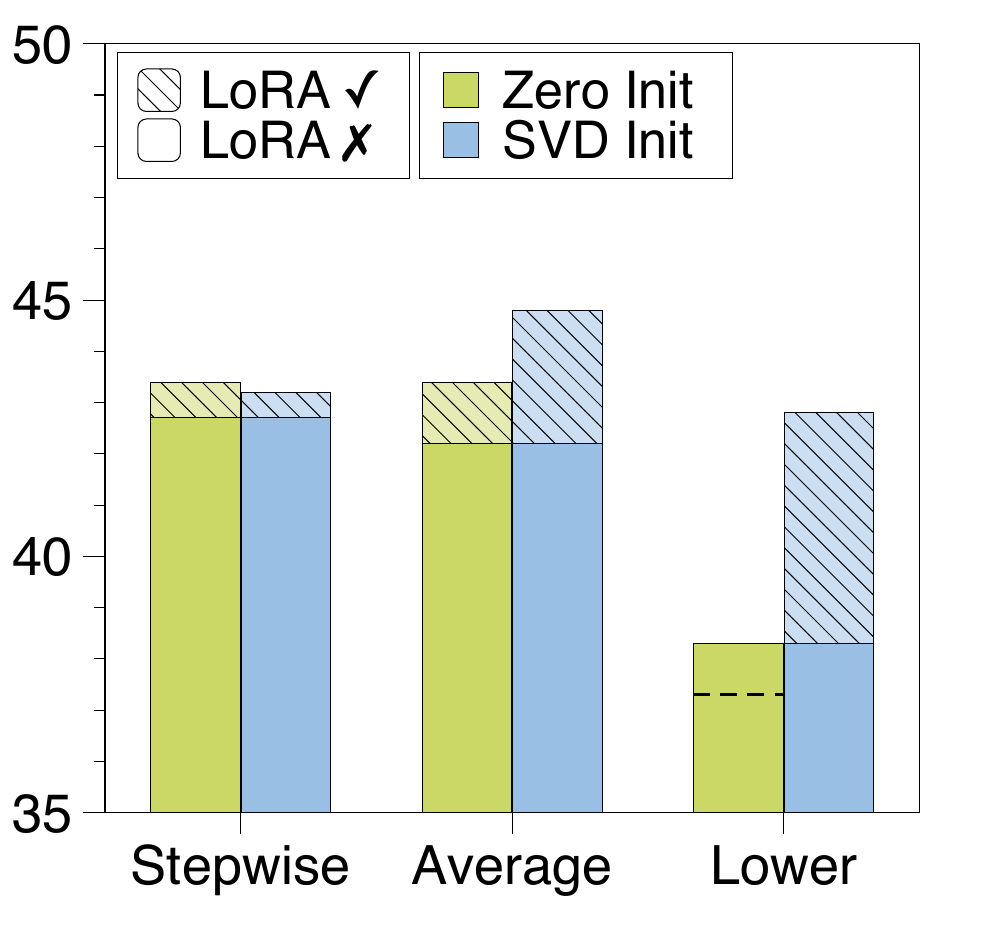}
        \subcaption{TinyLlama}
    \end{subfigure}
    \centering
    \begin{subfigure}[t]{0.31\textwidth}
        \includegraphics[width=\textwidth]{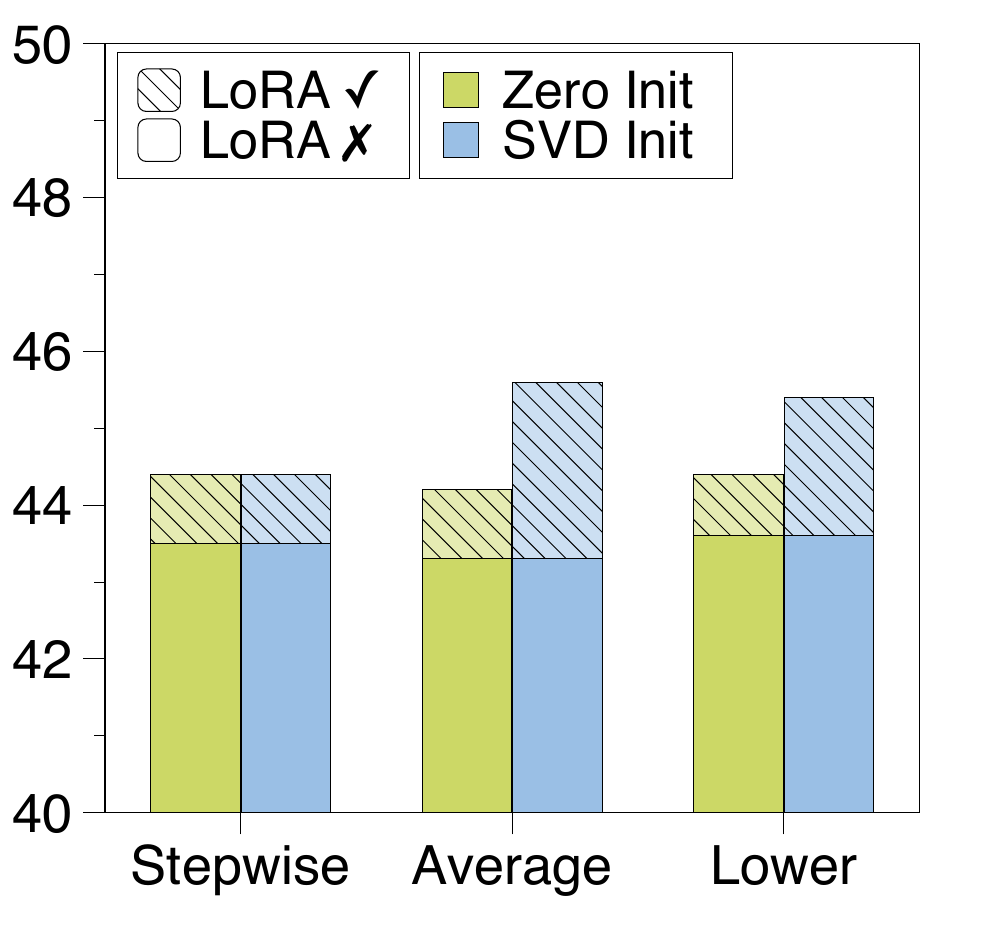}
        \subcaption{Pythia}
    \end{subfigure}
    \caption{Comparison of average few-shot accuracy between zero and SVD initialization methods across three models. Performance gains due to LoRA relaxation are indicated by hatched bars, while cases where performance is lower than the recursive counterpart (without LoRA modules) are represented by dotted lines.
    }
    \label{fig:zero_svd_init_app}
\end{figure}

\clearpage

\paragraph{Ablation study on the LoRA rank values} 
Our proposed SVD initialization ensures that the Relaxed Recursive Transformer can function as an interpolation between vanilla and Recursive Transformers. The approximation accuracy of SVD is directly influenced by the LoRA rank value; a higher rank leads to improved restoration of the pretrained model weights. In Figure\,\ref{fig:rank_fewshot_app}, we present a summary of the performance changes observed in the relaxed models by varying the LoRA ranks. As expected, for the Average and Lower looping initialization methods, a larger rank value results in enhanced performance. The Stepwise method, consistent with previous experimental findings, exhibited a U-shaped trend: with extremely low or high ranks, a clear performance increase results. However, with mid-range values, the approximation becomes less accurate, leading to a further performance decrease.

\begin{figure}[ht!]
    \centering
    \begin{subfigure}[t]{\textwidth}
        \includegraphics[width=\textwidth]{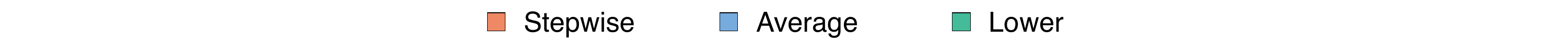}
    \end{subfigure}
    \centering
    \begin{subfigure}[t]{0.34\textwidth}
        \includegraphics[width=\textwidth]{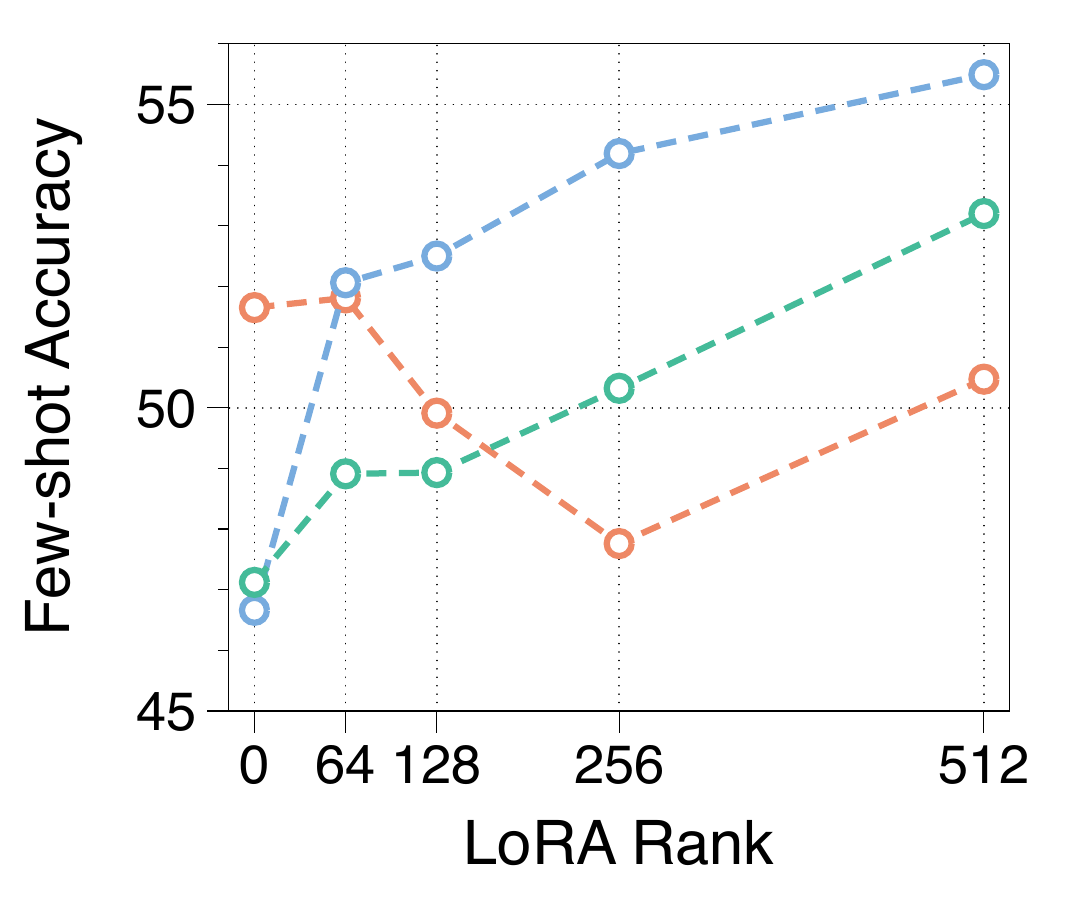}
        \subcaption{Gemma}
    \end{subfigure}
    \centering
    \begin{subfigure}[t]{0.31\textwidth}
        \includegraphics[width=\textwidth]{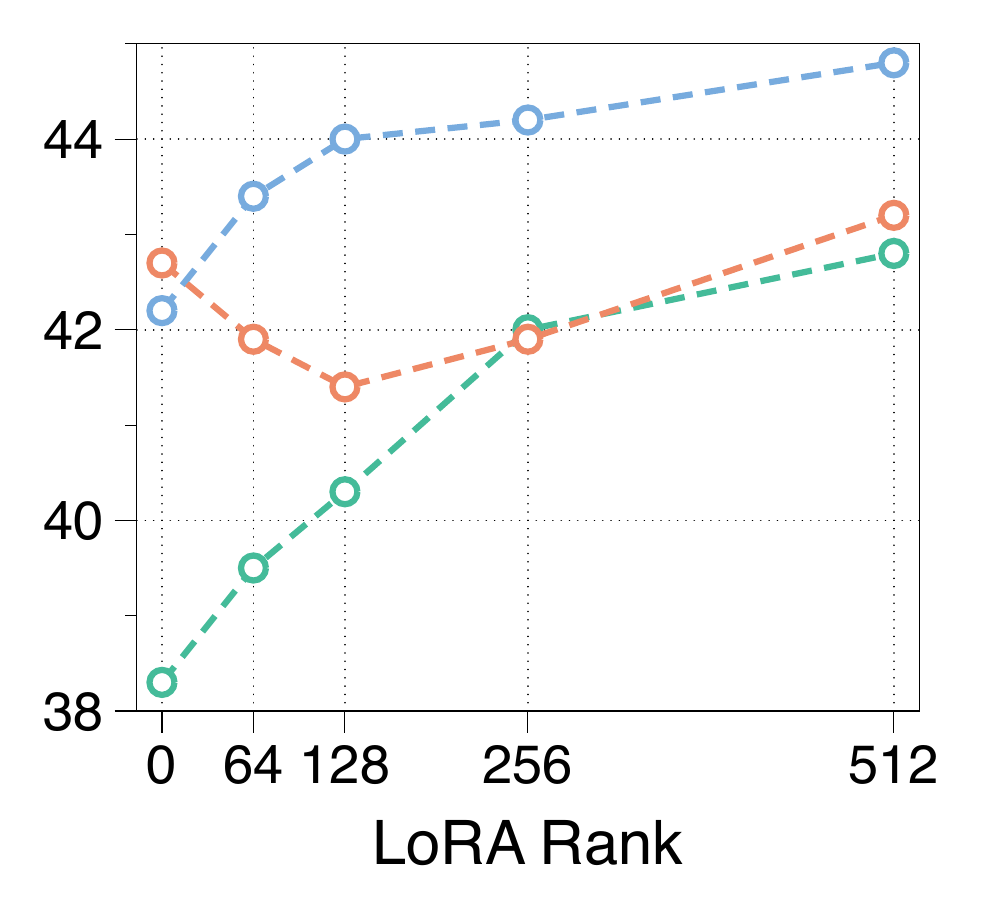}
        \subcaption{TinyLlama}
    \end{subfigure}
    \centering
    \begin{subfigure}[t]{0.31\textwidth}
        \includegraphics[width=\textwidth]{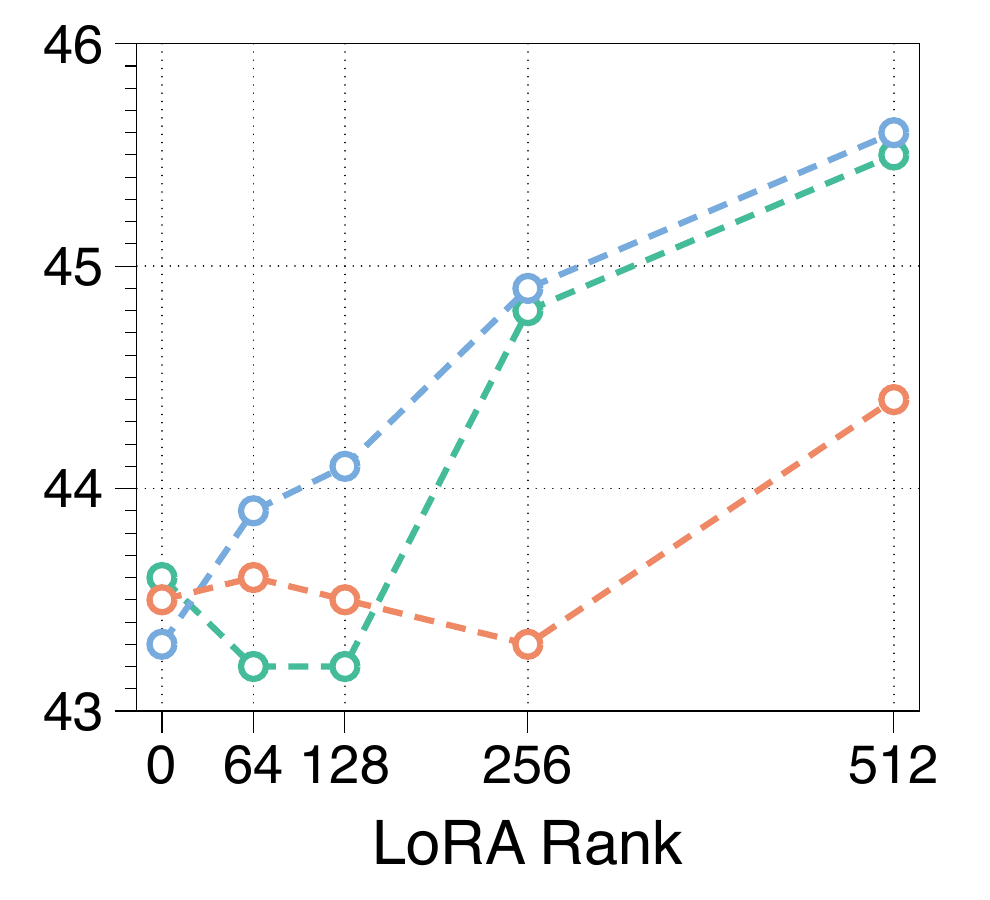}
        \subcaption{Pythia}
    \end{subfigure}
    \caption{Performance comparison with varying LoRA ranks under different initialization methods for looped layers. All LoRA weights are initialized using our proposed SVD initialization method.}
    \label{fig:rank_fewshot_app}
\end{figure}

We further experimented with assigning different ranks to LoRA modules associated with each linear weights. Given the computational overhead inherent to LoRA modules, allocating different ranks to each module can offer a better balance between performance and computational efficiency. The experimental results in Table\,\ref{tab:lora_rank_ablation_app} reveal a strong correlation between performance and overall model sizes. 
Due to the substantial hidden dimension of the linear weights within the FFN layer, reducing its rank led to the most significant performance drop. Conversely, the relatively smaller size of other attention weights resulted in less performance drop. 
It is intriguing that even minimal relaxation of key-value weights (achieved through small ranks) yielded comparable performance, despite the inherent strong sharing of key-value caches between attention heads in the Multi-Query attention structure~\citep{DBLP:conf/emnlp/AinslieLJZLS23}.

\clearpage

\begin{table}[ht]
    \small
    \centering
    \resizebox{\textwidth}{!}{
    \setlength{\tabcolsep}{3pt}
    \begin{tabular}{c|cc|cc|ccccc|rrr|ccccccc|cc}
    \toprule
     & \multicolumn{2}{c|}{\textbf{Uptrain}} & \multicolumn{2}{c|}{\textbf{Looping}} &  \multicolumn{5}{c|}{\textbf{LoRA}} & \multicolumn{3}{c|}{\textbf{Perplexity\,$\downarrow$}} & \multicolumn{8}{c}{\textbf{Few-shot Accuracy\,$\uparrow$}} \\
    \cmidrule(l{2pt}r{2pt}){2-3} \cmidrule(l{2pt}r{2pt}){4-5}  \cmidrule(l{2pt}r{2pt}){6-10} \cmidrule(l{2pt}r{2pt}){11-13}  \cmidrule(l{2pt}r{2pt}){14-21} 
    N-emb & PT & $N_{tok}$ & Block & Init  & Q & KV & Out & FFN & Init & SlimP & RedP & PG19 & LD & HS & PQ & WG & ARC-e & ARC-c & OB & Avg  \\
    \midrule
    1.99B &  \cmark & 15B & - & - & - & - & - & - & - & 10.76 & 8.47 & 13.08 & 63.5 & 68.5 & 77.0 & 63.5 & 67.6 & 38.1 & {42.6} & 60.1   \\
    0.99B &  \xmark & 15B & - & - & - & - & -  & - & - & 22.63 & 20.03 & 32.60 & 28.9 & 31.6 & 63.1 & 52.3 & 41.2 & 22.5 & 27.8 &  38.2  \\
    \midrule
    1.30B & \cmark & 15B & 2 & Avg  & 256 & 256 & 256 & 256 & SVD &  12.10 &  9.71 &  14.89 &  58.2 &  60.7 &  73.7 &  59.0 &  57.6 &  32.1 &  {38.0} &  54.2 \\
    1.28B & \cmark & 15B & 2 & Avg  & 128 & 256 & 128 & 256 & SVD &  12.27 &  9.81 &  15.10 &  57.4 &  60.2 &  72.5 &  58.9 &  58.1 &  32.6 &  37.8 &  53.9 \\
     1.29B & \cmark & 15B & 2 & Avg  & 256 & 128 & 256 & 256 & SVD &  12.33 &  9.90 &  15.25 &  58.5 &  59.7 &  73.3 &  58.3 &  56.6 &  32.0 &  40.0 &  54.1 \\
     1.18B & \cmark & 15B & 2 & Avg  & 256 & 256 & 256 & 128 & SVD &  12.56 &  10.12 &  15.59 &  57.0 &  58.7 &  73.0 &  57.4 &  57.0 &  31.6 &  38.2 &  53.3 \\
     1.27B & \cmark & 15B & 2 & Avg  & 128 & 128 & 128 & 256 & SVD &  12.36 &  9.92 &  15.31 &  57.2 &  59.2 &  73.1 &  57.3 &  58.0 &  32.2 &  38.6 &  53.7 \\
    \midrule
     1.15B & \cmark & 15B & 2 & Avg  & 128 & 128 & 128 & 128 & SVD &  12.52 &  10.07 &  15.51 &  56.1 &  58.2 &  72.3 &  55.8 &  57.1 &  30.7 &  37.2 &  52.5 \\
      1.14B & \cmark & 15B & 2 & Avg  & 64 & 128 & 64 & 128 & SVD &  12.61 &  10.14 &  15.69 &  55.0 &  57.8 &  73.0 &  57.5 &  56.7 &  30.9 &  38.8 &  52.8 \\
     1.14B & \cmark & 15B & 2 & Avg  & 128 & 64 & 128 &  128 & SVD &  12.72 &  10.18 &  15.76 &  55.5 &  57.7 &  72.7 &  57.0 &  56.9 &  30.1 &  38.2 &  52.6 \\
     1.08B& \cmark & 15B & 2 & Avg  & 128 & 128 & 128 & 64 & SVD &  12.80 &  10.33 &  15.97 &  55.3 &  56.7 &  72.9 &  57.7 &  55.0 &  29.6 &  36.0 &  51.9 \\
     1.13B & \cmark & 15B & 2 & Avg  & 64 & 64 & 64 & 128 & SVD &  12.77 &  10.29 &  15.95 &  55.2 &  57.4 &  73.0 &  56.7 &  56.5 &  30.5 &  37.2 &  52.3 \\
    \bottomrule
    \end{tabular}
    }
    \caption{
    Evaluation results of relaxed recursive Gemma models with varying LoRA ranks for different Transformer components. We adjusted the LoRA ranks attached to query, key-value, out, and FFN linear weights. Non-embedding parameter sizes include both the base layer parameters and the attached LoRA weights.
    }
    \label{tab:lora_rank_ablation_app}
\end{table}

\paragraph{Overall performance comparison of Relaxed Recursive Transformers}
A comprehensive performance comparison is presented in Table\,\ref{tab:lora_total_app}. This encompasses an evaluation of the performance across three models and various looping initialization methods, considering the degree of relaxation induced by the layer-wise LoRA module. The configuration utilizing the Average method for looped layer initialization, paired with SVD initialization for the LoRA module, consistently outperformed all other baselines. Furthermore, a clear trend of improved performance was observed with increasing rank.\looseness=-1

\begin{table}[ht!]
    \small
    \centering
    \resizebox{\textwidth}{!}{
    \setlength{\tabcolsep}{3pt}
    \begin{tabular}{l|c|cc|cc|cc|rrr|ccccccc|cc}
    \toprule
     & & \multicolumn{2}{c|}{\textbf{Uptrain}} & \multicolumn{2}{c|}{\textbf{Looping}} &  \multicolumn{2}{c|}{\textbf{LoRA}} & \multicolumn{3}{c|}{\textbf{Perplexity\,$\downarrow$}} & \multicolumn{9}{c}{\textbf{Few-shot Accuracy\,$\uparrow$}} \\
    \cmidrule(l{2pt}r{2pt}){3-4} \cmidrule(l{2pt}r{2pt}){5-6}  \cmidrule(l{2pt}r{2pt}){7-8} \cmidrule(l{2pt}r{2pt}){9-11}  \cmidrule(l{2pt}r{2pt}){12-20} 
     \textbf{Models} & N-emb & PT & $N_{tok}$ & Block & Init  & Rank & Init & SlimP & RedP & PG19 & LD & HS & PQ & WG & ARC-e & ARC-c & OB & Avg & $\Delta$  \\
    \midrule
    & 1.99B &  \cmark & 15B & - & - & - & - & 10.76 & 8.47 & 13.08 & 63.5 & 68.5 & 77.0 & 63.5 & 67.6 & 38.1 & {42.6} & 60.1  & - \\
     & 0.99B &  \xmark & 15B & - & - & - & - & 22.63 & 20.03 & 32.60 & 28.9 & 31.6 & 63.1 & 52.3 & 41.2 & 22.5 & 27.8 &  38.2 & -   \\  
     \cmidrule(l{2pt}r{2pt}){2-20} 
     & 0.99B & \cmark & 15B & 2 & Step & - & - &  {12.85} & {10.29} & {16.21} & {53.0} & {57.3} & {73.2} & {56.2} & {56.1} & {29.2} & {36.6} & {51.7} & - \\
    & 1.07B & \cmark & 15B & 2 & Step  & 64 & SVD &  12.76 &  10.21 &  15.99 &  52.1 &  57.2 &  73.0 &  57.8 &  56.9 &  28.9 &  36.8 &  51.8 & \textcolor{custom_green}{\textbf{+0.1}} \\
    & 1.15B & \cmark & 15B & 2 & Step  & 128 & SVD &  13.44 &  10.80 &  16.98 &  50.5 &  53.0 &  71.5 &  54.4 &  55.9 &  29.3 &  34.8 &  49.9 & \textcolor{custom_red}{\textbf{--\,1.8}} \\
    & 1.30B & \cmark & 15B & 2 & Step  & 256 & SVD &  14.02 &  11.44 &  18.09 &  46.1 &  49.1 &  71.8 &  53.2 &  52.8 &  27.8 &  33.4 &  47.8 & \textcolor{custom_red}{\textbf{--\,3.9}} \\
    & 1.60B & \cmark & 15B & 2 & Step  & 512 & SVD &  13.13 &  10.66 &  16.63 &  53.0 &  54.3 &  72.1 &  54.9 &  54.8 &  28.8 &  35.4 &  50.5 & \textcolor{custom_red}{\textbf{--\,1.2}} \\
     & 1.60B & \cmark & 15B & 2 & Step  & 512 & Zero &  12.46 &  9.97 &  15.58 &  54.9 &  58.8 &  \textbf{74.0} &  58.1 &  58.8 &  30.6 &  36.6 &  53.1 & \textcolor{custom_green}{\textbf{+1.4}} \\
     \cmidrule(l{2pt}r{2pt}){2-20} 
     Gemma & 0.99B & \cmark & 15B & 2 & Avg & - & - & 15.15 & 12.57 & 19.86 & 43.6 & 47.4 & 70.4 & 52.6 & 50.5 & 27.8 & 34.4 & 46.7 & -  \\
     \rowcolor[gray]{0.9} 
    \cellcolor{white!20} & 1.07B & \cmark & 15B & 2 & Avg  & 64 & SVD &  12.83 &  10.35 &  16.02 &  55.9 &  56.8 &  72.5 &  56.8 &  55.7 &  30.6 &  36.2 &  52.1 & \textcolor{custom_green}{\textbf{+5.4}} \\
     \rowcolor[gray]{0.9} 
    \cellcolor{white!20} & 1.15B & \cmark & 15B & 2 & Avg  & 128 & SVD &  12.52 &  10.07 &  15.51 &  56.1 &  58.2 &  72.3 &  55.8 &  57.1 &  30.7 &  37.2 &  52.5 & \textcolor{custom_green}{\textbf{+5.8}} \\
     \rowcolor[gray]{0.9} 
    \cellcolor{white!20} & 1.30B & \cmark & 15B & 2 & Avg  & 256 & SVD &  12.10 &  9.71 &  14.89 &  58.2 &  60.7 &  73.7 &  59.0 &  57.6 &  32.1 &  \textbf{38.0} &  54.2 & \textcolor{custom_green}{\textbf{+7.5}} \\
     \rowcolor[gray]{0.9} 
    \cellcolor{white!20} & 1.60B & \cmark & 15B & 2 & Avg  & 512 & SVD &  \textbf{11.83} &  \textbf{9.46} &  \textbf{14.57} &  \textbf{59.3} &  \textbf{62.8} &  \textbf{74.0 }&  \textbf{61.6} &  \textbf{60.1} &  \textbf{32.9} &  37.6 &  \textbf{55.5} & \textcolor{custom_green}{\textbf{+8.8}} \\
     & 1.60B & \cmark & 15B & 2 & Avg  & 512 & Zero &  13.78 &  11.31 &  17.71 &  49.8 &  52.4 &  71.7 &  53.3 &  51.2 &  29.4 &  35.0 &  49.0 & \textcolor{custom_green}{\textbf{+2.3}} \\
     \cmidrule(l{2pt}r{2pt}){2-20} 
     & 0.99B & \cmark & 15B & 2 & Lower & - & - & 15.03 & 12.46 & 19.63 & 42.5 & 48.0 & 71.0 & 54.6 & 52.2 & 27.7 & 33.8 & 47.1 & -  \\
    & 1.07B & \cmark & 15B & 2 & Lower  & 64 & SVD &  14.21 &  11.77 &  18.40 &  47.5 &  50.5 &  70.9 &  54.2 &  54.1 &  29.2 &  36.0 &  48.9 & \textcolor{custom_green}{\textbf{+1.8}} \\
    & 1.15B & \cmark & 15B & 2 & Lower  & 128 & SVD &  14.23 &  11.83 &  18.49 &  48.0 &  50.5 &  72.0 &  56.8 &  54.4 &  27.5 &  33.4 &  48.9 & \textcolor{custom_green}{\textbf{+1.8}} \\
    & 1.30B & \cmark & 15B & 2 & Lower  & 256 & SVD &  13.51 &  11.06 &  17.30 &  53.1 &  53.7 &  71.8 &  57.4 &  52.5 &  28.7 &  35.2 &  50.3 & \textcolor{custom_green}{\textbf{+3.2}} \\
    & 1.60B & \cmark & 15B & 2 & Lower  & 512 & SVD &  12.54 &  10.22 &  15.90 &  57.1 &  58.2 &  73.7 &  58.6 &  57.6 &  31.5 &  35.6 &  53.2 & \textcolor{custom_green}{\textbf{+6.1}} \\
     & 1.60B & \cmark & 15B & 2 & Lower  & 512 & Zero &  13.95 &  11.59 &  18.02 &  48.4 &  52.1 &  71.9 &  55.7 &  54.9 &  28.8 &  34.6 &  49.5 & \textcolor{custom_green}{\textbf{+2.4}} \\
     \midrule
    & 0.97B &  \cmark & - & - & - & - & - & 12.26 & 9.37 & 11.94 & 43.3 & 42.2 & 66.8 & 53.4 & 44.7 & 23.2 & 29.2 & 43.3  & - \\
     & 0.48B &  \xmark & 15B & - & - & - & - & 16.61 & 15.66 & 20.27 & 22.3 & 30.0 & 60.9 & 50.6 & 37.0 & 23.0 & 28.0 & 36.0 & -  \\ 
     \cmidrule(l{2pt}r{2pt}){2-20} 
     & 0.48B & \cmark & 15B & 2 & Step & - & - &  {11.61} & {9.89}  & {13.00} &{ 39.6} &{ 39.8} & {66.5} & {52.9} & {44.3} & {24.9} & {30.6} & {42.7} & -   \\
    & 0.53B & \cmark & 15B & 2 & Step  & 64 & SVD &  12.10 &  10.40 &  13.75 &  38.9 &  38.3 &  65.2 &  51.5 &  42.0 &  \textbf{26.0} &  31.0 &  41.9 & \textcolor{custom_red}{\textbf{--\,0.8}} \\
    & 0.58B & \cmark & 15B & 2 & Step  & 128 & SVD &  12.41 &  10.72 &  14.10 &  36.8 &  37.4 &  64.7 &  53.4 &  42.2 &  24.7 &  30.4 &  41.4 & \textcolor{custom_red}{\textbf{--\,1.3}} \\
    & 0.68B & \cmark & 15B & 2 & Step  & 256 & SVD &  11.96 &  10.35 &  13.48 &  38.9 &  38.3 &  65.8 &  51.9 &  43.1 &  25.4 &  29.8 &  41.9 & \textcolor{custom_red}{\textbf{--\,0.8}} \\
    & 0.86B & \cmark & 15B & 2 & Step  & 512 & SVD &  11.33 &  9.79 &  12.61 &  42.2 &  40.9 &  67.7 &  51.1 &  45.0 &  25.3 &  30.2 &  43.2 & \textcolor{custom_green}{\textbf{+0.5}} \\
     & 0.86B & \cmark & 15B & 2 & Step  & 512 & Zero &  11.24 &  9.60 &  12.56 &  42.0 &  41.0 &  67.4 &  52.2 &  44.5 &  25.9 &  \textbf{31.2} &  43.4 & \textcolor{custom_green}{\textbf{+0.7}} \\
     \cmidrule(l{2pt}r{2pt}){2-20} 
     TinyLlama & 0.48B & \cmark & 15B & 2 & Avg & - & - & 11.86 & 10.29  & 13.42 & 38.6 & 39.4 & 66.1 & 52.8 & 42.7 & 25.4 & {30.6} & 42.2 & -  \\
     \rowcolor[gray]{0.9} 
    \cellcolor{white!20} & 0.53B & \cmark & 15B & 2 & Avg  & 64 & SVD &  11.22 &  9.66 &  12.51 &  41.8 &  41.6 &  67.0 &  53.3 &  43.9 &  24.7 &  \textbf{31.2} &  43.4 & \textcolor{custom_green}{\textbf{+1.2}} \\
     \rowcolor[gray]{0.9} 
    \cellcolor{white!20} & 0.58B & \cmark & 15B & 2 & Avg  & 128 & SVD &  10.99 &  9.45 &  12.21 &  43.2 &  42.1 &  \textbf{68.3} &  53.2 &  44.8 &  25.9 &  30.4 &  44.0 & \textcolor{custom_green}{\textbf{+1.8}} \\
     \rowcolor[gray]{0.9} 
    \cellcolor{white!20} & 0.68B & \cmark & 15B & 2 & Avg  & 256 & SVD &  10.71 &  9.18 &  11.82 &  44.1 &  43.2 &  68.1 &  \textbf{53.5} &  44.4 &  25.7 &  30.4 &  44.2 & \textcolor{custom_green}{\textbf{+2.0}} \\
     \rowcolor[gray]{0.9} 
    \cellcolor{white!20} & 0.86B & \cmark & 15B & 2 & Avg  & 512 & SVD &  \textbf{10.46} &  \textbf{8.92} & \textbf{ 11.50} &  \textbf{46.0} &  \textbf{44.1} &  68.2 &  53.0 &  \textbf{45.8} &  25.1 &  \textbf{31.2} &  \textbf{44.8} & \textcolor{custom_green}{\textbf{+2.6}} \\
     & 0.86B & \cmark & 15B & 2 & Avg  & 512 & Zero &  11.28 &  9.75 &  12.69 &  41.5 &  41.0 &  66.8 &  53.2 &  44.8 &  25.5 &  \textbf{31.2} &  43.4 & \textcolor{custom_green}{\textbf{+1.2}} \\
     \cmidrule(l{2pt}r{2pt}){2-20} 
     & 0.48B & \cmark & 15B & 2 & Lower & - & - & 14.67 & 12.67  & 16.68 & 31.9 & 32.3 & 62.6 & 52.0 & 39.1 & 22.1 & 27.8 & 38.3 & -  \\
    & 0.53B & \cmark & 15B & 2 & Lower  & 64 & SVD &  13.68 &  11.77 &  15.48 &  35.5 &  34.0 &  63.8 &  51.0 &  40.0 &  24.6 &  28.0 &  39.5 & \textcolor{custom_green}{\textbf{+1.2}} \\
    & 0.58B & \cmark & 15B & 2 & Lower  & 128 & SVD &  13.00 &  11.14 &  14.61 &  37.6 &  35.4 &  65.3 &  51.5 &  40.4 &  24.5 &  27.6 &  40.3 & \textcolor{custom_green}{\textbf{+2.0}} \\
    & 0.68B & \cmark & 15B & 2 & Lower  & 256 & SVD &  12.14 &  10.39 &  13.59 &  40.0 &  37.7 &  66.1 &  52.6 &  42.5 &  24.8 &  30.0 &  42.0 & \textcolor{custom_green}{\textbf{+3.7}} \\
    & 0.86B & \cmark & 15B & 2 & Lower  & 512 & SVD &  11.31 &  9.61 &  12.49 &  43.2 &  40.5 &  66.0 &  50.8 &  43.9 &  24.8 &  30.0 &  42.8 & \textcolor{custom_green}{\textbf{+4.5}} \\
     & 0.86B & \cmark & 15B & 2 & Lower  & 512 & Zero &  14.56 &  12.69 &  16.57 &  21.2 &  32.9 &  63.9 &  52.6 &  39.5 &  22.9 &  27.8 &  37.3 & \textcolor{custom_red}{\textbf{--\,1.0}} \\
     \midrule
    & 0.81B &  \cmark & 15B & - & - & - & - & 13.46 & 9.95 & 13.38 &  55.0 & 49.0 & 71.0 & 53.6 & 51.8 & {28.2} & 32.8 & {48.8}  & - \\
     & 0.40B &  \xmark & 15B & - & - & - & - & 25.69 & 20.00 & 32.08 & 24.3 & 30.0 & 61.9 & 50.7 & 38.3 & 22.3 & 26.0  & 36.2 & -  \\ 
     \cmidrule(l{2pt}r{2pt}){2-20} 
     & 0.40B & \cmark & 15B & 2 & Step & - & - & {16.38} & {12.37} & {17.74} & 43.4 & {40.5} & 67.4 & 50.8 & {46.3} & 25.7 & 30.0 & 43.5 & -   \\
    & 0.44B & \cmark & 15B & 2 & Step  & 64 & SVD &  16.44 &  12.44 &  17.89 &  43.7 &  40.4 &  66.5 &  52.9 &  46.5 &  26.2 &  28.8 &  43.6 & \textcolor{custom_green}{\textbf{+0.1}} \\
    & 0.48B & \cmark & 15B & 2 & Step  & 128 & SVD &  16.63 &  12.61 &  18.35 &  42.4 &  39.3 &  68.0 &  51.5 &  46.3 &  26.7 &  30.6 &  43.5 & \textcolor{gray}{\textbf{+0.0}} \\
    & 0.55B & \cmark & 15B & 2 & Step  & 256 & SVD &  16.54 &  12.61 &  18.39 &  42.8 &  39.1 &  67.2 &  53.7 &  46.4 &  25.9 &  27.8 &  43.3 & \textcolor{custom_red}{\textbf{--\,0.2}} \\
    & 0.70B & \cmark & 15B & 2 & Step  & 512 & SVD &  15.68 &  11.88 &  17.25 &  45.4 &  41.3 &  \textbf{68.5} &  52.6 &  46.7 &  25.4 &  31.2 &  44.4 & \textcolor{custom_green}{\textbf{+0.9}} \\
     & 0.70B & \cmark & 15B & 2 & Step  & 512 & Zero &  15.88 &  12.01 &  17.16 &  45.5 &  41.8 &  68.0 &  52.6 &  46.6 &  \textbf{26.3} &  30.0 &  44.4 & \textcolor{custom_green}{\textbf{+0.9}} \\
     \cmidrule(l{2pt}r{2pt}){2-20} 
     Pythia & 0.40B & \cmark & 15B & 2 & Avg & - & - & 16.76 & 12.76 & 18.63 & 43.6 & 39.1 & {68.2} & 51.9 & 45.4 & 25.1 & 29.8 & 43.3 & -  \\
     \rowcolor[gray]{0.9} 
    \cellcolor{white!20} & 0.44B & \cmark & 15B & 2 & Avg  & 64 & SVD &  16.03 &  12.19 &  17.59 &  45.8 &  40.9 &  67.3 &  50.0 &  45.8 &  25.5 &  31.8 &  43.9 & \textcolor{custom_green}{\textbf{+0.6}} \\
     \rowcolor[gray]{0.9} 
    \cellcolor{white!20} & 0.48B & \cmark & 15B & 2 & Avg  & 128 & SVD &  15.67 &  11.93 &  17.10 &  46.9 &  41.9 &  67.4 &  51.2 &  45.4 &  24.8 &  31.2 &  44.1 & \textcolor{custom_green}{\textbf{+0.8}} \\
     \rowcolor[gray]{0.9} 
    \cellcolor{white!20} & 0.55B & \cmark & 15B & 2 & Avg  & 256 & SVD &  15.22 &  11.54 &  16.47 &  48.5 &  43.3 &  67.2 &  51.4 &  46.7 &  25.5 &  32.0 &  44.9 & \textcolor{custom_green}{\textbf{+1.6}} \\
     \rowcolor[gray]{0.9} 
    \cellcolor{white!20}    & 0.70B & \cmark & 15B & 2 & Avg  & 512 & SVD &  \textbf{14.70} &  \textbf{11.07} &  \textbf{15.71} &  \textbf{50.2} &  \textbf{44.7} &  68.2 &  51.6 &  \textbf{47.6} &  25.4 &  31.2 &  \textbf{45.6} & \textcolor{custom_green}{\textbf{+2.3}} \\
     & 0.70B & \cmark & 15B & 2 & Avg  & 512 & Zero &  15.97 &  12.14 &  17.65 &  45.7 &  41.5 &  68.1 &  51.7 &  46.5 &  25.7 &  30.0 &  44.2 & \textcolor{custom_green}{\textbf{+0.9}} \\
     \cmidrule(l{2pt}r{2pt}){2-20} 
     & 0.40B & \cmark & 15B & 2 & Lower & - & - & 17.04 & 12.62 & 18.44 & {43.9} & 39.2 & 66.3 & {53.4} & 45.4 & {25.8} & {31.2} & {43.6} & -  \\
    & 0.44B & \cmark & 15B & 2 & Lower  & 64 & SVD &  17.03 &  12.78 &  18.73 &  44.1 &  38.3 &  66.9 &  51.9 &  45.4 &  24.7 &  30.8 &  43.2 & \textcolor{custom_red}{\textbf{--\,0.4}} \\
    & 0.48B & \cmark & 15B & 2 & Lower  & 128 & SVD &  16.63 &  12.49 &  18.17 &  45.2 &  39.2 &  66.8 &  51.0 &  45.6 &  24.9 &  29.6 &  43.2 & \textcolor{custom_red}{\textbf{--\,0.4}} \\
    & 0.55B & \cmark & 15B & 2 & Lower  & 256 & SVD &  15.93 &  11.99 &  17.30 &  47.6 &  41.4 &  68.3 &  53.2 &  46.0 &  25.8 &  31.0 &  44.8 & \textcolor{custom_green}{\textbf{+1.2}} \\
    & 0.70B & \cmark & 15B & 2 & Lower  & 512 & SVD &  15.11 &  11.34 &  16.07 &  \textbf{50.2} &  43.5 &  67.8 &  51.8 &  47.2 &  25.2 &  32.0 &  45.4 & \textcolor{custom_green}{\textbf{+1.8}} \\
     & 0.70B & \cmark & 15B & 2 & Lower  & 512 & Zero &  16.45 &  12.25 &  17.76 &  45.2 &  40.4 &  66.4 &  \textbf{54.5} &  45.8 &  25.9 &  \textbf{32.6} &  44.4 & \textcolor{custom_green}{\textbf{+0.8}} \\
    \bottomrule
    \end{tabular}
    }
    \caption{Overall evaluation results of Relaxed Recursive Transformers. Delta\,($\Delta$) represent the accuracy differences between relaxed and non-relaxed models using the same looping initialization. 
    }
    \label{tab:lora_total_app}
\end{table}

\clearpage
\section{Alternative Approaches to Relaxation of Parameter Sharing}
\label{app:prefix_tuning_relaxation}

To mitigate the restrictive weight-tying inherent in parameter sharing, we employed LoRA modules as discussed in \S\ref{sec:methods:relaxed}, similar to prior work~\citep{DBLP:conf/emnlp/GeCW22, shim2024leveraging}. However, efficiently computing multiple LoRA modules requires specialized kernels and sequential computations of the base layers and LoRA modules, incurring computational overhead. 
Consequently, we explored layer-specific prompts~\citep{DBLP:journals/corr/abs-2110-07602} as an alternative. This approach integrates prompts specific to each layer as prefix tokens, generating layer-wise key and value states for self-attention, and is significantly more amenable to parallel computation.

Table\,\ref{tab:prefix_tuning_app} summarizes performance of the prefix tuning method. While offering computational advantages, its reliance on small, learnable prompts resulted in limited performance gains. Additionally, without leveraging the original pretrained weights, performance was significantly lower (52.1\% vs. 47.6\% with the Average method in 1.07B model size). Future work will explore enhancing the effectiveness of this parallel approach, as well as other strategies such as bias term-based relaxation~\citep{DBLP:conf/emnlp/GeCW22}.

\begin{table}[h]
    \small
    \centering
    \resizebox{\textwidth}{!}{
    \setlength{\tabcolsep}{3pt}
    \begin{tabular}{l|c|cc|cc|cc|rrr|ccccccc|cc}
    \toprule
     & & \multicolumn{2}{c|}{\textbf{Uptrain}} & \multicolumn{2}{c|}{\textbf{Looping}} &  \multicolumn{2}{c|}{\textbf{Prefix}} & \multicolumn{3}{c|}{\textbf{Perplexity\,$\downarrow$}} & \multicolumn{9}{c}{\textbf{Few-shot Accuracy\,$\uparrow$}} \\
    \cmidrule(l{2pt}r{2pt}){3-4} \cmidrule(l{2pt}r{2pt}){5-6}  \cmidrule(l{2pt}r{2pt}){7-8} \cmidrule(l{2pt}r{2pt}){9-11}  \cmidrule(l{2pt}r{2pt}){12-20} 
     \textbf{Models} & N-emb & PT & $N_{tok}$ & Block & Init  & Len & Size & SlimP & RedP & PG19 & LD & HS & PQ & WG & ARC-e & ARC-c & OB & Avg & $\Delta$  \\
    \midrule
    & 1.99B &  \cmark & 15B & - & - & - & - & 10.76 & 8.47 & 13.08 & 63.5 & 68.5 & 77.0 & 63.5 & 67.6 & 38.1 & {42.6} & 60.1  & - \\
     & 0.99B &  \xmark & 15B & - & - & - & - & 22.63 & 20.03 & 32.60 & 28.9 & 31.6 & 63.1 & 52.3 & 41.2 & 22.5 & 27.8 &  38.2 & -   \\
     \cmidrule(l{2pt}r{2pt}){2-20} 
     & 0.99B & \cmark & 15B & 2 & Step & - & - &  {12.85} & {10.29} & {16.21} & {53.0} & {57.3} & {73.2} & {56.2} & {56.1} & {29.2} & {36.6} & {51.7} & - \\
     & 1.00B & \cmark & 15B & 2 & Step  & \,\,\,256 & \,\,\,9.4M &  12.62 &  10.06 &  15.80 &  53.5 &  58.3 &  73.9 &  57.6 &  57.5 &  29.3 &  35.6 &  52.2 & \textcolor{custom_green}{\textbf{+0.5}} \\
    & 1.01B & \cmark & 15B & 2 & Step  & \,\,\,512 & 18.9M &  12.67 &  10.10 &  15.85 &  54.1 &  57.8 &  73.8 &  58.4 &  57.2 &  28.7 &  35.8 &  52.3 & \textcolor{custom_green}{\textbf{+0.6}} \\
    & 1.03B & \cmark & 15B & 2 & Step  & 1024 & 37.7M &  12.89 &  10.34 &  16.22 &  53.5 &  57.1 &  72.4 &  57.2 &  56.9 &  28.6 &  36.8 &  51.8 & \textcolor{custom_green}{\textbf{+0.1}} \\
    & 1.07B & \cmark & 15B & 2 & Step  & 2048 & 75.5M &  12.75 &  10.21 &  16.09 &  55.0 &  57.3 &  73.3 &  58.2 &  56.8 &  29.2 &  37.8 &  52.5 & \textcolor{custom_green}{\textbf{+0.8}} \\
     \cmidrule(l{2pt}r{2pt}){2-20} 
     Gemma & 0.99B & \cmark & 15B & 2 & Avg & - & - & 15.15 & 12.57 & 19.86 & 43.6 & 47.4 & 70.4 & 52.6 & 50.5 & 27.8 & 34.4 & 46.7 & -  \\
     & 1.00B & \cmark & 15B & 2 & Avg  & \,\,\,256 & \,\,\,9.4M &  14.85 &  12.31 &  19.41 &  46.9 &  48.3 &  70.4 &  52.7 &  51.4 &  27.2 &  34.0 &  47.3 & \textcolor{custom_green}{\textbf{+0.6}} \\
    & 1.01B & \cmark & 15B & 2 & Avg  & \,\,\,512 & 18.9M &  15.23 &  12.64 &  19.98 &  44.5 &  47.2 &  70.7 &  54.5 &  49.5 &  28.0 &  33.2 &  46.8 & \textcolor{custom_green}{\textbf{+0.1}} \\
    & 1.03B & \cmark & 15B & 2 & Avg  & 1024 & 37.7M &  14.60 &  12.02 &  18.89 &  46.9 &  49.7 &  71.1 &  52.3 &  51.0 &  28.6 &  34.2 &  47.7 & \textcolor{custom_green}{\textbf{+1.0}} \\
    & 1.07B & \cmark & 15B & 2 & Avg  & 2048 & 75.5M &  14.63 &  12.07 &  19.03 &  47.3 &  49.5 &  70.8 &  53.1 &  50.7 &  28.2 &  33.4 &  47.6 & \textcolor{custom_green}{\textbf{+0.9}} \\
     \cmidrule(l{2pt}r{2pt}){2-20} 
     & 0.99B & \cmark & 15B & 2 & Lower & - & - & 15.03 & 12.46 & 19.63 & 42.5 & 48.0 & 71.0 & 54.6 & 52.2 & 27.7 & 33.8 & 47.1 & -  \\
     & 1.00B & \cmark & 15B & 2 & Lower  & \,\,\,256 & \,\,\,9.4M &  14.59 &  12.12 &  19.02 &  46.3 &  49.7 &  71.5 &  55.1 &  52.9 &  29.0 &  34.0 &  48.4 & \textcolor{custom_green}{\textbf{+1.3}} \\
    & 1.01B & \cmark & 15B & 2 & Lower  & \,\,\,512 & 18.9M &  14.53 &  12.03 &  18.88 &  45.7 &  49.8 &  71.9 &  56.4 &  53.6 &  29.4 &  33.2 &  48.6 & \textcolor{custom_green}{\textbf{+1.5}} \\
    & 1.03B & \cmark & 15B & 2 & Lower  & 1024 & 37.7M &  14.43 &  11.98 &  18.74 &  46.3 &  50.0 &  71.9 &  55.1 &  54.3 &  29.7 &  33.8 &  48.7 & \textcolor{custom_green}{\textbf{+1.6}} \\
    & 1.07B & \cmark & 15B & 2 & Lower  & 2048 & 75.5M &  14.79 &  12.26 &  19.23 &  46.1 &  48.7 &  71.4 &  55.4 &  51.3 &  28.2 &  34.0 &  47.9 & \textcolor{custom_green}{\textbf{+0.8}} \\
     \midrule
    & 0.97B &  \cmark & - & - & - & - & - & 12.26 & 9.37 & 11.94 & 43.3 & 42.2 & 66.8 & 53.4 & 44.7 & 23.2 & 29.2 & 43.3  & - \\
     & 0.48B &  \xmark & 15B & - & - & - & - & 16.61 & 15.66 & 20.27 & 22.3 & 30.0 & 60.9 & 50.6 & 37.0 & 23.0 & 28.0 & 36.0 & -  \\
     \cmidrule(l{2pt}r{2pt}){2-20} 
     & 0.48B & \cmark & 15B & 2 & Step & - & - &  {11.61} & {9.89}  & {13.00} &{ 39.6} &{ 39.8} & {66.5} & {52.9} & {44.3} & {24.9} & {30.6} & {42.7} & -   \\
     & 0.49B & \cmark & 15B & 2 & Step  & \,\,\,256 & 11.5M &  11.61 &  9.89 &  13.00 &  39.6 &  39.9 &  66.5 &  53.9 &  44.4 &  25.3 &  30.6 &  42.9 & \textcolor{custom_green}{\textbf{+0.2}} \\
    & 0.50B & \cmark & 15B & 2 & Step  & \,\,\,512 & 23.1M &  11.61 &  9.89 &  13.01 &  39.5 &  39.9 &  66.7 &  53.4 &  44.1 &  25.3 &  29.8 &  42.7 & \textcolor{gray}{\textbf{+0.0}} \\
    & 0.53B & \cmark & 15B & 2 & Step  & 1024 & 46.1M &  11.60 &  9.88 &  13.00 &  39.7 &  39.9 &  66.7 &  53.0 &  44.3 &  25.1 &  30.6 &  42.8 & \textcolor{custom_green}{\textbf{+0.1}} \\
    & 0.57B & \cmark & 15B & 2 & Step  & 2048 & 92.3M &  11.58 &  9.87 &  13.01 &  40.1 &  39.9 &  66.8 &  53.4 &  44.4 &  24.9 &  30.0 &  42.8 & \textcolor{custom_green}{\textbf{+0.1}} \\
     \cmidrule(l{2pt}r{2pt}){2-20} 
     TinyLlama & 0.48B & \cmark & 15B & 2 & Avg & - & - & 11.86 & 10.29  & 13.42 & 38.6 & 39.4 & 66.1 & 52.8 & 42.7 & 25.4 & {30.6} & 42.2 & -  \\
     & 0.49B & \cmark & 15B & 2 & Avg  & \,\,\,256 & 11.5M &  11.86 &  10.28 &  13.41 &  38.5 &  39.4 &  66.2 &  52.5 &  42.8 &  25.9 &  30.8 &  42.3 & \textcolor{custom_green}{\textbf{+0.1}} \\
    & 0.50B & \cmark & 15B & 2 & Avg  & \,\,\,512 & 23.1M &  11.86 &  10.28 &  13.41 &  38.1 &  39.3 &  66.3 &  52.2 &  42.6 &  25.6 &  30.8 &  42.1 & \textcolor{custom_red}{\textbf{--\,0.1}} \\
    & 0.53B & \cmark & 15B & 2 & Avg  & 1024 & 46.1M &  11.86 &  10.28 &  13.42 &  38.4 &  39.2 &  65.7 &  52.7 &  42.5 &  25.5 &  31.0 &  42.1 & \textcolor{custom_red}{\textbf{--\,0.1}} \\
    & 0.57B & \cmark & 15B & 2 & Avg  & 2048 & 92.3M &  11.86 &  10.28 &  13.42 &  38.5 &  39.5 &  65.9 &  52.7 &  42.4 &  25.7 &  31.0 &  42.2 & \textcolor{gray}{\textbf{+0.0}} \\
     \cmidrule(l{2pt}r{2pt}){2-20} 
     & 0.48B & \cmark & 15B & 2 & Lower & - & - & 14.67 & 12.67  & 16.68 & 31.9 & 32.3 & 62.6 & 52.0 & 39.1 & 22.1 & 27.8 & 38.3 & -  \\
     & 0.49B & \cmark & 15B & 2 & Lower  & \,\,\,256 & 11.5M &  14.67 &  12.67 &  16.69 &  31.9 &  32.4 &  62.7 &  51.5 &  38.9 &  22.3 &  27.8 &  38.2 & \textcolor{custom_red}{\textbf{--\,0.1}} \\
    & 0.50B & \cmark & 15B & 2 & Lower  & \,\,\,512 & 23.1M &  14.67 &  12.67 &  16.69 &  31.9 &  32.3 &  62.8 &  51.7 &  38.9 &  22.2 &  27.8 &  38.2 & \textcolor{custom_red}{\textbf{--\,0.1}} \\
    & 0.53B & \cmark & 15B & 2 & Lower  & 1024 & 46.1M &  14.67 &  12.67 &  16.68 &  31.6 &  32.3 &  63.0 &  51.9 &  38.9 &  22.1 &  28.0 &  38.3 & \textcolor{gray}{\textbf{+0.0}} \\
    & 0.57B & \cmark & 15B & 2 & Lower  & 2048 & 92.3M &  14.67 &  12.67 &  16.67 &  34.1 &  32.5 &  62.8 &  52.4 &  38.5 &  23.0 &  27.6 &  38.7 & \textcolor{custom_green}{\textbf{+0.4}} \\
    \bottomrule
    \end{tabular}
    }
    \caption{Evaluation results of relaxation through prefix tuning methods. Prefix length denotes the sequence length of trainable vectors used to generate key-value states in each self-attention layer. Non-embedding parameter sizes include the sizes of these trainable prefixes.
    Delta\,($\Delta$) represent the accuracy differences to non-relaxed models using the same looping initialization. 
    }
    \label{tab:prefix_tuning_app}
\end{table}

\clearpage
\vspace{-10pt}
\section{Expanded Results of Extended Uptraining and Knowledge Distillation}
\label{app:further_techniques}

\paragraph{Ablation study on individual techniques}
To further enhance performance through uptraining, we increased the number of uptraining tokens and employed knowledge distillation loss~\citep{DBLP:journals/corr/HintonVD15, DBLP:conf/emnlp/KimR16}. Specifically, we expanded the token number from 15 billion to 60 billion. Furthermore, we designated the teacher model as the full-size model for each architecture, uptrained on 15 billion tokens from the SlimPajama dataset. Given the huge number of uptraining tokens, we adopted an \textit{online} approach to extract logits from the teacher model. Four loss functions were utilized: forward KL (FKL; \citet{DBLP:conf/emnlp/KimR16}), reverse KL (RKL; \citet{DBLP:conf/iclr/Gu0WH24}), Jensen–Shannon divergence (JSD; \citet{DBLP:conf/iclr/AgarwalVZSGGB24}), and total variation distance (TVD; \citet{wen-etal-2023-f}).
Table\,\ref{tab:kd_ablation_study} summarizes the controlled experimental results for each method. We finally selected forward KL as the distillation loss function due to its simplicity and superior performance. Especially, we observed a performance improvement of 1.7\%p attributed to the extended uptraining and up to 1.7\%p from the KD loss. This suggests that combining both techniques could yield even greater gains.

\begin{table}[h]
    \small
    \centering
    \resizebox{\textwidth}{!}{
    \setlength{\tabcolsep}{3pt}
    \begin{tabular}{c|cccc|cc|cc|rrr|ccccccc|cc}
    \toprule
     & \multicolumn{4}{c|}{\textbf{Uptrain}} & \multicolumn{2}{c|}{\textbf{Looping}} &  \multicolumn{2}{c|}{\textbf{LoRA}} & \multicolumn{3}{c|}{\textbf{Perplexity\,$\downarrow$}} & \multicolumn{9}{c}{\textbf{Few-shot Accuracy\,$\uparrow$}} \\
    \cmidrule(l{2pt}r{2pt}){2-5} \cmidrule(l{2pt}r{2pt}){6-7}  \cmidrule(l{2pt}r{2pt}){8-9} \cmidrule(l{2pt}r{2pt}){10-12}  \cmidrule(l{2pt}r{2pt}){13-21} 
      N-emb & PT & $N_{tok}$ & KD & Func & Block & Init  & Rank & Init & SlimP & RedP & PG19 & LD & HS & PQ & WG & ARC-e & ARC-c & OB & Avg & $\Delta$  \\
    \midrule
    0.99B &  \cmark & 15B & \xmark & - & 2 & Step & - & - &  {12.85} & {10.29} & {16.21} & {53.0} & {57.3} & {73.2} & {56.2} & {56.1} & {29.2} & {36.6} & {51.7} & - \\
    \rowcolor{gray!20}
    0.99B &  \cmark & 60B & \xmark & - & 2 & Step & - & - &  12.00 &  9.70 &  14.84 &  52.5 &  59.9 &  74.7 &  58.5 &  58.0 &  30.3 &  40.2 &  53.4 & \textcolor{custom_green}{\textbf{+1.7}} \\
    \midrule
    0.99B &  \cmark & 15B & \xmark & - & 2 & Step & - & - & {12.85} & {10.29} & {16.21} & {53.0} & {57.3} & {73.2} & {56.2} & {56.1} & {29.2} & {36.6} & {51.7} & - \\
    \rowcolor{gray!20}
    0.99B &  \cmark & 15B & \cmark & FKL & 2 & Step & - & - &  12.36 &  9.85 &  15.45 &  56.8 &  58.6 &  74.8 &  58.6 &  59.1 &  29.2 &  36.6 &  53.4 & \textcolor{custom_green}{\textbf{+1.7}} \\
    0.99B &  \cmark & 15B & \cmark & RKL & 2 & Step & - & - &  12.56 &  10.09 &  15.80 &  55.6 &  58.3 &  74.3 &  58.6 &  58.3 &  30.4 &  37.4 &  53.3 & \textcolor{custom_green}{\textbf{+1.6}} \\
    0.99B &  \cmark & 15B & \cmark & JSD & 2 & Step & - & - &  12.60 &  10.06 &  15.77 &  56.1 &  58.4 &  73.4 &  57.0 &  58.4 &  29.8 &  37.2 &  52.9 & \textcolor{custom_green}{\textbf{+1.2}} \\
    0.99B &  \cmark & 15B & \cmark & TVD & 2 & Step & - & - &  12.47 &  9.92 &  15.52 &  55.1 &  58.5 &  74.0 &  58.2 &  58.9 &  29.5 &  36.8 &  53.0 & \textcolor{custom_green}{\textbf{+1.3}} \\
    \midrule
    1.30B &  \cmark & 15B & \xmark & - & 2 & Avg & 256 & SVD & 12.10 &  9.71 &  14.89 &  58.2 &  60.7 &  73.7 &  59.0 &  57.6 &  32.1 &  {38.0} &  54.2 & - \\
    \rowcolor{gray!20}
    1.30B &  \cmark & 15B & \cmark & FKL & 2 & Avg & 256 & SVD &  11.90 &  9.52 &  14.63 &  59.9 &  62.0 &  74.1 &  60.0 &  58.6 &  33.2 &  38.0 &  55.1 & \textcolor{custom_green}{\textbf{+0.9}} \\
    1.30B &  \cmark & 15B & \cmark & RKL & 2 & Avg & 256 & SVD &  11.95 &  9.62 &  14.79 &  60.0 &  61.6 &  74.5 &  60.0 &  58.1 &  32.9 &  37.8 &  55.0 & \textcolor{custom_green}{\textbf{+0.8}} \\
    1.30B &  \cmark & 15B & \cmark & JSD & 2 & Avg & 256 & SVD &  12.09 &  9.65 &  14.81 &  58.1 &  61.1 &  73.1 &  60.8 &  59.0 &  33.2 &  38.6 &  54.8 & \textcolor{custom_green}{\textbf{+0.6}} \\
    1.30B &  \cmark & 15B & \cmark & TVD & 2 & Avg & 256 & SVD &  12.05 &  9.62 &  14.78 &  59.3 &  61.5 &  73.9 &  60.5 &  59.0 &  33.0 &  38.2 &  55.1 & \textcolor{custom_green}{\textbf{+0.9}} \\
    \bottomrule
    \end{tabular}
    }
    \caption{
    Evaluation results of ablation studies related to longer uptraining and knowledge distillation loss. Performance improvements, represented by Delta, were measured for each technique. For the knowledge distillation loss function, we experimented with four options: FKL, RKL, JSD, and TVD. Forward KL was selected as the final configuration due to its simplicity and superior performance.
    }
    \label{tab:kd_ablation_study}
\end{table}

\begin{figure}[h]
    \centering
    \begin{subfigure}[t]{\textwidth}
        \includegraphics[width=\textwidth]{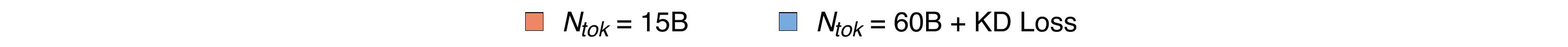}
    \end{subfigure}
    \centering
    \begin{subfigure}[t]{0.34\textwidth}
        \includegraphics[width=\textwidth]{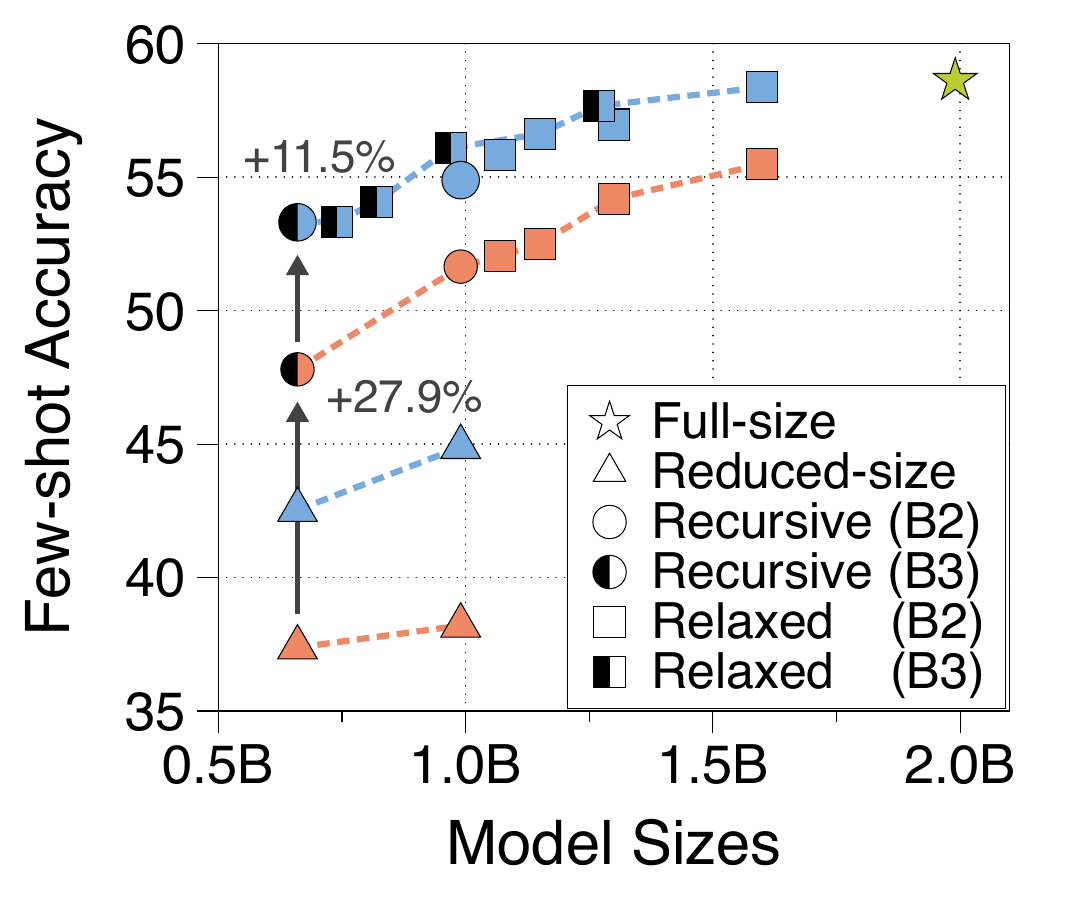}
        \subcaption{Gemma}
    \end{subfigure}
    \centering
    \begin{subfigure}[t]{0.31\textwidth}
        \includegraphics[width=\textwidth]{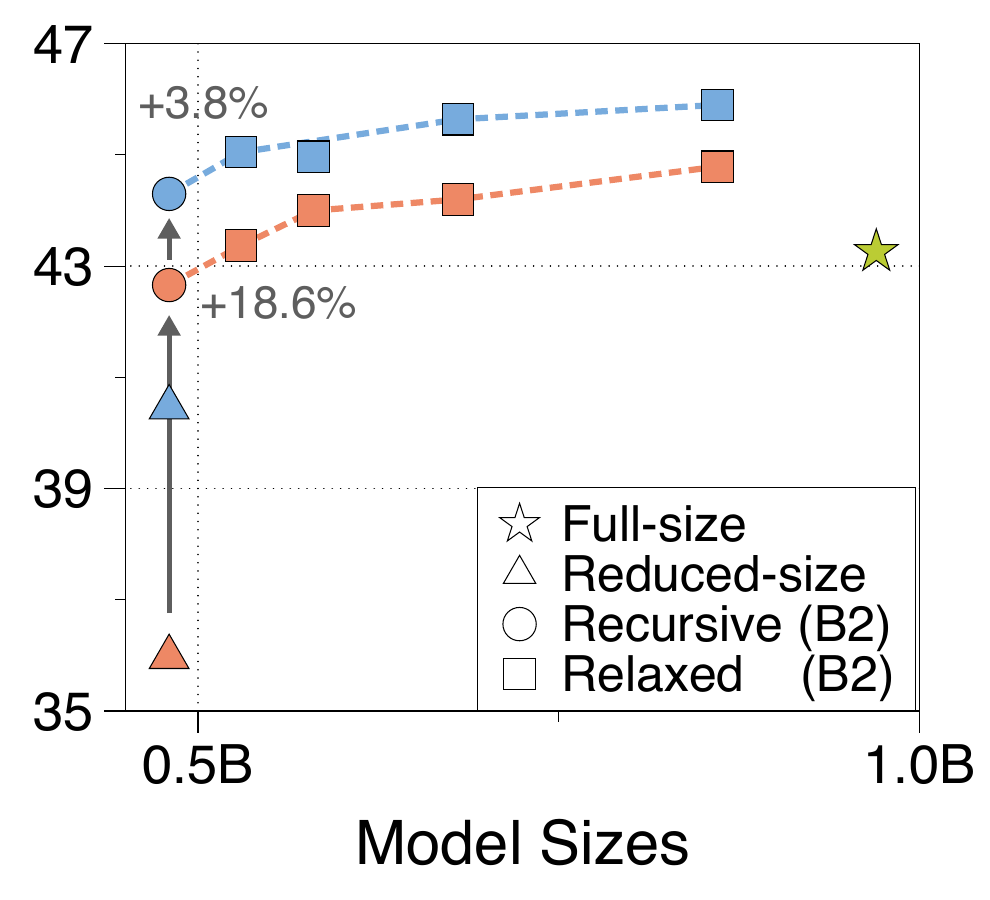}
        \subcaption{TinyLlama}
    \end{subfigure}
    \centering
    \begin{subfigure}[t]{0.31\textwidth}
        \includegraphics[width=\textwidth]{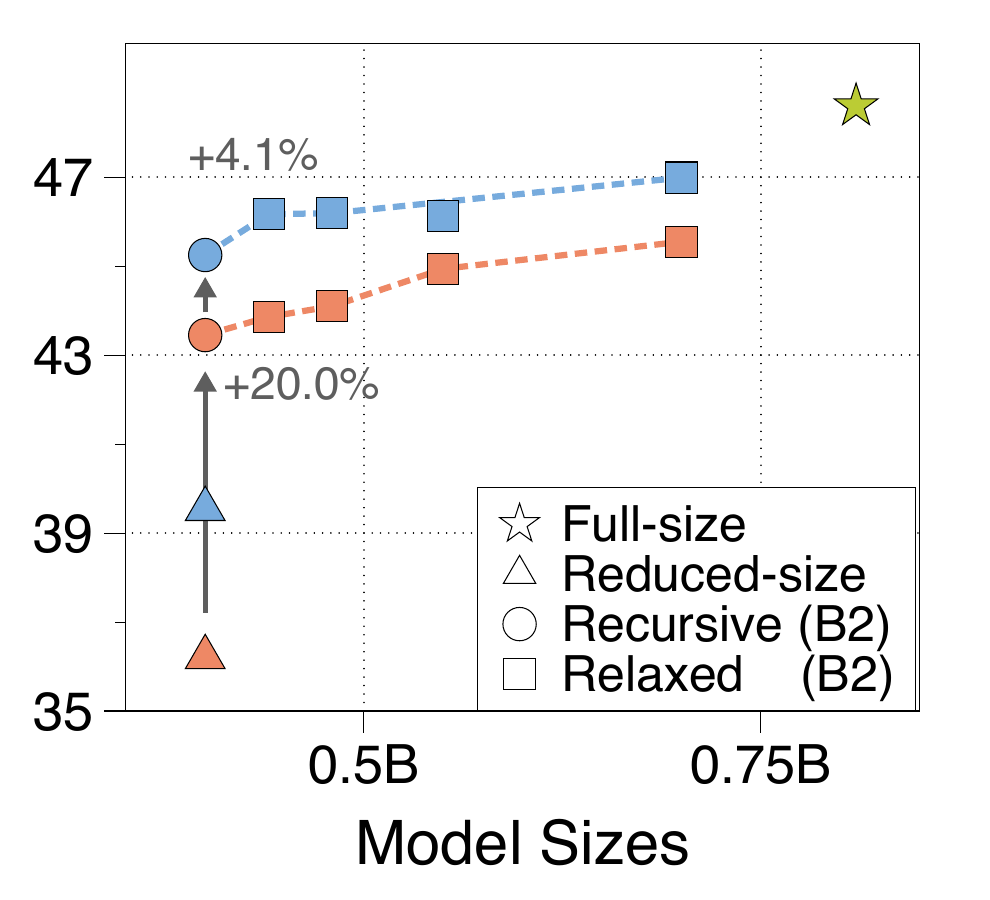}
        \subcaption{Pythia}
    \end{subfigure}
    \caption{
    Few-shot performance of three models with extended uptraining and knowledge distillation. Optimal configurations are used for each model size (Stepwise for recursive models and Average for relaxed models). 
    Dotted lines represent the Pareto frontier, showing the optimal trade-offs between model size and performance for each setting.
    }
    \label{fig:60b_kd_model_size_performance_app}
\end{figure}

\clearpage

\paragraph{Overall performance after longer training with distillation loss}
Figure\,\ref{fig:60b_kd_model_size_performance_app} and Table\,\ref{tab:long_training_kd_app} summarize the performance gains achieved by incorporating these advanced training techniques.
We consistently observed substantial improvements in few-shot performance across all architectures and with varying numbers of looping blocks. We anticipate that further performance enhancements can be achieved by utilizing a superior teacher model and increasing the uptraining cost.

\begin{table}[ht!]
    \small
    \centering
    \resizebox{\textwidth}{!}{
    \setlength{\tabcolsep}{3pt}
    \begin{tabular}{l|c|ccc|cc|cc|rrr|ccccccc|cc}
    \toprule
     & & \multicolumn{3}{c|}{\textbf{Uptrain}} & \multicolumn{2}{c|}{\textbf{Looping}} &  \multicolumn{2}{c|}{\textbf{LoRA}} & \multicolumn{3}{c|}{\textbf{Perplexity\,$\downarrow$}} & \multicolumn{9}{c}{\textbf{Few-shot Accuracy\,$\uparrow$}} \\
    \cmidrule(l{2pt}r{2pt}){3-5} \cmidrule(l{2pt}r{2pt}){6-7}  \cmidrule(l{2pt}r{2pt}){8-9} \cmidrule(l{2pt}r{2pt}){10-12}  \cmidrule(l{2pt}r{2pt}){13-21} 
     \textbf{Models} & N-emb & PT & $N_{tok}$ & KD & Block & Init  & Rank & Init & SlimP & RedP & PG19 & LD & HS & PQ & WG & ARC-e & ARC-c & OB & Avg & $\Delta$  \\
    \midrule
    & 1.99B &  \cmark & 60B & \xmark & - & - & - & - & 10.58 & 8.44 & 12.71 & 60.3 & 67.9 & 76.9 & 63.5 & 64.9 & 37.2 & {39.6} & 58.6  & - \\
     & 0.99B & \xmark & 60B & \cmark & - & - & - & - & 15.33 & 13.04 & 20.37 & 42.3 & 43.0 & 68.8 & 53.4 & 49.4 & 26.3 & 31.0 & 44.9 & - \\ 
     & 0.99B & \xmark & 15B & \xmark & - & - & - & - & 22.63 & 20.03 & 32.60 & 28.9 & 31.6 & 63.1 & 52.3 & 41.2 & 22.5 & 27.8 &  38.2 & - \\  
     & 0.66B & \xmark & 60B & \cmark & - & - & - & - & 16.79 & 14.39 & 22.85 & 37.5 & 38.4 & 68.7 & 50.4 & 46.5 & 24.6 & 31.6 & 42.5 & - \\ 
     & 0.66B & \xmark & 15B & \xmark & - & - & - & - & 24.44 & 21.69 & 36.03 & 27.2 & 30.6 & 63.8 & 50.5 & 40.6 & 22.0 & 27.0  & 37.4 & - \\ 
     \cmidrule(l{2pt}r{2pt}){2-21} 
     & 0.99B & \cmark & 15B & \xmark & 2 & Step & - & - &  {12.85} & {10.29} & {16.21} & {53.0} & {57.3} & {73.2} & {56.2} & {56.1} & {29.2} & {36.6} & {51.7} & - \\
     & 0.66B & \cmark & 15B & \xmark & 3 & Step & - & - &  {14.75} & {12.10} & {19.32} & {45.0} & {49.9} & {69.8} & {55.8} & {52.7} & {27.9} & {33.6} & {47.8} & - \\
     & 1.07B & \cmark & 15B & \xmark & 2 & Avg  & 64 & SVD &  12.83 &  10.35 &  16.02 &  55.9 &  56.8 &  72.5 &  56.8 &  55.7 &  30.6 &  36.2 &  52.1 & - \\
     & 1.15B & \cmark & 15B & \xmark & 2 & Avg  & 128 & SVD &  12.52 &  10.07 &  15.51 &  56.1 &  58.2 &  72.3 &  55.8 &  57.1 &  30.7 &  37.2 &  52.5 & - \\
     & 1.30B & \cmark & 15B & \xmark & 2 & Avg  & 256 & SVD &  12.10 &  9.71 &  14.89 &  58.2 &  60.7 &  73.7 &  59.0 &  57.6 &  32.1 &  {38.0} &  54.2 & - \\
    Gemma & 1.60B & \cmark & 15B & \xmark & 2 & Avg  & 512 & SVD &  {11.83} &  {9.46} &  {14.57} &  {59.3} &  {62.8} &  {74.0 }&  {61.6} &  {60.1} &  {32.9} &  37.6 &  {55.5} & - \\
     \cmidrule(l{2pt}r{2pt}){2-21} 
     \rowcolor[gray]{0.9} \cellcolor{white!20} 
     & 0.99B & \cmark & 60B & \cmark & 2 & Step  & - & - &  11.44 &  9.14 &  13.98 &  56.5 &  62.1 &  75.2 &  59.4 &  59.8 &  32.5 &  38.6 &  54.9 & \textcolor{custom_green}{\textbf{+3.2}} \\
     \rowcolor[gray]{0.9} \cellcolor{white!20} 
    & 1.07B & \cmark & 60B & \cmark & 2 & Avg  & 64 & SVD &  11.36 &  9.14 &  13.82 &  58.9 &  62.8 &  75.1 &  61.5 &  61.2 &  33.7 &  37.6 &  55.8 & \textcolor{custom_green}{\textbf{+3.7}} \\
    \rowcolor[gray]{0.9} \cellcolor{white!20} 
    & 1.15B & \cmark & 60B & \cmark & 2 & Avg  & 128 & SVD &  11.25 &  9.04 &  13.64 &  58.7 &  63.6 &  76.5 &  61.2 &  62.6 &  34.6 &  39.0 &  56.6 & \textcolor{custom_green}{\textbf{+4.1}} \\
    \rowcolor[gray]{0.9} \cellcolor{white!20} 
    & 1.30B & \cmark & 60B & \cmark & 2 & Avg  & 256 & SVD &  11.05 &  8.88 &  13.35 &  60.6 &  64.7 &  75.3 &  62.5 &  61.6 &  35.3 &  38.8 &  57.0 & \textcolor{custom_green}{\textbf{+2.8}} \\
    \rowcolor[gray]{0.9} \cellcolor{white!20} 
    & 1.60B & \cmark & 60B & \cmark & 2 & Avg  & 512 & SVD &  \textbf{10.81} &  \textbf{8.63} &  \textbf{12.94} &  61.4 &  \textbf{65.8} &  \textbf{76.3} & \textbf{63.5} &  \textbf{65.1} &  \textbf{37.1} &  39.4 & \textbf{58.4} & \textcolor{custom_green}{\textbf{+2.9}} \\
     \cmidrule(l{2pt}r{2pt}){2-21} 
     \rowcolor[gray]{0.9} \cellcolor{white!20} 
     & 0.66B & \cmark & 60B & \cmark & 3 & Step  & - & - &  12.27 &  9.90 &  15.24 &  55.6 &  58.1 &  73.1 &  60.2 &  58.8 &  30.2 &  37.2 &  53.3 & \textcolor{custom_green}{\textbf{+5.5}} \\
     \rowcolor[gray]{0.9} \cellcolor{white!20} 
    & 0.74B & \cmark & 60B & \cmark & 3 & Avg  & 64 & SVD &  12.13 &  9.80 &  14.95 &  55.5 &  58.3 &  73.5 &  60.1 &  58.0 &  31.1 &  36.8 &  53.3 & - \\
    \rowcolor[gray]{0.9} \cellcolor{white!20} 
    & 0.82B & \cmark & 60B & \cmark & 3 & Avg  & 128 & SVD &  11.83 &  9.53 &  14.51 &  56.7 &  60.2 &  74.2 &  59.8 &  59.1 &  33.0 &  35.4 &  54.1 & - \\
    \rowcolor[gray]{0.9} \cellcolor{white!20} 
    & 0.97B & \cmark & 60B & \cmark & 3 & Avg  & 256 & SVD &  11.43 &  9.17 &  13.87 &  59.3 &  62.6 &  74.7 &  61.2 &  61.6 &  32.9 &  \textbf{40.2} &  56.1 & - \\
    \rowcolor[gray]{0.9} \cellcolor{white!20} 
    & 1.27B & \cmark & 60B & \cmark & 3 & Avg  & 512 & SVD &  11.01 &  8.80 &  13.25 &  \textbf{61.5} &  64.9 &  76.2 &  62.0 &  64.3 &  35.6 &  39.2 &  57.7 & - \\
     \midrule
    & 0.97B &  \cmark & - & - & - & - & - & - & 12.26 & 9.37 & 11.94 & 43.3 & 42.2 & 66.8 & 53.4 & 44.7 & 23.2 & 29.2 & 43.3  & - \\
     & 0.48B &  \xmark & 60B & \cmark & - & - & - & - & 11.93 & 10.86 & 13.93 & 33.3 & 37.3 & 66.8 & 50.1 & 41.7 & 23.9 & 30.2 & 40.5 & - \\  
     & 0.48B &  \xmark & 15B & \xmark & - & - & - & - & 16.61 & 15.66 & 20.27 & 22.3 & 30.0 & 60.9 & 50.6 & 37.0 & 23.0 & 28.0 & 36.0 & - \\  
     \cmidrule(l{2pt}r{2pt}){2-21} 
     & 0.48B & \cmark & 15B & \xmark & 2 & Step & - & - &  {11.61} & {9.89}  & {13.00} &{ 39.6} &{ 39.8} & {66.5} & {52.9} & {44.3} & {24.9} & {30.6} & {42.7} & -   \\
      & 0.53B & \cmark & 15B & \xmark &  2 & Avg  & 64 & SVD &  11.22 &  9.66 &  12.51 &  41.8 &  41.6 &  67.0 &  53.3 &  43.9 &  24.7 &  {31.2} &  43.4 & - \\
    & 0.58B & \cmark & 15B &  \xmark & 2 & Avg  & 128 & SVD &  10.99 &  9.45 &  12.21 &  43.2 &  42.1 &  {68.3} &  53.2 &  44.8 &  25.9 &  30.4 &  44.0 & - \\
    TinyLlama & 0.68B & \cmark & 15B & \xmark &  2 & Avg  & 256 & SVD &  10.71 &  9.18 &  11.82 &  44.1 &  43.2 &  68.1 &  \textbf{53.5} &  44.4 &  25.7 &  30.4 &  44.2 & - \\
    & 0.86B & \cmark & 15B & \xmark &  2 & Avg  & 512 & SVD &  {10.46} &  {8.92} & { 11.50} &  {46.0} &  {44.1} &  68.2 &  53.0 &  {45.8} &  25.1 &  {31.2} &  {44.8} & - \\
     \cmidrule(l{2pt}r{2pt}){2-21} 
     \rowcolor[gray]{0.9} \cellcolor{white!20} 
    & 0.48B & \cmark & 60B & \cmark & 2 & Step  & - & - &  10.51 &  9.01 &  11.60 &  44.2 &  43.1 &  68.2 &  52.4 &  44.7 &  25.3 &  32.2 &  44.3 & \textcolor{custom_green}{\textbf{+1.6}} \\
    \rowcolor[gray]{0.9} \cellcolor{white!20} 
    & 0.53B & \cmark & 60B & \cmark & 2 & Avg  & 64 & SVD &  10.14 &  8.77 &  11.19 &  44.3 &  44.9 &  69.5 &  52.5 &  46.5 &  26.1 &  \textbf{31.6} &  45.1 & \textcolor{custom_green}{\textbf{+1.6}} \\
    \rowcolor[gray]{0.9} \cellcolor{white!20} 
    & 0.58B & \cmark & 60B & \cmark & 2 & Avg  & 128 & SVD &  10.07 &  8.68 &  11.07 &  45.9 &  45.1 &  69.4 &  50.5 &  46.8 &  25.4 &  \textbf{31.6} &  45.0 & \textcolor{custom_green}{\textbf{+1.0}} \\
    \rowcolor[gray]{0.9} \cellcolor{white!20} 
    & 0.68B & \cmark & 60B & \cmark & 2 & Avg  & 256 & SVD &  9.96 &  8.56 &  10.93 &  46.2 &  45.7 &  69.0 &  53.2 &  \textbf{47.9} &  25.9 &  \textbf{31.6} &  45.6 & \textcolor{custom_green}{\textbf{+1.4}} \\
    \rowcolor[gray]{0.9} \cellcolor{white!20} 
    & 0.86B & \cmark & 60B & \cmark & 2 & Avg  & 512 & SVD &  \textbf{9.85} &  \textbf{8.44} &  \textbf{10.76} &  \textbf{47.4} &  \textbf{46.3} &  \textbf{69.7} &  52.8 &  47.5 &  \textbf{26.3} &  31.4 &  \textbf{45.9} & \textcolor{custom_green}{\textbf{+1.1}} \\
     \midrule
    & 0.81B &  \cmark & 60B & \xmark & - & - & - & - & 12.83 & 9.76 & 13.57 &  53.0 & 50.2 & 71.1 & 54.8 & 51.9 & {27.7} & 31.6 & {48.6}  & - \\
     & 0.40B &  \xmark & 60B & \cmark & - & - & - & - & 18.27 & 14.39 & 21.93 & 32.1 & 35.0 & 66.1 & 49.6 & 42.9 & 24.2 & 27.0  & 39.5 & -  \\
     & 0.40B &  \xmark & 15B & \xmark & - & - & - & - & 25.69 & 20.00 & 32.08 & 24.3 & 30.0 & 61.9 & 50.7 & 38.3 & 22.3 & 26.0  & 36.2 & -  \\ 
     \cmidrule(l{2pt}r{2pt}){2-21} 
     & 0.40B & \cmark & 15B & \xmark & 2 & Step & - & - & {16.38} & {12.37} & {17.74} & 43.4 & {40.5} & 67.4 & 50.8 & {46.3} & 25.7 & 30.0 & 43.5 & -   \\
     & 0.44B & \cmark & 15B & \xmark  & 2 & Avg  & 64 & SVD &  16.03 &  12.19 &  17.59 &  45.8 &  40.9 &  67.3 &  50.0 &  45.8 &  25.5 &  31.8 &  43.9 & - \\
     & 0.48B & \cmark & 15B & \xmark  & 2 & Avg  & 128 & SVD &  15.67 &  11.93 &  17.10 &  46.9 &  41.9 &  67.4 &  51.2 &  45.4 &  24.8 &  31.2 &  44.1 & - \\
    Pythia & 0.55B & \cmark & 15B & \xmark & 2 & Avg  & 256 & SVD &  15.22 &  11.54 &  16.47 &  48.5 &  43.3 &  67.2 &  51.4 &  46.7 &  25.5 &  32.0 &  44.9 & - \\
    & 0.70B & \cmark & 15B & \xmark & 2 & Avg  & 512 & SVD &  {14.70} &  {11.07} &  {15.71} &  {50.2} &  {44.7} &  68.2 &  51.6 &  {47.6} &  25.4 &  31.2 &  {45.6} & - \\
     \cmidrule(l{2pt}r{2pt}){2-21} 
    \rowcolor[gray]{0.9} \cellcolor{white!20} 
     & 0.40B & \cmark & 60B & \cmark & 2 & Step  & - & - &  14.59 &  11.13 &  15.79 &  47.8 &  43.8 &  69.3 &  52.0 &  48.1 &  25.4 &  30.4 &  45.2 & \textcolor{custom_green}{\textbf{+1.7}} \\
     \rowcolor[gray]{0.9} \cellcolor{white!20} 
    & 0.44B & \cmark & 60B & \cmark & 2 & Avg  & 64 & SVD &  14.24 &  10.89 &  15.52 &  50.0 &  44.5 &  68.9 &  \textbf{54.1} &  48.0 &  26.5 &  31.2 &  46.2 & \textcolor{custom_green}{\textbf{+2.3}} \\
    \rowcolor[gray]{0.9} \cellcolor{white!20} 
    & 0.48B & \cmark & 60B & \cmark & 2 & Avg  & 128 & SVD &  14.10 &  10.79 &  15.27 &  50.1 &  45.5 &  69.0 &  52.6 &  48.3 &  25.8 &  32.0 &  46.2 & \textcolor{custom_green}{\textbf{+2.1}} \\
    \rowcolor[gray]{0.9} \cellcolor{white!20} 
    & 0.55B & \cmark & 60B & \cmark & 2 & Avg  & 256 & SVD &  13.91 &  10.61 &  14.91 &  50.5 &  45.6 &  68.7 &  51.2 &  48.4 &  25.7 &  \textbf{32.8} &  46.1 & \textcolor{custom_green}{\textbf{+1.2}} \\
    \rowcolor[gray]{0.9} \cellcolor{white!20} 
    & 0.70B & \cmark & 60B & \cmark & 2 & Avg  & 512 & SVD &  \textbf{13.59} &  \textbf{10.38} &  \textbf{14.43} &  \textbf{52.0} &  \textbf{47.0} &  \textbf{69.6} &  53.4 &  \textbf{48.9} &  \textbf{26.9} &  31.2 &  \textbf{47.0} & \textcolor{custom_green}{\textbf{+1.4}} \\
    \bottomrule
    \end{tabular}
    }
    \caption{Evaluation results of our Recursive Transformers with 60 billion token uptraining and knowledge distillation loss. We utilized the forward KL loss as the knowledge distillation loss function. 
    Full-size model baselines for Gemma and Pythia are the pretrained models further uptrained on 60 billion tokens, accounting for distribution shifts between Slimapajama and their pretraining datasets.
    Delta\,($\Delta$) represents the accuracy differences between the longer uptrained models with KD and their 15 billion uptrained counterparts. 
    }
    \label{tab:long_training_kd_app}
\end{table}

\clearpage

\section{Expanded Results of Early-Exit Training}
\label{app:early_exit}

\paragraph{Ablation study on early-exit training strategy}
To enable early-exiting capabilities, all models require additional training to align intermediate representations with classifier heads. In this study, we conduct ablation studies on various strategies, demonstrating Recursive Transformers can be transformed into early-exiting models without compromising final loop output's performance. 
Table\,\ref{tab:early_exit_ablation_app} presents a comprehensive summary of experimental results across various categories, including training procedures, loss functions, and early-exit training data. Our key findings are as follows:
\begin{itemize}[leftmargin=*]
    \item Post-training after uptraining is essential for preserving final loop performance. Jointly training intermediate loop output during the uptraining phase (co-training) significantly degraded the final output performance, even with an aggressive loss coefficient strategy.
    \item While freezing all parameters, we attempted to train intermediate loop outputs by attaching trainable LoRA modules to frozen classifier head. However, we found that this proved ineffective.
    \item The aggressive coefficient strategy successfully maintained final loop output performance while enhancing intermediate output performance. Moreover, incorporating knowledge distillation from detached final outputs further enhanced intermediate performance.
    \item No significant performance differences were observed when using the overlapped uptraining dataset versus new SlimPajama tokens for post-training.
\end{itemize}
Finally, we opted to perform post-training with new tokens sourced from the same SlimPajama dataset. Moreover, we incorporated a distillation loss from the final loop output, while aggressively reducing the loss coefficients of intermediate outputs.

\paragraph{Early-exit training results on final models}
We applied the aggressive coefficient strategy with distillation loss to our final models (uptrained on 60 billion tokens with knowledge distillation). Tables\,\ref{tab:final_early_exit_gemma_app} and \ref{tab:final_early_exit_tinyllama_pythia_app} summarize the performance of intermediate loops and the final loop across three models. For fair comparison, the full-size models (Gemma and Pythia) were also uptrained with 60 billion tokens and then post-trained with 15 billion tokens.
Consistent with previous findings, the aggressive coefficient strategy yielded the best performance across both intermediate and final outputs.

However, we find that intermediate loop outputs in LoRA-relaxed models underperformed their non-relaxed counterparts (recursive models). This could potentially reduce throughput gain, as early loop performance directly influences the number of tokens eligible for early-exit. 
In perfectly tied looping blocks, intermediate outputs seem to be distilled from the last, as all gradients are backpropagated to the same parameters. 
Conversely, since LoRA modules allow each layer to specialize based on its depth, intermediate representations appear to be optimized to facilitate performance of the final output, not for their own sake. 
Hence, relaxation introduces a trade-off between final performance and early-exiting benefits. 
As the optimal strategy derived from the non-relaxed models was directly applied to the relaxed models, further exploration of optimal strategies specifically for relaxed models is left for future work.

\begin{table}[ht]
    \small
    \centering
    \resizebox{\textwidth}{!}{
    \setlength{\tabcolsep}{3pt}
    \begin{tabular}{c|cc|cc|cccccc|rrr|ccccccc|cc}
    \toprule
     & \multicolumn{2}{c|}{\textbf{Uptrain}} & \multicolumn{2}{c|}{\textbf{Looping}} &  \multicolumn{6}{c|}{\textbf{Early-Exit Train}} & \multicolumn{3}{c|}{\textbf{Perplexity\,$\downarrow$}} & \multicolumn{9}{c}{\textbf{Few-shot Accuracy\,$\uparrow$}} \\
    \cmidrule(l{2pt}r{2pt}){2-3} \cmidrule(l{2pt}r{2pt}){4-5}  \cmidrule(l{2pt}r{2pt}){6-11} \cmidrule(l{2pt}r{2pt}){12-14}  \cmidrule(l{2pt}r{2pt}){15-23} 
     N-emb & PT & $N_{tok}$ & Block & Init  & Train & Freeze& $N_{tok}$ & CE & KD & Data & SlimP & RedP & PG19 & LD & HS & PQ & WG & ARC-e & ARC-c & OB & Avg & $\Delta$  \\
    \midrule
    1.99B &  \cmark & 15B  & - & - & - & - & - & - & - & - & 10.76 & 8.47 & 13.08 & 63.5 & 68.5 & 77.0 & 63.5 & 67.6 & 38.1 & {42.6} & 60.1  & - \\
     0.99B & \xmark & 15B  & - & - & - & - & - & - & - & - & 22.63 & 20.03 & 32.60 & 28.9 & 31.6 & 63.1 & 52.3 & 41.2 & 22.5 & 27.8 &  38.2 & - \\ 
     0.99B & \cmark & 15B  & 2 & Step &  - &- & - & - & - & - & {12.85} & {10.29} & {16.21} & {53.0} & {57.3} & {73.2} & {56.2} & {56.1} & {29.2} & {36.6} & {51.7} & - \\
    \midrule
     \multirow{2}{*}{0.99B} & \multirow{2}{*}{\cmark} & \multirow{2}{*}{15B}  & \multirow{2}{*}{2} & \multirow{2}{*}{Step}  &  \multirow{2}{*}{Post-} &  \multirow{2}{*}{\xmark} & \multirow{2}{*}{15B}&  \multirow{2}{*}{Weighted} &  \multirow{2}{*}{\xmark} &  \multirow{2}{*}{Ovlp}  &  12.97 &  10.51 &  16.55 &  48.9 &  55.5 &  72.7 &  55.3 &  54.9 &  30.1 &  36.0 &  50.5 & \textcolor{custom_red}{\textbf{--\,1.2}} \\
     &  &  &  & &  &  &  & &    &  &  13.11 &  10.59 &  16.71 &  49.5 &  54.8 &  72.0 &  53.4 &  54.1 &  29.1 &  35.6 &  49.8 & - \\
     \addlinespace[-1pt]
     \cmidrule(l{2pt}r{2pt}){12-23} 
     \addlinespace[-1pt]
     \multirow{2}{*}{0.99B} & \multirow{2}{*}{\cmark} & \multirow{2}{*}{15B} & \multirow{2}{*}{2} & \multirow{2}{*}{Step} &  \multirow{2}{*}{Post-} &  \multirow{2}{*}{\xmark}& \multirow{2}{*}{15B} &  \multirow{2}{*}{Agg\,(0.3)} &  \multirow{2}{*}{\xmark} &  \multirow{2}{*}{Ovlp} &  12.60 &  10.21 &  15.75 &  51.8 &  58.2 &  73.7 &  56.8 &  57.0 &  29.9 &  37.8 &  52.2 & \textcolor{custom_green}{\textbf{+0.5}} \\
     &  &  &  & &  &  &   &  & &  &  13.63 &  11.02 &  17.55 &  47.5 &  53.0 &  71.2 &  54.9 &  50.2 &  28.2 &  34.8 &  48.5 & - \\
     \addlinespace[-1pt]
     \cmidrule(l{2pt}r{2pt}){12-23} 
     \addlinespace[-1pt]
     \rowcolor[gray]{0.9}
      &  &  &  &  &  &   &   &  &   &  &  12.37 &  9.94 &  15.37 &  53.0 &  59.1 &  73.9 &  55.4 &  57.4 &  30.6 &  37.8 &  52.5 & \textcolor{custom_green}{\textbf{+0.8}} \\
      \rowcolor[gray]{0.9}
     \multirow{-2}{*}{0.99B} & \multirow{-2}{*}{\cmark} &  \multirow{-2}{*}{15B} & \multirow{-2}{*}{2} & \multirow{-2}{*}{Step} & \multirow{-2}{*}{Post-} & \multirow{-2}{*}{\xmark} &  \multirow{-2}{*}{15B}  & \multirow{-2}{*}{Agg\,(0.1)} & \multirow{-2}{*}{\xmark} & \multirow{-2}{*}{Ovlp} &  14.55 &  11.87 &  19.00 &  45.9 &  51.2 &  71.4 &  54.5 &  48.1 &  26.8 &  32.0 &  47.1 & - \\
     \addlinespace[-1pt]
     \cmidrule(l{2pt}r{2pt}){12-23} 
     \addlinespace[-1pt]
     \multirow{2}{*}{0.99B} & \multirow{2}{*}{\cmark} & \multirow{2}{*}{15B} & \multirow{2}{*}{2} & \multirow{2}{*}{Step} &  \multirow{2}{*}{Post-} &  \multirow{2}{*}{\xmark} &  \multirow{2}{*}{15B}  &  \multirow{2}{*}{Agg\,(0.05)} &  \multirow{2}{*}{\xmark} &  \multirow{2}{*}{Ovlp} &  12.33 &  9.90 &  15.31 &  52.8 &  59.2 &  73.6 &  57.5 &  57.7 &  30.5 &  37.2 &  52.6 & \textcolor{custom_green}{\textbf{+0.9}} \\
     &  &  &  & &  &  &   &  & & &  15.70 &  12.93 &  20.69 &  43.1 &  49.8 &  69.8 &  55.2 &  46.0 &  26.9 &  31.2 &  46.0 & - \\
     \addlinespace[-1pt]
     \cmidrule(l{2pt}r{2pt}){12-23} 
     \addlinespace[-1pt]
      &  & &  &  &  &   &   &  &  &   &  12.28 &  9.80 &  15.23 &  52.9 &  59.5 &  73.3 &  56.5 &  57.2 &  30.1 &  37.2 &  52.4 & \textcolor{custom_green}{\textbf{+0.7}} \\
     \multirow{-2}{*}{0.99B} & \multirow{-2}{*}{\cmark} &  \multirow{-2}{*}{15B} & \multirow{-2}{*}{2} & \multirow{-2}{*}{Step} & \multirow{-2}{*}{Post-} & \multirow{-2}{*}{\xmark} &  \multirow{-2}{*}{15B}  & \multirow{-2}{*}{Agg\,(0.01)} & \multirow{-2}{*}{\xmark} & \multirow{-2}{*}{Ovlp} &  22.76 &  20.37 &  30.39 &  32.2 &  45.2 &  67.5 &  53.9 &  40.3 &  26.3 &  29.2 &  42.1 & - \\
     \midrule
     \multirow{2}{*}{0.99B} & \multirow{2}{*}{\cmark} & \multirow{2}{*}{15B} & \multirow{2}{*}{2} & \multirow{2}{*}{Step} &  \multirow{2}{*}{Post-} &  \multirow{2}{*}{\xmark} &  \multirow{2}{*}{15B}  &  \multirow{2}{*}{Weighted} &  \multirow{2}{*}{\cmark} &  \multirow{2}{*}{Ovlp} &  13.04 &  10.57 &  16.66 &  47.7 &  55.1 &  73.2 &  55.6 &  54.5 &  29.1 &  37.2 &  50.4 & \textcolor{custom_red}{\textbf{--\,1.3}} \\
     &  &  &  & &  &  &   &  & & &  13.04 &  10.54 &  16.66 &  48.3 &  54.9 &  72.1 &  55.9 &  54.3 &  28.4 &  35.4 &  49.9 & - \\
     \addlinespace[-1pt]
     \cmidrule(l{2pt}r{2pt}){12-23} 
     \addlinespace[-1pt]
     \rowcolor[gray]{0.9}
      &  &  &  & &  &   &   &  & &    &  12.40 &  9.97 &  15.42 &  52.9 &  58.9 &  73.7 &  55.7 &  57.5 &  31.1 &  38.2 &  52.6 & \textcolor{custom_green}{\textbf{+0.9}} \\
    \rowcolor[gray]{0.9}
     \multirow{-2}{*}{0.99B} & \multirow{-2}{*}{\cmark} &  \multirow{-2}{*}{15B} & \multirow{-2}{*}{2} & \multirow{-2}{*}{Step} & \multirow{-2}{*}{Post-} & \multirow{-2}{*}{\xmark} &  \multirow{-2}{*}{15B}  & \multirow{-2}{*}{Agg\,(0.1)} & \multirow{-2}{*}{\cmark} & \multirow{-2}{*}{Ovlp} &  14.11 &  11.47 &  18.32 &  46.3 &  52.1 &  71.6 &  55.3 &  49.2 &  28.5 &  32.6 &  48.0 & - \\
     \midrule
     \multirow{2}{*}{0.99B} & \multirow{2}{*}{\cmark} & \multirow{2}{*}{15B} & \multirow{2}{*}{2} & \multirow{2}{*}{Step} &  \multirow{2}{*}{Post-} &  \multirow{2}{*}{\cmark} &  \multirow{2}{*}{15B}  &  \multirow{2}{*}{Standard} &  \multirow{2}{*}{\xmark} &  \multirow{2}{*}{Ovlp} & {12.85} & {10.29} & {16.21} & {53.0} & {57.3} & {73.2} & {56.2} & {56.1} & {29.2} & {36.6} & {51.7} & \textcolor{gray}{\textbf{+0.0}} \\
     &  &  &  & &  &  &   &  & & & 43.74 &  41.63 &  56.78 &  5.3 &  37.9 &  61.4 &  52.6 &  35.3 &  24.0 &  29.0 &  35.0 & - \\
     \addlinespace[-1pt]
     \cmidrule(l{2pt}r{2pt}){12-23} 
     \addlinespace[-1pt]
     \multirow{2}{*}{0.99B} & \multirow{2}{*}{\cmark} & \multirow{2}{*}{15B}  & \multirow{2}{*}{2} & \multirow{2}{*}{Step} &  \multirow{2}{*}{Post-} &  \multirow{2}{*}{\cmark} &  \multirow{2}{*}{15B}  &  \multirow{2}{*}{Standard} &  \multirow{2}{*}{\cmark} &  \multirow{2}{*}{Ovlp} & {12.85} & {10.29} & {16.21} & {53.0} & {57.3} & {73.2} & {56.2} & {56.1} & {29.2} & {36.6} & {51.7} & \textcolor{gray}{\textbf{+0.0}} \\
     &  &  &  & &  &  &   &  & &  &  43.09 &  39.97 &  55.37 &  5.6 &  37.7 &  62.5 &  52.7 &  34.5 &  24.7 &  29.2 &  35.3 & - \\
     \midrule
     \multirow{2}{*}{0.99B} & \multirow{2}{*}{\cmark} & \multirow{2}{*}{15B}  & \multirow{2}{*}{2} & \multirow{2}{*}{Step} &  \multirow{2}{*}{Co-} &  \multirow{2}{*}{\xmark} &  \multirow{2}{*}{15B}  &  \multirow{2}{*}{Agg\,(0.1)} &  \multirow{2}{*}{\xmark} &  \multirow{2}{*}{Ovlp} &  13.24 &  10.67 &  16.98 &  50.1 &  54.2 &  72.2 &  53.7 &  54.7 &  28.9 &  37.4 &  50.2 & \textcolor{custom_red}{\textbf{--\,1.5}} \\
     &  &  &  & &  &  &   &  & &  &  13.59 &  10.89 &  17.42 &  50.6 &  52.7 &  71.2 &  54.4 &  53.0 &  27.5 &  35.0 &  49.2 & - \\
     \midrule
     \rowcolor[gray]{0.9}
      &  &  &  &  &  &    &  &   &  & &  12.34 &  9.92 &  15.31 &  52.3 &  59.0 &  73.8 &  57.6 &  55.5 &  30.4 &  37.2 &  52.3 & \textcolor{custom_green}{\textbf{+0.6}} \\
     \rowcolor[gray]{0.9}
    \multirow{-2}{*}{0.99B} & \multirow{-2}{*}{\cmark} &  \multirow{-2}{*}{15B} & \multirow{-2}{*}{2} & \multirow{-2}{*}{Step} & \multirow{-2}{*}{Post-} & \multirow{-2}{*}{\xmark} &  \multirow{-2}{*}{15B}  & \multirow{-2}{*}{Agg\,(0.1)} & \multirow{-2}{*}{\xmark} & \multirow{-2}{*}{New} &  14.49 &  11.86 &  18.89 &  43.9 &  51.3 &  71.0 &  54.9 &  48.1 &  27.5 &  31.4 &  46.9 & - \\
    \bottomrule
    \end{tabular}
    }
    \caption{
    Ablation studies on early-exit training for recursive Gemma models. 
    We evaluated performance in a static-exiting scenario\,\citep{DBLP:conf/nips/SchusterFG0B0TM22, DBLP:conf/emnlp/BaeKSY23}, where all tokens exit at either first or second iteration loops (9th or 18th depths).
    We explored post-training (after uptraining) and co-training (during uptraining) approaches. Moreover, we explored freezing uptrained weights and adding LoRA with the rank of 128 to the classifier head. Different coefficient values were tested for the aggressive CE loss function. Early-exit training utilized 15 billion tokens, either overlapping with uptraining data or entirely new.
    Delta\,($\Delta$) indicates the performance changes of the final loop outputs.
    We highlight the final configuration: post-training with aggressive CE and KD loss on 15 billion new tokens.
    }
    \label{tab:early_exit_ablation_app}
\end{table}

\begin{table}[ht!]
    \small
    \centering
    \resizebox{\textwidth}{!}{
    \setlength{\tabcolsep}{3pt}
    \begin{tabular}{c|ccc|cc|cc|ccc|rrr|ccccccc|cc}
    \toprule
     & \multicolumn{3}{c|}{\textbf{Uptrain}} & \multicolumn{2}{c|}{\textbf{Looping}} &  \multicolumn{2}{c|}{\textbf{LoRA}} &  \multicolumn{3}{c|}{\textbf{Early-exit\,Train}} & \multicolumn{3}{c|}{\textbf{Perplexity\,$\downarrow$}} & \multicolumn{9}{c}{\textbf{Few-shot Accuracy\,$\uparrow$}} \\
    \cmidrule(l{2pt}r{2pt}){2-4} \cmidrule(l{2pt}r{2pt}){5-6}  \cmidrule(l{2pt}r{2pt}){7-8} \cmidrule(l{2pt}r{2pt}){9-11} \cmidrule(l{2pt}r{2pt}){12-14}  \cmidrule(l{2pt}r{2pt}){15-23} 
     N-emb & PT & $N_{tok}$ & KD & Block & Init  & Rank & Init & $N_{tok}$ & CE & KD & SlimP & RedP & PG19 & LD & HS & PQ & WG & ARC-e & ARC-c & OB & Avg & $\Delta$  \\
    \midrule
    1.99B & \cmark & 60B & \xmark & - & -  & - & - &  - & -  & - &  10.58 & 8.44 & 12.71 & 60.3 & 67.9 & 76.9 & 63.5 & 64.9 & 37.2 & {39.6} & 58.6  & - \\
    1.99B & \cmark & 75B & \xmark & - & -  & - & - &  - & -  & - &  11.03 &  8.88 &  13.33 &  57.0 &  65.9 &  76.2 &  63.9 &  63.0 &  35.9 &  38.8 &  57.3 & \textcolor{custom_red}{\textbf{--\,1.3}} \\
    \midrule
    0.99B & \cmark & 60B & \cmark & 2 & Step  & - & - &  - & -  & -&  11.44 &  9.14 &  13.98 &  56.5 &  62.1 &  75.2 &  59.4 &  59.8 &  32.5 &  38.6 &  54.9 & - \\
    1.07B & \cmark & 60B & \cmark & 2 & Avg  & 64 & SVD &  - & - & - &  11.36 &  9.14 &  13.82 &  58.9 &  62.8 &  75.1 &  61.5 &  61.2 &  33.7 &  37.6 &  55.8 & - \\
    1.15B & \cmark & 60B & \cmark & 2 & Avg  & 128 & SVD &  - & - & - &  11.25 &  9.04 &  13.64 &  58.7 &  63.6 &  76.5 &  61.2 &  62.6 &  34.6 &  39.0 &  56.6 & - \\
    1.30B & \cmark & 60B & \cmark & 2 & Avg  & 256 & SVD &  - & - & - &  11.05 &  8.88 &  13.35 &  60.6 &  64.7 &  75.3 &  62.5 &  61.6 &  35.3 &  38.8 &  57.0 & - \\
    1.60B & \cmark & 60B & \cmark & 2 & Avg  & 512 & SVD &  - & - & - &  {10.81} &  {8.63} &  {12.94} &  61.4 &  {65.8} &  {76.3} & { 63.5} &  {65.1} &  {37.1} &  39.4 & {58.4} & - \\
    \midrule
       & & & & & & & & &  & &  11.71 &  9.56 &  14.46 &  54.0 &  61.7 &  75.1 &  58.9 &  58.6 &  31.9 &  37.6 &  54.0 & \textcolor{custom_red}{\textbf{--\,0.9}} \\
    \multirow{-2}{*}{0.99B} & \multirow{-2}{*}{\cmark} & \multirow{-2}{*}{60B} & \multirow{-2}{*}{\cmark} & \multirow{-2}{*}{2} & \multirow{-2}{*}{Step} & \multirow{-2}{*}{-} & \multirow{-2}{*}{-} & \multirow{-2}{*}{15B}  & \multirow{-2}{*}{Agg\,(0.1)} & \multirow{-2}{*}{\cmark}  &  13.68 &  11.39 &  17.60 &  45.0 &  54.1 &  71.9 &  58.5 &  49.8 &  28.8 &  33.4 &  48.8 & - \\
    \addlinespace[-1pt] \cmidrule(l{2pt}r{2pt}){12-23} \addlinespace[-1pt]
      & & & & & & & & & & & 11.79 &  9.70 &  14.52 &  53.7 &  60.8 &  73.6 &  61.1 &  58.7 &  32.9 &  37.2 &  54.0 &\textcolor{custom_red}{\textbf{--\,1.8}} \\
    \multirow{-2}{*}{1.07B} & \multirow{-2}{*}{\cmark} & \multirow{-2}{*}{60B} & \multirow{-2}{*}{\cmark} & \multirow{-2}{*}{2} & \multirow{-2}{*}{Avg} & \multirow{-2}{*}{64} & \multirow{-2}{*}{SVD}  & \multirow{-2}{*}{15B}  & \multirow{-2}{*}{Agg\,(0.1)} & \multirow{-2}{*}{\cmark}  &  19.45 &  16.46 &  26.10 &  30.7 &  37.9 &  66.5 &  55.3 &  42.2 &  25.3 &  27.6 &  40.8 & - \\
    \addlinespace[-1pt] \cmidrule(l{2pt}r{2pt}){12-23} \addlinespace[-1pt]
      & & & & & & & & & & &  11.66 &  9.59 &  14.32 &  53.3 &  62.1 &  74.9 &  60.0 &  59.9 &  33.4 &  38.8 &  54.6 &\textcolor{custom_red}{\textbf{--\,2.0}} \\
    \multirow{-2}{*}{1.15B} & \multirow{-2}{*}{\cmark} & \multirow{-2}{*}{60B} & \multirow{-2}{*}{\cmark} & \multirow{-2}{*}{2} & \multirow{-2}{*}{Avg} & \multirow{-2}{*}{128} & \multirow{-2}{*}{SVD} & \multirow{-2}{*}{15B}   & \multirow{-2}{*}{Agg\,(0.1)} & \multirow{-2}{*}{\cmark}  &  19.65 &  16.77 &  26.44 &  29.7 &  37.7 &  66.8 &  52.6 &  41.4 &  25.3 &  28.0 &  40.2 & - \\
    \addlinespace[-1pt] \cmidrule(l{2pt}r{2pt}){12-23} \addlinespace[-1pt]
      & & & & & & & & & & &  11.47 &  9.39 &  14.03 &  54.9 &  63.0 &  74.5 &  61.7 &  60.5 &  33.1 &  38.4 &  55.2 & \textcolor{custom_red}{\textbf{--\,1.8}} \\
    \multirow{-2}{*}{1.30B} & \multirow{-2}{*}{\cmark} & \multirow{-2}{*}{60B} & \multirow{-2}{*}{\cmark} & \multirow{-2}{*}{2} & \multirow{-2}{*}{Avg} & \multirow{-2}{*}{256} & \multirow{-2}{*}{SVD} & \multirow{-2}{*}{15B}   & \multirow{-2}{*}{Agg\,(0.1)} & \multirow{-2}{*}{\cmark}  &  19.67 &  16.82 &  26.40 &  29.7 &  38.3 &  66.4 &  53.1 &  43.8 &  24.7 &  27.6 &  40.5 & - \\
    \addlinespace[-1pt] \cmidrule(l{2pt}r{2pt}){12-23} \addlinespace[-1pt]
      & & & & & & & & & & &  11.20 &  9.14 &  13.58 &  57.2 &  64.1 &  75.2 &  61.7 &  62.1 &  34.6 &  38.2 &  56.2 & \textcolor{custom_red}{\textbf{--\,2.2}} \\
    \multirow{-2}{*}{1.60B} & \multirow{-2}{*}{\cmark} & \multirow{-2}{*}{60B} & \multirow{-2}{*}{\cmark} & \multirow{-2}{*}{2} & \multirow{-2}{*}{Avg} & \multirow{-2}{*}{512} & \multirow{-2}{*}{SVD}  & \multirow{-2}{*}{15B}  & \multirow{-2}{*}{Agg\,(0.1)} & \multirow{-2}{*}{\cmark}  &  19.29 &  16.47 &  25.73 &  32.0 &  39.6 &  67.6 &  53.3 &  43.2 &  25.8 &  30.2 &  41.7 & -\\
    \midrule
       & & & & & & & & & & & 12.11 &  9.98 &  14.97 &  52.6 &  59.8 &  74.4 &  59.4 &  57.6 &  31.1 &  37.0 &  53.1 & \textcolor{custom_red}{\textbf{--\,2.7}}\\
    \multirow{-2}{*}{1.07B} & \multirow{-2}{*}{\cmark} & \multirow{-2}{*}{60B} & \multirow{-2}{*}{\cmark} & \multirow{-2}{*}{2} & \multirow{-2}{*}{Avg} & \multirow{-2}{*}{64} & \multirow{-2}{*}{SVD} & \multirow{-2}{*}{15B}   & \multirow{-2}{*}{Agg\,(0.3)} & \multirow{-2}{*}{\cmark}  &  16.09 &  13.54 &  21.19 &  35.4 &  42.8 &  69.8 &  52.8 &  45.8 &  25.8 &  31.0 &  43.3 & -\\
    \addlinespace[-1pt] \cmidrule(l{2pt}r{2pt}){12-23} \addlinespace[-1pt]
      & & & & & & & & & & & 11.96 &  9.87 &  14.76 &  52.3 &  60.5 &  74.2 &  59.1 &  58.9 &  33.0 &  37.2 &  53.6 & \textcolor{custom_red}{\textbf{--\,3.0}} \\
    \multirow{-2}{*}{1.15B} & \multirow{-2}{*}{\cmark} & \multirow{-2}{*}{60B} & \multirow{-2}{*}{\cmark} & \multirow{-2}{*}{2} & \multirow{-2}{*}{Avg} & \multirow{-2}{*}{128} & \multirow{-2}{*}{SVD}  & \multirow{-2}{*}{15B}  & \multirow{-2}{*}{Agg\,(0.3)} & \multirow{-2}{*}{\cmark}  &  16.28 &  13.77 &  21.45 &  35.2 &  42.1 &  69.8 &  53.5 &  46.5 &  25.8 &  31.2 &  43.4 & -\\
    \addlinespace[-1pt] \cmidrule(l{2pt}r{2pt}){12-23} \addlinespace[-1pt]
      & & & & & & & & & & & 11.73 &  9.63 &  14.43 &  54.3 &  61.4 &  75.0 &  60.7 &  58.8 &  33.1 &  38.6 &  54.6 & \textcolor{custom_red}{\textbf{--\,2.4}} \\
    \multirow{-2}{*}{1.30B} & \multirow{-2}{*}{\cmark} & \multirow{-2}{*}{60B} & \multirow{-2}{*}{\cmark} & \multirow{-2}{*}{2} & \multirow{-2}{*}{Avg} & \multirow{-2}{*}{256} & \multirow{-2}{*}{SVD}  & \multirow{-2}{*}{15B}  & \multirow{-2}{*}{Agg\,(0.3)} & \multirow{-2}{*}{\cmark}  &  16.41 &  13.89 &  21.68 &  35.6 &  42.3 &  69.0 &  52.7 &  46.8 &  26.4 &  29.8 &  43.2 & - \\
    \addlinespace[-1pt] \cmidrule(l{2pt}r{2pt}){12-23} \addlinespace[-1pt]
      & & & & & & & & & & & 11.47 &  9.36 &  13.93 &  56.2 &  62.7 &  75.4 &  60.9 &  60.4 &  34.0 &  37.0 &  55.2  & \textcolor{custom_red}{\textbf{--\,3.2}}\\
    \multirow{-2}{*}{1.60B} & \multirow{-2}{*}{\cmark} & \multirow{-2}{*}{60B} & \multirow{-2}{*}{\cmark} & \multirow{-2}{*}{2} & \multirow{-2}{*}{Avg} & \multirow{-2}{*}{512} & \multirow{-2}{*}{SVD}  & \multirow{-2}{*}{15B}  & \multirow{-2}{*}{Agg\,(0.3)} & \multirow{-2}{*}{\cmark}  &  16.24 &  13.72 &  21.42 &  37.8 &  43.6 &  69.0 &  54.4 &  45.5 &  26.4 &  31.2 &  44.0 & - \\
    \midrule
    0.66B & \cmark & 60B & \cmark & 3 & Step  & - & - &  - & - & - & 12.27 &  9.90 &  15.24 &  55.6 &  58.1 &  73.1 &  60.2 &  58.8 &  30.2 &  37.2 &  53.3 & -\\
    0.74B & \cmark & 60B & \cmark & 3 & Avg  & 64 & SVD &  - & - & - & 12.13 &  9.80 &  14.95 &  55.5 &  58.3 &  73.5 &  60.1 &  58.0 &  31.1 &  36.8 &  53.3 & - \\
    0.82B & \cmark & 60B & \cmark & 3 & Avg  & 128 & SVD &  - & - & - & 11.83 &  9.53 &  14.51 &  56.7 &  60.2 &  74.2 &  59.8 &  59.1 &  33.0 &  35.4 &  54.1 & -\\
    0.97B & \cmark & 60B & \cmark & 3 & Avg  & 256 & SVD &  - & - & - & 11.43 &  9.17 &  13.87 &  59.3 &  62.6 &  74.7 &  61.2 &  61.6 &  32.9 &  {40.2} &  56.1 & - \\
    1.27B & \cmark & 60B & \cmark & 3 & Avg  & 512 & SVD &  - & - & - & 11.01 &  8.80 &  13.25 &  {61.5} &  64.9 &  76.2 &  62.0 &  64.3 &  35.6 &  39.2 &  57.7  & - \\
    \midrule
      & & & & & & & & & & & 12.75 &  10.48 &  16.01 &  50.2 &  57.0 &  72.7 &  58.6 &  56.7 &  30.0 &  38.2 &  51.9 & \textcolor{custom_red}{\textbf{--\,1.4}} \\
      & & & & & & & & & & & 13.81 &  11.47 &  17.80 &  48.4 &  53.0 &  72.4 &  55.6 &  51.6 &  27.2 &  35.2 &  49.0 & - \\
    \multirow{-3}{*}{0.66B} & \multirow{-3}{*}{\cmark} & \multirow{-3}{*}{60B} & \multirow{-3}{*}{\cmark} & \multirow{-3}{*}{3} & \multirow{-3}{*}{Step} & \multirow{-3}{*}{-} & \multirow{-3}{*}{-}  & \multirow{-3}{*}{15B}  & \multirow{-3}{*}{Agg\,(0.1)} & \multirow{-3}{*}{\cmark}  &  16.72 &  14.23 &  22.97 &  37.7 &  44.2 &  69.8 &  53.6 &  44.2 &  24.6 &  30.2 &  43.5 & -\\
    \addlinespace[-1pt] \cmidrule(l{2pt}r{2pt}){12-23} \addlinespace[-1pt]
      & & & & & & & & & & & 12.64 &  10.43 &  15.81 &  51.4 &  56.3 &  72.2 &  57.9 &  56.7 &  30.4 &  35.0 &  51.4 & \textcolor{custom_red}{\textbf{--\,1.9}} \\
      & & & & & & & & & & & 19.90 &  16.88 &  26.26 &  30.4 &  39.3 &  66.3 &  54.1 &  41.2 &  24.8 &  29.2 &  40.8 & - \\
    \multirow{-3}{*}{0.74B} & \multirow{-3}{*}{\cmark} & \multirow{-3}{*}{60B} & \multirow{-3}{*}{\cmark} & \multirow{-3}{*}{3} & \multirow{-3}{*}{Avg} & \multirow{-3}{*}{64} & \multirow{-3}{*}{SVD}  & \multirow{-3}{*}{15B} & \multirow{-3}{*}{Agg\,(0.1)} & \multirow{-3}{*}{\xmark}  &  26.31 &  22.49 &  36.10 &  20.9 &  31.2 &  62.6 &  50.8 &  37.2 &  22.0 &  28.0 &  36.1 & - \\
    \addlinespace[-1pt] \cmidrule(l{2pt}r{2pt}){12-23} \addlinespace[-1pt]
      & & & & & & & & & & &  12.37 &  10.21 &  15.38 &  52.0 &  58.0 &  72.0 &  56.5 &  58.4 &  30.0 &  35.2 &  51.7 & \textcolor{custom_red}{\textbf{--\,2.4}}\\
      & & & & & & & & & & & 20.07 &  17.09 &  26.47 &  30.9 &  40.5 &  66.3 &  55.4 &  40.8 &  24.4 &  29.6 &  41.1 & - \\
    \multirow{-3}{*}{0.82B} & \multirow{-3}{*}{\cmark} & \multirow{-3}{*}{60B} & \multirow{-3}{*}{\cmark} & \multirow{-3}{*}{3} & \multirow{-3}{*}{Avg} & \multirow{-3}{*}{128} & \multirow{-3}{*}{SVD}  & \multirow{-3}{*}{15B} & \multirow{-3}{*}{Agg\,(0.1)} & \multirow{-3}{*}{\xmark}  &  26.15 &  22.46 &  35.98 &  21.3 &  31.2 &  62.7 &  51.8 &  36.4 &  22.9 &  26.2 &  36.1 & - \\
    \addlinespace[-1pt] \cmidrule(l{2pt}r{2pt}){12-23} \addlinespace[-1pt]
      & & & & & & & & & & & 11.92 &  9.78 &  14.71 &  54.8 &  60.6 &  74.6 &  60.1 &  60.1 &  31.8 &  36.6 &  54.1 & \textcolor{custom_red}{\textbf{--\,2.0}}\\
      & & & & & & & & & & & 19.29 &  16.49 &  25.51 &  35.2 &  42.5 &  65.8 &  55.6 &  41.5 &  25.6 &  29.4 &  42.2 & - \\
    \multirow{-3}{*}{0.97B} & \multirow{-3}{*}{\cmark} & \multirow{-3}{*}{60B} & \multirow{-3}{*}{\cmark} & \multirow{-3}{*}{3} & \multirow{-3}{*}{Avg} & \multirow{-3}{*}{256} & \multirow{-3}{*}{SVD}  & \multirow{-3}{*}{15B} & \multirow{-3}{*}{Agg\,(0.1)} & \multirow{-3}{*}{\xmark}  &  25.12 &  21.53 &  34.53 &  23.1 &  32.0 &  63.2 &  49.7 &  36.1 &  23.0 &  25.2 &  36.1 & - \\
    \addlinespace[-1pt] \cmidrule(l{2pt}r{2pt}){12-23} \addlinespace[-1pt]
      & & & & & & & & & & & 11.49 &  9.38 &  14.00 &  56.1 &  62.7 &  74.4 &  60.5 &  62.1 &  34.9 &  38.8 &  55.7 & \textcolor{custom_red}{\textbf{--\,3.0}}\\
      & & & & & & & & & & & 18.52 &  15.79 &  24.34 &  36.7 &  44.9 &  67.2 &  55.3 &  43.8 &  26.0 &  30.4 &  43.5 & - \\
    \multirow{-3}{*}{1.27B} & \multirow{-3}{*}{\cmark} & \multirow{-3}{*}{60B} & \multirow{-3}{*}{\cmark} & \multirow{-3}{*}{3} & \multirow{-3}{*}{Avg} & \multirow{-3}{*}{512} & \multirow{-3}{*}{SVD}  & \multirow{-3}{*}{15B} & \multirow{-3}{*}{Agg\,(0.1)} & \multirow{-3}{*}{\xmark}  &  24.19 &  20.70 &  33.20 &  24.4 &  32.4 &  63.9 &  50.8 &  37.9 &  21.9 &  27.4 &  37.0 & - \\
    \midrule
      & & & & & & & & & & & 13.07 &  10.84 &  16.49 &  47.7 &  54.4 &  71.7 &  56.1 &  55.9 &  29.4 &  35.2 &  50.1 & \textcolor{custom_red}{\textbf{--\,3.2}}\\
      & & & & & & & & & & & 16.68 &  14.08 &  21.86 &  35.4 &  42.4 &  68.2 &  53.8 &  44.6 &  26.3 &  29.4 &  42.9 & - \\
    \multirow{-3}{*}{0.74B} & \multirow{-3}{*}{\cmark} & \multirow{-3}{*}{60B} & \multirow{-3}{*}{\cmark} & \multirow{-3}{*}{3} & \multirow{-3}{*}{Avg} & \multirow{-3}{*}{64} & \multirow{-3}{*}{SVD} & \multirow{-3}{*}{15B}  & \multirow{-3}{*}{Agg\,(0.3)} & \multirow{-3}{*}{\xmark}  &  21.43 &  18.26 &  29.12 &  24.4 &  34.1 &  64.3 &  50.5 &  40.7 &  22.3 &  27.8 &  37.7 & - \\
    \addlinespace[-1pt] \cmidrule(l{2pt}r{2pt}){12-23} \addlinespace[-1pt]
      & & & & & & & & & & & 12.71 &  10.54 &  15.92 &  50.4 &  55.9 &  73.1 &  57.5 &  56.8 &  30.1 &  34.8 &  51.2 & \textcolor{custom_red}{\textbf{--\,2.9}}\\
      & & & & & & & & & & & 16.90 &  14.32 &  22.18 &  37.6 &  43.5 &  67.6 &  54.5 &  45.0 &  25.3 &  29.0 &  43.2 & - \\
    \multirow{-3}{*}{0.82B} & \multirow{-3}{*}{\cmark} & \multirow{-3}{*}{60B} & \multirow{-3}{*}{\cmark} & \multirow{-3}{*}{3} & \multirow{-3}{*}{Avg} & \multirow{-3}{*}{128} & \multirow{-3}{*}{SVD}  & \multirow{-3}{*}{15B} & \multirow{-3}{*}{Agg\,(0.3)} & \multirow{-3}{*}{\xmark}  &  21.23 &  18.13 &  28.88 &  25.3 &  34.0 &  64.6 &  51.7 &  40.7 &  23.0 &  26.4 &  38.0 & - \\
    \addlinespace[-1pt] \cmidrule(l{2pt}r{2pt}){12-23} \addlinespace[-1pt]
      & & & & & & & & & & & 12.26 &  10.15 &  15.23 &  53.5 &  58.5 &  73.5 &  58.8 &  58.3 &  30.6 &  37.6 &  53.0 & \textcolor{custom_red}{\textbf{--\,3.1}}\\
      & & & & & & & & & & & 16.56 &  14.09 &  21.68 &  42.6 &  45.1 &  68.2 &  57.7 &  45.7 &  25.9 &  28.8 &  44.8 & - \\
    \multirow{-3}{*}{0.97B} & \multirow{-3}{*}{\cmark} & \multirow{-3}{*}{60B} & \multirow{-3}{*}{\cmark} & \multirow{-3}{*}{3} & \multirow{-3}{*}{Avg} & \multirow{-3}{*}{256} & \multirow{-3}{*}{SVD}  & \multirow{-3}{*}{15B} & \multirow{-3}{*}{Agg\,(0.3)} & \multirow{-3}{*}{\xmark}  &  20.78 &  17.72 &  28.29 &  27.9 &  34.3 &  66.3 &  52.2 &  39.7 &  23.6 &  26.8 &  38.7 & - \\
    \addlinespace[-1pt] \cmidrule(l{2pt}r{2pt}){12-23} \addlinespace[-1pt]
      & & & & & & & & & & & 11.80 &  9.68 &  14.45 &  54.1 &  61.2 &  74.0 &  59.0 &  59.9 &  32.9 &  38.0 &  54.1 & \textcolor{custom_red}{\textbf{--\,3.6}}\\
      & & & & & & & & & & & 16.02 &  13.53 &  20.86 &  43.5 &  47.5 &  68.3 &  56.2 &  47.1 &  27.0 &  30.4 &  45.7 & - \\
    \multirow{-3}{*}{1.27B} & \multirow{-3}{*}{\cmark} & \multirow{-3}{*}{60B} & \multirow{-3}{*}{\cmark} & \multirow{-3}{*}{3} & \multirow{-3}{*}{Avg} & \multirow{-3}{*}{512} & \multirow{-3}{*}{SVD}  & \multirow{-3}{*}{15B} & \multirow{-3}{*}{Agg\,(0.3)} & \multirow{-3}{*}{\xmark}  &  20.20 &  17.21 &  27.50 &  28.9 &  35.2 &  65.6 &  52.9 &  41.9 &  23.2 &  26.6 &  39.2 & - \\
    \bottomrule
    \end{tabular}
    }
    \caption{
    Evaluation results of Gemma models after early-exit training. 
    Delta\,($\Delta$) represent the accuracy changes in original last loop outputs after early-exit post-training. These changes should be compared in reference to the performance drops observed in 75B and 60B uptraining for the full-size model.
    The relaxed model with three blocks shows a more significant performance drop because KD loss (from final loop output) could not be utilized due to out-of-memory issues.
    }
    \label{tab:final_early_exit_gemma_app}
\end{table}

\begin{table}[ht!]
    \small
    \centering
    \resizebox{\textwidth}{!}{
    \setlength{\tabcolsep}{3pt}
    \begin{tabular}{l|c|ccc|cc|cc|ccc|rrr|ccccccc|cc}
    \toprule
     & & \multicolumn{3}{c|}{\textbf{Uptrain}} & \multicolumn{2}{c|}{\textbf{Looping}} &  \multicolumn{2}{c|}{\textbf{LoRA}} &  \multicolumn{3}{c|}{\textbf{Early-exit\,Train}} & \multicolumn{3}{c|}{\textbf{Perplexity\,$\downarrow$}} & \multicolumn{9}{c}{\textbf{Few-shot Accuracy\,$\uparrow$}} \\
    \cmidrule(l{2pt}r{2pt}){3-5} \cmidrule(l{2pt}r{2pt}){6-7}  \cmidrule(l{2pt}r{2pt}){8-9} \cmidrule(l{2pt}r{2pt}){10-12} \cmidrule(l{2pt}r{2pt}){13-15}  \cmidrule(l{2pt}r{2pt}){16-24} 
    \textbf{Models} & N-emb & PT & $N_{tok}$ & KD & Block & Init  & Rank & Init & $N_{tok}$ & CE & KD & SlimP & RedP & PG19 & LD & HS & PQ & WG & ARC-e & ARC-c & OB & Avg & $\Delta$  \\
    \midrule
    & 0.97B & \cmark & - & \xmark & - & -  & - & - &  - & - & - &  12.26 & 9.37 & 11.94 & 43.3 & 42.2 & 66.8 & 53.4 & 44.7 & 23.2 & 29.2 & 43.3  & - \\
    \cmidrule(l{2pt}r{2pt}){2-24}
      & 0.48B & \cmark & 60B & \cmark & 2 & Step  & - & - &  - & - & - &  10.51 &  9.01 &  11.60 &  44.2 &  43.1 &  68.2 &  52.4 &  44.7 &  25.3 &  32.2 &  44.3 & - \\
    & 0.53B & \cmark & 60B & \cmark & 2 & Avg  & 64 & SVD &  - & - & - & 10.14 &  8.77 &  11.19 &  44.3 &  44.9 &  69.5 &  52.5 &  46.5 &  26.1 &  {31.6} &  45.0 & - \\
    & 0.58B & \cmark & 60B & \cmark & 2 & Avg  & 128 & SVD &  - & - & - & 10.07 &  8.68 &  11.07 &  45.9 &  45.1 &  69.4 &  50.5 &  46.8 &  25.4 &  {31.6} &  45.0 & - \\
    & 0.68B & \cmark & 60B & \cmark & 2 & Avg  & 256 & SVD &  - & - & - & 9.96 &  8.56 &  10.93 &  46.2 &  45.7 &  69.0 &  53.2 &  {47.9} &  25.9 &  {31.6} &  45.6 & - \\
    & 0.86B & \cmark & 60B & \cmark & 2 & Avg  & 512 & SVD &  - & - & - & {9.85} &  {8.44} &  {10.76} &  {47.4} &  {46.3} &  {69.7} &  52.8 &  47.5 &  {26.3} &  31.4 &  {45.9} & - \\
    \cmidrule(l{2pt}r{2pt}){2-24}
      & & & & & & & & & & & & 10.55 &  9.16 &  11.68 &  45.0 &  43.7 &  68.9 &  53.4 &  44.8 &  25.3 &  32.2 &  44.8 & \textcolor{custom_green}{\textbf{+\,0.5}} \\
     & \multirow{-2}{*}{0.48B} & \multirow{-2}{*}{\cmark} & \multirow{-2}{*}{60B} & \multirow{-2}{*}{\cmark} & \multirow{-2}{*}{2} & \multirow{-2}{*}{Step} & \multirow{-2}{*}{-} & \multirow{-2}{*}{-}  & \multirow{-2}{*}{15B} & \multirow{-2}{*}{Agg\,(0.1)} & \multirow{-2}{*}{\cmark}  &  12.28 &  10.62 &  13.83 &  38.2 &  39.4 &  65.8 &  52.3 &  41.5 &  24.7 &  30.6 &  41.8 & -\\
    \addlinespace[-1pt] \cmidrule(l{2pt}r{2pt}){13-24} \addlinespace[-1pt]
     & & & & & & & & & & & & 10.34 &  9.08 &  11.50 &  43.4 &  44.8 &  69.5 &  53.4 &  46.9 &  25.6 &  32.0 &  45.1 & \textcolor{custom_green}{\textbf{+\,0.1}}\\
     & \multirow{-2}{*}{0.53B} & \multirow{-2}{*}{\cmark} & \multirow{-2}{*}{60B} & \multirow{-2}{*}{\cmark} & \multirow{-2}{*}{2} & \multirow{-2}{*}{Avg} & \multirow{-2}{*}{64} & \multirow{-2}{*}{SVD}  & \multirow{-2}{*}{15B} & \multirow{-2}{*}{Agg\,(0.1)} & \multirow{-2}{*}{\cmark}  &  21.23 &  18.63 &  24.85 &  16.8 &  29.0 &  57.6 &  48.9 &  33.2 &  23.1 &  27.0 &  33.7 & - \\
    \addlinespace[-1pt] \cmidrule(l{2pt}r{2pt}){13-24} \addlinespace[-1pt]
     & & & & & & & & & & & & 10.25 &  8.97 &  11.36 &  45.2 &  45.5 &  68.8 &  54.0 &  46.5 &  25.0 &  31.6 &  45.2 & \textcolor{custom_green}{\textbf{+\,0.2}}\\
     & \multirow{-2}{*}{0.58B} & \multirow{-2}{*}{\cmark} & \multirow{-2}{*}{60B} & \multirow{-2}{*}{\cmark} & \multirow{-2}{*}{2} & \multirow{-2}{*}{Avg} & \multirow{-2}{*}{128} & \multirow{-2}{*}{SVD}  & \multirow{-2}{*}{15B} & \multirow{-2}{*}{Agg\,(0.1)} & \multirow{-2}{*}{\cmark}  &  21.30 &  18.56 &  24.75 &  18.5 &  28.9 &  58.4 &  48.0 &  34.1 &  21.8 &  27.4 &  33.9 & - \\
    \addlinespace[-1pt] \cmidrule(l{2pt}r{2pt}){13-24} \addlinespace[-1pt]
    TinyLlama & & & & & & & & & & & &  10.13 &  8.84 &  11.23 &  45.2 &  45.9 &  69.6 &  53.6 &  46.9 &  25.9 &  32.0 &  45.6 & \textcolor{gray}{\textbf{+\,0.0}}\\
     & \multirow{-2}{*}{0.68B} & \multirow{-2}{*}{\cmark} & \multirow{-2}{*}{60B} & \multirow{-2}{*}{\cmark} & \multirow{-2}{*}{2} & \multirow{-2}{*}{Avg} & \multirow{-2}{*}{256} & \multirow{-2}{*}{SVD}  & \multirow{-2}{*}{15B} & \multirow{-2}{*}{Agg\,(0.1)} & \multirow{-2}{*}{\cmark}  &  20.95 &  18.16 &  24.22 &  20.1 &  28.8 &  57.8 &  48.9 &  33.8 &  22.5 &  25.0 &  33.9 & - \\
    \addlinespace[-1pt] \cmidrule(l{2pt}r{2pt}){13-24} \addlinespace[-1pt]
     & & & & & & & & & & & & 10.02 &  8.74 &  11.04 &  46.6 &  46.5 &  68.6 &  54.5 &  47.9 &  26.3 &  32.2 &  46.1 & \textcolor{custom_green}{\textbf{+\,0.2}}\\
     & \multirow{-2}{*}{0.86B} & \multirow{-2}{*}{\cmark} & \multirow{-2}{*}{60B} & \multirow{-2}{*}{\cmark} & \multirow{-2}{*}{2} & \multirow{-2}{*}{Avg} & \multirow{-2}{*}{512} & \multirow{-2}{*}{SVD}  & \multirow{-2}{*}{15B} & \multirow{-2}{*}{Agg\,(0.1)} & \multirow{-2}{*}{\cmark}  &  20.38 &  17.70 &  23.57 &  19.9 &  28.8 &  58.2 &  49.0 &  34.7 &  22.8 &  25.8 &  34.2 & - \\
    \cmidrule(l{2pt}r{2pt}){2-24}
     & & & & & & & & & & & & 10.61 &  9.36 &  11.87 &  42.1 &  43.7 &  68.6 &  54.1 &  46.1 &  26.0 &  31.2 &  44.6 & \textcolor{custom_red}{\textbf{--\,0.4}}\\
     & \multirow{-2}{*}{0.53B} & \multirow{-2}{*}{\cmark} & \multirow{-2}{*}{60B} & \multirow{-2}{*}{\cmark} & \multirow{-2}{*}{2} & \multirow{-2}{*}{Avg} & \multirow{-2}{*}{64} & \multirow{-2}{*}{SVD}  & \multirow{-2}{*}{15B} & \multirow{-2}{*}{Agg\,(0.3)} & \multirow{-2}{*}{\cmark}  &  16.83 &  14.88 &  19.77 &  22.0 &  30.3 &  60.7 &  50.7 &  36.9 &  24.1 &  27.8 &  36.1 & - \\
    \addlinespace[-1pt] \cmidrule(l{2pt}r{2pt}){13-24} \addlinespace[-1pt]
     & & & & & & & & & & & & 10.50 &  9.22 &  11.71 &  44.2 &  44.2 &  69.2 &  53.0 &  46.0 &  25.5 &  31.2 &  44.8 & \textcolor{custom_red}{\textbf{--\,0.2}}\\
     & \multirow{-2}{*}{0.58B} & \multirow{-2}{*}{\cmark} & \multirow{-2}{*}{60B} & \multirow{-2}{*}{\cmark} & \multirow{-2}{*}{2} & \multirow{-2}{*}{Avg} & \multirow{-2}{*}{128} & \multirow{-2}{*}{SVD}  & \multirow{-2}{*}{15B} & \multirow{-2}{*}{Agg\,(0.3)} & \multirow{-2}{*}{\cmark}  &  17.10 &  15.03 &  19.99 &  23.5 &  30.1 &  60.8 &  51.3 &  36.5 &  23.8 &  26.4 &  36.0 & - \\
    \addlinespace[-1pt] \cmidrule(l{2pt}r{2pt}){13-24} \addlinespace[-1pt]
     & & & & & & & & & & & & 10.34 &  9.07 &  11.51 &  44.0 &  45.0 &  68.4 &  53.0 &  45.8 &  26.0 &  31.2 &  44.8 & \textcolor{custom_red}{\textbf{--\,0.8}}\\
     & \multirow{-2}{*}{0.68B} & \multirow{-2}{*}{\cmark} & \multirow{-2}{*}{60B} & \multirow{-2}{*}{\cmark} & \multirow{-2}{*}{2} & \multirow{-2}{*}{Avg} & \multirow{-2}{*}{256} & \multirow{-2}{*}{SVD}  & \multirow{-2}{*}{15B} & \multirow{-2}{*}{Agg\,(0.3)} & \multirow{-2}{*}{\cmark}  &  17.06 &  14.92 &  19.82 &  24.2 &  30.4 &  59.9 &  51.7 &  36.2 &  23.9 &  27.2 &  36.2 & -\\
    \addlinespace[-1pt] \cmidrule(l{2pt}r{2pt}){13-24} \addlinespace[-1pt]
     & & & & & & & & & & & & 10.21 &  8.94 &  11.28 &  45.1 &  45.8 &  69.3 &  54.5 &  46.7 &  25.9 &  33.4 &  45.8 & \textcolor{custom_green}{\textbf{--\,0.1}}\\
     & \multirow{-2}{*}{0.86B} & \multirow{-2}{*}{\cmark} & \multirow{-2}{*}{60B} & \multirow{-2}{*}{\cmark} & \multirow{-2}{*}{2} & \multirow{-2}{*}{Avg} & \multirow{-2}{*}{512} & \multirow{-2}{*}{SVD}  & \multirow{-2}{*}{15B} & \multirow{-2}{*}{Agg\,(0.3)} & \multirow{-2}{*}{\cmark}  &  16.76 &  14.68 &  19.43 &  24.4 &  30.0 &  61.1 &  51.9 &  37.1 &  22.9 &  28.2 &  36.5 & -\\
    \midrule
    & 0.81B & \cmark & 60B & \xmark & - & -  & - & - &  - & - & - & 12.83 & 9.76 & 13.57 &  53.0 & 50.2 & 71.1 & 54.8 & 51.9 & {27.7} & 31.6 & {48.6}  & - \\
    & 0.81B & \cmark & 75B & \xmark & - & -  & - & - &  - & - & - & 12.86 & 9.86 & 13.74 &  54.8 & 50.3 & 70.5 & 55.3 & 52.2 & {28.8} & 33.0 & {49.3}  & \textcolor{custom_green}{\textbf{+\,0.7}} \\
    \cmidrule(l{2pt}r{2pt}){2-24}
    & 0.40B & \cmark & 60B & \cmark & 2 & Step  & - & - &  - & - & - & 14.59 &  11.13 &  15.79 &  47.8 &  43.8 &  69.3 &  52.0 &  48.1 &  25.4 &  30.4 &  45.2 & -\\
    & 0.44B & \cmark & 60B & \cmark & 2 & Avg  & 64 & SVD &  - & - & - & 14.24 &  10.89 &  15.52 &  50.0 &  44.5 &  68.9 &  {54.1} &  48.0 &  26.5 &  31.2 &  46.2 & - \\
    & 0.48B & \cmark & 60B & \cmark & 2 & Avg  & 128 & SVD &  - & - & - & 14.10 &  10.79 &  15.27 &  50.1 &  45.5 &  69.0 &  52.6 &  48.3 &  25.8 &  32.0 &  46.2 & -\\
    & 0.55B & \cmark & 60B & \cmark & 2 & Avg  & 256 & SVD &  - & - & - & 13.91 &  10.61 &  14.91 &  50.5 &  45.6 &  68.7 &  51.2 &  48.4 &  25.7 &  {32.8} &  46.1 & - \\
    & 0.70B & \cmark & 60B & \cmark & 2 & Avg  & 512 & SVD &  - & - & - & {13.59} &  {10.38} &  {14.43} &  {52.0} &  {47.0} &  {69.6} &  53.4 &  {48.9} &  {26.9} &  31.2 &  {47.0}  & - \\
    \cmidrule(l{2pt}r{2pt}){2-24}
      & & & & & & & & & & & & 14.72 &  11.38 &  16.31 &  47.0 &  44.2 &  69.2 &  53.4 &  48.6 &  24.7 &  30.4 &  45.4 & \textcolor{custom_green}{\textbf{+\,0.2}} \\
     & \multirow{-2}{*}{0.40B} & \multirow{-2}{*}{\cmark} & \multirow{-2}{*}{60B} & \multirow{-2}{*}{\cmark} & \multirow{-2}{*}{2} & \multirow{-2}{*}{Step} & \multirow{-2}{*}{-} & \multirow{-2}{*}{-}  & \multirow{-2}{*}{15B} & \multirow{-2}{*}{Agg\,(0.1)} & \multirow{-2}{*}{\cmark}  &  18.61 &  14.11 &  20.96 &  38.4 &  38.1 &  67.0 &  53.7 &  43.3 &  24.4 &  29.0 &  42.0 & -\\
    \addlinespace[-1pt] \cmidrule(l{2pt}r{2pt}){13-24} \addlinespace[-1pt]
     & & & & & & & & & & & & 14.49 &  11.22 &  16.12 &  49.1 &  43.9 &  69.8 &  53.8 &  48.6 &  26.1 &  31.2 &  46.1 & \textcolor{custom_red}{\textbf{--\,0.1}}\\
     & \multirow{-2}{*}{0.44B} & \multirow{-2}{*}{\cmark} & \multirow{-2}{*}{60B} & \multirow{-2}{*}{\cmark} & \multirow{-2}{*}{2} & \multirow{-2}{*}{Avg} & \multirow{-2}{*}{64} & \multirow{-2}{*}{SVD}  & \multirow{-2}{*}{15B} & \multirow{-2}{*}{Agg\,(0.1)} & \multirow{-2}{*}{\cmark}  &  24.43 &  18.19 &  27.89 &  26.7 &  31.6 &  61.6 &  50.8 &  38.2 &  22.9 &  27.6 &  37.1 & - \\
    \addlinespace[-1pt] \cmidrule(l{2pt}r{2pt}){13-24} \addlinespace[-1pt]
     & & & & & & & & & & & & 14.35 &  11.17 &  15.93 &  50.1 &  44.7 &  69.0 &  52.1 &  49.9 &  25.3 &  32.6 &  46.2 & \textcolor{gray}{\textbf{+\,0.0}}\\
     & \multirow{-2}{*}{0.48B} & \multirow{-2}{*}{\cmark} & \multirow{-2}{*}{60B} & \multirow{-2}{*}{\cmark} & \multirow{-2}{*}{2} & \multirow{-2}{*}{Avg} & \multirow{-2}{*}{128} & \multirow{-2}{*}{SVD}  & \multirow{-2}{*}{15B} & \multirow{-2}{*}{Agg\,(0.1)} & \multirow{-2}{*}{\cmark}  &  24.33 &  18.09 &  27.96 &  28.2 &  32.3 &  61.1 &  53.0 &  38.8 &  23.7 &  27.4 &  37.8 & -\\
    \addlinespace[-1pt] \cmidrule(l{2pt}r{2pt}){13-24} \addlinespace[-1pt]
     Pythia & & & & & & & & & & & & 14.14 &  10.96 &  15.54 &  50.8 &  45.5 &  68.2 &  53.9 &  48.8 &  25.3 &  32.8 &  46.5 & \textcolor{custom_green}{\textbf{+\,0.4}}\\
     & \multirow{-2}{*}{0.55B} & \multirow{-2}{*}{\cmark} & \multirow{-2}{*}{60B} & \multirow{-2}{*}{\cmark} & \multirow{-2}{*}{2} & \multirow{-2}{*}{Avg} & \multirow{-2}{*}{256} & \multirow{-2}{*}{SVD}  & \multirow{-2}{*}{15B} & \multirow{-2}{*}{Agg\,(0.1)} & \multirow{-2}{*}{\cmark}  &  24.18 &  17.87 &  27.48 &  28.1 &  32.3 &  61.9 &  54.1 &  38.1 &  22.9 &  28.6 &  38.0 & -\\
    \addlinespace[-1pt] \cmidrule(l{2pt}r{2pt}){13-24} \addlinespace[-1pt]
     & & & & & & & & & & & & 13.81 &  10.72 &  15.11 &  52.4 &  47.0 &  69.3 &  52.7 &  50.1 &  26.9 &  32.0 &  47.2 & \textcolor{custom_green}{\textbf{+\,0.2}}\\
     & \multirow{-2}{*}{0.70B} & \multirow{-2}{*}{\cmark} & \multirow{-2}{*}{60B} & \multirow{-2}{*}{\cmark} & \multirow{-2}{*}{2} & \multirow{-2}{*}{Avg} & \multirow{-2}{*}{512} & \multirow{-2}{*}{SVD}  & \multirow{-2}{*}{15B} & \multirow{-2}{*}{Agg\,(0.1)} & \multirow{-2}{*}{\cmark}  &  23.50 &  17.49 &  26.72 &  29.5 &  32.8 &  63.2 &  52.3 &  38.8 &  22.8 &  27.8 &  38.2 & -\\
    \cmidrule(l{2pt}r{2pt}){2-24}
     & & & & & & & & & & & & 14.87 &  11.53 &  16.61 &  47.0 &  43.1 &  68.7 &  53.0 &  47.4 &  25.7 &  31.0 &  45.1 & \textcolor{custom_red}{\textbf{--\,0.9}}\\
     & \multirow{-2}{*}{0.44B} & \multirow{-2}{*}{\cmark} & \multirow{-2}{*}{60B} & \multirow{-2}{*}{\cmark} & \multirow{-2}{*}{2} & \multirow{-2}{*}{Avg} & \multirow{-2}{*}{64} & \multirow{-2}{*}{SVD}  & \multirow{-2}{*}{15B} & \multirow{-2}{*}{Agg\,(0.3)} & \multirow{-2}{*}{\cmark}  &  20.62 &  15.60 &  23.57 &  32.6 &  33.6 &  63.4 &  51.2 &  40.7 &  23.3 &  28.0 &  39.0 & - \\
    \addlinespace[-1pt] \cmidrule(l{2pt}r{2pt}){13-24} \addlinespace[-1pt]
     & & & & & & & & & & & & 14.69 &  11.46 &  16.36 &  48.9 &  43.8 &  68.4 &  53.0 &  49.1 &  26.2 &  31.6 &  45.9 & \textcolor{custom_red}{\textbf{--\,0.3}}\\
     & \multirow{-2}{*}{0.48B} & \multirow{-2}{*}{\cmark} & \multirow{-2}{*}{60B} & \multirow{-2}{*}{\cmark} & \multirow{-2}{*}{2} & \multirow{-2}{*}{Avg} & \multirow{-2}{*}{128} & \multirow{-2}{*}{SVD}  & \multirow{-2}{*}{15B} & \multirow{-2}{*}{Agg\,(0.3)} & \multirow{-2}{*}{\cmark}  &  20.60 &  15.56 &  23.63 &  33.2 &  33.6 &  62.7 &  51.1 &  41.3 &  23.6 &  27.8 &  39.0 & - \\
    \addlinespace[-1pt] \cmidrule(l{2pt}r{2pt}){13-24} \addlinespace[-1pt]
     & & & & & & & & & & & & 14.44 &  11.20 &  15.94 &  50.0 &  44.7 &  69.2 &  52.3 &  48.1 &  25.4 &  32.2 &  46.0 & \textcolor{custom_red}{\textbf{--\,0.1}}\\
     & \multirow{-2}{*}{0.55B} & \multirow{-2}{*}{\cmark} & \multirow{-2}{*}{60B} & \multirow{-2}{*}{\cmark} & \multirow{-2}{*}{2} & \multirow{-2}{*}{Avg} & \multirow{-2}{*}{256} & \multirow{-2}{*}{SVD}  & \multirow{-2}{*}{15B} & \multirow{-2}{*}{Agg\,(0.3)} & \multirow{-2}{*}{\cmark}  &  20.61 &  15.48 &  23.45 &  33.3 &  34.2 &  63.4 &  52.2 &  40.8 &  23.0 &  28.8 &  39.4 & - \\
    \addlinespace[-1pt] \cmidrule(l{2pt}r{2pt}){13-24} \addlinespace[-1pt]
     & & & & & & & & & & & & 14.08 &  10.94 &  15.44 &  51.1 &  46.4 &  68.7 &  52.2 &  50.0 &  26.9 &  31.6 &  46.7 & \textcolor{custom_red}{\textbf{--\,0.3}} \\
     & \multirow{-2}{*}{0.70B} & \multirow{-2}{*}{\cmark} & \multirow{-2}{*}{60B} & \multirow{-2}{*}{\cmark} & \multirow{-2}{*}{2} & \multirow{-2}{*}{Avg} & \multirow{-2}{*}{512} & \multirow{-2}{*}{SVD}  & \multirow{-2}{*}{15B} & \multirow{-2}{*}{Agg\,(0.3)} & \multirow{-2}{*}{\cmark}  &  20.20 &  15.25 &  22.98 &  34.6 &  34.1 &  63.5 &  53.0 &  41.5 &  23.6 &  27.6 &  39.7 & - \\
    \bottomrule
    \end{tabular}
    }
    \caption{
    Evaluation results of TinyLlama and Pythia models after early-exit training. 
    Delta\,($\Delta$) represents the accuracy change in the original last loop outputs after early-exit post-training. 
    }
    \label{tab:final_early_exit_tinyllama_pythia_app}
\end{table}

\clearpage

\section{Expanded Results of Hypothetical Generation Speedup}
\label{app:hypothetical_generation_speedup}

\paragraph{Measuring the average per-token generation time}
First, we measured the generation time with various model configurations using dummy weights and inputs. We measured the elapsed time for each components, such as embedding matrices, Transformer blocks, and the classifier head. 
Especially, we calculated the time per token by dividing the total time by the decoding length. Default prefix and decoding lengths are set to 512 and 2048, but we also used shorter context lengths, like 64 and 256 to simulate scenarios where parameter memory sizes become limiting.
We measured decoding speed using FlashDecoding~\citep{DBLP:conf/nips/DaoFERR22}, a technique that has recently become standard in serving LLMs.
Using a single A100 40GB GPU, we measured generation times by increasing batch sizes until an out-of-memory error occurred or memory usuage reached the predefined limit.

In Table\,\ref{tab:measured_generation_time_a100}, generation time was measured up to the maximum batch size that a single A100 GPU could accommodate before encountering out-of-memory errors, with prefix and decoding lengths set to 512 and 2048, respectively. Meanwhile, Table\,\ref{tab:measured_generation_time_a100_16gb} presents generation times measured in a more memory-constrained deployment scenario, where the prefix and decoding lengths were reduced to 64 and 256, and the memory limit was set to 16GB.
As anticipated, under severe memory constraints, the reduced parameter memory footprint of Recursive Transformers enabled substantially larger batch sizes. This observation indicates that Recursive Transformers, even without continuous batching techniques, can achieve higher throughput than vanilla Transformers due to their memory efficiency.

When comparing the speed of the three models, Gemma 2B was the fastest, followed by TinyLlama 1.1B and then Pythia 1B. This order is the exact inverse of their non-embedding parameter sizes. This speed difference is attributed to the Grouped-Query~\citep{DBLP:journals/corr/abs-1911-02150} and Multi-Query attention~\citep{DBLP:conf/emnlp/AinslieLJZLS23}. The main decoding bottleneck in Transformers is memory access of heavy key-value caches. Hence, Gemma that effectively reduces the key-value cache size through the MQA mechanism, achieves fastest speeds. Despite using GQA, TinyLlama 1.1B has a similar speed to Gemma 2B due to its shallow and deep architecture (22 layers compared to Gemma's 18 layers). This deeper architecture likely offsets the speed gains from the attention mechanism.

\vspace{-12pt}
\paragraph{Comparison of hypothetical generation throughput}
We conducted early-exiting simulations using language modeling datasets (test sets of SlimPajama, RedPajama, and PG19), assuming our models generated those tokens. We used 20K samples to obtain their exit trajectories, and we employed an oracle-exiting algorithm to determine the earliest possible exit point for each token. 
Combining this trajectory data with previously measured per-token generation times across various batch sizes (considering only Transformer block computations), we estimated the hypothetical throughput for various settings and datasets.
The results are detailed in Tables\,\ref{tab:final_performance_throughput} and \ref{tab:final_performance_throughput_16gb}.

Our analysis reveals that Recursive Transformers achieve a 2-3$\times$ throughput gain over vanilla model counterparts. Relaxed models also demonstrated significant speedup despite their unoptimized LoRA computations. Currently, we merge multiple LoRAs into a single, large LoRA module to enable parallel computation of samples at different looping iterations. 
However, this introduces extra overhead due to redundant computations, resulting in reduced throughput gains in memory-constrained scenarios (shorter context lengths and lower memory limits). This degradation stems from the increased proportion of LoRA computation time relative to overall processing time. Since attention computation has quadratic complexity with respect to sequence length, it becomes less expensive at shorter lengths, while the complexity of LoRA computation remains constant. This highlights the need for highly optimized LoRA computations to achieve substantial throughput gains in all scenarios.  Nevertheless, these findings also suggest that relaxed models will yield even greater performance and throughput improvements with longer contexts where attention computation dominates.

\clearpage

\paragraph{Approximation errors in our hypothetical throughput}
Since our throughput estimations are based on theoretical estimation, they may introduce certain approximation errors as follows:
\begin{itemize}[leftmargin=*]
    \item 
    As our models are not fine-tuned for any downstream task, we simulated the exit trajectories of language modeling datasets by assuming they were generated by our models. While this approach is expected to closely approximate actual generation, empirical validation is necessary to confirm its accuracy.
    \item Throughput gains should be measured using realistic (confidence-based) early-exiting algorithms, rather than relying on the oracle-exiting algorithm. While early-exiting algorithms can introduce performance degradation due to inherent errors in confidence estimation, they also incur additional computational costs for estimating prediction confidence, necessitating further efficiency improvements.
    \item Our analysis solely focused on speed improvements within Transformer blocks. However, upon early exiting, the exited tokens require separate processing through the embedding layer or the classifier head for subsequent sequence generation. This necessitates non-exited tokens to wait for others, potentially reducing efficiency as the embedding layer computation may not fully utilize the maximum batch size.
    \item 
    Early-exiting architectures require computing key-value caches in remaining layers for already exited tokens to prevent performance degradation~\citep{DBLP:conf/emnlp/BaeKSY23}. While this adds negligible overhead in memory-bound scenarios, it inevitably increases overhead in compute-bound scenarios where the maximum batch size is fully utilized.  Our throughput estimation, however, excludes the computation time for these key-value caches in later loops (though we did account for their memory size).  Incorporating these computations into a more realistic analysis of early-exiting generation is a direction for future work.
\end{itemize}

\definecolor{Gray}{gray}{0.9}
\newcolumntype{a}{>{\columncolor{Gray}}c}

\begin{table}[ht!]
    \small
    \centering
    \resizebox{\textwidth}{!}{
    \setlength{\tabcolsep}{6pt}
    \begin{tabular}{l|ccccc|c|cc|c|ccac}
    \toprule
      &  \multicolumn{5}{c|}{\textbf{Model Architecture}} &  & \multicolumn{2}{c|}{\textbf{Recursive}} & & \multicolumn{4}{c}{\textbf{Time\,(ms) per token}} \\
    \cmidrule(l{2pt}r{2pt}){2-6} \cmidrule(l{2pt}r{2pt}){8-9}
     \cmidrule(l{2pt}r{2pt}){11-14}
     \textbf{Models} & $N_L$ & $d_{model}$ & $N_{head}$ & $N_{KV}$ & Vocab & N-emb & Block & Rank & Batch & Total & Emb & \cellcolor{white} Transformer & Head  \\
    \midrule
     & & & & & & & & & 1 & 22.994 & 0.087 & 21.344 & 0.803 \\
     & \multirow{-2}{*}{18} & \multirow{-2}{*}{2048} & \multirow{-2}{*}{8} & \multirow{-2}{*}{1} & \multirow{-2}{*}{256K} & \multirow{-2}{*}{1.98B} &  \multirow{-2}{*}{-}  & \multirow{-2}{*}{-} & \textbf{43} & \,\,\,0.657 & 0.002 & \,\,\,0.616 & 0.023  \\
    \addlinespace[-1pt] \cmidrule(l{2pt}r{2pt}){2-14} \addlinespace[-1pt]
     & & & & & & & & & 1 & 13.918 & 0.088 & 11.059 & 0.827 \\
     & \multirow{-2}{*}{18} & \multirow{-2}{*}{2048} & \multirow{-2}{*}{8} & \multirow{-2}{*}{1} & \multirow{-2}{*}{256K} & \multirow{-2}{*}{0.99B}  & \multirow{-2}{*}{2}  & \multirow{-2}{*}{-} & \textbf{43} & \,\,\,0.336 & 0.002 & \,\,\,0.265 & 0.023  \\
    \addlinespace[-1pt] \cmidrule(l{2pt}r{2pt}){2-14} \addlinespace[-1pt]
     & & & & & & & & & 1 & 15.858 & 0.080 & 13.096 & 0.825 \\
     & \multirow{-2}{*}{18} & \multirow{-2}{*}{2048} & \multirow{-2}{*}{8} & \multirow{-2}{*}{1} & \multirow{-2}{*}{256K} & \multirow{-2}{*}{1.07B}  & \multirow{-2}{*}{2}  & \multirow{-2}{*}{64} & \textbf{41} & \,\,\,0.398 & 0.002 & \,\,\,0.323 & 0.024  \\
    \addlinespace[-1pt] \cmidrule(l{2pt}r{2pt}){2-14} \addlinespace[-1pt]
     & & & & & & & & & 1 & 15.708 & 0.080 & 12.969 & 0.822 \\
     & \multirow{-2}{*}{18} & \multirow{-2}{*}{2048} & \multirow{-2}{*}{8} & \multirow{-2}{*}{1} & \multirow{-2}{*}{256K} & \multirow{-2}{*}{1.15B}  & \multirow{-2}{*}{2} & \multirow{-2}{*}{128} & \textbf{41} & \,\,\,0.398 & 0.002 & \,\,\,0.324 & 0.024  \\
    \addlinespace[-1pt] \cmidrule(l{2pt}r{2pt}){2-14} \addlinespace[-1pt]
     & & & & & & & & & 1 & 15.456 & 0.083 & 12.721 & 0.818 \\
     & \multirow{-2}{*}{18} & \multirow{-2}{*}{2048} & \multirow{-2}{*}{8} & \multirow{-2}{*}{1} & \multirow{-2}{*}{256K} & \multirow{-2}{*}{1.30B}  & \multirow{-2}{*}{2}  & \multirow{-2}{*}{256} & \textbf{39} & \,\,\,0.450 & 0.002 & \,\,\,0.372 & 0.025  \\
    \addlinespace[-1pt] \cmidrule(l{2pt}r{2pt}){2-14} \addlinespace[-1pt]
    Gemma & & & & & & & & & 1 & 15.489 & 0.078 & 12.775 & 0.817 \\
     & \multirow{-2}{*}{18} & \multirow{-2}{*}{2048} & \multirow{-2}{*}{8} & \multirow{-2}{*}{1} & \multirow{-2}{*}{256K} & \multirow{-2}{*}{1.60B} & \multirow{-2}{*}{2} & \multirow{-2}{*}{512} & \textbf{39} & \,\,\,0.499 & 0.002 & \,\,\,0.422 & 0.025  \\
    \addlinespace[-1pt] \cmidrule(l{2pt}r{2pt}){2-14} \addlinespace[-1pt]
     & & & & & & & & & 1 & 10.546 & 0.081 & \,\,\,7.394 & 0.827 \\
     & \multirow{-2}{*}{18} & \multirow{-2}{*}{2048} & \multirow{-2}{*}{8} & \multirow{-2}{*}{1} & \multirow{-2}{*}{256K} & \multirow{-2}{*}{0.66B} &  \multirow{-2}{*}{3}  & \multirow{-2}{*}{-} & \textbf{43} & \,\,\,0.263 & 0.002 & \,\,\,0.182 & 0.023  \\
    \addlinespace[-1pt] \cmidrule(l{2pt}r{2pt}){2-14} \addlinespace[-1pt]
     & & & & & & & & & 1 & 11.871 & 0.080 & \,\,\,8.724 & 0.827 \\
     & \multirow{-2}{*}{18} & \multirow{-2}{*}{2048} & \multirow{-2}{*}{8} & \multirow{-2}{*}{1} & \multirow{-2}{*}{256K} & \multirow{-2}{*}{0.74B} &  \multirow{-2}{*}{3}  & \multirow{-2}{*}{64} & \textbf{43} & \,\,\,0.306 & 0.002 & \,\,\,0.182 & 0.023  \\
    \addlinespace[-1pt] \cmidrule(l{2pt}r{2pt}){2-14} \addlinespace[-1pt]
     & & & & & & & & & 1 & 11.768 & 0.080 & \,\,\,8.649 & 0.825 \\
     & \multirow{-2}{*}{18} & \multirow{-2}{*}{2048} & \multirow{-2}{*}{8} & \multirow{-2}{*}{1} & \multirow{-2}{*}{256K} & \multirow{-2}{*}{0.82B} &  \multirow{-2}{*}{3}  & \multirow{-2}{*}{128} & \textbf{43} & \,\,\,0.294 & 0.002 & \,\,\,0.221 & 0.023  \\
    \addlinespace[-1pt] \cmidrule(l{2pt}r{2pt}){2-14} \addlinespace[-1pt]
     & & & & & & & & & 1 & 12.018 & 0.081 & \,\,\,8.848 & 0.823 \\
     & \multirow{-2}{*}{18} & \multirow{-2}{*}{2048} & \multirow{-2}{*}{8} & \multirow{-2}{*}{1} & \multirow{-2}{*}{256K} & \multirow{-2}{*}{0.97B} &  \multirow{-2}{*}{3}  & \multirow{-2}{*}{256} & \textbf{41} & \,\,\,0.311 & 0.002 & \,\,\,0.226 & 0.024  \\
    \addlinespace[-1pt] \cmidrule(l{2pt}r{2pt}){2-14} \addlinespace[-1pt]
     & & & & & & & & & 1 & 12.087 & 0.082 & \,\,\,8.932 & 0.822 \\
     & \multirow{-2}{*}{18} & \multirow{-2}{*}{2048} & \multirow{-2}{*}{8} & \multirow{-2}{*}{1} & \multirow{-2}{*}{256K} & \multirow{-2}{*}{1.27B} &  \multirow{-2}{*}{3}  & \multirow{-2}{*}{512} & \textbf{39} & \,\,\,0.325 & 0.002 & \,\,\,0.237 & 0.025  \\
    \midrule
     & & & & & & & & & 1 & 22.016 & 0.082 & 21.010 & 0.188 \\
     & \multirow{-2}{*}{22} & \multirow{-2}{*}{2048} & \multirow{-2}{*}{32} & \multirow{-2}{*}{4} & \multirow{-2}{*}{32K} & \multirow{-2}{*}{0.97B} &  \multirow{-2}{*}{-}  & \multirow{-2}{*}{-} & \textbf{329} & \,\,\,0.819 & 0.000 & \,\,\,0.815 & 0.001  \\
    \addlinespace[-1pt] \cmidrule(l{2pt}r{2pt}){2-14} \addlinespace[-1pt]
     & & & & & & & & & 1 & 12.657 & 0.077 & 10.370 & 0.209 \\
     & \multirow{-2}{*}{22} & \multirow{-2}{*}{2048} & \multirow{-2}{*}{32} & \multirow{-2}{*}{4} & \multirow{-2}{*}{32K} & \multirow{-2}{*}{0.48B} &  \multirow{-2}{*}{2}  & \multirow{-2}{*}{-} & \textbf{233} & \,\,\,0.446 & 0.000 & \,\,\,0.413 & 0.001  \\
    \addlinespace[-1pt] \cmidrule(l{2pt}r{2pt}){2-14} \addlinespace[-1pt]
     & & & & & & & & & 1 & 15.243 & 0.079 & 12.908 & 0.211 \\
     TinyLlama & \multirow{-2}{*}{22} & \multirow{-2}{*}{2048} & \multirow{-2}{*}{32} & \multirow{-2}{*}{4} & \multirow{-2}{*}{32K} & \multirow{-2}{*}{0.53B} &  \multirow{-2}{*}{2}  & \multirow{-2}{*}{64} & \textbf{211} & \,\,\,0.454 & 0.000 & \,\,\,0.421 & 0.002  \\
    \addlinespace[-1pt] \cmidrule(l{2pt}r{2pt}){2-14} \addlinespace[-1pt]
     & & & & & & & & & 1 & 15.456 & 0.082 & 13.118 & 0.213 \\
     & \multirow{-2}{*}{22} & \multirow{-2}{*}{2048} & \multirow{-2}{*}{32} & \multirow{-2}{*}{4} & \multirow{-2}{*}{32K} & \multirow{-2}{*}{0.58B} &  \multirow{-2}{*}{2}  & \multirow{-2}{*}{128} & \textbf{209} & \,\,\,0.454 & 0.000 & \,\,\,0.421 & 0.002  \\
    \addlinespace[-1pt] \cmidrule(l{2pt}r{2pt}){2-14} \addlinespace[-1pt]
     & & & & & & & & & 1 & 15.223 & 0.081 & 12.908 & 0.208 \\
     & \multirow{-2}{*}{22} & \multirow{-2}{*}{2048} & \multirow{-2}{*}{32} & \multirow{-2}{*}{4} & \multirow{-2}{*}{32K} & \multirow{-2}{*}{0.68B} &  \multirow{-2}{*}{2}  & \multirow{-2}{*}{256} & \textbf{209} & \,\,\,0.457 & 0.000 & \,\,\,0.423 & 0.002  \\
    \addlinespace[-1pt] \cmidrule(l{2pt}r{2pt}){2-14} \addlinespace[-1pt]
     & & & & & & & & & 1 & 15.383 & 0.080 & 13.062 & 0.211 \\
     & \multirow{-2}{*}{22} & \multirow{-2}{*}{2048} & \multirow{-2}{*}{32} & \multirow{-2}{*}{4} & \multirow{-2}{*}{32K} & \multirow{-2}{*}{0.86B} &  \multirow{-2}{*}{2}  & \multirow{-2}{*}{512} & \textbf{209} & \,\,\,0.461 & 0.000 & \,\,\,0.428 & 0.002  \\
    \midrule
     & & & & & & & & & 1 & 13.280 & 0.080 & 12.286 & 0.235 \\
     & \multirow{-2}{*}{16} & \multirow{-2}{*}{2048} & \multirow{-2}{*}{8} & \multirow{-2}{*}{8} & \multirow{-2}{*}{50K} & \multirow{-2}{*}{0.81B} &  \multirow{-2}{*}{-}  & \multirow{-2}{*}{-} & \textbf{53} & \,\,\,1.227 & 0.002 & \,\,\,1.206 & 0.005  \\
    \addlinespace[-1pt] \cmidrule(l{2pt}r{2pt}){2-14} \addlinespace[-1pt]
     & & & & & & & & & 1 & \,\,\,8.423 & 0.081 & \,\,\,6.378 & 0.262 \\
     & \multirow{-2}{*}{16} & \multirow{-2}{*}{2048} & \multirow{-2}{*}{8} & \multirow{-2}{*}{8} & \multirow{-2}{*}{50K} & \multirow{-2}{*}{0.40B} &  \multirow{-2}{*}{2}  & \multirow{-2}{*}{-} & \textbf{61} & \,\,\,0.856 & 0.001 & \,\,\,0.606 & 0.005  \\
    \addlinespace[-1pt] \cmidrule(l{2pt}r{2pt}){2-14} \addlinespace[-1pt]
     & & & & & & & & & 1 & 10.554 & 0.082 & \,\,\,8.519 & 0.260 \\
    Pythia & \multirow{-2}{*}{16} & \multirow{-2}{*}{2048} & \multirow{-2}{*}{8} & \multirow{-2}{*}{8} & \multirow{-2}{*}{50K} & \multirow{-2}{*}{0.44B} &  \multirow{-2}{*}{2}  & \multirow{-2}{*}{64} & \textbf{63} & \,\,\,0.875 & 0.001 & \,\,\,0.626 & 0.005  \\
    \addlinespace[-1pt] \cmidrule(l{2pt}r{2pt}){2-14} \addlinespace[-1pt]
     & & & & & & & & & 1 & 10.167 & 0.076 & \,\,\,8.196 & 0.256 \\
     & \multirow{-2}{*}{16} & \multirow{-2}{*}{2048} & \multirow{-2}{*}{8} & \multirow{-2}{*}{8} & \multirow{-2}{*}{50K} & \multirow{-2}{*}{0.48B} &  \multirow{-2}{*}{2}  & \multirow{-2}{*}{128} & \textbf{59} & \,\,\,0.892 & 0.001 & \,\,\,0.642 & 0.005  \\
    \addlinespace[-1pt] \cmidrule(l{2pt}r{2pt}){2-14} \addlinespace[-1pt]
     & & & & & & & & & 1 & 10.410 & 0.079 & \,\,\,8.402 & 0.258 \\
     & \multirow{-2}{*}{16} & \multirow{-2}{*}{2048} & \multirow{-2}{*}{8} & \multirow{-2}{*}{8} & \multirow{-2}{*}{50K} & \multirow{-2}{*}{0.55B} &  \multirow{-2}{*}{2}  & \multirow{-2}{*}{256} & \textbf{59} & \,\,\,0.913 & 0.001 & \,\,\,0.662 & 0.005  \\
    \addlinespace[-1pt] \cmidrule(l{2pt}r{2pt}){2-14} \addlinespace[-1pt]
     & & & & & & & & & 1 & 12.609 & 0.091 & 10.311 & 0.267 \\
     & \multirow{-2}{*}{16} & \multirow{-2}{*}{2048} & \multirow{-2}{*}{8} & \multirow{-2}{*}{8} & \multirow{-2}{*}{50K} & \multirow{-2}{*}{0.70B} &  \multirow{-2}{*}{2}  & \multirow{-2}{*}{512} & \textbf{53} & \,\,\,0.956 & 0.002 & \,\,\,0.702 & 0.006  \\
    \bottomrule
    \end{tabular}
    }
    \caption{
    Measurements of generation time across three models using a single A100 40GB GPU. We measured time per token for both a batch size of 1 and the maximum batch size achievable by each model. The prefix length was set to 512 tokens, and the decoded output length to 2048 tokens. We then averaged the total elapsed time by the output length of 2048. Dummy input and dummy tensors were used for measurement. Both Gemma, employing multi-query attention, and TinyLlama, utilizing grouped-query attention, demonstrated fast generation speeds and large maximum batch sizes relative to their model sizes. TinyLlama's deep and narrow architecture allowed for a significantly large maximum batch size, although its generation speed was slower due to the increased number of layers.
    }
    \label{tab:measured_generation_time_a100}
\end{table}

\begin{table}[ht!]
    \small
    \centering
    \resizebox{\textwidth}{!}{
    \setlength{\tabcolsep}{3pt}
    \begin{tabular}{l|c|ccc|cc|cc|ccc|cc|ccc|c|rrrcc}
    \toprule
     & &  \multicolumn{3}{c|}{\textbf{Uptrain}} & \multicolumn{2}{c|}{\textbf{Looping}} &  \multicolumn{2}{c|}{\textbf{LoRA}} &  \multicolumn{3}{c|}{\textbf{Early-Exit\,Train}}  &  \multicolumn{2}{c|}{\textbf{Batching}} & \multicolumn{3}{c|}{\textbf{Few-shot Accuracy}} &  & \multicolumn{5}{c}{\textbf{Throughput\,$\uparrow$}} \\
     \cmidrule(l{2pt}r{2pt}){3-5} \cmidrule(l{2pt}r{2pt}){6-7}  \cmidrule(l{2pt}r{2pt}){8-9} \cmidrule(l{2pt}r{2pt}){10-12} \cmidrule(l{2pt}r{2pt}){13-14} \cmidrule(l{2pt}r{2pt}){15-17} \cmidrule(l{2pt}r{2pt}){19-23}
    \textbf{Models} & N-emb & PT & $N_{tok}$ & KD & Block & Init  & Rank & Init & $N_{tok}$ & CE & KD & Type & Exit & Last & Mid\,1 & Mid\,2 & Batch & SlimP & RedP & PG19 & $\Delta_{V}$ & $\Delta_{Seq}$ \\
    \midrule
      & 1.99B & \cmark & 75B & \xmark & - & -  & - & - & - & - & - & - & \xmark & 57.3 & - & - & 43 & 655 &  1228 &  1357 & \textcolor{gray}{{$\mathbf{\times 1.00}$}} & \textcolor{custom_red}{{$\mathbf{\times 0.71}$}} \\
      & 1.99B & \cmark & 75B & \xmark & - & -  & - & - &  - & - & - & CSB & \xmark & 57.3 & - & - & 43  &  1622 &  1604 &  1357 & \textcolor{custom_green}{{$\mathbf{\times 1.41}$}} & \textcolor{gray}{{$\mathbf{\times 1.00}$}} \\
      \cmidrule(l{2pt}r{2pt}){2-23}
      & 0.99B & \cmark & 60B & \cmark & 2 & Step  & - & - &  15B & Agg\,(0.1) & \cmark & CDB & \cmark & 54.0 & 48.8 & - & 43  &  3159 &  3050 &  2421 & \textcolor{custom_green}{{$\mathbf{\times {2.66}}$}} & \textcolor{custom_green}{{$\mathbf{\times {1.88}}$}} \\
      & 1.07B & \cmark & 60B & \cmark & 2 & Avg  & 64 & SVD &  15B & Agg\,(0.1) & \cmark & CDB & \cmark & 54.0 & 40.8 & - &  41  & 2357 &  2255 &  1858 & \textcolor{custom_green}{{$\mathbf{\times {2.00}}$}} & \textcolor{custom_green}{{$\mathbf{\times {1.41}}$}} \\
      & 1.15B & \cmark & 60B & \cmark & 2 & Avg  & 128 & SVD &  15B & Agg\,(0.1) & \cmark & CDB & \cmark & 54.6 & 40.2 & - & 41  &  2355 &  2250 &  1844 
 & \textcolor{custom_green}{{$\mathbf{\times {1.99}}$}} & \textcolor{custom_green}{{$\mathbf{\times {1.41}}$}} \\
      & 1.30B & \cmark & 60B & \cmark & 2 & Avg  & 256 & SVD &  15B & Agg\,(0.1) & \cmark & CDB & \cmark & 55.2 & 40.5 & - & 39  &  2047 &  1976 &  1740 
 & \textcolor{custom_green}{{$\mathbf{\times {1.78}}$}} & \textcolor{custom_green}{{$\mathbf{\times {1.26}}$}} \\
      & 1.60B & \cmark & 60B & \cmark & 2 & Avg  & 512 & SVD &  15B & Agg\,(0.1) & \cmark & CDB & \cmark & 56.2 & 41.7 & - & 39  &  1806 &  1754 &  1598 
 & \textcolor{custom_green}{{$\mathbf{\times {1.59}}$}} & \textcolor{custom_green}{{$\mathbf{\times {1.13}}$}} \\
      \cmidrule(l{2pt}r{2pt}){2-23}
      & 1.07B & \cmark & 60B & \cmark & 2 & Avg  & 64 & SVD &  15B & Agg\,(0.3) & \cmark & CDB & \cmark & 53.1 & 43.3 & - & 41  &  2454 &  2357 &  1929 
 & \textcolor{custom_green}{{$\mathbf{\times {2.08}}$}} & \textcolor{custom_green}{{$\mathbf{\times {1.47}}$}} \\
      & 1.15B & \cmark & 60B & \cmark & 2 & Avg  & 128 & SVD &  15B & Agg\,(0.3) & \cmark & CDB & \cmark & 53.6 & 43.4 & - & 41  &  2445 &  2346 &  1926 
 & \textcolor{custom_green}{{$\mathbf{\times {2.07}}$}} & \textcolor{custom_green}{{$\mathbf{\times {1.47}}$}} \\
      & 1.30B & \cmark & 60B & \cmark & 2 & Avg  & 256 & SVD &  15B & Agg\,(0.3) & \cmark & CDB & \cmark & 54.6 & 43.2 & - & 39  &  2123 &  2056 &  1804 
 & \textcolor{custom_green}{{$\mathbf{\times {1.85}}$}} & \textcolor{custom_green}{{$\mathbf{\times {1.31}}$}} \\
      Gemma & 1.60B & \cmark & 60B & \cmark & 2 & Avg  & 512 & SVD &  15B & Agg\,(0.3) & \cmark & CDB & \cmark & 55.2 & 44.0 & - & 39  &  1870 &  1819 &  1655 
 & \textcolor{custom_green}{{$\mathbf{\times {1.65}}$}} & \textcolor{custom_green}{{$\mathbf{\times {1.17}}$}} \\
      \cmidrule(l{2pt}r{2pt}){2-23}
      & 0.66B & \cmark & 60B & \cmark & 3 & Step  & - & - &  15B & Agg\,(0.1) & \cmark & CDB & \cmark & 51.9 & 49.0 & 43.5 & 43  &  3120 &  3041 &  2729 
 & \textcolor{custom_green}{{$\mathbf{\times {2.74}}$}} & \textcolor{custom_green}{{$\mathbf{\times {1.94}}$}} \\
      & 0.74B & \cmark & 60B & \cmark & 3 & Avg  & 64 & SVD &  15B & Agg\,(0.1) & \xmark & CDB & \cmark & 51.4 & 40.8 & 36.1 & 43  &  2334 &  2274 &  2059 
 & \textcolor{custom_green}{{$\mathbf{\times {2.06}}$}} & \textcolor{custom_green}{{$\mathbf{\times {1.45}}$}} \\
      & 0.82B & \cmark & 60B & \cmark & 3 & Avg  & 128 & SVD &  15B & Agg\,(0.1) & \xmark & CDB & \cmark & 51.7 & 41.1 & 36.1 & 43  &  2290 &  2230 &  2007 
 & \textcolor{custom_green}{{$\mathbf{\times {2.02}}$}} & \textcolor{custom_green}{{$\mathbf{\times {1.42}}$}} \\
      & 0.97B & \cmark & 60B & \cmark & 3 & Avg  & 256 & SVD &  15B & Agg\,(0.1) & \xmark & CDB & \cmark & 54.1 & 42.2 & 36.1 & 41  &  2281 &  2219 &  1984 
 & \textcolor{custom_green}{{$\mathbf{\times {2.00}}$}} & \textcolor{custom_green}{{$\mathbf{\times {1.41}}$}} \\
      & 1.27B & \cmark & 60B & \cmark & 3 & Avg  & 512 & SVD &  15B & Agg\,(0.1) & \xmark & CDB & \cmark & 55.7 & 43.5 & 37.0 &  39  & 2181 &  2122 &  1900 
 & \textcolor{custom_green}{{$\mathbf{\times {1.91}}$}} & \textcolor{custom_green}{{$\mathbf{\times {1.35}}$}} \\
      \cmidrule(l{2pt}r{2pt}){2-23}
      & 0.74B & \cmark & 60B & \cmark & 3 & Avg  & 64 & SVD &  15B & Agg\,(0.3) & \xmark & CDB & \cmark & 50.1 & 42.9 & 37.7 & 43  &  2427 &  2372 &  2143 
 & \textcolor{custom_green}{{$\mathbf{\times {2.14}}$}} & \textcolor{custom_green}{{$\mathbf{\times {1.51}}$}} \\
      & 0.82B & \cmark & 60B & \cmark & 3 & Avg  & 128 & SVD &  15B & Agg\,(0.3) & \xmark & CDB & \cmark & 51.2 & 43.2 & 38.0 & 43  &  2376 &  2321 &  2084 
 & \textcolor{custom_green}{{$\mathbf{\times {2.09}}$}} & \textcolor{custom_green}{{$\mathbf{\times {1.48}}$}} \\
      & 0.97B & \cmark & 60B & \cmark & 3 & Avg  & 256 & SVD &  15B & Agg\,(0.3) & \xmark & CDB & \cmark & 53.0 & 44.8 & 38.7 &  41  & 2359 &  2300 &  2039 
 & \textcolor{custom_green}{{$\mathbf{\times {2.07}}$}} & \textcolor{custom_green}{{$\mathbf{\times {1.46}}$}} \\
      & 1.27B & \cmark & 60B & \cmark & 3 & Avg  & 512 & SVD &  15B & Agg\,(0.3) & \xmark & CDB & \cmark & 54.1 & 45.7 & 39.2 & 39  &  2251 &  2191 &  1975 
 & \textcolor{custom_green}{{$\mathbf{\times {1.98}}$}} & \textcolor{custom_green}{{$\mathbf{\times {1.40}}$}} \\
      \midrule
      & 0.97B & \cmark & - & - & - & -  & - & - & - & - & - & - & \xmark & 43.3 & - & - & 329  &  1205 &  1220 &  1194 & \textcolor{gray}{{$\mathbf{\times 1.00}$}} & \textcolor{custom_red}{{$\mathbf{\times 0.99}$}}\\
      & 0.97B & \cmark & - & - & - & -  & - & - &  - & - & - & CSB & \xmark & 43.3 & - & - & 329  &  1227 &  1225 &  1194 & \textcolor{custom_green}{{$\mathbf{\times 1.01}$}} & \textcolor{gray}{{$\mathbf{\times 1.00}$}}\\
      \cmidrule(l{2pt}r{2pt}){2-23}
      & 0.48B & \cmark & 60B & \cmark & 2 & Step  & - & - &  15B & Agg\,(0.1) & \cmark & CDB & \cmark & 44.8 & 41.8 & - & 233  &  2038 &  2023 &  1933 
 & \textcolor{custom_green}{{$\mathbf{\times {1.66}}$}} & \textcolor{custom_green}{{$\mathbf{\times {1.64}}$}} \\
      & 0.53B & \cmark & 60B & \cmark & 2 & Avg  & 64 & SVD &  15B & Agg\,(0.1) & \cmark & CDB & \cmark & 45.1 & 33.7 & - & 211  &  1733 &  1719 &  1617  
 & \textcolor{custom_green}{{$\mathbf{\times {1.40}}$}} & \textcolor{custom_green}{{$\mathbf{\times {1.39}}$}} \\
      & 0.58B & \cmark & 60B & \cmark & 2 & Avg  & 128 & SVD &  15B & Agg\,(0.1) & \cmark & CDB & \cmark & 45.2 & 33.9 & - & 209  &  1733 &  1717 &  1609 
 & \textcolor{custom_green}{{$\mathbf{\times {1.40}}$}} & \textcolor{custom_green}{{$\mathbf{\times {1.39}}$}} \\
      TinyLlama & 0.68B & \cmark & 60B & \cmark & 2 & Avg  & 256 & SVD &  15B & Agg\,(0.1) & \cmark & CDB & \cmark & 45.6 & 33.9 & - & 209  &  1728 &  1714 &  1606  & \textcolor{custom_green}{{$\mathbf{\times {1.39}}$}} & \textcolor{custom_green}{{$\mathbf{\times {1.38}}$}} \\
      & 0.86B & \cmark & 60B & \cmark & 2 & Avg  & 512 & SVD &  15B & Agg\,(0.1) & \cmark & CDB & \cmark & 46.1 & 34.2 & - & 209  &  1716 &  1702 &  1581 
    & \textcolor{custom_green}{{$\mathbf{\times {1.38}}$}} & \textcolor{custom_green}{{$\mathbf{\times {1.37}}$}} \\    
      \cmidrule(l{2pt}r{2pt}){2-23}
      & 0.53B & \cmark & 60B & \cmark & 2 & Avg  & 64 & SVD &  15B & Agg\,(0.3) & \cmark & CDB & \cmark & 44.6 & 36.1 & - &  211  & 1810 &  1796 &  1688 
 & \textcolor{custom_green}{{$\mathbf{\times {1.46}}$}} & \textcolor{custom_green}{{$\mathbf{\times {1.45}}$}} \\
      & 0.58B & \cmark & 60B & \cmark & 2 & Avg  & 128 & SVD &  15B & Agg\,(0.3) & \cmark & CDB & \cmark & 44.8 & 36.0 & - & 209  &  1802 &  1787 &  1668  
 & \textcolor{custom_green}{{$\mathbf{\times {1.45}}$}} & \textcolor{custom_green}{{$\mathbf{\times {1.44}}$}} \\
      & 0.68B & \cmark & 60B & \cmark & 2 & Avg  & 256 & SVD &  15B & Agg\,(0.3) & \cmark & CDB & \cmark & 44.8 & 36.2 & - & 209  &  1793 &  1779 &  1668 
 & \textcolor{custom_green}{{$\mathbf{\times {1.45}}$}} & \textcolor{custom_green}{{$\mathbf{\times {1.44}}$}} \\
      & 0.86B & \cmark & 60B & \cmark & 2 & Avg  & 512 & SVD &  15B & Agg\,(0.3) & \cmark & CDB & \cmark & 45.8 & 36.5 & - & 209  &  1778 &  1763 &  1637  
 & \textcolor{custom_green}{{$\mathbf{\times {1.43}}$}} & \textcolor{custom_green}{{$\mathbf{\times {1.42}}$}} \\
      \midrule
      & 0.81B & \cmark & 75B & \xmark & - & -  & - & - & - & - & - & - & \xmark & 49.3 & - & - & 53  &  702 &  785 &  822 &\textcolor{gray}{{$\mathbf{\times 1.00}$}} & \textcolor{custom_red}{{$\mathbf{\times 0.93}$}} \\
      & 0.81B & \cmark & 75B & \xmark & - & -  & - & - &  - & - & - & CSB & \xmark & 49.3 & - & - & 53  &  829 &  827 &  822 & \textcolor{custom_green}{{$\mathbf{\times 1.07}$}} & \textcolor{gray}{{$\mathbf{\times 1.00}$}} \\
      \cmidrule(l{2pt}r{2pt}){2-23}
      & 0.40B & \cmark & 60B & \cmark & 2 & Step  & - & - &  15B & Agg\,(0.1) & \cmark & CDB & \cmark & 45.4 & 42.0 & - & 61  &  1339 &  1333 &  1281 
 & \textcolor{custom_green}{{$\mathbf{\times {1.71}}$}} & \textcolor{custom_green}{{$\mathbf{\times {1.60}}$}} \\
      & 0.44B & \cmark & 60B & \cmark & 2 & Avg  & 64 & SVD &  15B & Agg\,(0.1) & \cmark & CDB & \cmark & 46.1 & 37.1 & - & 63  &  1205 &  1203 &  1140 
 & \textcolor{custom_green}{{$\mathbf{\times {1.54}}$}} & \textcolor{custom_green}{{$\mathbf{\times {1.43}}$}} \\
      & 0.48B & \cmark & 60B & \cmark & 2 & Avg  & 128 & SVD &  15B & Agg\,(0.1) & \cmark & CDB & \cmark & 46.2 & 37.8 & - & 59  &  1156 &  1180 &  1108 
 & \textcolor{custom_green}{{$\mathbf{\times {1.49}}$}} & \textcolor{custom_green}{{$\mathbf{\times {1.39}}$}} \\
      Pythia & 0.55B & \cmark & 60B & \cmark & 2 & Avg  & 256 & SVD &  15B & Agg\,(0.1) & \cmark & CDB & \cmark & 46.5 & 38.0 & - & 59  &  1138 &  1139 &  1071 
 & \textcolor{custom_green}{{$\mathbf{\times {1.45}}$}} & \textcolor{custom_green}{{$\mathbf{\times {1.35}}$}} \\
      & 0.70B & \cmark & 60B & \cmark & 2 & Avg  & 512 & SVD &  15B & Agg\,(0.1) & \cmark & CDB & \cmark & 47.2 & 38.2 & - & 53  &  1051 &  1077 &  1021  
 & \textcolor{custom_green}{{$\mathbf{\times {1.36}}$}} & \textcolor{custom_green}{{$\mathbf{\times {1.27}}$}} \\
      \cmidrule(l{2pt}r{2pt}){2-23}
      & 0.44B & \cmark & 60B & \cmark & 2 & Avg  & 64 & SVD &  15B & Agg\,(0.3) & \cmark & CDB & \cmark & 45.1 & 39.0 & - & 63  &  1254 &  1252 &  1190  
 & \textcolor{custom_green}{{$\mathbf{\times {1.60}}$}} & \textcolor{custom_green}{{$\mathbf{\times {1.49}}$}} \\
      & 0.48B & \cmark & 60B & \cmark & 2 & Avg  & 128 & SVD &  15B & Agg\,(0.3) & \cmark & CDB & \cmark & 45.9 & 39.0 & - & 59  &  1200 &  1226 &  1153 
 & \textcolor{custom_green}{{$\mathbf{\times {1.55}}$}} & \textcolor{custom_green}{{$\mathbf{\times {1.45}}$}} \\
      & 0.55B & \cmark & 60B & \cmark & 2 & Avg  & 256 & SVD &  15B & Agg\,(0.3) & \cmark & CDB & \cmark & 46.0 & 39.4 & - & 59  &  1180 &  1180 &  1112 
 & \textcolor{custom_green}{{$\mathbf{\times {1.50}}$}} & \textcolor{custom_green}{{$\mathbf{\times {1.40}}$}} \\
      & 0.70B & \cmark & 60B & \cmark & 2 & Avg  & 512 & SVD &  15B & Agg\,(0.3) & \cmark & CDB & \cmark & 46.7 & 39.7 & - & 53  &  1088 &  1114 &  1058 
 & \textcolor{custom_green}{{$\mathbf{\times {1.41}}$}} & \textcolor{custom_green}{{$\mathbf{\times {1.32}}$}} \\
    \bottomrule
    \end{tabular}
    }
    \caption{
    Hypothetical generation speedup of Recursive Transformers across three models. We utilized the measurements of per-token generation time calculated in Table\,\ref{tab:measured_generation_time_a100}, which were calculated using an A100 40GB GPU, a prefix length of 512, and a decoding length of 2048. We only considered the time spent within Transformer blocks, simulating generation on the SlimPajama, RedPajama, and PG19 test sets. We used vanilla Transformer models, both with and without continuous sequence-wise batching (CSB), as our baselines. Our recursive models further enhance throughput by applying continuous depth-wise batching (CDB), leveraging looping and early-exiting techniques. The throughput improvements over the vanilla Transformer without and with sequence-wise batching are denoted as $\Delta_V$ and $\Delta_{Seq}$, respectively. To aid in understanding the speedup, we also provide the performance of intermediate layers and the maximum batch size.
    }
    \label{tab:final_performance_throughput}
\end{table}

\begin{table}[ht!]
    \small
    \centering
    \resizebox{\textwidth}{!}{
    \setlength{\tabcolsep}{6pt}
    \begin{tabular}{l|ccccc|c|cc|c|ccac}
    \toprule
    &  \multicolumn{5}{c|}{\textbf{Model Architecture}} &  & \multicolumn{2}{c|}{\textbf{Recursive}} & & \multicolumn{4}{c}{\textbf{Time\,(ms) per token}} \\
    \cmidrule(l{2pt}r{2pt}){2-6} \cmidrule(l{2pt}r{2pt}){8-9} \cmidrule(l{2pt}r{2pt}){11-14}
    \textbf{Models} & $N_L$ & $d_{model}$ & $N_{head}$ & $N_{KV}$ & Vocab & N-emb & Block & Rank & Batch & Total & Emb & \cellcolor{white} Transformer & Head  \\
    \midrule
    & & & & & & & & & 1 & 22.577 & 0.084 & 20.937 & 0.801\\
    &  \multirow{-2}{*}{18} & \multirow{-2}{*}{2048} & \multirow{-2}{*}{8} & \multirow{-2}{*}{1} & \multirow{-2}{*}{256K} & \multirow{-2}{*}{1.98B} &  \multirow{-2}{*}{-}  & \multirow{-2}{*}{-} & \textbf{111} & 
    \,\,\,0.207 & 0.001 & \,\,\,0.188 & 0.010\\
    \addlinespace[-1pt] \cmidrule(l{2pt}r{2pt}){2-14} \addlinespace[-1pt]
    &  & & & & & & & & 1 & 13.576 & 0.079 & 10.819 & 0.815\\
    &  \multirow{-2}{*}{18} & \multirow{-2}{*}{2048} & \multirow{-2}{*}{8} & \multirow{-2}{*}{1} & \multirow{-2}{*}{256K} & \multirow{-2}{*}{0.99B} &  \multirow{-2}{*}{2}  & \multirow{-2}{*}{-} & \textbf{123} & 
    \,\,\,0.118 & 0.001 & \,\,\,0.091 & 0.009\\
    \addlinespace[-1pt] \cmidrule(l{2pt}r{2pt}){2-14} \addlinespace[-1pt]
    & & & & & & & & & 1 & 15.372 & 0.080 & 12.675 & 0.813\\
    & \multirow{-2}{*}{18} & \multirow{-2}{*}{2048} & \multirow{-2}{*}{8} & \multirow{-2}{*}{1} & \multirow{-2}{*}{256K} & \multirow{-2}{*}{1.07B} &  \multirow{-2}{*}{2}  & \multirow{-2}{*}{64} & \textbf{117} & 
    \,\,\,0.140 & 0.001 & \,\,\,0.112 & 0.009\\
    \addlinespace[-1pt] \cmidrule(l{2pt}r{2pt}){2-14} \addlinespace[-1pt]
    & & & & & & & & & 1 & 15.631 & 0.082 & 12.899 & 0.816\\
    & \multirow{-2}{*}{18} & \multirow{-2}{*}{2048} & \multirow{-2}{*}{8} & \multirow{-2}{*}{1} & \multirow{-2}{*}{256K} & \multirow{-2}{*}{1.15B} &  \multirow{-2}{*}{2}  & \multirow{-2}{*}{128} & \textbf{115} & 
    \,\,\,0.141 & 0.001 & \,\,\,0.113 & 0.010\\
    \addlinespace[-1pt] \cmidrule(l{2pt}r{2pt}){2-14} \addlinespace[-1pt]
    &  & & & & & & & & 1 & 15.317 & 0.079 & 12.639 & 0.811\\
    & \multirow{-2}{*}{18} & \multirow{-2}{*}{2048} & \multirow{-2}{*}{8} & \multirow{-2}{*}{1} & \multirow{-2}{*}{256K} & \multirow{-2}{*}{1.30B} &  \multirow{-2}{*}{2}  & \multirow{-2}{*}{256} & \textbf{111} & 
    \,\,\,0.143 & 0.001 & \,\,\,0.115 & 0.010\\
    \addlinespace[-1pt] \cmidrule(l{2pt}r{2pt}){2-14} \addlinespace[-1pt]
    Gemma &  & & & & & & & & 1 & 15.379 & 0.080 & 12.692 & 0.807\\
    & \multirow{-2}{*}{18} & \multirow{-2}{*}{2048} & \multirow{-2}{*}{8} & \multirow{-2}{*}{1} & \multirow{-2}{*}{256K} & \multirow{-2}{*}{1.60B} &  \multirow{-2}{*}{2}  & \multirow{-2}{*}{512} & \textbf{103} & 
    \,\,\,0.158 & 0.001 & \,\,\,0.127 & 0.011\\
    \addlinespace[-1pt] \cmidrule(l{2pt}r{2pt}){2-14} \addlinespace[-1pt]
    &  & & & & & & & & 1 & 10.528 & 0.080 & \,\,\,7.411 & 0.817\\
    &  \multirow{-2}{*}{18} & \multirow{-2}{*}{2048} & \multirow{-2}{*}{8} & \multirow{-2}{*}{1} & \multirow{-2}{*}{256K} & \multirow{-2}{*}{0.66B} &  \multirow{-2}{*}{3}  & \multirow{-2}{*}{-} & \textbf{131} & 
    \,\,\,0.087 & 0.001 & \,\,\,0.058 & 0.010\\
    \addlinespace[-1pt] \cmidrule(l{2pt}r{2pt}){2-14} \addlinespace[-1pt]
    &  & & & & & & & & 1 & 11.957 & 0.081 & \,\,\,8.855 & 0.815\\
    & \multirow{-2}{*}{18} & \multirow{-2}{*}{2048} & \multirow{-2}{*}{8} & \multirow{-2}{*}{1} & \multirow{-2}{*}{256K} & \multirow{-2}{*}{0.74B} &  \multirow{-2}{*}{3}  & \multirow{-2}{*}{64} & \textbf{123} & 
    \,\,\,0.105 & 0.001 & \,\,\,0.075 & 0.009\\
    \addlinespace[-1pt] \cmidrule(l{2pt}r{2pt}){2-14} \addlinespace[-1pt]
    & & & & & & & & & 1 & 11.898 & 0.080 & \,\,\,8.787 & 0.816\\
    & \multirow{-2}{*}{18} & \multirow{-2}{*}{2048} & \multirow{-2}{*}{8} & \multirow{-2}{*}{1} & \multirow{-2}{*}{256K} & \multirow{-2}{*}{0.82B} &  \multirow{-2}{*}{3}  & \multirow{-2}{*}{128} & \textbf{121} & 
    \,\,\,0.103 & 0.001 & \,\,\,0.074 & 0.009\\
    \addlinespace[-1pt] \cmidrule(l{2pt}r{2pt}){2-14} \addlinespace[-1pt]
    & & & & & & & & & 1 & 11.734 & 0.079 & \,\,\,8.654 & 0.813\\
    & \multirow{-2}{*}{18} & \multirow{-2}{*}{2048} & \multirow{-2}{*}{8} & \multirow{-2}{*}{1} & \multirow{-2}{*}{256K} & \multirow{-2}{*}{0.97B} &  \multirow{-2}{*}{3}  & \multirow{-2}{*}{256} & \textbf{117} & 
    \,\,\,0.106 & 0.001 & \,\,\,0.076 & 0.009\\
    \addlinespace[-1pt] \cmidrule(l{2pt}r{2pt}){2-14} \addlinespace[-1pt]
    & & & & & & & & & 1 & 11.986 & 0.080 & \,\,\,8.856 & 0.809\\
    & \multirow{-2}{*}{18} & \multirow{-2}{*}{2048} & \multirow{-2}{*}{8} & \multirow{-2}{*}{1} & \multirow{-2}{*}{256K} & \multirow{-2}{*}{1.27B} &  \multirow{-2}{*}{3}  & \multirow{-2}{*}{512} & \textbf{107} & 
    \,\,\,0.125 & 0.001 & \,\,\,0.090 & 0.010\\
    \midrule
    & & & & & & & & & 1 & 23.898 & 0.080 & 22.909 & 0.189\\
    & \multirow{-2}{*}{22} & \multirow{-2}{*}{2048} & \multirow{-2}{*}{32} & \multirow{-2}{*}{4} & \multirow{-2}{*}{32K} & \multirow{-2}{*}{0.97B} &  \multirow{-2}{*}{-}  & \multirow{-2}{*}{-} & \textbf{1049} & 
    \,\,\,0.131 & 0.000 & \,\,\,0.129 & 0.001\\
    \addlinespace[-1pt] \cmidrule(l{2pt}r{2pt}){2-14} \addlinespace[-1pt]
    & & & & & & & & & 1 & 14.129 & 0.080 & 11.846 & 0.202\\
    & \multirow{-2}{*}{22} & \multirow{-2}{*}{2048} & \multirow{-2}{*}{32} & \multirow{-2}{*}{4} & \multirow{-2}{*}{32K} & \multirow{-2}{*}{0.48B} &  \multirow{-2}{*}{2}  & \multirow{-2}{*}{-} & \textbf{1121} & 
    \,\,\,0.070 & 0.000 & \,\,\,0.064 & 0.001\\
    \addlinespace[-1pt] \cmidrule(l{2pt}r{2pt}){2-14} \addlinespace[-1pt]
    & & & & & & & & & 1 & 14.897 & 0.080 & 12.627 & 0.202\\
    TinyLlama & \multirow{-2}{*}{22} & \multirow{-2}{*}{2048} & \multirow{-2}{*}{32} & \multirow{-2}{*}{4} & \multirow{-2}{*}{32K} & \multirow{-2}{*}{0.53B} &  \multirow{-2}{*}{2}  & \multirow{-2}{*}{64} & \textbf{1105} & 
    \,\,\,0.073 & 0.000 & \,\,\,0.068 & 0.001\\
    \addlinespace[-1pt] \cmidrule(l{2pt}r{2pt}){2-14} \addlinespace[-1pt]
    & & & & & & & & & 1 & 15.090 & 0.081 & 12.778 & 0.205\\
    & \multirow{-2}{*}{22} & \multirow{-2}{*}{2048} & \multirow{-2}{*}{32} & \multirow{-2}{*}{4} & \multirow{-2}{*}{32K} & \multirow{-2}{*}{0.58B} &  \multirow{-2}{*}{2}  & \multirow{-2}{*}{128} & \textbf{1089} & 
    \,\,\,0.074 & 0.000 & \,\,\,0.069 & 0.001\\
    \addlinespace[-1pt] \cmidrule(l{2pt}r{2pt}){2-14} \addlinespace[-1pt]
    & & & & & & & & & 1 & 14.962 & 0.081 & 12.659 & 0.201\\
    & \multirow{-2}{*}{22} & \multirow{-2}{*}{2048} & \multirow{-2}{*}{32} & \multirow{-2}{*}{4} & \multirow{-2}{*}{32K} & \multirow{-2}{*}{0.68B} &  \multirow{-2}{*}{2}  & \multirow{-2}{*}{256} & \textbf{1065} & 
    \,\,\,0.076 & 0.000 & \,\,\,0.071 & 0.001\\
    \addlinespace[-1pt] \cmidrule(l{2pt}r{2pt}){2-14} \addlinespace[-1pt]
    & & & & & & & & & 1 & 15.284 & 0.083 & 12.950 & 0.206\\
    & \multirow{-2}{*}{22} & \multirow{-2}{*}{2048} & \multirow{-2}{*}{32} & \multirow{-2}{*}{4} & \multirow{-2}{*}{32K} & \multirow{-2}{*}{0.86B} &  \multirow{-2}{*}{2}  & \multirow{-2}{*}{512} & \textbf{1017} & 
    \,\,\,0.080 & 0.000 & \,\,\,0.075 & 0.001\\
    \midrule
    & & & & & & & & & 1 & 13.341 & 0.081 & 12.326 & 0.239\\
    & \multirow{-2}{*}{16} & \multirow{-2}{*}{2048} & \multirow{-2}{*}{8} & \multirow{-2}{*}{8} & \multirow{-2}{*}{50K} & \multirow{-2}{*}{0.81B} &  \multirow{-2}{*}{-}  & \multirow{-2}{*}{-} & \textbf{229} & 
    \,\,\,0.176 & 0.000 & \,\,\,0.171 & 0.002\\
    \addlinespace[-1pt] \cmidrule(l{2pt}r{2pt}){2-14} \addlinespace[-1pt]
    & & & & & & & & & 1 & \,\,\,8.336 & 0.079 & \,\,\,6.303 & 0.261\\
    & \multirow{-2}{*}{16} & \multirow{-2}{*}{2048} & \multirow{-2}{*}{8} & \multirow{-2}{*}{8} & \multirow{-2}{*}{50K} & \multirow{-2}{*}{0.40B} &  \multirow{-2}{*}{2}  & \multirow{-2}{*}{-} & \textbf{241} & 
    \,\,\,0.121 & 0.000 & \,\,\,0.086 & 0.002\\
    \addlinespace[-1pt] \cmidrule(l{2pt}r{2pt}){2-14} \addlinespace[-1pt]
    & & & & & & & & & 1 & 10.408 & 0.081 & \,\,\,8.353 & 0.262\\
    Pythia & \multirow{-2}{*}{16} & \multirow{-2}{*}{2048} & \multirow{-2}{*}{8} & \multirow{-2}{*}{8} & \multirow{-2}{*}{50K} & \multirow{-2}{*}{0.44B} &  \multirow{-2}{*}{2}  & \multirow{-2}{*}{64} & \textbf{233} & 
    \,\,\,0.133 & 0.000 & \,\,\,0.097 & 0.002\\
    \addlinespace[-1pt] \cmidrule(l{2pt}r{2pt}){2-14} \addlinespace[-1pt]
    & & & & & & & & & 1 & 10.426 & 0.082 & \,\,\,8.378 & 0.259\\
    & \multirow{-2}{*}{16} & \multirow{-2}{*}{2048} & \multirow{-2}{*}{8} & \multirow{-2}{*}{8} & \multirow{-2}{*}{50K} & \multirow{-2}{*}{0.48B} &  \multirow{-2}{*}{2}  & \multirow{-2}{*}{128} & \textbf{221} & 
    \,\,\,0.137 & 0.000 & \,\,\,0.101 & 0.002\\
    \addlinespace[-1pt] \cmidrule(l{2pt}r{2pt}){2-14} \addlinespace[-1pt]
    & & & & & & & & & 1 & 10.509 & 0.080 & \,\,\,8.471 & 0.256\\
    & \multirow{-2}{*}{16} & \multirow{-2}{*}{2048} & \multirow{-2}{*}{8} & \multirow{-2}{*}{8} & \multirow{-2}{*}{50K} & \multirow{-2}{*}{0.55B} &  \multirow{-2}{*}{2}  & \multirow{-2}{*}{256} & \textbf{205} & 
    \,\,\,0.151 & 0.000 & \,\,\,0.115 & 0.002\\
    \addlinespace[-1pt] \cmidrule(l{2pt}r{2pt}){2-14} \addlinespace[-1pt]
    & & & & & & & & & 1 & 11.254 & 0.080 & \,\,\,9.241 & 0.257\\
    & \multirow{-2}{*}{16} & \multirow{-2}{*}{2048} & \multirow{-2}{*}{8} & \multirow{-2}{*}{8} & \multirow{-2}{*}{50K} & \multirow{-2}{*}{0.70B} &  \multirow{-2}{*}{2}  & \multirow{-2}{*}{512} & \textbf{165} & 
    \,\,\,0.177 & 0.001 & \,\,\,0.139 & 0.002\\
    \bottomrule
    \end{tabular}
    }
    \caption{
    Generation time measurements of Gemma models on a single A100 40GB GPU with 16GB memory constraint. 
    We measured generation time per token for both a batch size of 1 and the maximum batch size achievable by each model. 
    The prefix length was set to 64 tokens, and the decoded output length to 256 tokens. We then averaged the total elapsed time by the output length of 256. Dummy input and dummy tensors were used for measurement. 
    }
    \label{tab:measured_generation_time_a100_16gb}
\end{table}

\begin{table}[ht!]
    \small
    \centering
    \resizebox{\textwidth}{!}{
    \setlength{\tabcolsep}{3pt}
    \begin{tabular}{l|c|ccc|cc|cc|ccc|cc|ccc|c|rrrcc}
    \toprule
     & &  \multicolumn{3}{c|}{\textbf{Uptrain}} & \multicolumn{2}{c|}{\textbf{Looping}} &  \multicolumn{2}{c|}{\textbf{LoRA}} &  \multicolumn{3}{c|}{\textbf{Early-Exit\,Train}}  &  \multicolumn{2}{c|}{\textbf{Batching}} & \multicolumn{3}{c|}{\textbf{Few-shot Accuracy}} &  & \multicolumn{5}{c}{\textbf{Throughput\,$\uparrow$}} \\
     \cmidrule(l{2pt}r{2pt}){3-5} \cmidrule(l{2pt}r{2pt}){6-7}  \cmidrule(l{2pt}r{2pt}){8-9} \cmidrule(l{2pt}r{2pt}){10-12} \cmidrule(l{2pt}r{2pt}){13-14} \cmidrule(l{2pt}r{2pt}){15-17} \cmidrule(l{2pt}r{2pt}){19-23}
    \textbf{Models} & N-emb & PT & $N_{tok}$ & KD & Block & Init  & Rank & Init & $N_{tok}$ & CE & KD & Type & Exit & Last & Mid\,1 & Mid\,2 & Batch & SlimP & RedP & PG19 & $\Delta_{V}$ & $\Delta_{Seq}$ \\
    \midrule
      & 1.99B & \cmark & 75B & \xmark & - & -  & - & - & - & - & - & - & \xmark & 57.3 & - & - & 111 & 1740 &  3059 &  4796 & \textcolor{gray}{{$\mathbf{\times 1.00}$}} & \textcolor{custom_red}{{$\mathbf{\times 0.63}$}} \\
      & 1.99B & \cmark & 75B & \xmark & - & -  & - & - &  - & - & - & CSB & \xmark & 57.3 & - & - & 111  &  5287 &  5060 &  4796 & \textcolor{custom_green}{{$\mathbf{\times 1.58}$}} & \textcolor{gray}{{$\mathbf{\times 1.00}$}} \\
      \cmidrule(l{2pt}r{2pt}){2-23}
      & 0.99B & \cmark & 60B & \cmark & 2 & Step  & - & - &  15B & Agg\,(0.1) & \cmark & CDB & \cmark & 54.0 & 48.8 & - & 43  &  3159 &  3050 &  2421 & \textcolor{custom_green}{{$\mathbf{\times {2.50}}$}} & \textcolor{custom_green}{{$\mathbf{\times {1.59}}$}} \\
      & 1.07B & \cmark & 60B & \cmark & 2 & Avg  & 64 & SVD &  15B & Agg\,(0.1) & \cmark & CDB & \cmark & 54.0 & 40.8 & - &  41  & 2357 &  2255 &  1858 
 & \textcolor{custom_green}{{$\mathbf{\times {1.87}}$}} & \textcolor{custom_green}{{$\mathbf{\times {1.19}}$}} \\
      & 1.15B & \cmark & 60B & \cmark & 2 & Avg  & 128 & SVD &  15B & Agg\,(0.1) & \cmark & CDB & \cmark & 54.6 & 40.2 & - & 41  &  2355 &  2250 &  1844 
 & \textcolor{custom_green}{{$\mathbf{\times {1.87}}$}} & \textcolor{custom_green}{{$\mathbf{\times {1.19}}$}} \\
      & 1.30B & \cmark & 60B & \cmark & 2 & Avg  & 256 & SVD &  15B & Agg\,(0.1) & \cmark & CDB & \cmark & 55.2 & 40.5 & - & 39  &  2047 &  1976 &  1740 
 & \textcolor{custom_green}{{$\mathbf{\times {1.86}}$}} & \textcolor{custom_green}{{$\mathbf{\times {1.18}}$}} \\
      & 1.60B & \cmark & 60B & \cmark & 2 & Avg  & 512 & SVD &  15B & Agg\,(0.1) & \cmark & CDB & \cmark & 56.2 & 41.7 & - & 39  &  1806 &  1754 &  1598 
 & \textcolor{custom_green}{{$\mathbf{\times {1.73}}$}} & \textcolor{custom_green}{{$\mathbf{\times {1.10}}$}} \\
      \cmidrule(l{2pt}r{2pt}){2-23}
      & 1.07B & \cmark & 60B & \cmark & 2 & Avg  & 64 & SVD &  15B & Agg\,(0.3) & \cmark & CDB & \cmark & 53.1 & 43.3 & - & 41  &  2454 &  2357 &  1929 
 & \textcolor{custom_green}{{$\mathbf{\times {1.95}}$}} & \textcolor{custom_green}{{$\mathbf{\times {1.24}}$}} \\
      & 1.15B & \cmark & 60B & \cmark & 2 & Avg  & 128 & SVD &  15B & Agg\,(0.3) & \cmark & CDB & \cmark & 53.6 & 43.4 & - & 41  &  2445 &  2346 &  1926 
 & \textcolor{custom_green}{{$\mathbf{\times {1.95}}$}} & \textcolor{custom_green}{{$\mathbf{\times {1.24}}$}} \\
      & 1.30B & \cmark & 60B & \cmark & 2 & Avg  & 256 & SVD &  15B & Agg\,(0.3) & \cmark & CDB & \cmark & 54.6 & 43.2 & - & 39  &  2123 &  2056 &  1804 
 & \textcolor{custom_green}{{$\mathbf{\times {1.93}}$}} & \textcolor{custom_green}{{$\mathbf{\times {1.22}}$}} \\
      Gemma & 1.60B & \cmark & 60B & \cmark & 2 & Avg  & 512 & SVD &  15B & Agg\,(0.3) & \cmark & CDB & \cmark & 55.2 & 44.0 & - & 39  &  1870 &  1819 &  1655 
 & \textcolor{custom_green}{{$\mathbf{\times {1.79}}$}} & \textcolor{custom_green}{{$\mathbf{\times {1.14}}$}} \\
      \cmidrule(l{2pt}r{2pt}){2-23}
      & 0.66B & \cmark & 60B & \cmark & 3 & Step  & - & - &  15B & Agg\,(0.1) & \cmark & CDB & \cmark & 51.9 & 49.0 & 43.5 & 43  &  3120 &  3041 &  2729 
 & \textcolor{custom_green}{{$\mathbf{\times {2.62}}$}} & \textcolor{custom_green}{{$\mathbf{\times {1.66}}$}} \\
      & 0.74B & \cmark & 60B & \cmark & 3 & Avg  & 64 & SVD &  15B & Agg\,(0.1) & \xmark & CDB & \cmark & 51.4 & 40.8 & 36.1 & 43  &  2334 &  2274 &  2059 
 & \textcolor{custom_green}{{$\mathbf{\times {1.87}}$}} & \textcolor{custom_green}{{$\mathbf{\times {1.19}}$}} \\
      & 0.82B & \cmark & 60B & \cmark & 3 & Avg  & 128 & SVD &  15B & Agg\,(0.1) & \xmark & CDB & \cmark & 51.7 & 41.1 & 36.1 & 43  &  2290 &  2230 &  2007 
 & \textcolor{custom_green}{{$\mathbf{\times {1.90}}$}} & \textcolor{custom_green}{{$\mathbf{\times {1.20}}$}} \\
      & 0.97B & \cmark & 60B & \cmark & 3 & Avg  & 256 & SVD &  15B & Agg\,(0.1) & \xmark & CDB & \cmark & 54.1 & 42.2 & 36.1 & 41  &  2281 &  2219 &  1984 
 & \textcolor{custom_green}{{$\mathbf{\times {1.86}}$}} & \textcolor{custom_green}{{$\mathbf{\times {1.18}}$}} \\
      & 1.27B & \cmark & 60B & \cmark & 3 & Avg  & 512 & SVD &  15B & Agg\,(0.1) & \xmark & CDB & \cmark & 55.7 & 43.5 & 37.0 &  39  & 2181 &  2122 &  1900 
 & \textcolor{custom_green}{{$\mathbf{\times {1.62}}$}} & \textcolor{custom_green}{{$\mathbf{\times {1.03}}$}} \\
      \cmidrule(l{2pt}r{2pt}){2-23}
      & 0.74B & \cmark & 60B & \cmark & 3 & Avg  & 64 & SVD &  15B & Agg\,(0.3) & \xmark & CDB & \cmark & 50.1 & 42.9 & 37.7 & 43  &  2427 &  2372 &  2143 
 & \textcolor{custom_green}{{$\mathbf{\times {1.94}}$}} & \textcolor{custom_green}{{$\mathbf{\times {1.23}}$}} \\
      & 0.82B & \cmark & 60B & \cmark & 3 & Avg  & 128 & SVD &  15B & Agg\,(0.3) & \xmark & CDB & \cmark & 51.2 & 43.2 & 38.0 & 43  &  2376 &  2321 &  2084
 & \textcolor{custom_green}{{$\mathbf{\times {1.97}}$}} & \textcolor{custom_green}{{$\mathbf{\times {1.25}}$}} \\
      & 0.97B & \cmark & 60B & \cmark & 3 & Avg  & 256 & SVD &  15B & Agg\,(0.3) & \xmark & CDB & \cmark & 53.0 & 44.8 & 38.7 &  41  & 2359 &  2300 &  2039 
 & \textcolor{custom_green}{{$\mathbf{\times {1.92}}$}} & \textcolor{custom_green}{{$\mathbf{\times {1.22}}$}} \\
      & 1.27B & \cmark & 60B & \cmark & 3 & Avg  & 512 & SVD &  15B & Agg\,(0.3) & \xmark & CDB & \cmark & 54.1 & 45.7 & 39.2 & 39  &  2251 &  2191 &  1975
 & \textcolor{custom_green}{{$\mathbf{\times {1.67}}$}} & \textcolor{custom_green}{{$\mathbf{\times {1.06}}$}} \\
      \midrule
      & 0.97B & \cmark & - & - & - & -  & - & - & - & - & - & - & \xmark & 43.3 & - & - & 1049  &  6856 &  7481 &  4090 & \textcolor{gray}{{$\mathbf{\times 1.00}$}} & \textcolor{custom_red}{{$\mathbf{\times 0.96}$}}\\
      & 0.97B & \cmark & - & - & - & -  & - & - &  - & - & - & CSB & \xmark & 43.3 & - & - & 1049  &  7709 &  7481 &  4090 & \textcolor{custom_green}{{$\mathbf{\times 1.05}$}} & \textcolor{gray}{{$\mathbf{\times 1.00}$}}\\
      \cmidrule(l{2pt}r{2pt}){2-23}
      & 0.48B & \cmark & 60B & \cmark & 2 & Step  & - & - &  15B & Agg\,(0.1) & \cmark & CDB & \cmark & 44.8 & 41.8 & - & 233  &  2038 &  2023 &  1933 
 & \textcolor{custom_green}{{$\mathbf{\times {1.70}}$}} & \textcolor{custom_green}{{$\mathbf{\times {1.62}}$}} \\
      & 0.53B & \cmark & 60B & \cmark & 2 & Avg  & 64 & SVD &  15B & Agg\,(0.1) & \cmark & CDB & \cmark & 45.1 & 33.7 & - & 211  &  1733 &  1719 &  1617  
 & \textcolor{custom_green}{{$\mathbf{\times {1.38}}$}} & \textcolor{custom_green}{{$\mathbf{\times {1.32}}$}} \\
      & 0.58B & \cmark & 60B & \cmark & 2 & Avg  & 128 & SVD &  15B & Agg\,(0.1) & \cmark & CDB & \cmark & 45.2 & 33.9 & - & 209  &  1733 &  1717 &  1609 
 & \textcolor{custom_green}{{$\mathbf{\times {1.36}}$}} & \textcolor{custom_green}{{$\mathbf{\times {1.30}}$}} \\
      TinyLlama & 0.68B & \cmark & 60B & \cmark & 2 & Avg  & 256 & SVD &  15B & Agg\,(0.1) & \cmark & CDB & \cmark & 45.6 & 33.9 & - & 209  &  1728 &  1714 &  1606 
 & \textcolor{custom_green}{{$\mathbf{\times {1.34}}$}} & \textcolor{custom_green}{{$\mathbf{\times {1.28}}$}} \\
      & 0.86B & \cmark & 60B & \cmark & 2 & Avg  & 512 & SVD &  15B & Agg\,(0.1) & \cmark & CDB & \cmark & 46.1 & 34.2 & - & 209  &  1716 &  1702 &  1581 
 & \textcolor{custom_green}{{$\mathbf{\times {1.28}}$}} & \textcolor{custom_green}{{$\mathbf{\times {1.23}}$}} \\
      \cmidrule(l{2pt}r{2pt}){2-23}
      & 0.53B & \cmark & 60B & \cmark & 2 & Avg  & 64 & SVD &  15B & Agg\,(0.3) & \cmark & CDB & \cmark & 44.6 & 36.1 & - &  211  & 1810 &  1796 &  1688 
 & \textcolor{custom_green}{{$\mathbf{\times {1.45}}$}} & \textcolor{custom_green}{{$\mathbf{\times {1.38}}$}} \\
      & 0.58B & \cmark & 60B & \cmark & 2 & Avg  & 128 & SVD &  15B & Agg\,(0.3) & \cmark & CDB & \cmark & 44.8 & 36.0 & - & 209  &  1802 &  1787 &  1668  
 & \textcolor{custom_green}{{$\mathbf{\times {1.41}}$}} & \textcolor{custom_green}{{$\mathbf{\times {1.35}}$}} \\
      & 0.68B & \cmark & 60B & \cmark & 2 & Avg  & 256 & SVD &  15B & Agg\,(0.3) & \cmark & CDB & \cmark & 44.8 & 36.2 & - & 209  &  1793 &  1779 &  1668 
 & \textcolor{custom_green}{{$\mathbf{\times {1.39}}$}} & \textcolor{custom_green}{{$\mathbf{\times {1.33}}$}} \\
      & 0.86B & \cmark & 60B & \cmark & 2 & Avg  & 512 & SVD &  15B & Agg\,(0.3) & \cmark & CDB & \cmark & 45.8 & 36.5 & - & 209  &  1778 &  1763 &  1637  
 & \textcolor{custom_green}{{$\mathbf{\times {1.33}}$}} & \textcolor{custom_green}{{$\mathbf{\times {1.27}}$}} \\
      \midrule
      & 0.81B & \cmark & 75B & \xmark & - & -  & - & - & - & - & - & - & \xmark & 49.3 & - & - & 229  &  4273 &  5346 &  5149 &\textcolor{gray}{{$\mathbf{\times 1.00}$}} & \textcolor{custom_red}{{$\mathbf{\times 0.89}$}} \\
      & 0.81B & \cmark & 75B & \xmark & - & -  & - & - &  - & - & - & CSB & \xmark & 49.3 & - & - & 229  &  5813 &  5724 &  5149 & \textcolor{custom_green}{{$\mathbf{\times 1.13}$}} & \textcolor{gray}{{$\mathbf{\times 1.00}$}} \\
      \cmidrule(l{2pt}r{2pt}){2-23}
      & 0.40B & \cmark & 60B & \cmark & 2 & Step  & - & - &  15B & Agg\,(0.1) & \cmark & CDB & \cmark & 45.4 & 42.0 & - & 61  &  1339 &  1333 &  1281 
 & \textcolor{custom_green}{{$\mathbf{\times {1.77}}$}} & \textcolor{custom_green}{{$\mathbf{\times {1.57}}$}} \\
      & 0.44B & \cmark & 60B & \cmark & 2 & Avg  & 64 & SVD &  15B & Agg\,(0.1) & \cmark & CDB & \cmark & 46.1 & 37.1 & - & 63  &  1205 &  1203 &  1140 
 & \textcolor{custom_green}{{$\mathbf{\times {1.44}}$}} & \textcolor{custom_green}{{$\mathbf{\times {1.28}}$}} \\
      & 0.48B & \cmark & 60B & \cmark & 2 & Avg  & 128 & SVD &  15B & Agg\,(0.1) & \cmark & CDB & \cmark & 46.2 & 37.8 & - & 59  &  1156 &  1180 &  1108 
 & \textcolor{custom_green}{{$\mathbf{\times {1.32}}$}} & \textcolor{custom_green}{{$\mathbf{\times {1.17}}$}} \\
      Pythia & 0.55B & \cmark & 60B & \cmark & 2 & Avg  & 256 & SVD &  15B & Agg\,(0.1) & \cmark & CDB & \cmark & 46.5 & 38.0 & - & 59  &  1138 &  1139 &  1071
 & \textcolor{custom_green}{{$\mathbf{\times {1.22}}$}} & \textcolor{custom_green}{{$\mathbf{\times {1.08}}$}} \\
      & 0.70B & \cmark & 60B & \cmark & 2 & Avg  & 512 & SVD &  15B & Agg\,(0.1) & \cmark & CDB & \cmark & 47.2 & 38.2 & - & 53  &  1051 &  1077 &  1021  
 & \textcolor{custom_red}{{$\mathbf{\times {0.98}}$}} & \textcolor{custom_red}{{$\mathbf{\times {0.87}}$}} \\
      \cmidrule(l{2pt}r{2pt}){2-23}
      & 0.44B & \cmark & 60B & \cmark & 2 & Avg  & 64 & SVD &  15B & Agg\,(0.3) & \cmark & CDB & \cmark & 45.1 & 39.0 & - & 63  &  1254 &  1252 &  1190  
 & \textcolor{custom_green}{{$\mathbf{\times {1.50}}$}} & \textcolor{custom_green}{{$\mathbf{\times {1.33}}$}} \\
      & 0.48B & \cmark & 60B & \cmark & 2 & Avg  & 128 & SVD &  15B & Agg\,(0.3) & \cmark & CDB & \cmark & 45.9 & 39.0 & - & 59  &  1200 &  1226 &  1153 
 & \textcolor{custom_green}{{$\mathbf{\times {1.37}}$}} & \textcolor{custom_green}{{$\mathbf{\times {1.22}}$}} \\
      & 0.55B & \cmark & 60B & \cmark & 2 & Avg  & 256 & SVD &  15B & Agg\,(0.3) & \cmark & CDB & \cmark & 46.0 & 39.4 & - & 59  &  1180 &  1180 &  1112 
 & \textcolor{custom_green}{{$\mathbf{\times {1.27}}$}} & \textcolor{custom_green}{{$\mathbf{\times {1.12}}$}} \\
      & 0.70B & \cmark & 60B & \cmark & 2 & Avg  & 512 & SVD &  15B & Agg\,(0.3) & \cmark & CDB & \cmark & 46.7 & 39.7 & - & 53  &  1088 &  1114 &  1058 
 & \textcolor{custom_green}{{$\mathbf{\times {1.02}}$}} & \textcolor{custom_red}{{$\mathbf{\times {0.90}}$}} \\
    \bottomrule
    \end{tabular}
    }
    \caption{
    Hypothetical generation speedup of Recursive Transformers across three models. We utilized the measurements of per-token generation time calculated in Table\,\ref{tab:measured_generation_time_a100_16gb}, which were calculated using an A100 GPU with 16GB memory constraint, a prefix length of 64, and a decoding length of 256. We only considered the time spent within Transformer blocks, simulating generation on the SlimPajama, RedPajama, and PG19 test sets. We used vanilla transformer models, both with and without continuous sequence-wise batching (CSB), as our baselines. Our recursive models further enhance throughput by applying continuous depth-wise batching (CDB), leveraging looping and early-exiting techniques. The throughput improvements over the vanilla Transformer without and with sequence-wise batching are denoted as $\Delta_V$ and $\Delta_{Seq}$, respectively. To aid in understanding the speedup, we also provide the performance of intermediate layers and the maximum batch size.
    }
    \label{tab:final_performance_throughput_16gb}
\end{table}

}

\end{document}